\documentclass{vgtc}                          

\ifpdf
  \pdfoutput=1\relax                   
  \usepackage{graphicx}                
  \DeclareGraphicsExtensions{.pdf,.png,.jpg,.jpeg} 
\else
  \usepackage{graphicx}                
  \DeclareGraphicsExtensions{.eps}     
\fi%

\graphicspath{{figures/}{pictures/}{images/}{./}} 

\usepackage{microtype}                 
\PassOptionsToPackage{warn}{textcomp}  
\usepackage{textcomp}                  
\usepackage{mathptmx}                  
\usepackage{times}                     
\usepackage{cite}                      
\usepackage{tabu}                      
\usepackage{booktabs}                  
\usepackage{nicefrac}				   
\usepackage{mathptmx}
\usepackage{graphicx}
\usepackage{times}
\usepackage[linesnumbered,ruled]{algorithm2e}
\usepackage{amsfonts}
\usepackage{booktabs}         
\usepackage{gensymb}
\usepackage{amsmath}
\usepackage{comment}
\usepackage{times} 
\usepackage{caption}
\usepackage{bm}
\usepackage{multirow}
\usepackage{soul}
\usepackage{amssymb}
\newcommand{\hot}[1]{{\color{black} #1}}

\newcommand{\ndm}[1]{\frac{n}{m}}

\newenvironment{myitemize}{
\begin{itemize}
 \setlength{\itemsep}{1pt}
 \setlength{\parskip}{0pt}
 \setlength{\parsep}{0pt}}{\end{itemize}
}

\usepackage{float}
\DeclareMathOperator*{\IN}{in}
\DeclareMathOperator*{\OUT}{out}

\onlineid{0}

\vgtccategory{Research}

\vgtcinsertpkg

\title{ECNR: Efficient Compressive Neural Representation of Time-Varying Volumetric Datasets}

\author{Kaiyuan Tang\thanks{e-mail: ktang2@nd.edu} %
\and Chaoli Wang\thanks{e-mail: chaoli.wang@nd.edu}} %
\affiliation{\scriptsize University of Notre Dame}

\abstract{Due to its conceptual simplicity and generality, compressive neural representation has emerged as a promising alternative to traditional compression methods for managing massive volumetric datasets. The current practice of neural compression utilizes a single large multilayer perceptron (MLP) to encode the global volume, incurring slow training and inference. This paper presents an efficient compressive neural representation (ECNR) solution for time-varying data compression, utilizing the Laplacian pyramid for adaptive signal fitting. Following a multiscale structure, we leverage multiple small MLPs at each scale for fitting local content or residual blocks. By assigning similar blocks to the same MLP via size uniformization, we enable balanced parallelization among MLPs to significantly speed up training and inference. Working in concert with the multiscale structure, we tailor a deep compression strategy to compact the resulting model. We show the effectiveness of ECNR with multiple datasets and compare it with state-of-the-art compression methods (mainly SZ3, TTHRESH, and neurcomp). The results position ECNR as a promising solution for volumetric data compression.} 

\begin{document}


\maketitle

\vspace{-0.05in}
\section{Introduction}


Over the past few years, deep learning techniques have surfaced as a variable solution in solving different scientific visualization problems, including data compression~\cite{Wang-TVCG23}. Lu et al.\ \cite{Lu-CGF21} proposed neurcomp, a deep learning-based method for neural compression of volumetric datasets. They applied a {\em multilayer perceptron} (MLP) to fit a volume, where the network takes the spatial coordinates $(x,y,z)$ as input upon training and generates the corresponding voxel values to reconstruct the volume. As a result, only the trained model must be stored since the data in the entire volume can be inferred. This provides an excellent compression opportunity as the model size is orders of magnitude smaller than the size of the original volumetric data. Furthermore, the network can be extended to ingest spatiotemporal coordinates $(x,y,z,t)$ to represent time-varying volumetric data naturally.

Despite its great promise, neurcomp suffers several limitations. First, as a solution based on a coordinate-based network, neurcomp is slow in training and inference as an {\em entire} feedforward pass through the network must be computed for {\em every} sample (for training) and coordinate (for inference). Second, the current solution selects {\em random} voxel samples during training, ignoring that volumetric data often exhibit spatial features or regions of interest that demand greater attention than less important ones. Third, the temporal aspect is treated as an additional input coordinate to the network without a careful design that leverages their spatiotemporal {\em similarities} to speed up the training.

In this paper, we propose ECNR, an \underline{\bf e}fficient \underline{\bf c}ompressive \underline{\bf n}eural \underline{\bf r}epresentation for time-varying volumetric datasets. 
Unlike neurcomp, which employs a single large MLP to fit the entire volume, 
ECNR advocates MINER, a multiscale approach~\cite{saragadam-ECCV22} proposed for {\em implicit neural representation} (INR) of image and point cloud data. 
Similar to MINER, ECNR adaptively decomposes the spatiotemporal volume into blocks via the Laplacian pyramid, starting from the coarsest scale. As such, a block is only partitioned further if its residual remains significant, demanding the capture of finer space-time details for accurate signal reconstruction. To fit the local spatiotemporal blocks at each scale, we utilize multiple small MLPs, permitting fast encoding and decoding, reduced memory consumption, and enhanced reconstruction quality. 

Different from MINER, 
\hot{ECNR handles 4D (3D+time) volumetric datasets, while MINER only processes 2D static images or 3D mesh. We treat spatial and temporal dimensions equally and produce the same number of scales during data partitioning.} 
At each scale, we group similar blocks into clusters, and each cluster consists of nearly the same number of blocks. We then assign each cluster to an MLP and effectively train them in parallel. 
Furthermore, we leverage a deep compression strategy (including \hot{block-guided pruning, global quantization}, and entropy encoding) to compact the resulting model with a high compression rate (CR). 
Finally, we propose a lightweight convolutional neural network (CNN) to mitigate the possible boundary artifacts in the MLP-decoded results during inference. 

We demonstrate the effectiveness and efficiency of ECNR by comparing it with three state-of-the-art solutions: deep learning-based neurcomp and conventional compression methods SZ3~\cite{Liang-TBD22} and TTHRESH~\cite{Ripoll-TVCG20}. 
Compared with conventional compression methods (SZ3 and TTHRESH), ECNR achieves better quality (on average, +5.87 dB and +7.18 dB in PSNR) under high CRs. 
ECNR can handle time-varying volumetric datasets with a large spatial extent and/or long temporal sequence. Compared with neurcomp, it speeds up the encoding (up to 3.18$\times$) and decoding (up to 29.57$\times$) process, making it a more practical choice. 

\begin{table}[htb]
\caption{\hot{Comparison of different methods over encoding time (ET), decoding time (DT), compression rate (CR), and data quality (DQ).}}
\vspace{-0.1in}
\centering
{\scriptsize \hot{
\begin{tabular}{c|cccc}
             method  & ET & DT & CR & DQ   \\ \hline
 		SZ3  & very fast    & very fast    &  high  & medium \\   
		TTHRESH  & fast    & fast    &  high  & medium \\   
		neurcomp  & very slow    & slow    &  very high  & high \\
                ECNR  & slow   & fast   & very high  & high \\
\end{tabular}
}}
\label{tab:comp-methods}
\end{table}

\hot{Table~\ref{tab:comp-methods} summarizes the comparison among these four methods. As deep learning-based solutions that depend on GPUs, ECNR's {\em efficiency} advantage over neurcomp lies in its encoding and decoding speed improvement. Nevertheless, it is still slow compared with SZ3 and TTHRESH, which are CPU-based solutions. This renders ECNR unsuitable for time-critical scenarios like co-processing and in situ processing unless massive GPU parallelism (i.e., a large-scale GPU cluster) becomes commonplace. Our value proposition is that ECNR can be an ideal choice if one needs to archive a large time-varying dataset during post-processing when the speed is not the primary concern, aiming for highly compressive results while preserving the high fidelity of decompressed data.}


The contributions of this work can be summarized as follows. 
First, we present ECNR, a new INR-based compression method that \hot{uses a unified space-time partitioning strategy to} adaptively compresses the time-varying volumetric dataset using multiple small MLPs for fitting local blocks \hot{(Section~\ref{subsec:usta})}. 
Second, we propose a novel block assignment scheme that leads to balanced parallelization among MLPs, significantly speeding up training and inference (Section~\ref{subsec:ba}). 
Third, our deep MLP compression strategy features \hot{block-guided pruning} that adjusts the sparsity of each MLP based on the average block loss and \hot{global quantization} that supports fine-tuning shared parameters \hot{of all MLPs within the same scale} (Section~\ref{sec:deep-comp}). 
Fourth, we incorporate a lightweight CNN to mitigate block boundary artifacts in the MLP-decoded results (Section~\ref{subsec:bam}). 
All these additions (\hot{unified 4D partition}, block assignment, deep compression, boundary artifact mitigation) are not presented in MINER. 
They contribute to performance and quality gains, highlighting the differences and novelty of ECNR.

\vspace{-0.05in}
\section{Related Work}


{\bf Volume data compression.}
Given the prevalent need for data reduction in scientific computing, much research has been done to compress large-scale scientific simulation data. 
The earlier works leveraged wavelet transform~\cite{Muraki-VIS92} or discrete cosine transform (DCT)~\cite{Yeo-TVCG95}.  
Later, researchers developed TAMRESH~\cite{Suter-CGF13} and TTHRESH~\cite{Ripoll-TVCG20} that utilize data decomposition for tensor compression of multidimensional data over regular grids. 
Another compression technique is data fitting using pre-conditioners or predictors~\cite{lakshminarasimhan-ECPP11,Liang-ICBD18}. 
Other models seek to save different data features, such as graph-based models~\cite{Iverson-ECPP12}, topological features~\cite{Soler-PVIS18}, and dictionaries~\cite{Diaz-C&G20}. 
Researchers also explored various statistical approaches,  
such as frequency statistics and Gaussian mixture-based techniques and sampling, 
for solving similar problems~\cite{Wang-TVCG08,Dutta-PVIS17,Wang-PVIS17,Rapp-TVCG20,Biswas-TVCG21}. 
In a bigger scope, Li et al.\ \cite{Li-CGF18} surveyed data reduction techniques for simulation, visualization, and data analysis. 
Lu et al.\ \cite{Lu-CGF21} presented neurcomp to achieve volume data fitting via MLPs. 
Our work addresses its limitations by encoding time-varying data using multiple small MLPs that fit local spatiotemporal blocks, \hot{inspired by MINER~\cite{saragadam-ECCV22} and KiloNeRF~\cite{Reiser-ICCV21}.} 

\begin{figure*}[htb]
\centering
\includegraphics[width=\linewidth]{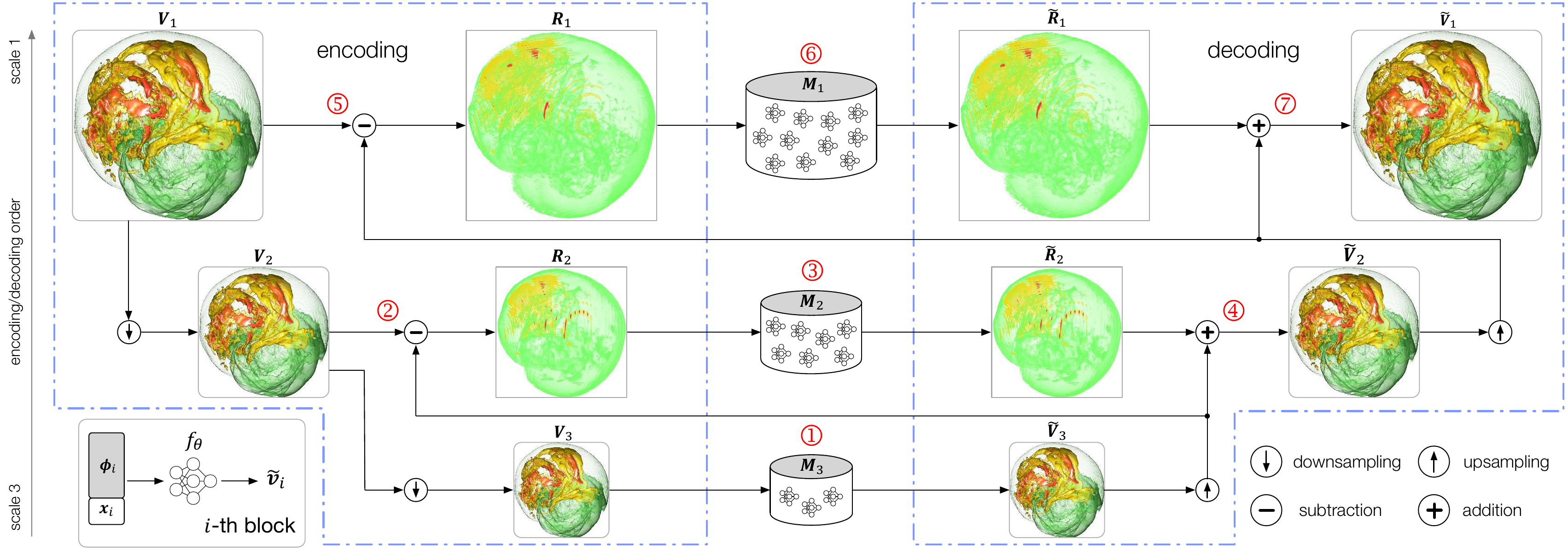}
\vspace{-.25in}
\caption{Employing the Laplacian pyramid, ECNR decomposes a volume into blocks in terms of their low-resolution content (coarsest scale) and residuals (finer scales). A three-scale ($s=3$) example with the simplified 3D version (i.e., the temporal dimension is omitted) is sketched. 
A group of MLPs encodes all blocks at the same scale.} 
\label{fig:ECNR-overview}
\end{figure*}

\hot{{\bf Neural field representation.}
In computer vision, neural field representation has become a popular topic due to its versatility in fitting various signals or tasks~\cite{Xie-CGF22}. 
%
%
Martel et al.\ \cite{Martel-TOG21} presented ACORN, an INR that optimizes the multiscale block hierarchy to represent large-scale or complex scenes. 
M{\"u}ller et al.\ \cite{Muller-TOG22} introduced Instant-NGP, which leverages multiresolution hash tables (MHTs) to dramatically downsize the number of network parameters. 
Signals across various scales often exhibit inherent similarities. 
This property has been leveraged in 3D volume representation using octrees~\cite{takikawa-CVPR21,Martel-TOG21,Li-TVCG}. 
Saragadam et al.\ \cite{saragadam-ECCV22} proposed MINER, a multiscale INR based on a Laplacian pyramid for more efficient presenting multiscale signals. 
%
%
Instant-NGP requires larger space than regular implicit networks to store explicit 4D MHTs for time-varying data, rendering it unsuitable for our data compression goal. 
Unlike ACORN~\cite{Martel-TOG21}, our ECNR follows MINER and employs the Laplacian pyramid to decompose signals into low-resolution content at the coarsest scale and residuals at subsequent finer scales (refer to Figure~\ref{fig:ECNR-overview} for an example). 
ECNR eliminates the need for time-consuming on-the-fly maintenance and tree structure updating. 
At the beginning of processing each scale, a one-time computation of block partitioning suffices.

In volume visualization, neural representation methods have been utilized for generative rendering model~\cite{Berger-TVCG19}, visualization synthesis~\cite{He-VIS19,Han-TVCG,Wu-arXiv23}, and image compression~\cite{Gu-CG23}. 
They have also been leveraged in interactive neural rendering via scene representation network (SRN), such as fV-SRN~\cite{Weiss-CGF22}, \hot{APMGSRN}~\cite{Wurster-VIS23}, and MHT-based INR~\cite{Wu-TVCG}. 
Other works include super-resolution generation~\cite{Han-TVCG22,Han-VIS19,Han-VIS21,Han-TVCG,Wu-arXiv23,Tang-CG}. 
Wurster et al.\ \cite{Wurster-TVCG} presented a hierarchical super-resolution solution that upscales volumetric data to a high resolution with minimal seam artifacts along boundaries of different resolutions. 
However, this hierarchical structure trains several large networks for upscaling, which can be expensive to store. In contrast, ENCR utilizes straightforward downsampling and upsampling with small MLPs to represent the volumetric data cost-effectively.}

\hot{{\bf Model compression.}
The common practice for compressing deep network models is a three-step process: network pruning, weight quantization, and entropy encoding~\cite{Han-arXiv15}. 
Wen et al.\ \cite{Wen-NIPS16} proposed a structured sparsity learning (SSL) method to regularize the network structures (i.e., filters, channels, filter shapes, and layer depth). 
Ye et al.\ \cite{Ye-arXiv18} utilized the alternating direction method of multipliers (ADMM) to perform weight pruning and clustering/quantization in a unified way. Additional techniques, including iterative weight quantization and retraining, joint weight clustering training and centroid updating, and weight clustering retraining, were also employed for further performance improvement. 
Molchanov et al.\ \cite{Molchanov-CVPR19} estimated the contribution of a neuron (filter) to the final loss and iteratively removed those with smaller scores. 
Lin et al.\ \cite{Lin-ICLR20} obtained a performant sparse model in one single training pass by dynamically allocating the sparsity pattern and incorporating feedback signal to reactivate prematurely pruned weights. 
%
Our deep compression of MLPs follows that of the common three-step~\cite{Han-arXiv15}, but we reconsider several design choices given the multiple-MLP architecture. Specifically, we propose the block-guided pruning and global quantization strategy for small MLPs within our model to further improve the performance.}

\begin{figure}[htb]
\centering
$\begin{array}{c@{\hspace{0.1in}}c}
 \includegraphics[width=0.45\linewidth]{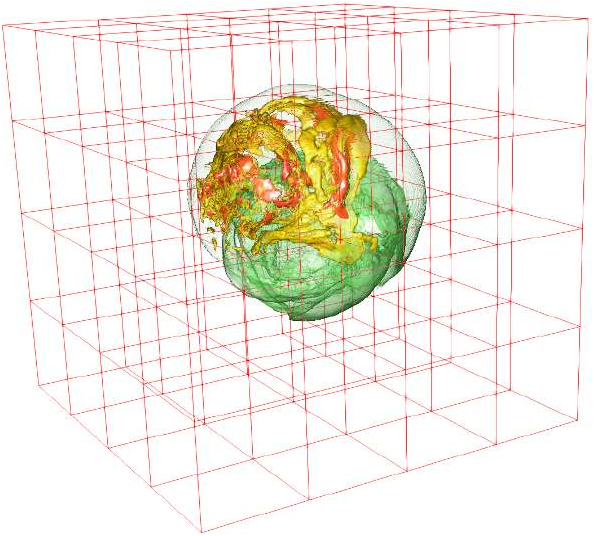}&
 \includegraphics[width=0.45\linewidth]{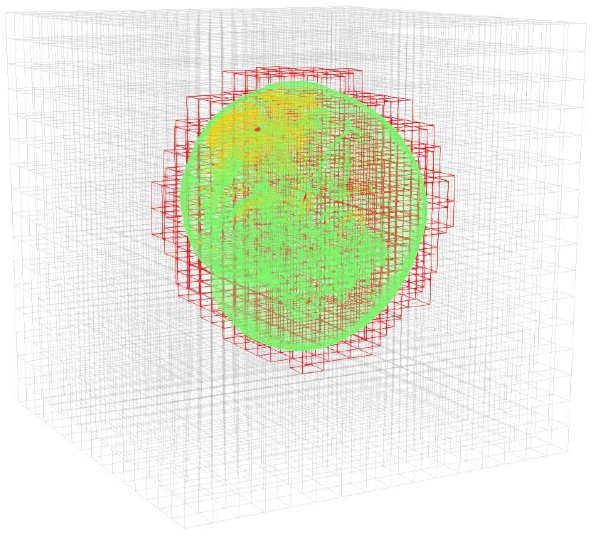}\\
 \mbox{(a) scale 3 partition} &\mbox{(b) scale 1 partition}\\
\end{array}$
\vspace{-.1in}
\caption{At each scale, the encoding target is split into equal-sized blocks, and only effective blocks with large residual values (shown in red bounding boxes) are processed.} 
\label{fig:ECNR-block-partition}
\end{figure}


\vspace{-0.05in}
\section{ECNR} 

Given a time-varying volume dataset, INR encodes the data by leveraging MLP, which inputs the coordinates to fit the volume. 
When employing a {\em single large global} MLP for the encoding task, a single forward pass through such an MLP may require millions of operations for one voxel. 
Thus, the cost of computing a complete volume with merely $100^3$ voxels already inflates to trillions of operations. 
ECNR employs {\em multiple small local} MLPs to encode the volume data to circumvent this issue. 
Since each MLP is small in parameter size and represents all voxels belonging to a disjoint cluster of local blocks, the computation time for a single MLP can be significantly reduced. 
Furthermore, parallel processing of MLPs boosts the efficiency of encoding the entire volume. 
Besides using multiple MLPs, ECNR integrates three main steps to improve the overall performance: 
a {\em spatiotemporal multiscale method} that decomposes the spatiotemporal volume into blocks hierarchically (Section~\ref{subsec:mbp}) and achieves fast convergence using a balanced block assignment scheme (Section~\ref{subsec:ba}), 
a {\em multiple-MLP deep compression strategy} that works in concert with the multiscale structure to compact the resulting model (Section~\ref{sec:deep-comp}), 
and a {\em lightweight CNN module} to mitigate the possible block boundary artifacts in the MLP-decoded results (Section~\ref{subsec:bam}). 

Our ECNR is an adaptive solution that utilizes MLPs with sinusoidal activation functions but focuses on time-varying volumetric data compression. Compared with the existing practice, ECNR offers several advantages. 
First, instead of relying on a predetermined network capacity (i.e., a fixed number of parameters) for encoding, ECNR exhibits superior robustness by adaptively adjusting network capacity to encode large-scale time-varying volumetric data, ensuring stable compression of data with various spatiotemporal characteristics. 
Second, ECNR employs a distinct group of MLPs at each scale for compressive neural representation, resulting in fast encoding and decoding, reduced memory consumption, and enhanced reconstruction quality. 
Third, 
\hot{ECNR leverages {\em block-guided pruning} that adjusts the sparsity of each MLP differently according to their nonuniform blocks fitting error, {\em global quantization} that allows representing MLP parameter with the shared parameters of all MLPs within the same scale, and {\em entropy encoding} to achieve a higher CR without quality loss.}

\vspace{-0.05in}
\subsection{\hot{Unified Space-Time Approach}}
\label{subsec:usta}

We advocate a unified approach that treats spatial and temporal domains equally and applies ECNR to the 4D space-time data. Both domains produce the same number of scales as we partition the data. Each partition divides each of the $(x, y, z, t)$ dimensions in half. At each scale, we split these 4D space-time blocks into individual 3D spatial blocks for efficient MLP fitting, which also enables us to handle datasets with long temporal sequences. These spatial blocks have the same dimension regardless of which scale they reside in. Assuming the volume dimension, block dimension, and number of scales are $(x, y, z, t)$, $(x_b, y_b, z_b)$, and $s$, respectively, then the number of spatial blocks at scale $i$ (where $i=s$ is the coarsest scale) will be $x/(2^{i-1} x_b) \times y/(2^{i-1} y_b) \times z/(2^{i-1} z_b) \times t/2^{i-1}$. \hot{As $i$ decreases, we have increasingly more blocks to discard at a fine scale if their residuals are small,} leading to a dramatically reduced number of effective blocks that need to be fitted. Since we process scale by scale, only the effective blocks at the current scale need to be loaded into memory simultaneously. With this spatiotemporal treatment, ECNR can handle volumetric datasets with a large spatial extent and/or long temporal sequence, as long as the memory can hold the effective blocks at the finest scale.

\vspace{-0.05in}
\subsection{Multiscale Block Partitioning}
\label{subsec:mbp}

Like MINER, ECNR represents the large-scale volume dataset as the sum of blocks from different scales.
As sketched in Figure~\ref{fig:ECNR-overview}, the volume is encoded/decoded sequentially 
scale by scale, starting from the coarsest scale. 
A distinct group of MLPs handles each scale to achieve fast convergence. 
Take the coarsest scale $s$, for example, 
let $\mathbf{V}_s$ be the volume,
$\widetilde{\mathbf{V}}_s$ be the ECNR-decoded (i.e., reconstructed) volume, and 
$M_s$ be the corresponding MLPs, 
we have $M_s(\mathbf{V}_s) \rightarrow \widetilde{\mathbf{V}}_s$, indicating that $\mathbf{V}_s$ is encoded by $M_s$ during training and then decoded to $\widetilde{\mathbf{V}}_s$ during inference.
The downsampling operator $D(\cdot)$, denoted as \textcircled{$\downarrow$} in Figure~\ref{fig:ECNR-overview}, subsamples the original data $\mathbf{V}$ once along each spatiotemporal dimension, and $\mathbf{V}_s$ is the application of $D(\cdot)$ to $\mathbf{V}$ for $(s-1)$ times. 
The upsampling operator $U(\cdot)$, denoted as \textcircled{$\uparrow$} in Figure~\ref{fig:ECNR-overview}, upsamples the reconstructed volume $\widetilde{\mathbf{V}}_s$ once along each spatiotemporal dimension by first conducting trilinear interpolation in the spatial domain and then linear interpolation in the temporal domain. 
The entire encoding and decoding process can be represented as 
\begin{small} 
\begin{align}
\label{eqn:multi-scale}
   \mbox{ \hot{\raisebox{.5pt}{\textcircled{\raisebox{-.9pt} {1}}}} }
   M_s(\mathbf{V}_s) &\rightarrow \widetilde{\mathbf{V}}_s;  \mbox{ }   
   \nonumber \\
    \mbox{ \hot{\raisebox{.5pt}{\textcircled{\raisebox{-.9pt} {2}}}} }
   \mathbf{V}_{s-1} \ominus U(\widetilde{\mathbf{V}}_s) \rightarrow \mathbf{R}_{s-1}, 
   \mbox{ \hot{\raisebox{.5pt}{\textcircled{\raisebox{-.9pt} {3}}}} }
  M_{s-1}(\mathbf{R}_{s-1}) &\rightarrow \widetilde{\mathbf{R}}_{s-1}, \mbox{ }
   \mbox{ \hot{\raisebox{.5pt}{\textcircled{\raisebox{-.9pt} {4}}}} }
  U(\widetilde{\mathbf{V}}_s) \oplus \widetilde{\mathbf{R}}_{s-1} \rightarrow \widetilde{\mathbf{V}}_{s-1};
 \nonumber \\
  			&\cdots   \nonumber \\
   \mbox{ \hot{\raisebox{.5pt}{\textcircled{\raisebox{-.9pt} {5}}}} }
   \mathbf{V}_1 \ominus U(\widetilde{\mathbf{V}}_2)  \rightarrow {\mathbf{R}}_1, \mbox{ }
   \mbox{ \hot{\raisebox{.5pt}{\textcircled{\raisebox{-.9pt} {6}}}} }
   M_1(\mathbf{R}_1) &\rightarrow \widetilde{\mathbf{R}}_1, \mbox{ }
      \mbox{ \hot{\raisebox{.5pt}{\textcircled{\raisebox{-.9pt} {7}}}} }
   U(\widetilde{\mathbf{V}}_2) \oplus \widetilde{\mathbf{R}}_1 \rightarrow \widetilde{\mathbf{V}}_1;
    \nonumber
\end{align}
\end{small}
where $\mathbf{V}_{s-1} \ominus U(\widetilde{\mathbf{V}}_s) \rightarrow \mathbf{R}_{s-1}$ indicates that residual $\mathbf{R}_{s-1}$ is the difference between $\mathbf{V}_{s-1}$ and the upscaled version of $\widetilde{\mathbf{V}}_s$. 
Given each scale's encoding target, we split these space-time target signals into individual 3D spatial blocks (refer to Figure~\ref{fig:ECNR-block-partition} for example). 
Each block is then assigned to a local MLP for encoding. 
Even though the fitting target covers different resolutions across scales, the block dimensions are identical. 
Therefore, a coarse scale will contain fewer blocks for fast encoding, while a fine scale will contain more blocks to handle the content-rich parts not preserved in the coarse scale. 
Still, a finer scale may also have content-poor blocks as the encoded residuals could be sparse. 
We discard these blocks (i.e., representing them as empty blocks) without MLP encoding if the $L_2$ norm of their residual is below a threshold. 
To increase the CR of ECNR, we initialize and assign each effective block a learnable latent code. We then concatenate the block's latent code with spatial coordinates as the input to each MLP, enabling it to identify and reconstruct multiple blocks based on different input latent codes.

\vspace{-0.05in}
\subsection{Block Assignment}
\label{subsec:ba}

MINER assigns each block to a different MLP for encoding. 
Unlike MINER, which only handles a single image, we tackle a sequence of volumes, demanding an MLP to encode multiple blocks to achieve good compression performance. 
This poses the issue of block assignment. 
After splitting the encoding target into $n_i$ 3D spatial blocks at scale $i$, we assume that ECNR leverages $m_i$ MLPs to encode these blocks ($n_i > m_i$). 
We are interested in investigating whether an effective assignment scheme exists to arrange $n_i$ blocks to $m_i$ MLPs. 
In our scenario, an MLP at a scale will be optimized based on blocks at the same scale across various spatiotemporal locations. While these blocks utilize distinct latent codes for encoding, the expressive capacity of these latent codes is considerably limited compared to MLPs. Consequently, it is necessary for the blocks assigned to a particular MLP to exhibit intrinsic similarities to achieve effective optimization. 

This paper addresses {\em block assignment} via {\em block clustering} using the k-means clustering method. 
%
Note that standard k-means clustering can lead to imbalanced block assignment among MLPs. 
In extreme cases, this could result in one MLP only getting a single block assigned and another MLP getting thousands of blocks. 
Such an assignment is not ideal due to the following reasons. 
First, MLPs getting a few blocks achieve a low CR, even with aggressive pruning in the subsequent deep compression.
Second, MLPs getting excessive blocks could not ensure high-quality reconstruction due to their limited network capacity. 
Third, imbalanced block assignment leads to an imbalanced workload that could severely impact parallel efficiency during encoding and decoding. 

\begin{algorithm}[htb]
    \caption{K-means cluster size uniformization}
    \label{alg:k-means-uniform}
        \KwIn{number of clusters $k$, set of $n$ points $D=\{d_1, d_2, \ldots, d_n\}$.} 
        \KwOut{cluster set $S=\{S_1,S_2, \ldots, S_k\}$ with nearly uniform cluster size and corresponding centroids $C=\{C_1, C_2, \ldots, C_k\}$}
      
        Initialize $S$ and $C$ using a standard k-means: $(S, C) \leftarrow Kmeans(D, k)$\; 
        Create distance matrix $\mathcal{D}[1,\ldots,n][1,\ldots,k]$\; 
        Create maximum gain array $\delta[1,\ldots,n]$\; 
                        \ForEach{$(d_i, C_j)$ in $\mathcal{D}$}{
               	$\mathcal{D}[i][j] = ||d_i - C_j||^2$\; 
            }          
            \tcc{calculate the assignment priority}
                        \ForEach{$d_i$ in $D$}{
               	$\delta[i] = \max(\mathcal{D}[i]) - \min(\mathcal{D}[i])$\; 
            }          
            Sort points in $D$ by $\delta$ in descending order\; 
            \tcc{reassign points to achieve nearly uniform cluster size}
            \ForEach{$d_i$ in sorted $D$}{
            select $\hat{S}$ from $S$ with cluster size less than $\lfloor \frac{n}{k} \rfloor$\; 
            	\lIf{$\hat{S} \neq \varnothing$}{assign $d_i$ to the nearest cluster in $\hat{S}$}
            	\lElse{assign $d_i$ to the nearest cluster in $S$ with cluster size less than $\lceil \frac{n}{k} \rceil$}
            }
        \Return $(S, C)$
\end{algorithm}

The ideal block assignment strategy should ensure each MLP processes a similar number of blocks. In other words, each cluster's size should be roughly the same. 
Given the target number of blocks $b_i$ that each MLP should fit at scale $i$, we set $m_i$ to $\lceil \frac{n_i}{b_i} \rceil$.
To this end, we propose a k-means cluster size uniformization algorithm (refer to Algorithm~\ref{alg:k-means-uniform}). 
The algorithm is initialized with a non-uniform clustering result from a standard k-means, where the distance between two points (in our case, blocks) is defined as the Euclidean distance between their corresponding voxel values (following the rationale that similar blocks should go to the same MLP).  
Then, for each point, we reassign it to its nearest unfull cluster in $\hat{S}$, which has not reached the desired uniform size (i.e., $b_i$). 
Note that the assignment order influences the result. 
In particular, the data point having the longest distance between its nearest ($\min(\mathcal{D}[i])$) and farthest ($\max(\mathcal{D}[i])$) clusters should be reassigned first, as reallocating this point correctly can maximize the gain (i.e., minimizing the inner-cluster distance and maximizing the inter-cluster distance). 
Our solution is easy to implement and runs efficiently to address the imbalance workload among MLPs. 
We provide further implementation details in the appendix. 

\vspace{-0.05in}
\subsection{Loss Function}

To optimize each MLP $f_{\theta}$, we compute for each block $f_{\theta}$ fits, the mean squared error (MSE) between its decoded version $\widetilde{\mathbf{v}}_i$ and GT version $\mathbf{v}_i$ as the loss function, which is defined as
\vspace{-0.1in}
\begin{equation}
	\label{eqn:loss}
	\mathcal{L}_{f_{\theta}}	= \frac{1}{j}\sum^{j}_{i=1}\| \widetilde{\mathbf{v}}_i - \mathbf{v}_i \|_2,
\vspace{-0.05in}
\end{equation}
where $j$ is the number of blocks for $f_{\theta}$ to fit, and $\widetilde{\mathbf{v}}_i = f_{\theta}(\mathbf{x}_i;\phi_i)$ with local coordinates $\mathbf{x}_i$ and corresponding latent code $\phi_i$ as input (refer to the bottom-left corner of Figure~\ref{fig:ECNR-overview}). 
MLPs are optimized sequentially from the coarsest to finest scales, while MLPs at the same scale are trained in parallel.

\vspace{-0.05in}
\subsection{Deep Compression of MLPs}
\label{sec:deep-comp}


We arrange similar blocks to the same MLP, and each MLP gets a similar number of blocks. This promotes a balanced workload as we input all voxels (content or residuals) in the blocks in batches for training (content-rich blocks do not take more time than content-poor ones). 
\hot{Furthermore, such an assignment allows us to prune each MLP with varying intensities during training, based on the fitting error of their respective blocks.}
As a result, one MLP handling content-poor blocks can prune more aggressively and consistently with high CR without quality downgrade, and another MLP tackling content-rich ones can prune more conservatively. After network training, \hot{global quantization} and entropy encoding follow to increase CR further. 

\hot{{\bf Block-guided pruning.}}
For MLPs at a scale of the multiscale representation's encoding phase, 
we prune unimportant neurons globally across all MLPs at the scale by permanently setting their parameters to zero. 
After pruning, the network may perform worse due to the reduction of parameter numbers. 
We fine-tune the pruned network with extra epochs to lessen the performance drop using a reduced learning rate. 
Specifically, we conduct \hot{{\em iterative pruning}~\cite{Han-NIPS15,Ye-arXiv18,Lin-ICLR20}}: instead of one-shot pruning, we prune and fine-tune MLPs multiple times to achieve a higher CR without significant performance degradation. 
For MLPs used in our work, we observe the following.
First, the number of parameters in the last layer for each MLP is much fewer than in other layers, so pruning parameters in the last layer cannot obtain significant model size reduction.
Second, MLPs that optimize simple blocks tend to have a sparser structure than those handling complex blocks.
Third, the bias in the first layer of each MLP is non-sparse and crucial for achieving good performance. 

Based on these insights, we prune all network parameters, excluding the weights in the last layer and biases in the first and last layers. 
For each pruning round, we rank each parameter by its importance. 
With a target sparsity, we start pruning from the least important parameters and iteratively prune the parameters until the target sparsity is met. 
The importance $\mathcal{I}$ of each candidate parameter $p$ is represented by
\vspace{-0.1in}
\begin{equation}
\label{eqn:importance-map}
	\mathcal{I}(p) = \mathcal{N}(|p|) + \lambda_b \mathcal{N}(\mathcal{L}_{b}(p)),
\vspace{-0.05in}
\end{equation}
where $|p|$ is absolute value of $p$, 
$\mathcal{L}_{b}(p)$ is the average block loss (we simply use MSE) for the MLP that contains $p$, 
$\lambda_b$ is a weight parameter, 
and $\mathcal{N}(\cdot)$ denotes min-max normalization. 
The two terms emphasize parameters that need pruning with high priority: 
$\mathcal{N}(|p|)$ indicates that parameters with small magnitude should be pruned first; 
$\mathcal{N}(\mathcal{L}_{b}(p))$ suggests that parameters in an MLP that achieve low average block loss should be pruned more aggressively.
As such, even though the difficulty of fitting the target blocks varies for each MLP, pruning can adaptively compact the MLPs. 

\hot{{\bf Global quantization.}}
After \hot{block-guided pruning}, we apply \hot{global quantization} to compress network parameters further. 
Network quantization has shown its effectiveness in neurcomp~\cite{Lu-CGF21}, and we follow a similar strategy. 
For example, given weight quantization precision $\beta$ bits, we employ standard k-means to group one layer of unpruned weights in an MLP to $2^\beta$ clusters. MLP weights in this quantized layer can then be represented as $\beta$-bit indices that map to $2^\beta$ floating-point shared parameters. These parameters are shared among unpruned weights in the layer. 

Besides the above quantization, 
we also make three improvements based on the structure of ECNR.
First, each scale consists of multiple MLPs. While each MLP encodes distinct blocks, these blocks often contain similar content that the same parameters can represent. 
Therefore, we apply \hot{global quantization} across all MLPs per scale rather than quantizing each MLP individually. 
Second, after quantization, we fix the parameter indices and fine-tune the corresponding 
shared parameters through backpropagation to achieve minimal performance degradation at the same precision. 
Third, 
instead of only quantizing the MLP's weights, 
we quantize MLPs' biases and weights, respectively, to boost CR under similar reconstruction accuracy.

{\bf Entropy encoding.} 
In the last step, we leverage entropy encoding to compress the resulting model file further using Huffman coding~\cite{Huffman-IRE52}, a lossless encoding scheme. Empirically, entropy encoding brings an additional 10\% model size reduction.

\vspace{-0.05in}
\subsection{Boundary Artifact Mitigation}
\label{subsec:bam}

ECNR decomposes volumetric data into disjoint blocks for encoding, which are then composed to form the complete volume. While this ensures high reconstruction accuracy at the {\em data level}, it does not guarantee high rendering fidelity at the {\em image level}, especially for data with intricate structures. The inconsistent outputs between adjacent blocks along their boundary could lead to artifacts in rendering. 
To mitigate this problem, we apply a {\em lightweight CNN module} to the output of the finest scale MLPs. The CNN's parameter size should be small for efficient compression: we use five convolution layers, each with 32 channels. Since the CNN is shallow, there is no need to add skip connections between layers. The loss function uses the MSE between the MLP-decoded results and ground-truth (GT) volume. 
Considering the trade-off between efficiency and CR, we only apply quantization with 9 bits at the end of CNN training without fine-tuning. This treatment reduces boundary artifacts with slightly increased encoding time and decreased CR. 


\begin{table}[htb]
\caption{The dimensions of experimented datasets.}
\vspace{-0.1in}
\centering
\resizebox{\columnwidth}{!}{
\begin{tabular}{c|cccc}
             & volume dimension                    & \hot{data} & block dimension  & \\ 
dataset & ($x \times y \times z \times t$) & \hot{size}  & ($x_b \times y_b \times z_b$) & scale \\ \hline
combustion & $480 \times 720 \times 120 \times 70$ & \hot{10.81 GB} & $40 \times 45 \times 15$ & 3\\ 
half-cylinder & $640 \times 240 \times 80 \times 100$ & \hot{4.57 GB} & $40 \times 30 \times 20$  & 3\\ 
solar plume & $256 \times 256 \times 1024 \times 28$ & \hot{7 GB} & $32 \times 32 \times 32$   & 3 \\ 
Targaroa &$300 \times 180 \times 120 \times 150$ & \hot{3.62 GB}  & $25 \times 15 \times 10$  & 3 \\ \hline
asteroids & $500 \times 500 \times 500 \times 1$ & \hot{476.83 MB} & $25 \times 25 \times 25$  & 3 \\ 
supernova-1 & $432 \times 432 \times 432 \times 1$ & \hot{307.54 MB}  & $18 \times 18 \times 18$ & 3 \\ 
supernova-2 & $1200 \times 1200 \times 1200 \times 1$ & \hot{6.43 GB} & $50 \times 50 \times 50$  & 4 \\ 
\hot{supernova-3} & \hot{$1728 \times 1728 \times 1728 \times 1$} & \hot{19.22 GB} & \hot{$72 \times 72 \times 72$}  & \hot{3} \\ 
\end{tabular}
}
\label{tab:dataset}
\end{table}

\vspace{-0.05in}
\section{Results and Discussion}

\subsection{Datasets and Network Training}

Table~\ref{tab:dataset} displays the datasets we experimented with. 
We primarily used four time-varying datasets for comparison with baseline methods. 
\hot{Four} static datasets were used in additional comparisons. 
We only used the first 70 of 100 timesteps for the combustion dataset because neurcomp could not handle more than 70 timesteps. 
A single NVIDIA A40 GPU, Intel Xeon CPU with 3.5 GHz, and 256 GB RAM was used for network training and inference. 
We set the block dimensions based on the volume dimensions. 
\hot{We set $s=3$ for most datasets to effectively filter out non-effective blocks. For large spatial datasets, we set $s=4$ for supernova-2 due to its spatial extent and $s=3$ for supernova-3 to conduct a comparative study with supernova-1.}
We removed a block at a finer scale if its residual's $L_2$ norm is below $10^{-4}$. 
From the coarsest to finest scales, we set the number of target blocks per MLP to 8, 16, and 32, respectively. 
All MLPs at each scale comprise three layers with sinusoidal activation functions followed by one linear layer for output.   The sinusoidal activation factor $\omega_0$ is set to 30. 
Each MLP contains 24 neurons in hidden layers. 
We initialized all layers in MLPs following Sitzmann et al.\ \cite{sitzmann-NIPS20} and utilized Adam optimizer with $\beta_1=0.9$, $\beta_2=0.999$, and a weight decay of $2\times10^{-5}$ to optimize the network parameters.

\begin{figure*}[htb]
	\begin{center}
		$\begin{array}{c@{\hspace{0.05in}}c@{\hspace{0.05in}}c@{\hspace{0.05in}}c}
				\includegraphics[height=1.0in]{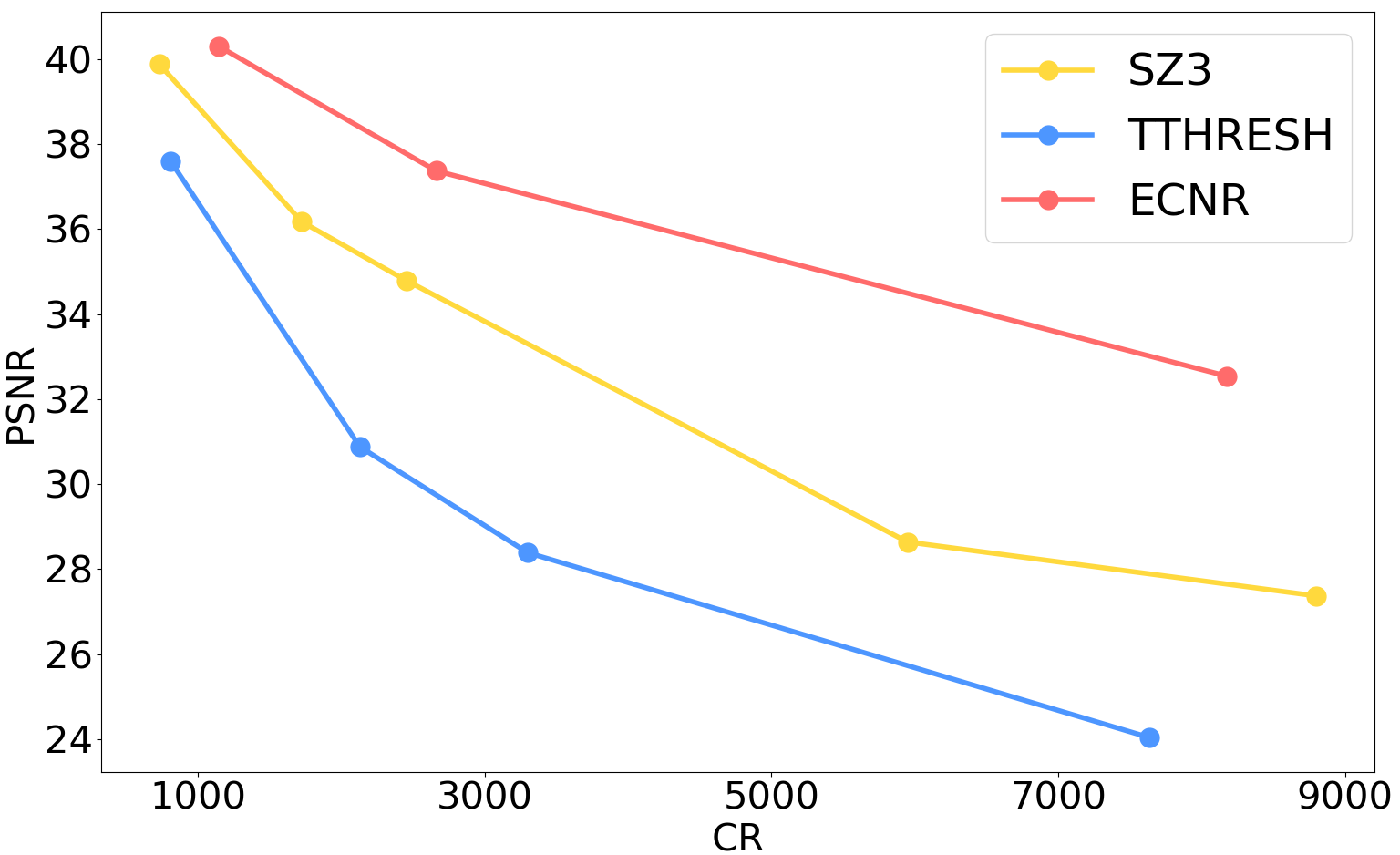} &
				\includegraphics[height=1.0in]{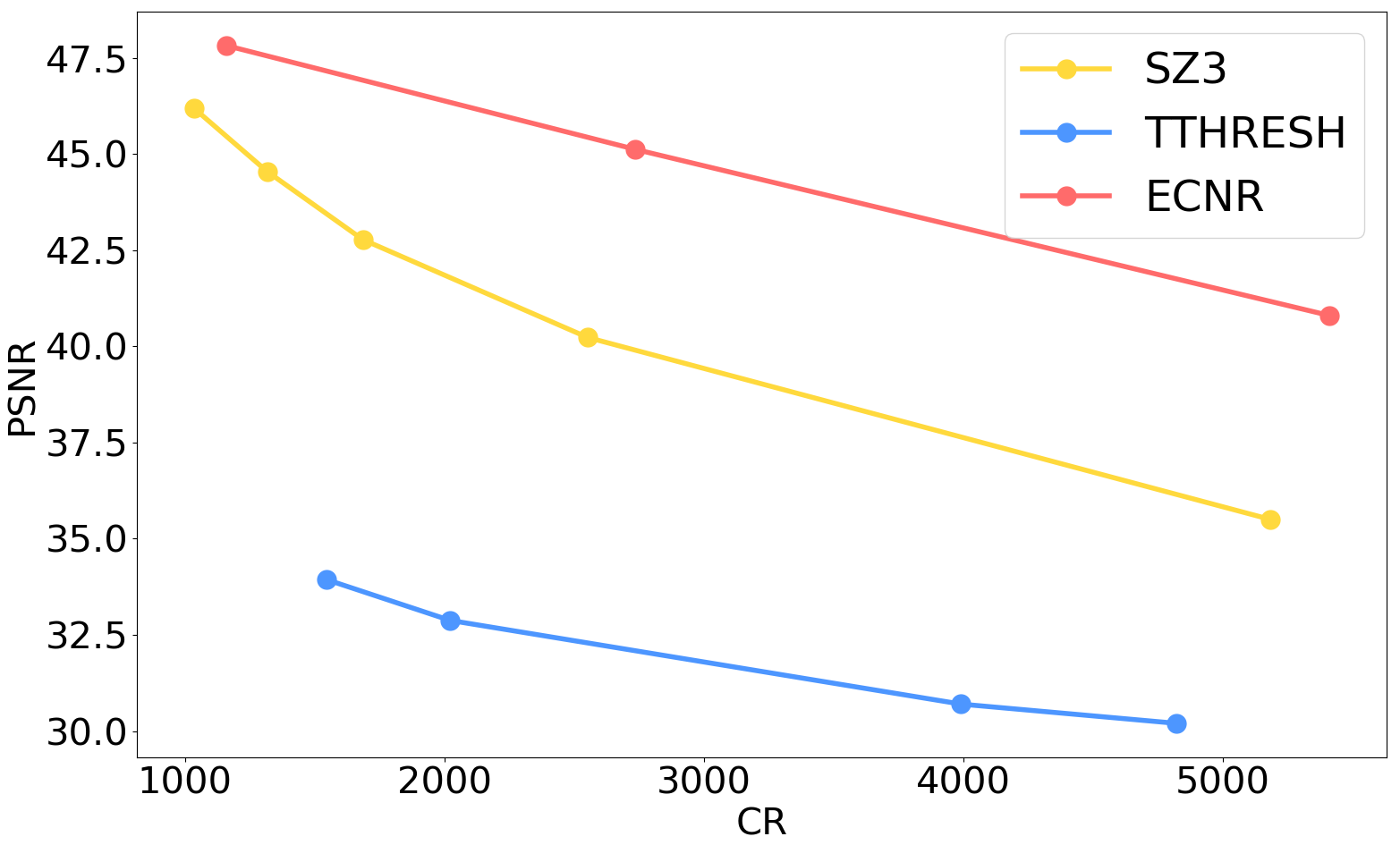} &
				\includegraphics[height=1.0in]{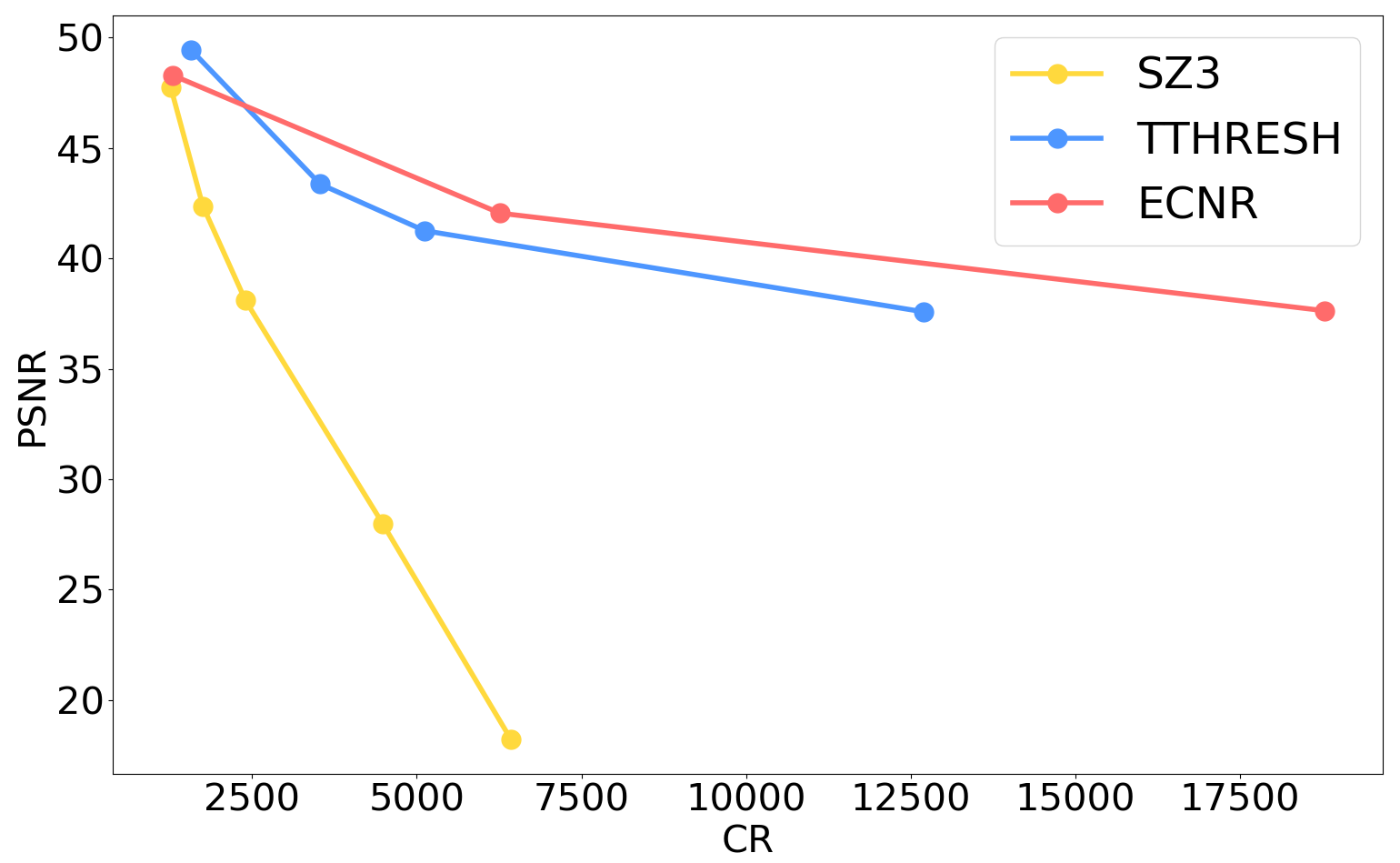} &
				\includegraphics[height=1.0in]{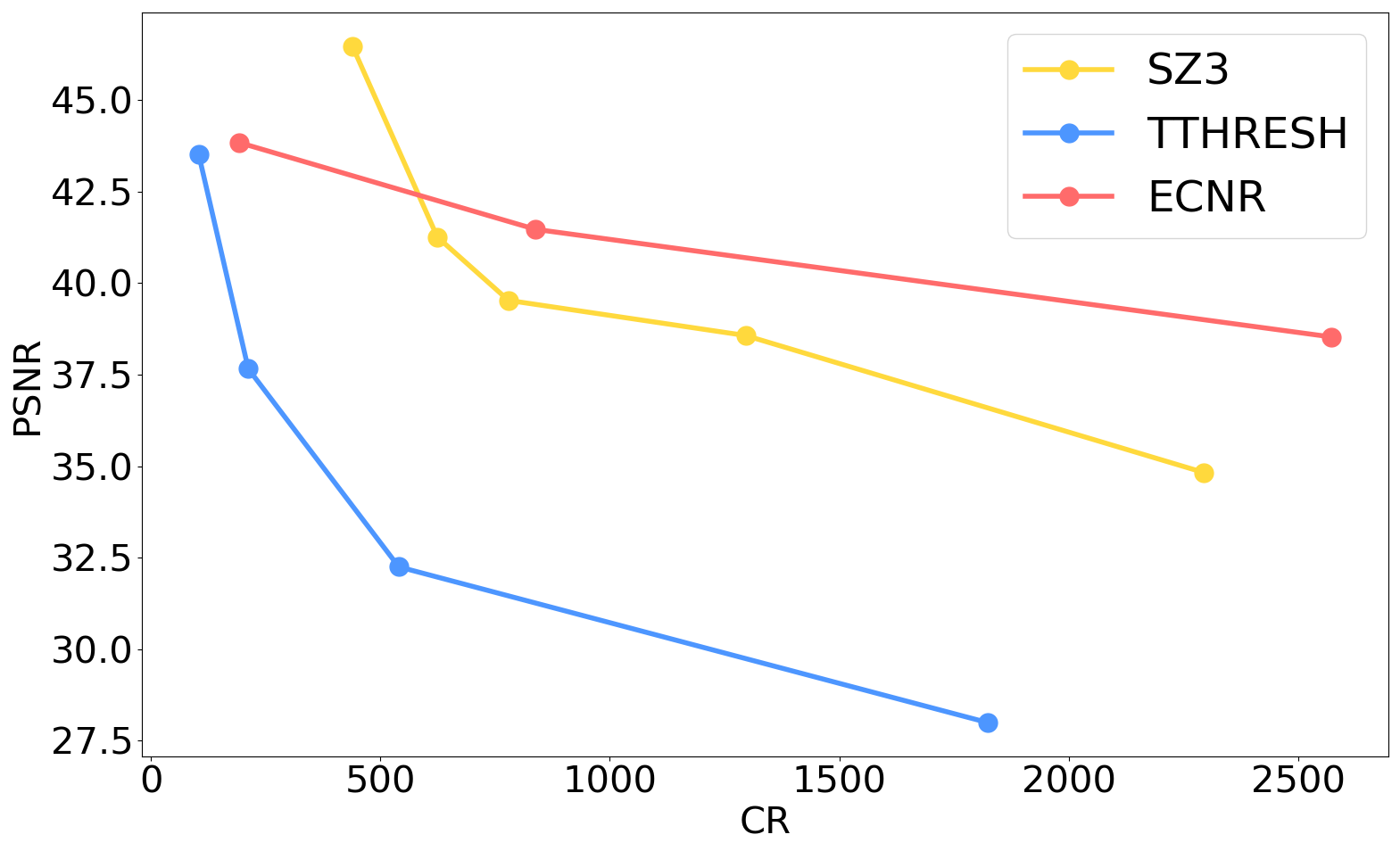} \\
				\mbox{\footnotesize (a) combustion}
				& \mbox{\footnotesize (b) half-cylinder}
				& \mbox{\footnotesize (c) solar plume}
				& \mbox{\footnotesize (d) Tangaroa}
			\end{array}$
	\end{center}
	\vspace{-.25in}
	\caption{Average PSNR (dB) values across all timesteps under different CRs.}
	\label{fig:diff-CR}
\end{figure*}

\begin{table}[htb]
	\caption{Average PSNR (dB), LPIPS, and CD values across all timesteps, as well as encoding time (ET), decoding time (DT), and \hot{compression rate (CR)}. 
	\hot{The isovalue $v$ reported is for CD calculation.}
	s, m, and h represent seconds, minutes, and hours. 
	The best performance is highlighted in bold.}
	\vspace{-0.1in}
	\centering
	\resizebox{\columnwidth}{!}{
		\begin{tabular}{c|c|lll|ll|l}
			dataset  & method   & PSNR$\uparrow$ & LPIPS$\downarrow$ & CD$\downarrow$ & ET$\downarrow$ & DT$\downarrow$ & CR$\uparrow$ \\
			\hline 
			               	        & SZ3      &34.79  &0.142   &1.67 &\textbf{70s} &\textbf{89s} &2,453 \\
 			combustion		& TTHRESH  &30.89 &\textbf{0.051}  &1.75    &32.4m &6.9m  &2,129      \\
			($v = -0.7$)	& neurcomp &33.15  &0.155   &2.54  &23.9h & 2.8h & 2,521 \\
			  				& ECNR 	   &\textbf{37.37}  &0.114   &\textbf{1.58} &11.4h &7.8m  &\textbf{2,662}    \\
			\hline
			               	& SZ3      	& 40.23    &0.063   &2.28  	& \textbf{19.2s}   & \textbf{10.3s} & 2,553 	 \\
			half-cylinder  	& TTHRESH  	& 32.87    &0.131   &12.82   & 118.3s 	 & 50.4s & 2,022  \\
			($v = 0.1$)		& neurcomp 	& 44.32    &0.031   &0.96   & 15.3h	 &38.6m & 2,633  \\
			   				& ECNR 	  	& \textbf{45.12}    &\textbf{0.026}   &\textbf{0.87}  & 4.8h & 1.3m & \textbf{2,736}	\\ 
			\hline
						& SZ3 &27.98 &0.343 &3.63 &\textbf{44.2s} &\textbf{1.1m} &4,483 \\ 
			solar plume	& TTHRESH  &41.25 &0.067 &\textbf{1.05} &23.3m &4.1m &5,119 \\
			($v = -0.5$) & neurcomp &41.98 &0.051 &2.15 &10.8h &1.5h &5,778\\
						& ECNR &\textbf{42.05} &\textbf{0.039} &1.98 &4.6h &5.1m &\textbf{6,262} \\
			\hline

			             	& SZ3      	&39.53 &0.085 &1.66 &\textbf{25.3s} &\textbf{32.8s} &779 \\
			Tangaroa		& TTHRESH  	&32.25 &0.240 &1.83 &6.2m &2.3m &540   \\
			($v = -0.85$)	& neurcomp 	&36.58 &0.097 &3.06 &3.6h &28.8m &772 \\
			  				& ECNR 	  	&\textbf{41.47} &\textbf{0.046} &\textbf{1.50} &98.5m &2.2m &\textbf{838}\\ 

		\end{tabular}
	}
	\label{tab:comp-time-varying}
\end{table}

For time-varying datasets, we trained the MLPs and latent codes at each scale for 500 epochs. 
We started with a learning rate of 0.001 and decayed it exponentially with a factor of 0.75 after each round of pruning. 
\hot{Block-guided pruning} occurs four rounds during training, first at 150 epochs, then at every 75 subsequent epochs until 375 epochs. 
$\lambda_b$ is set to 0.1 to compute $\mathcal{I}$ in Equation~\ref{eqn:importance-map} for pruning. 
Because the network contains more potentially sparse structures at the beginning of training, we pruned candidate network parameters to meet a target sparsity of 30\% at the first round of pruning and then 40\%, 45\%, and 50\%.  
Once trained, we quantized the MLPs with $\beta=8$ bits and then fine-tuned 75 epochs with a learning rate of $10^{-5}$. 
Subsequently, we trained a lightweight CNN using a learning rate of $10^{-5}$ for 100 epochs to mitigate boundary artifacts in the MLP-decoded results. 
For static datasets, we train the number of target blocks per MLP from the coarsest to finest scales was adjusted to 1, 2, and 4 because the dataset is static. We trained the MLPs and latent codes from the coarsest to finest scales for 200, 150, and 125 epochs. \hot{Block-guided pruning} happened at 75 and 115 epochs with a target sparsity of 20\% and 30\%. 

Note that ECNR involves extra space for storing the metadata, including data structure for block assignments and network configurations, which is encapsulated in the header of our output file and factors in the CR computation. Refer to the appendix for a detailed investigation.


\begin{figure*}[htb]
 \begin{center}
 $\begin{array}{c@{\hspace{0.025in}}c@{\hspace{0.025in}}c@{\hspace{0.025in}}c@{\hspace{0.025in}}c}
 \includegraphics[width=0.1925\linewidth]{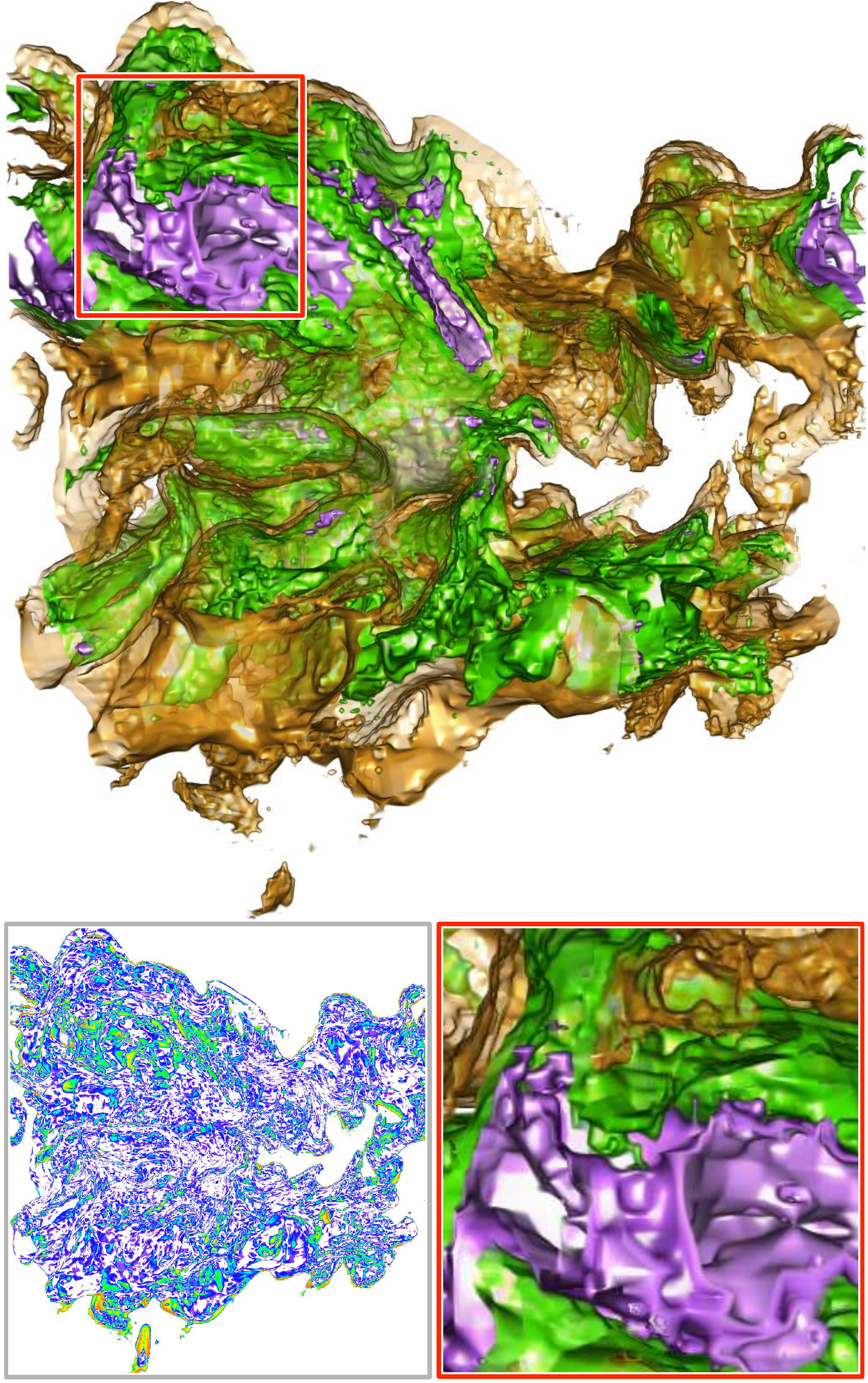}&
 \includegraphics[width=0.1925\linewidth]{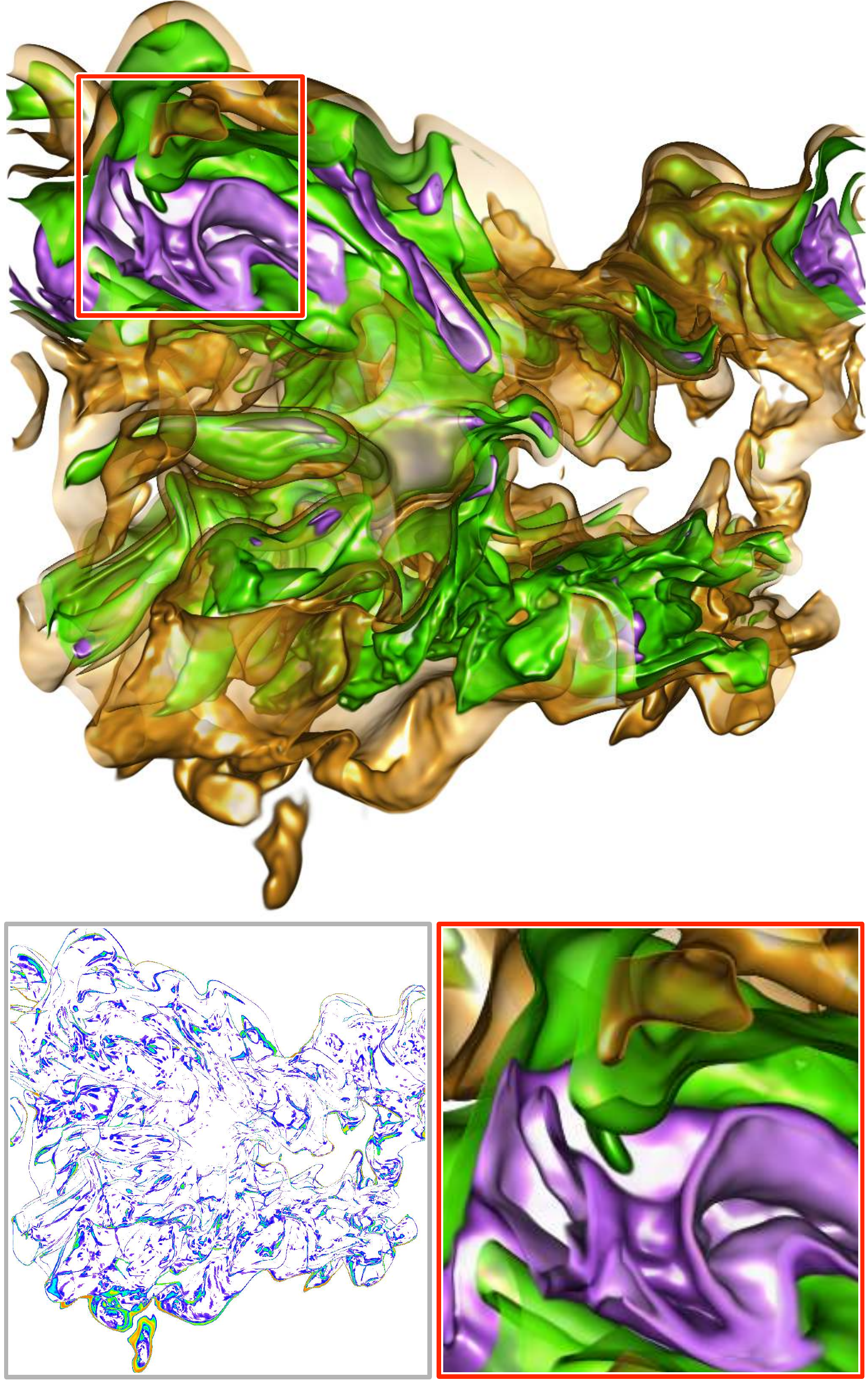}&
 \includegraphics[width=0.1925\linewidth]{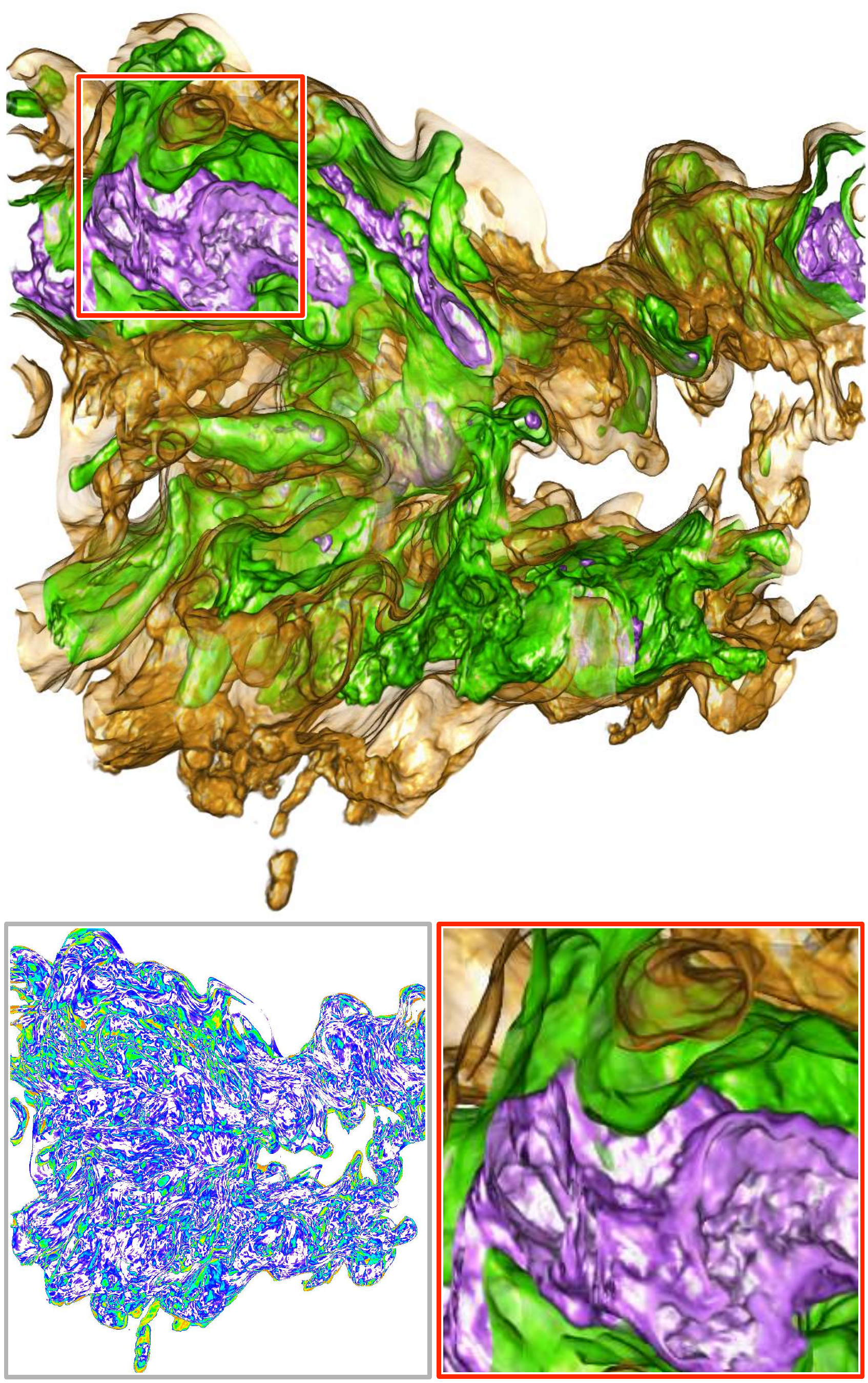}&
  \includegraphics[width=0.1925\linewidth]{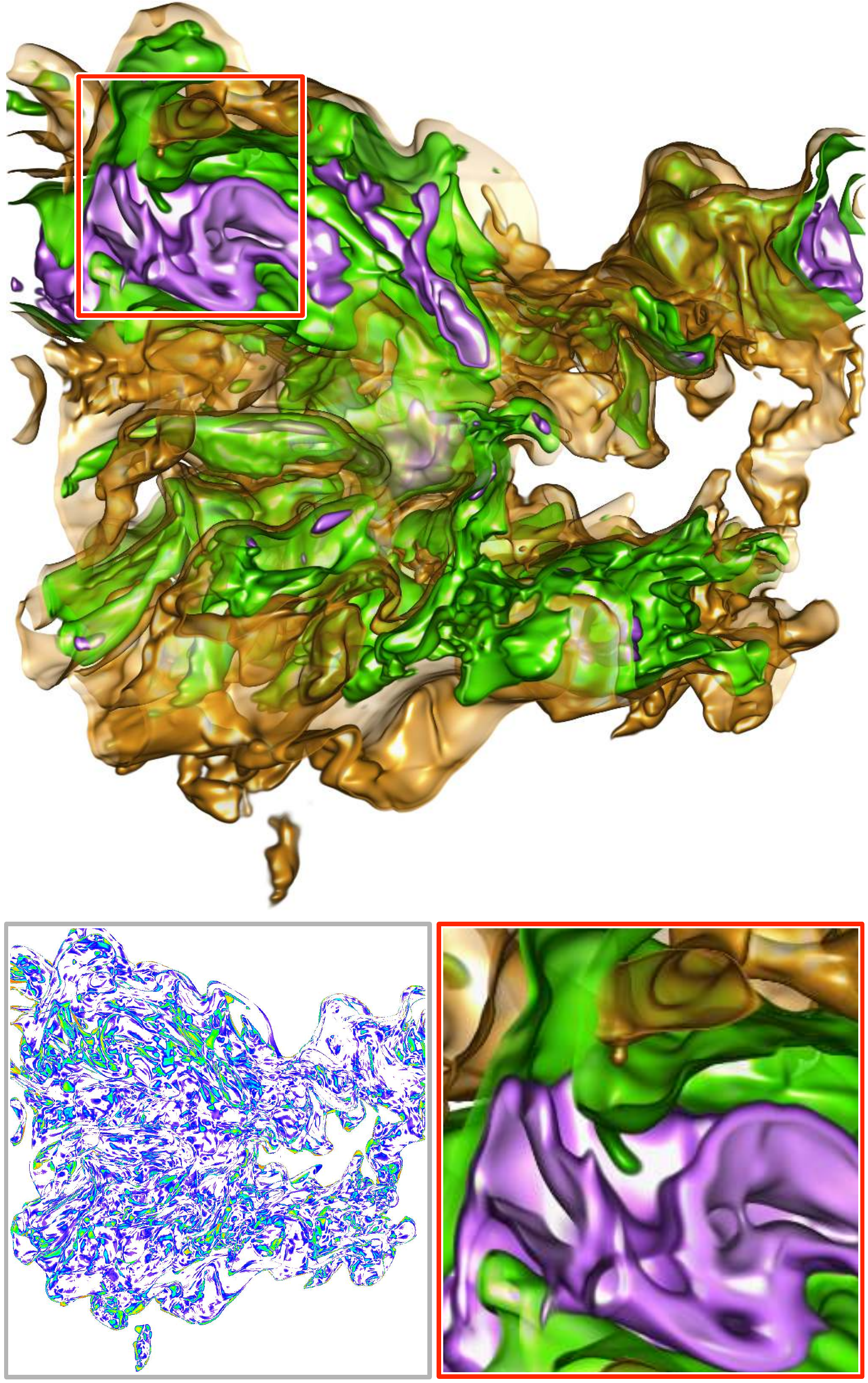}&
\includegraphics[width=0.1925\linewidth]{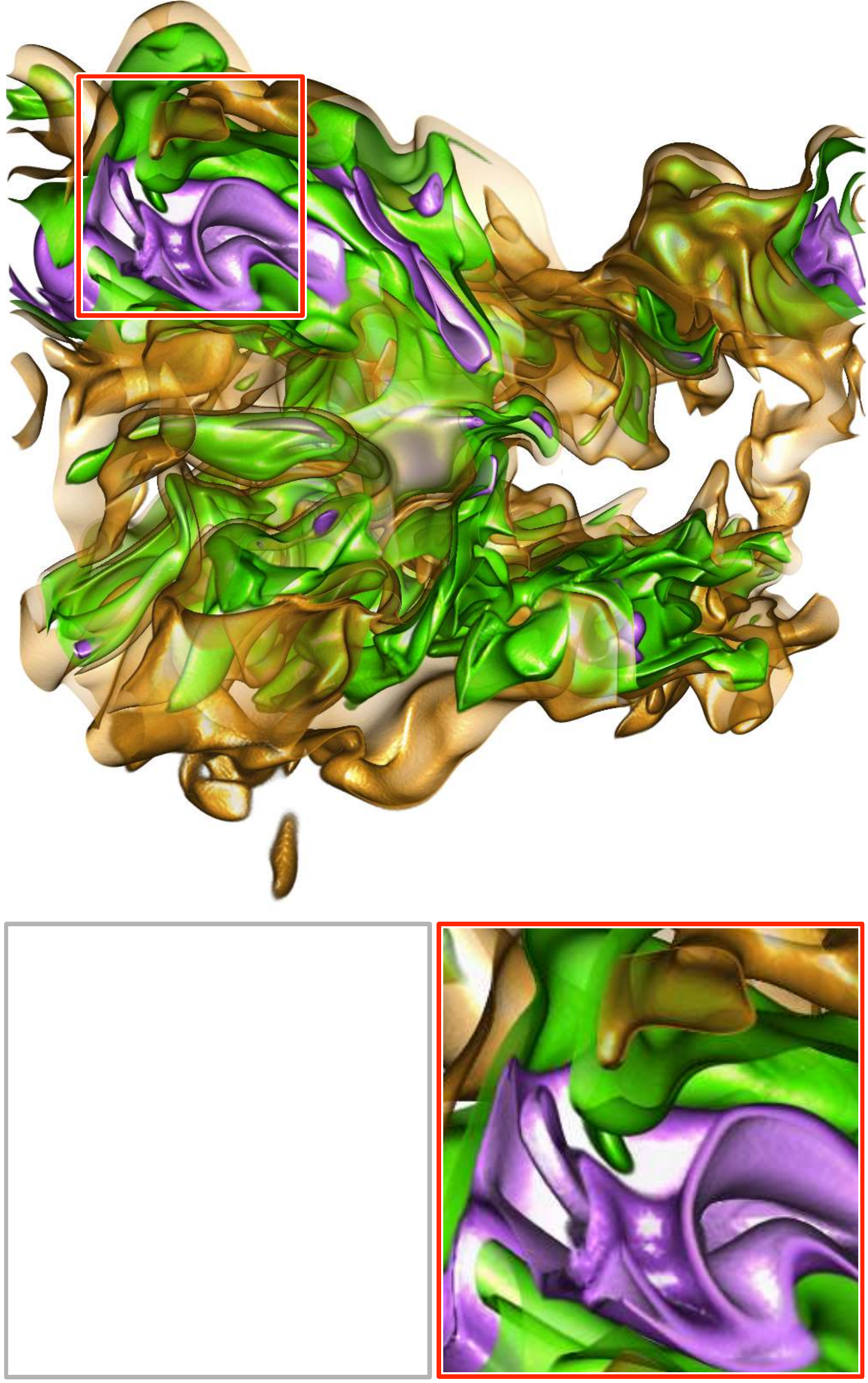}\\
 \includegraphics[width=0.1925\linewidth]{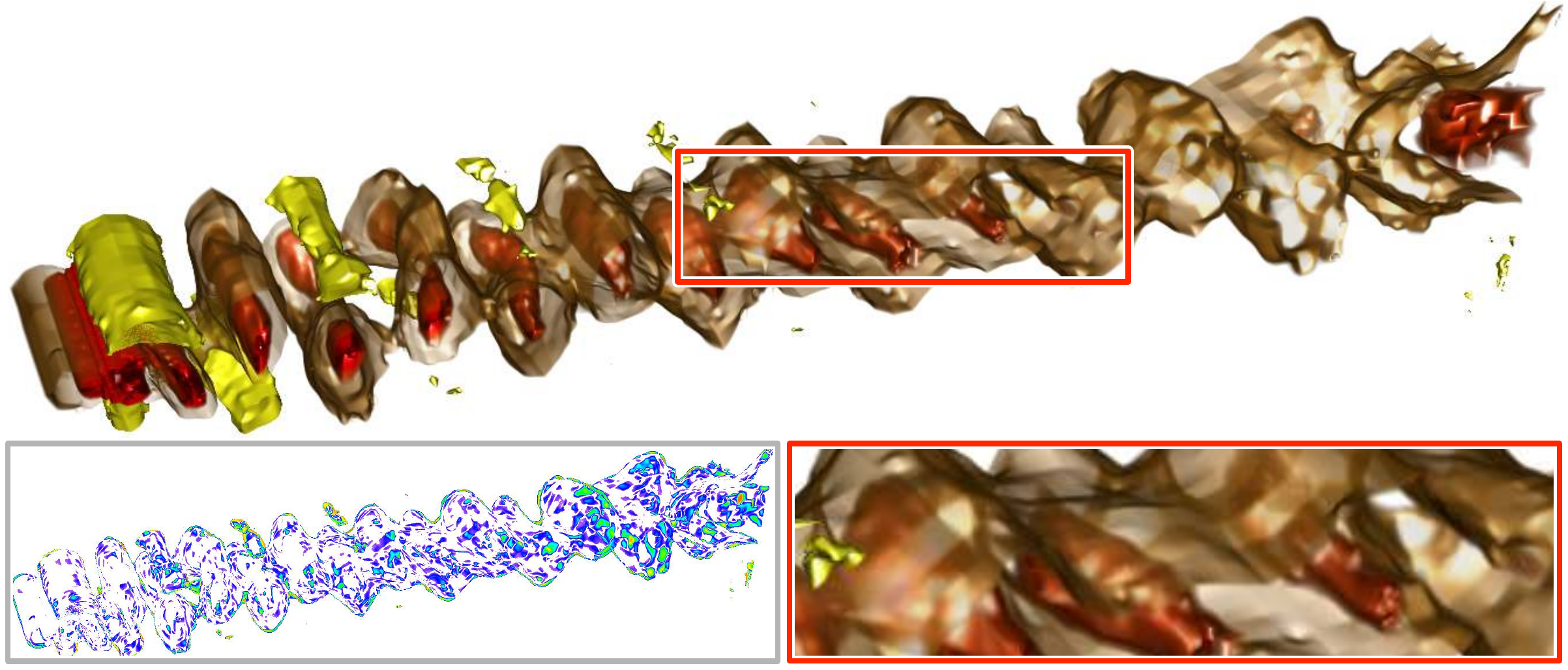}&
 \includegraphics[width=0.1925\linewidth]{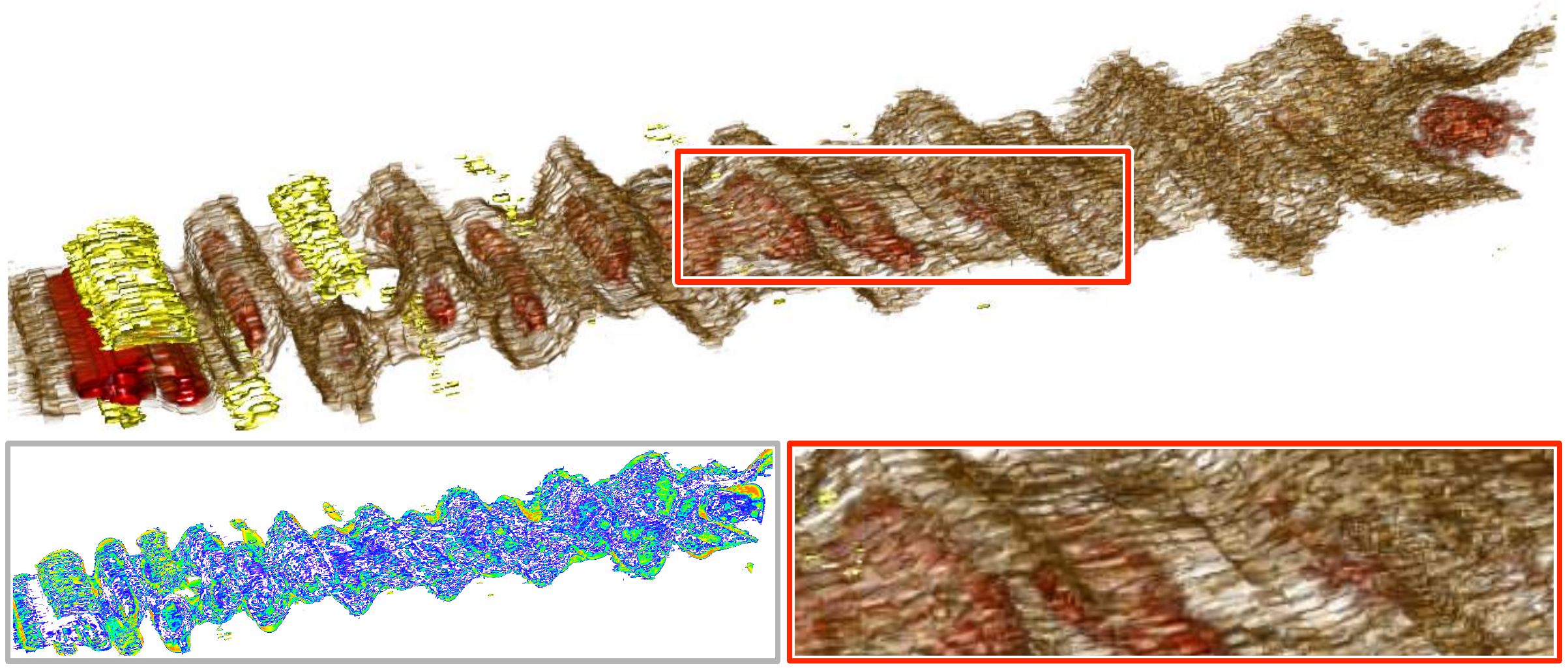}&
 \includegraphics[width=0.1925\linewidth]{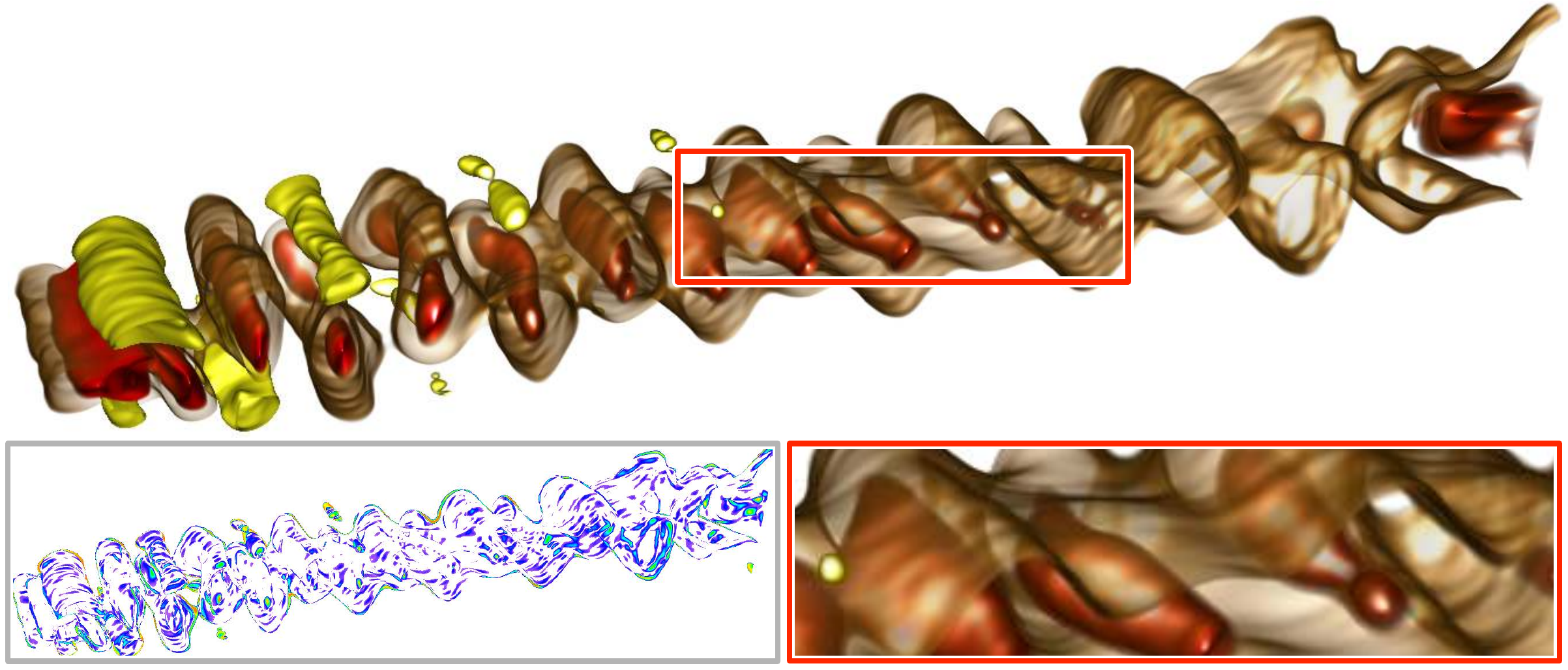}&
  \includegraphics[width=0.1925\linewidth]{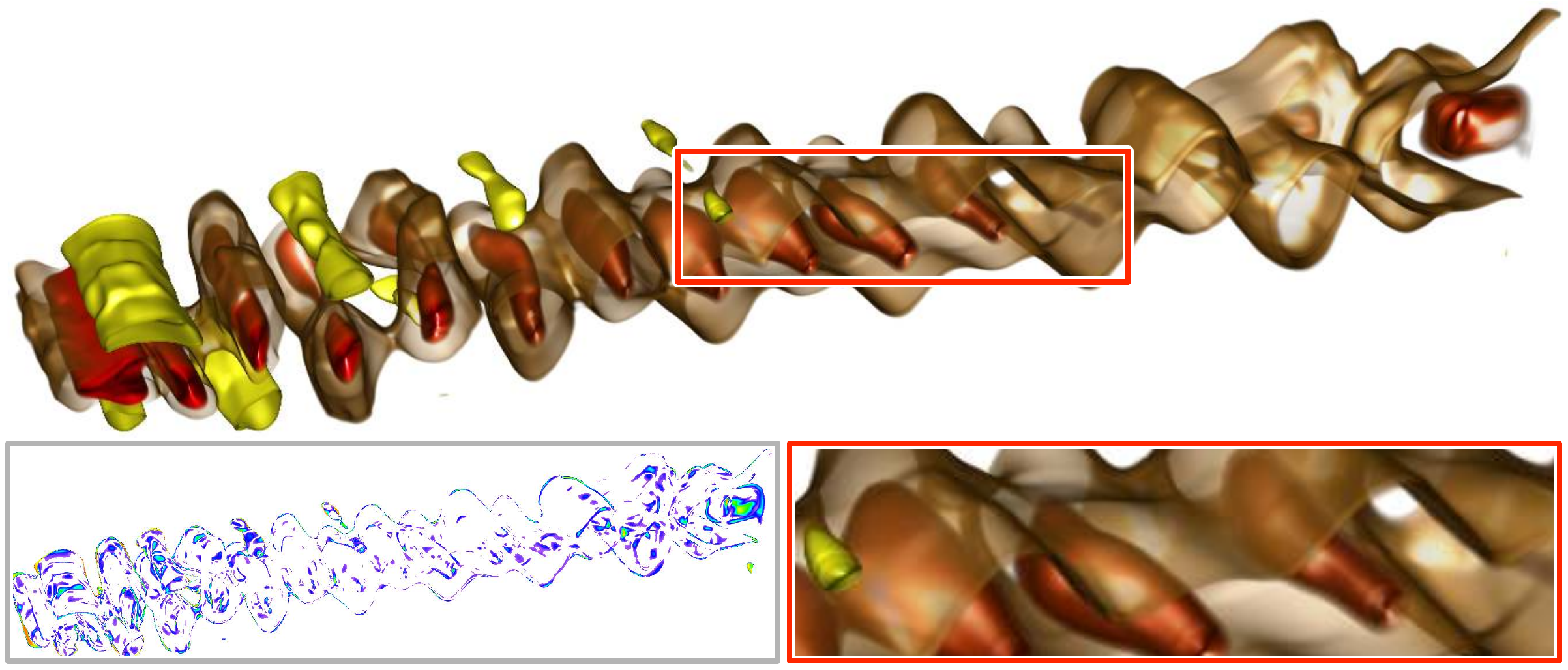}&
\includegraphics[width=0.1925\linewidth]{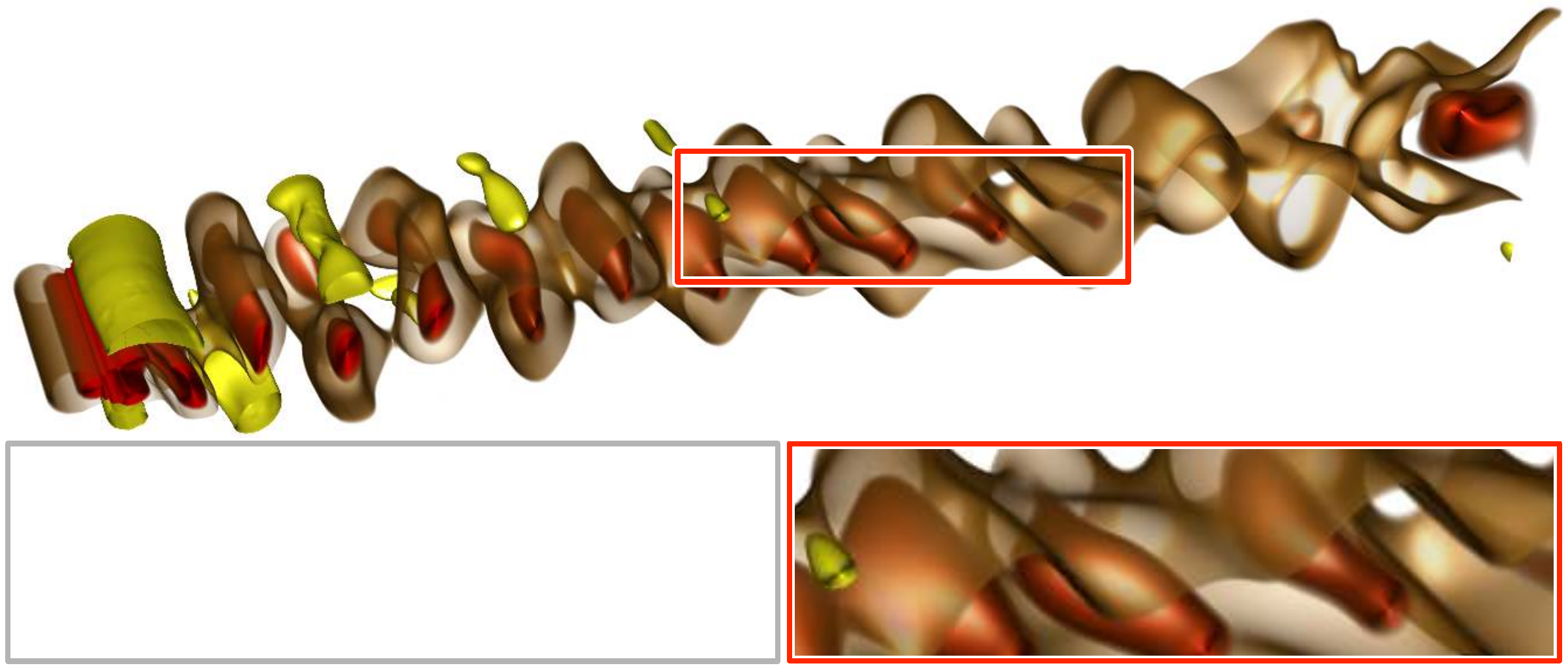}\\
 \includegraphics[width=0.1925\linewidth]{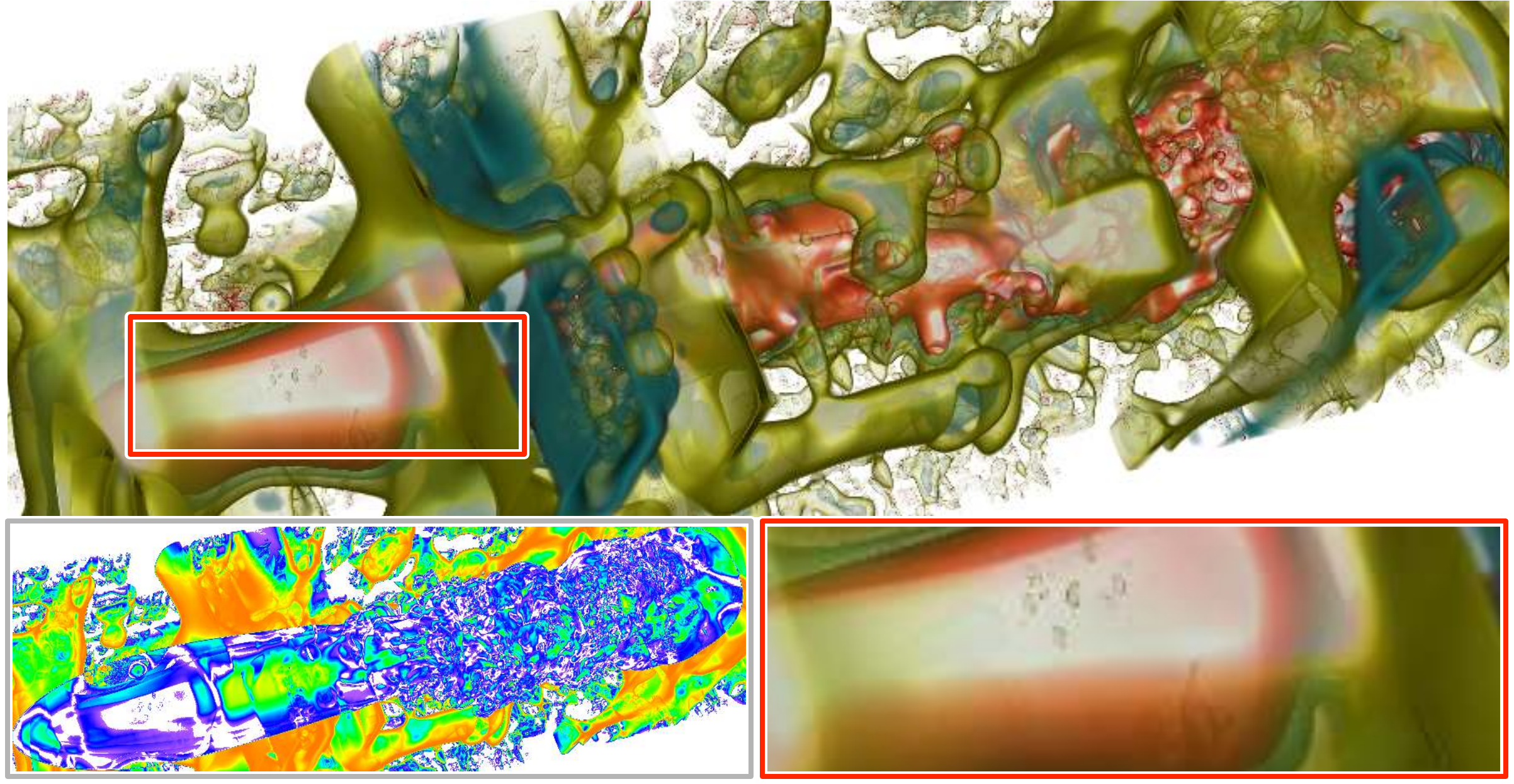}&
 \includegraphics[width=0.1925\linewidth]{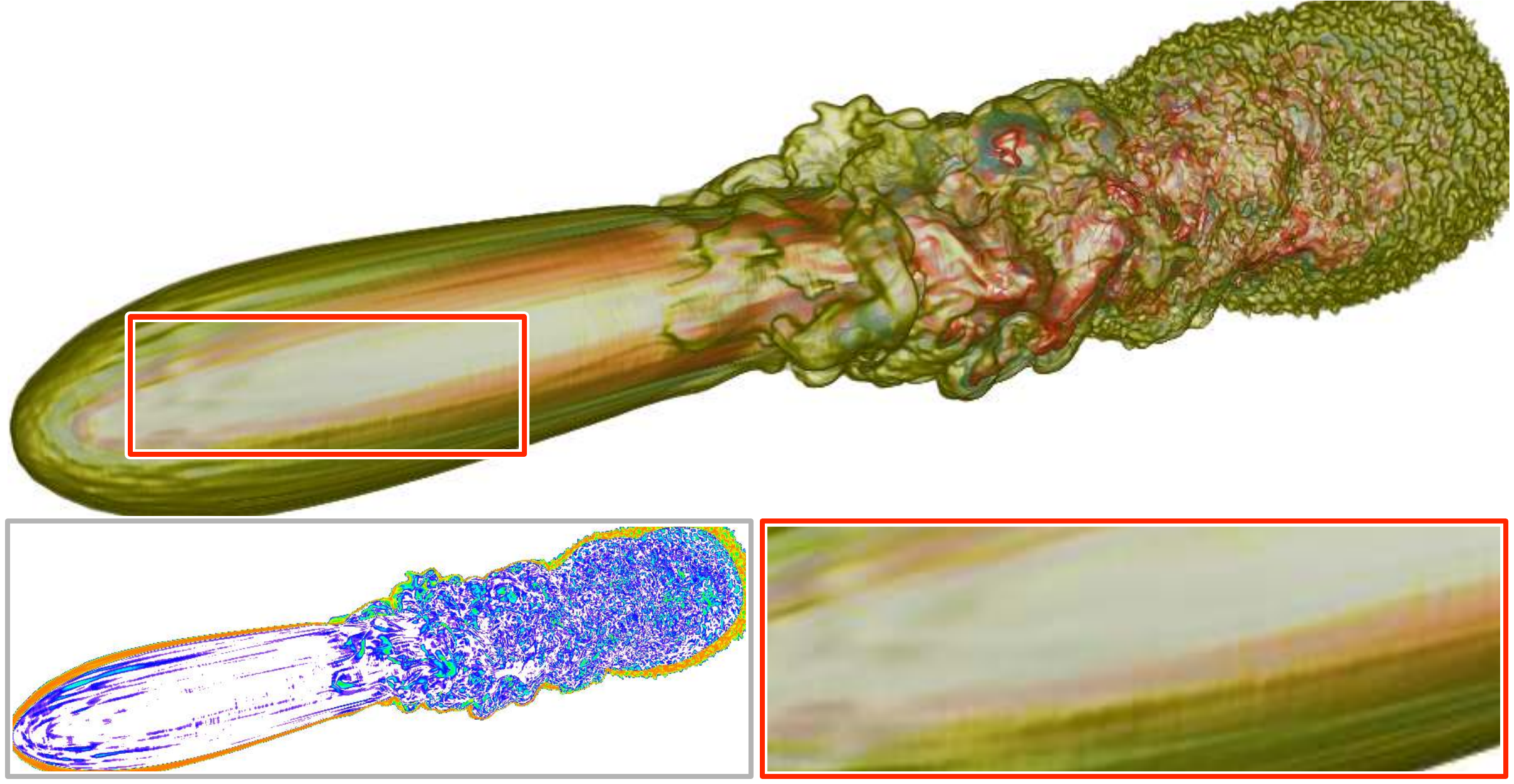}&
 \includegraphics[width=0.1925\linewidth]{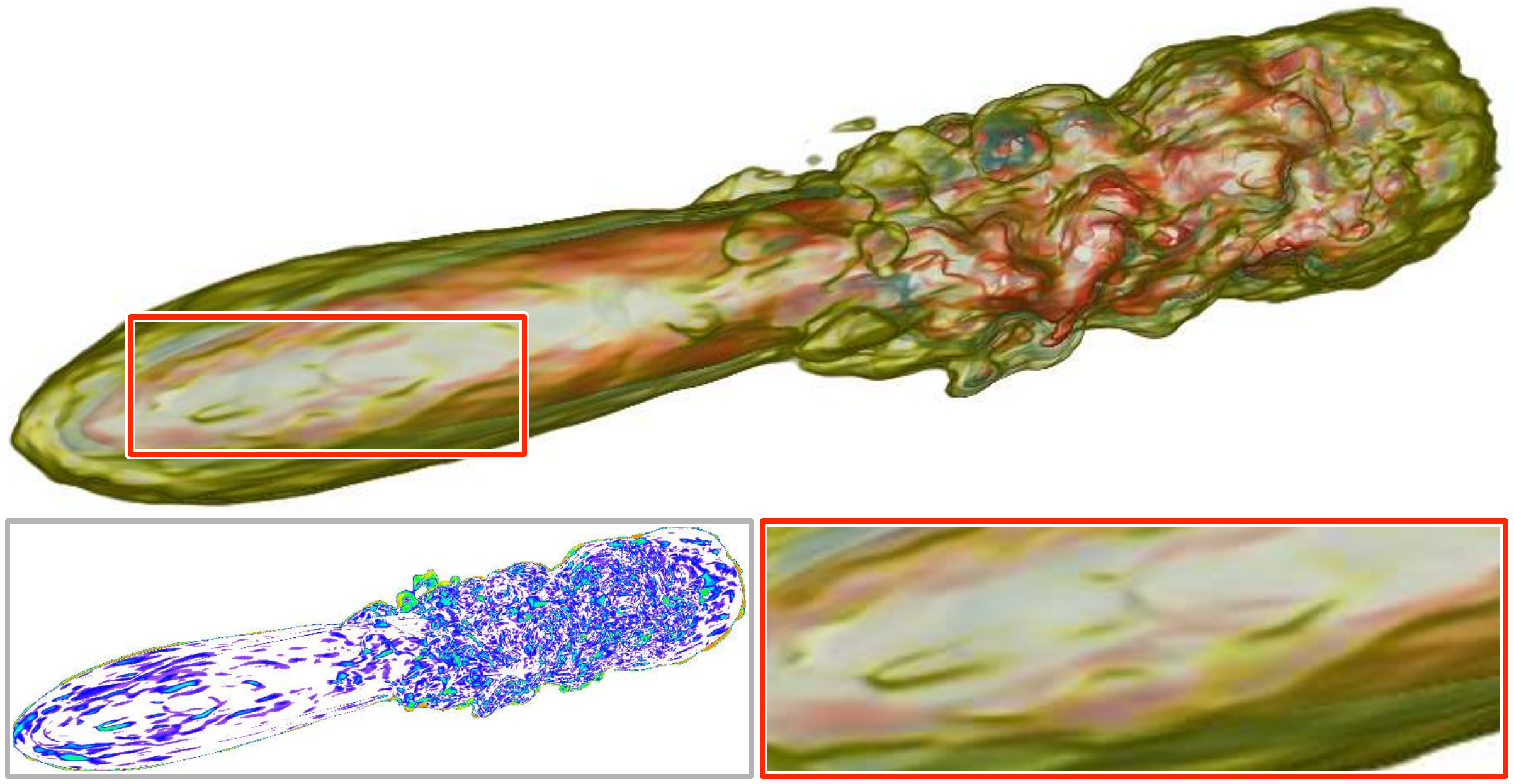}&
  \includegraphics[width=0.1925\linewidth]{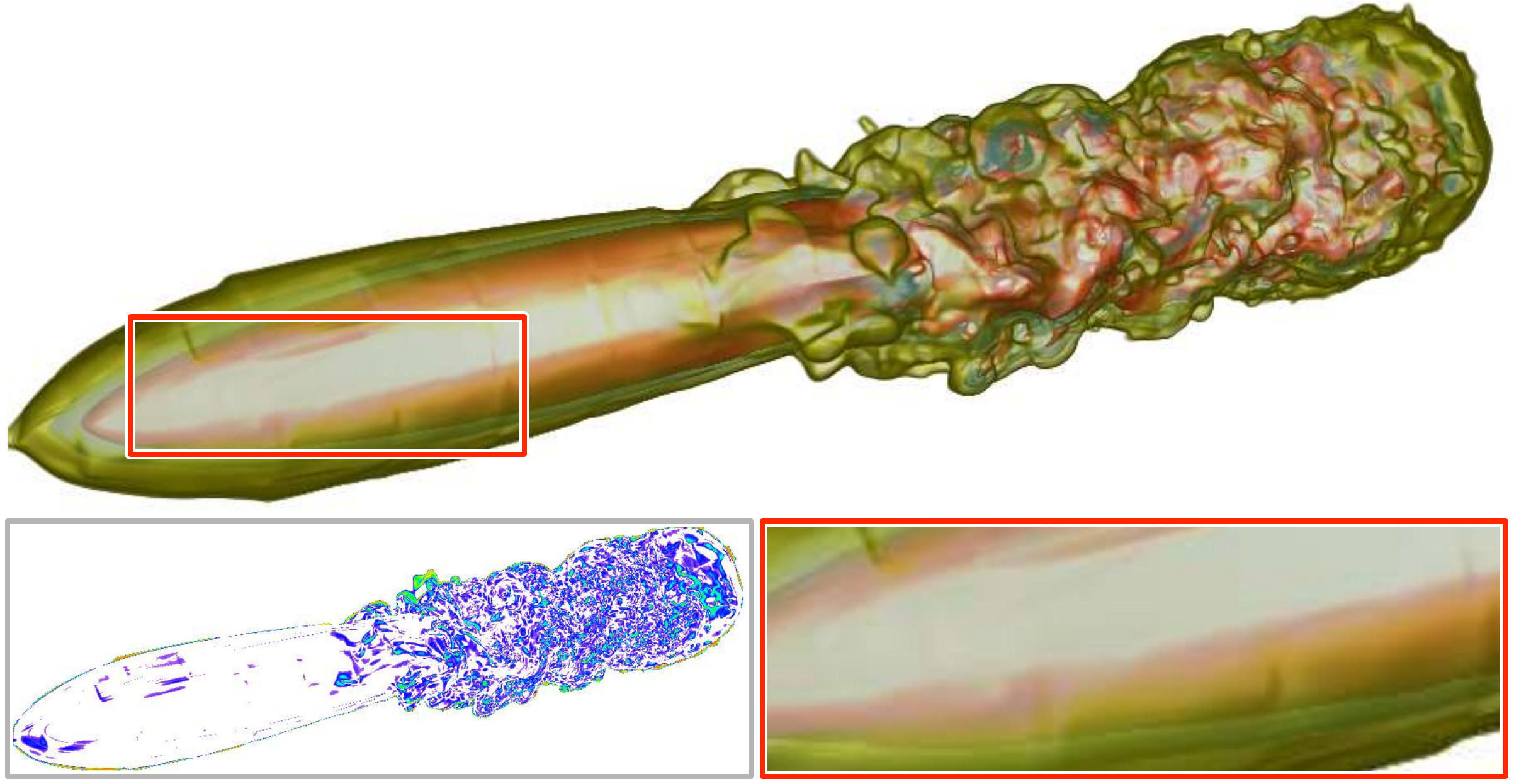}&
\includegraphics[width=0.1925\linewidth]{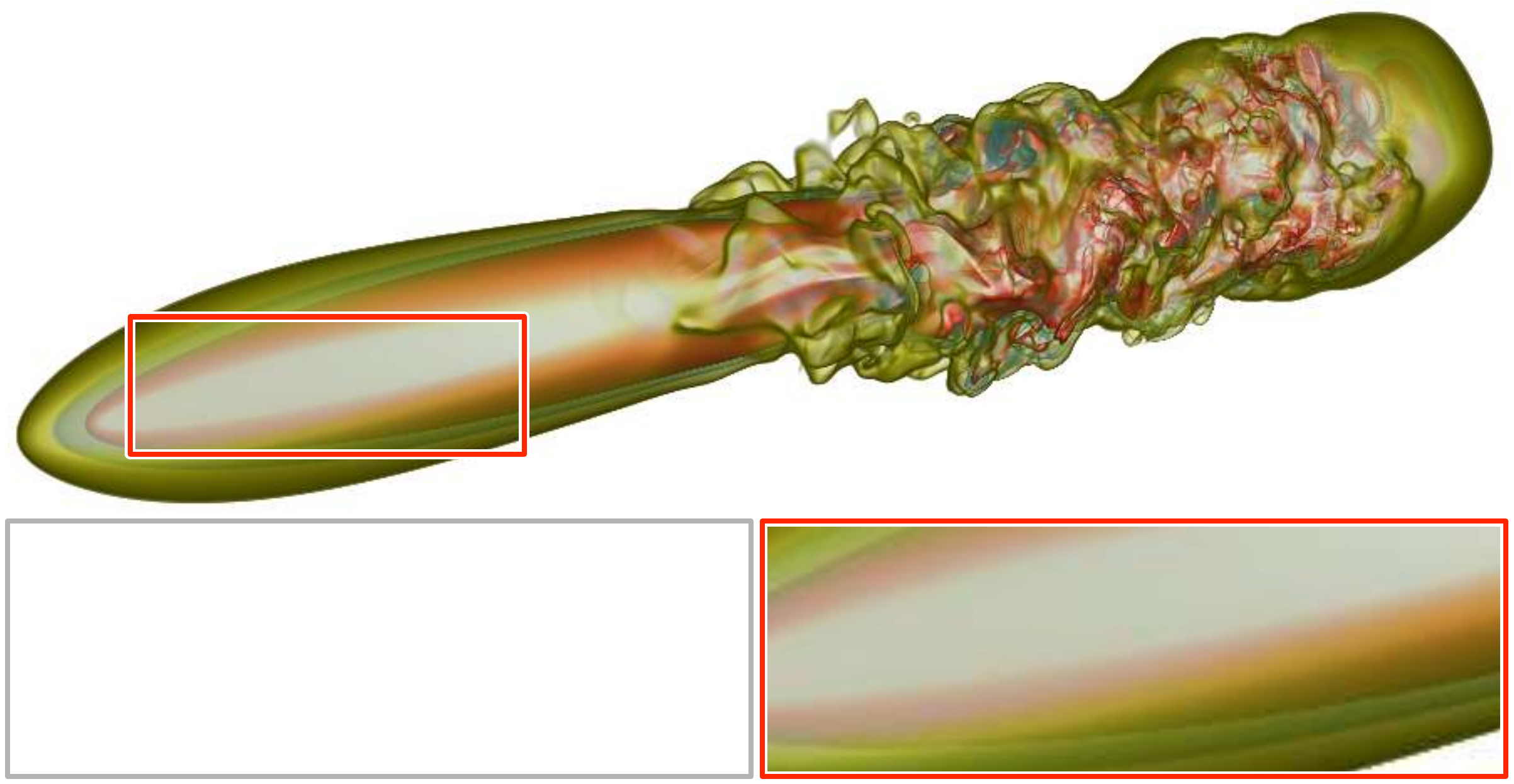}\\
 \includegraphics[width=0.1925\linewidth]{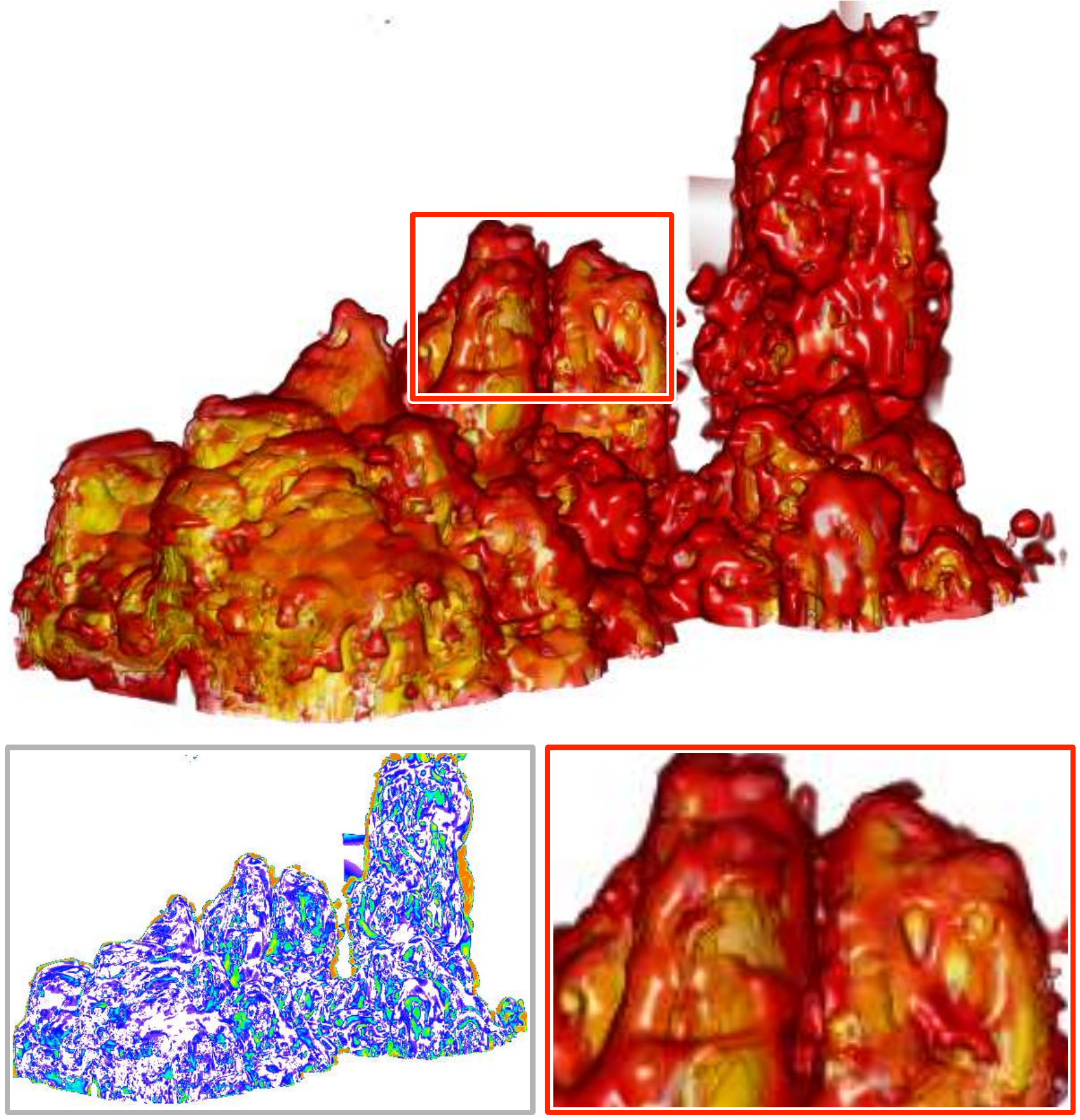}&
 \includegraphics[width=0.1925\linewidth]{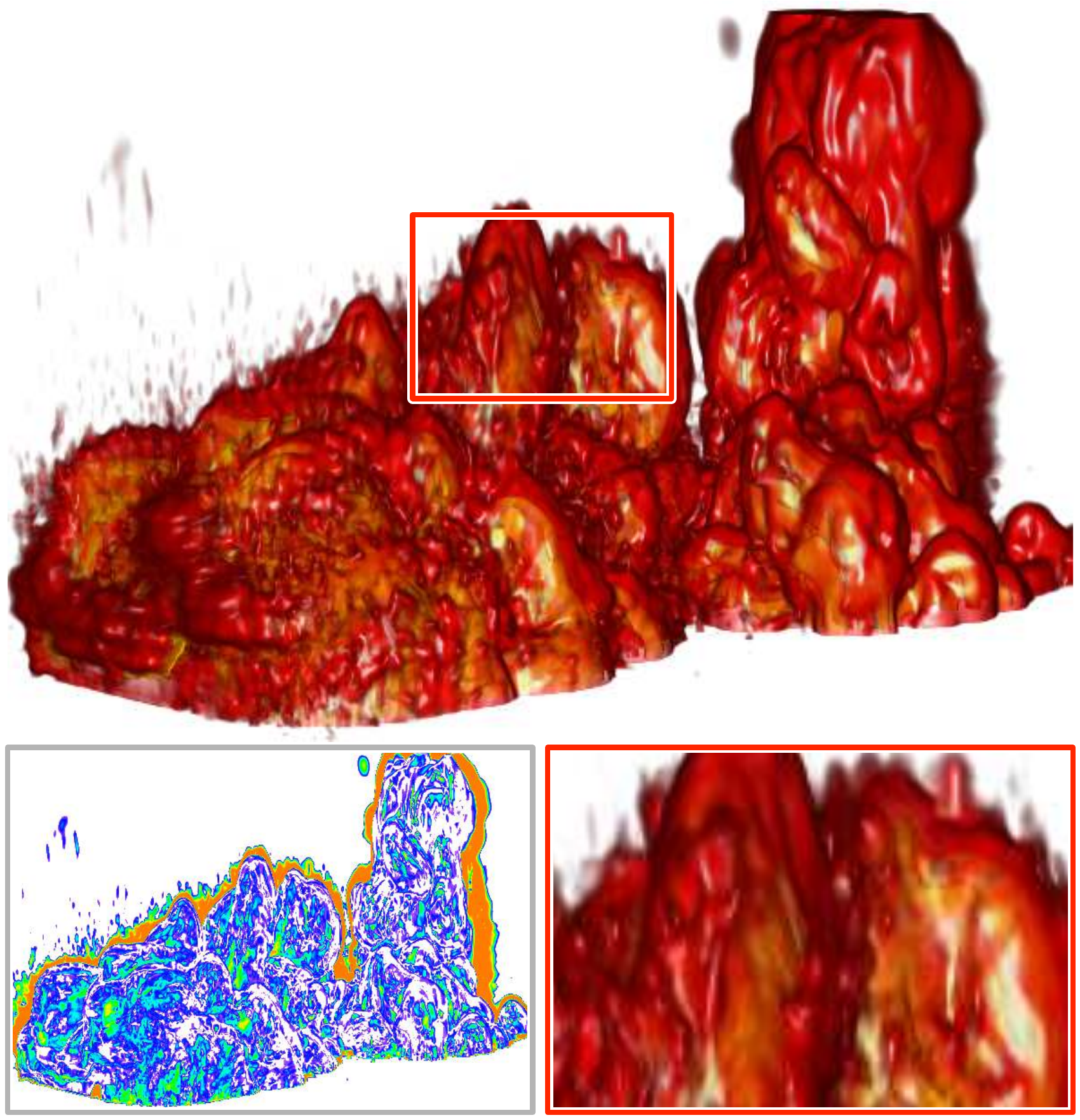}&
 \includegraphics[width=0.1925\linewidth]{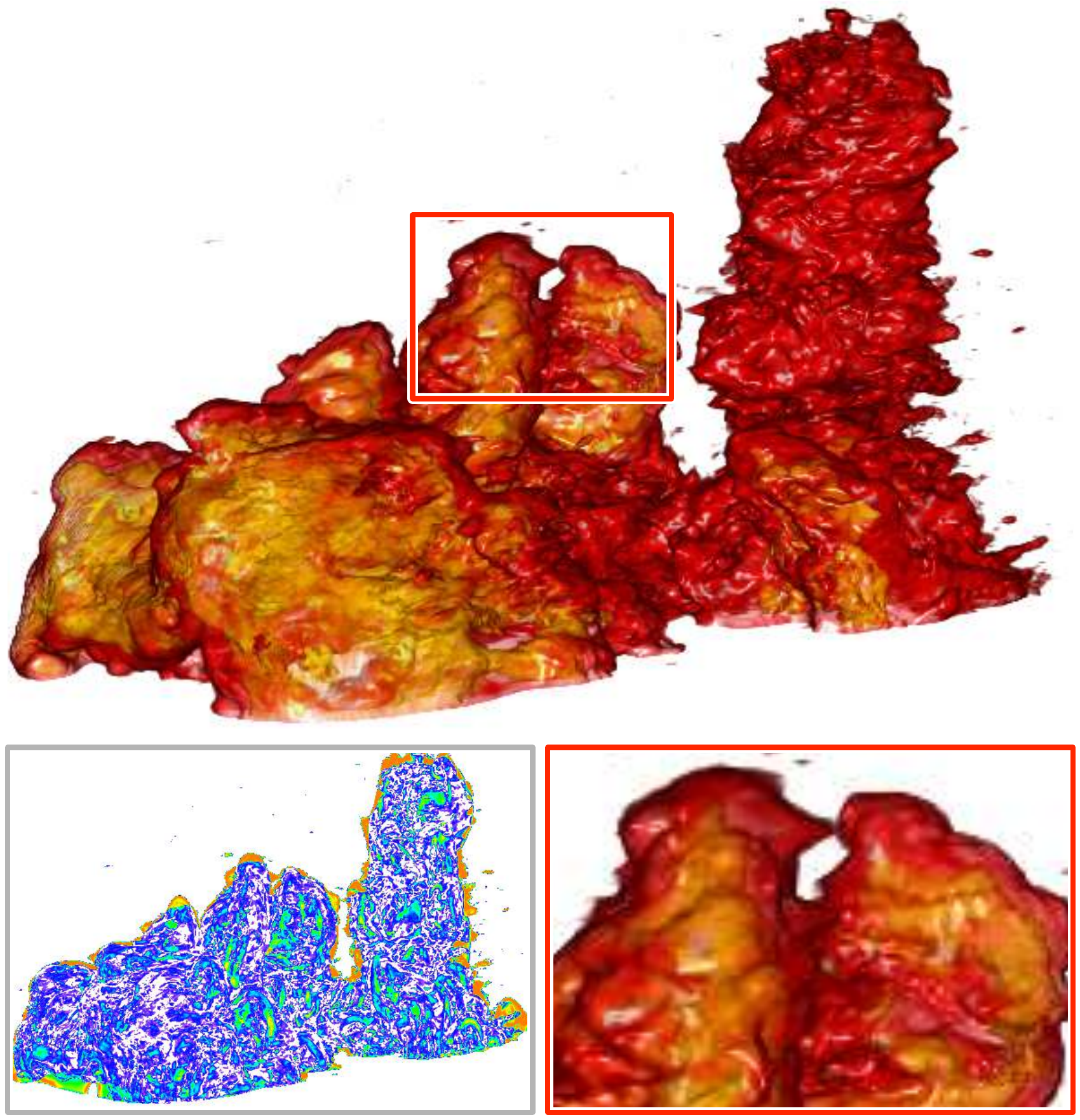}&
  \includegraphics[width=0.1925\linewidth]{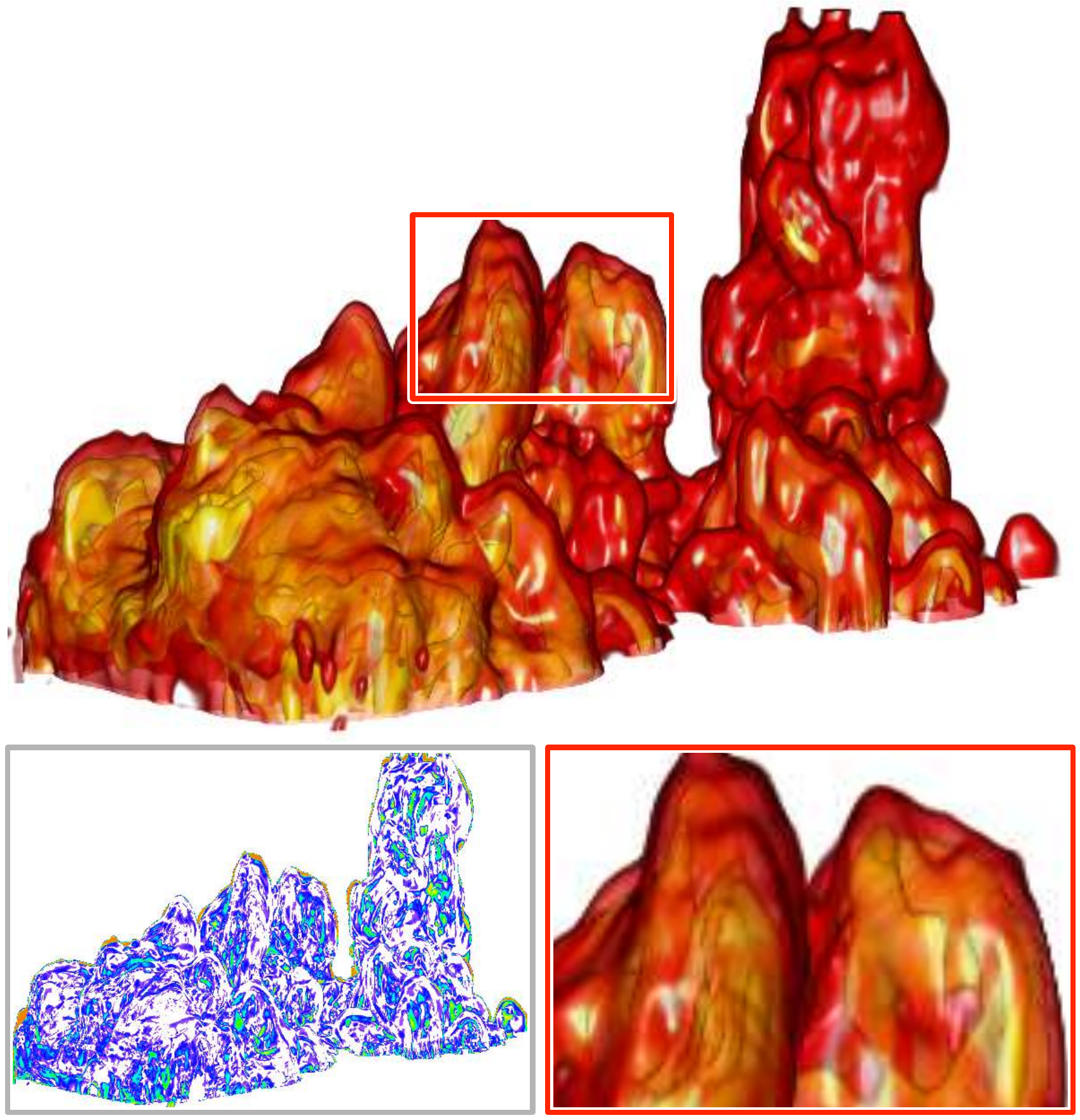}&
\includegraphics[width=0.1925\linewidth]{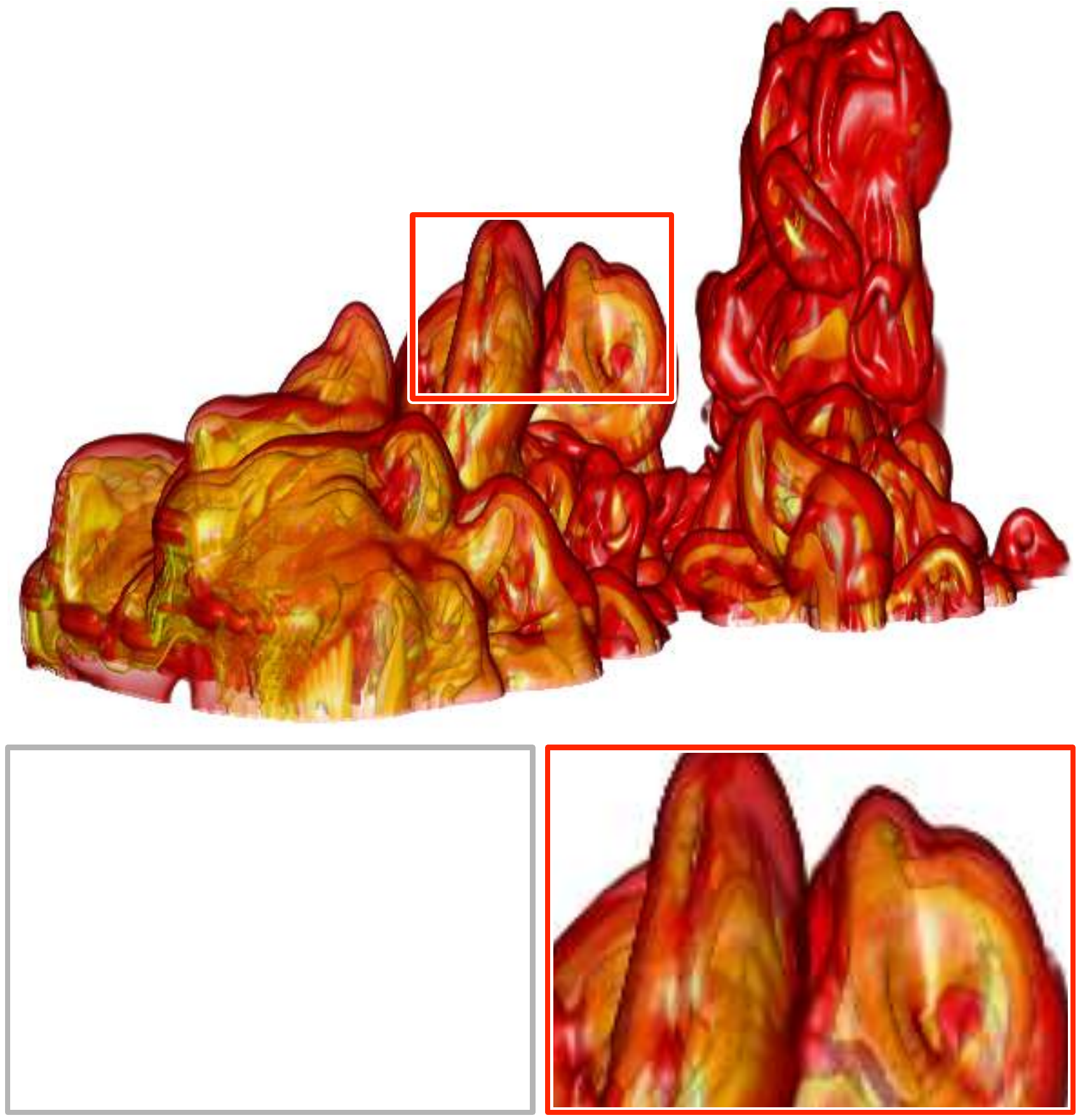}\\
\mbox{(a) SZ3} & \mbox{(b) TTHRESH} & \mbox{(c) neurcomp} &\mbox{(d) ECNR} &\mbox{(e) GT} 
\end{array}$
\end{center}
\vspace{-.25in} 
\caption{Volume rendering results of SZ3, TTHRESH, neurcomp, ECNR, and GT. From top to bottom: combustion, half-cylinder, solar plume, and Tangaroa.} 
\label{fig:vr-results}
\end{figure*}

\begin{figure*}[htb]
 \begin{center}
 $\begin{array}{c@{\hspace{0.025in}}c@{\hspace{0.025in}}c@{\hspace{0.025in}}c@{\hspace{0.025in}}c}
 \includegraphics[width=0.1925\linewidth]{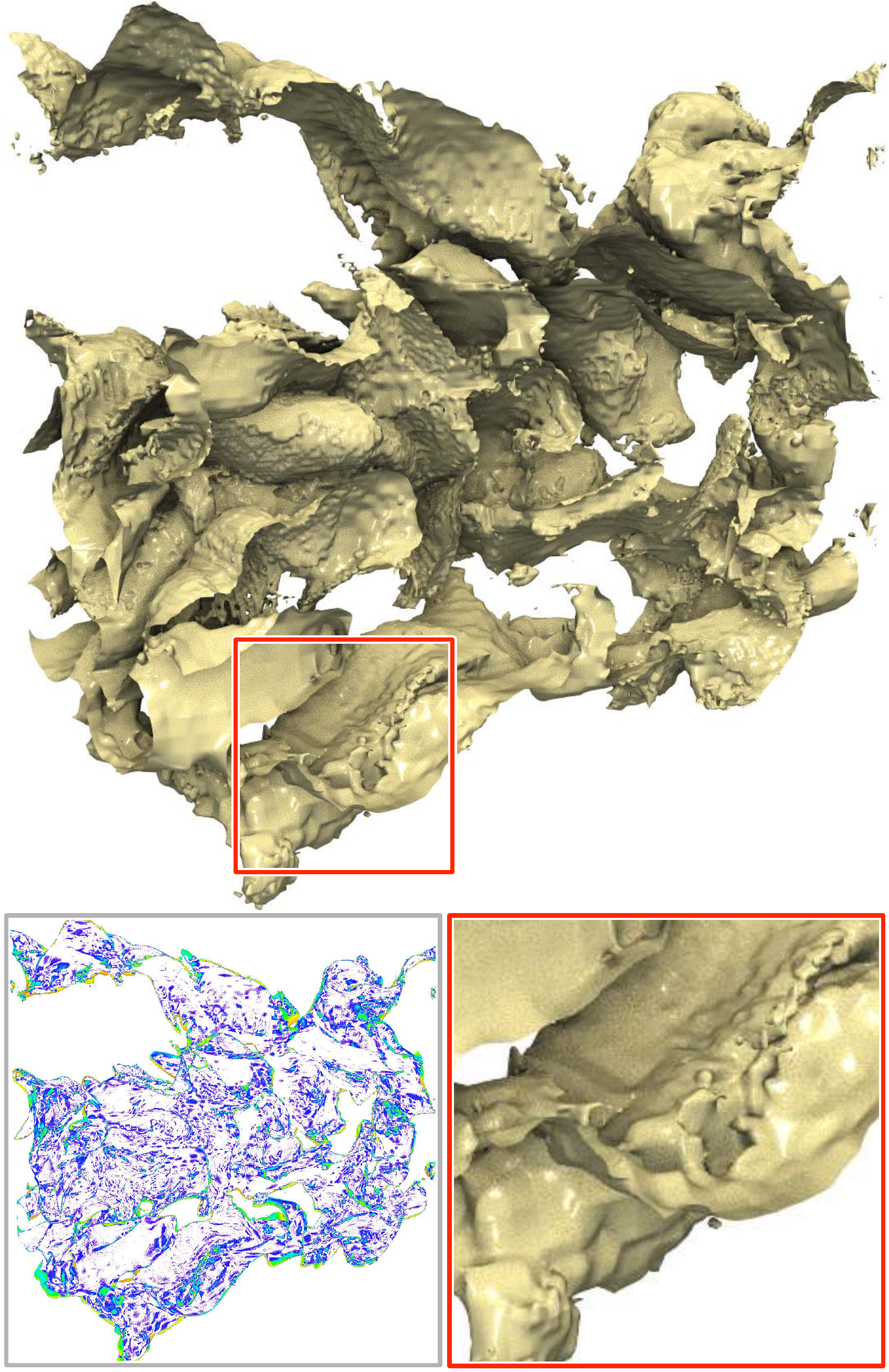}&
 \includegraphics[width=0.1925\linewidth]{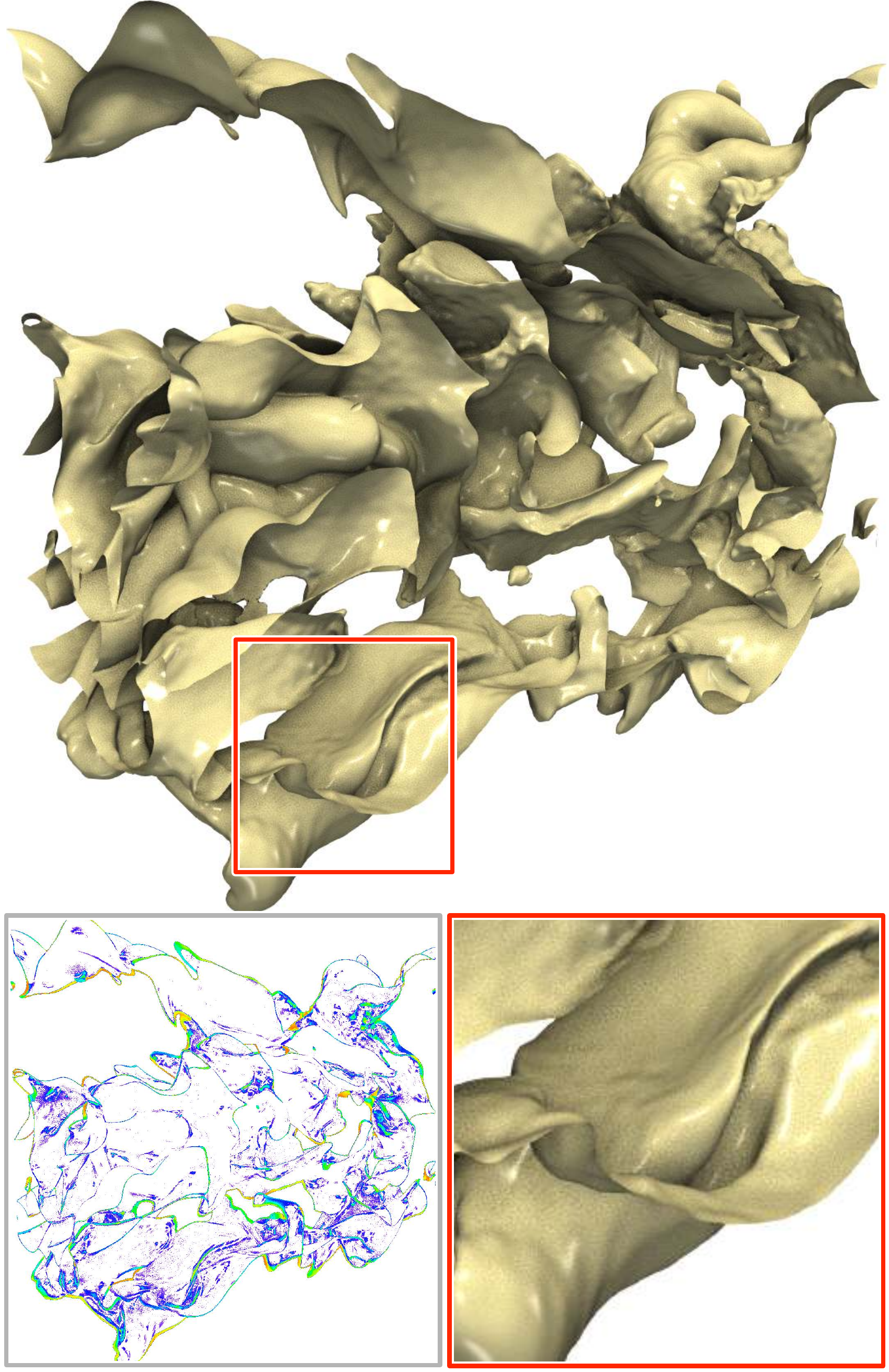}&
 \includegraphics[width=0.1925\linewidth]{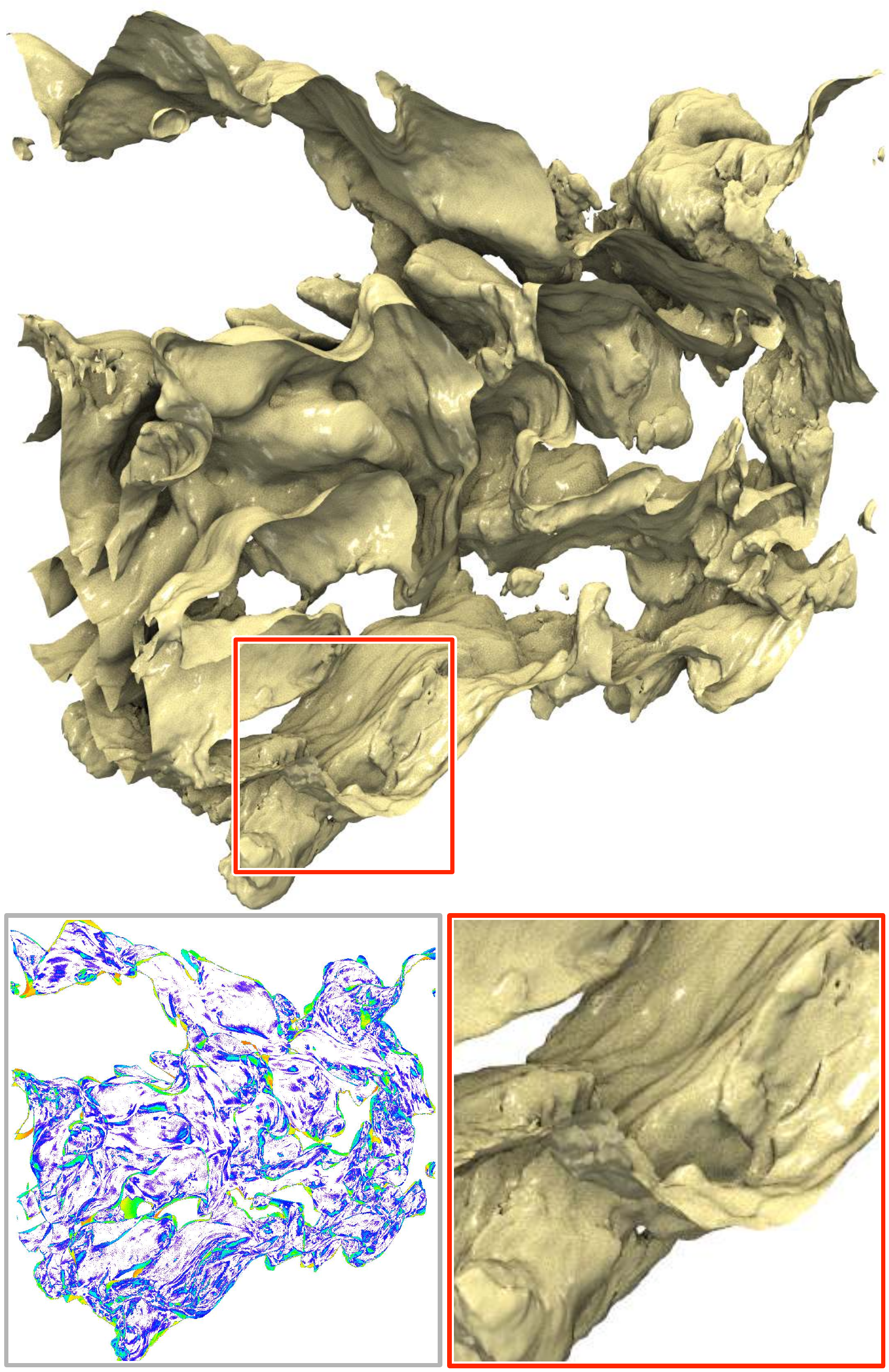}&
  \includegraphics[width=0.1925\linewidth]{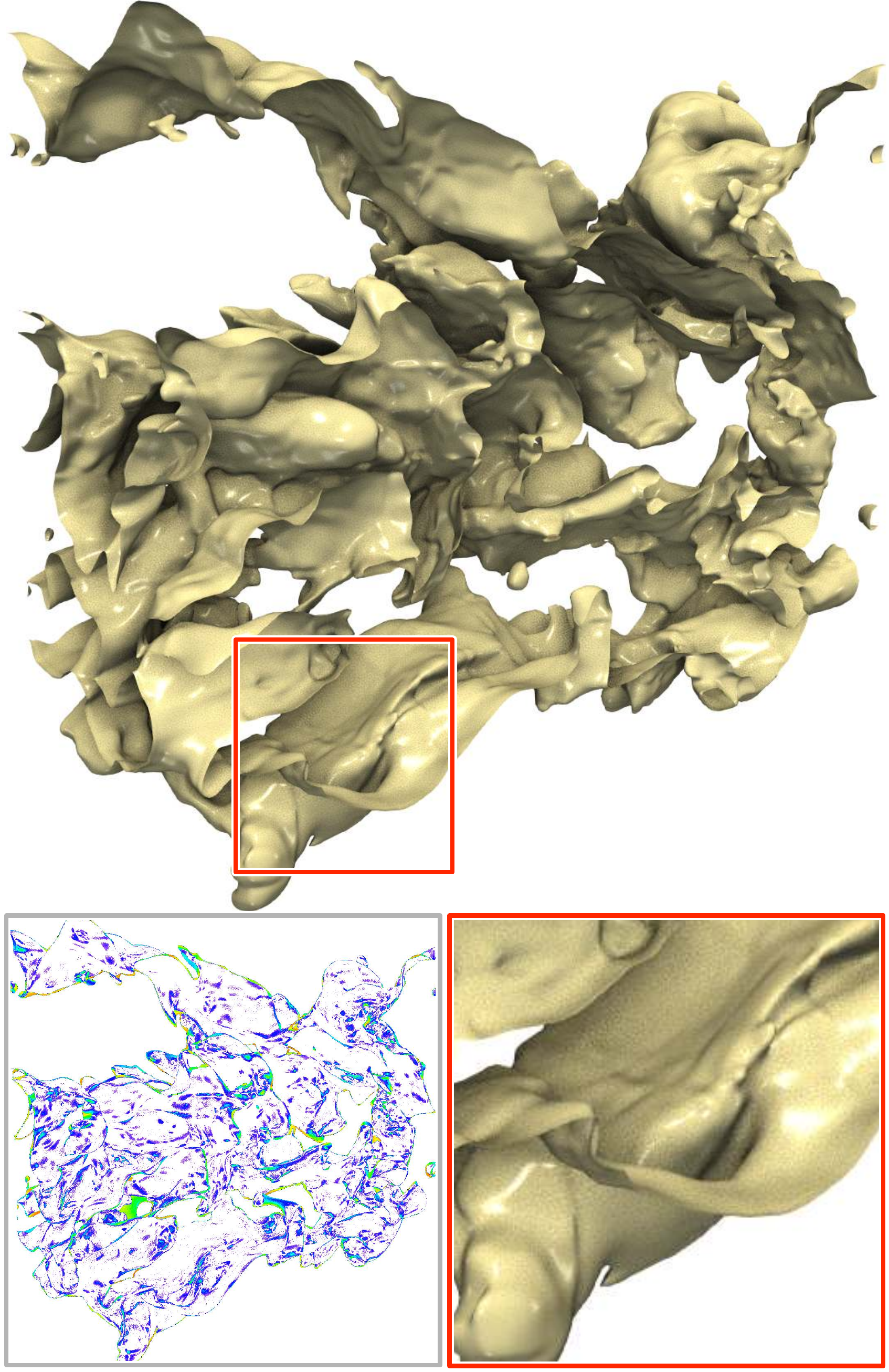}&
\includegraphics[width=0.1925\linewidth]{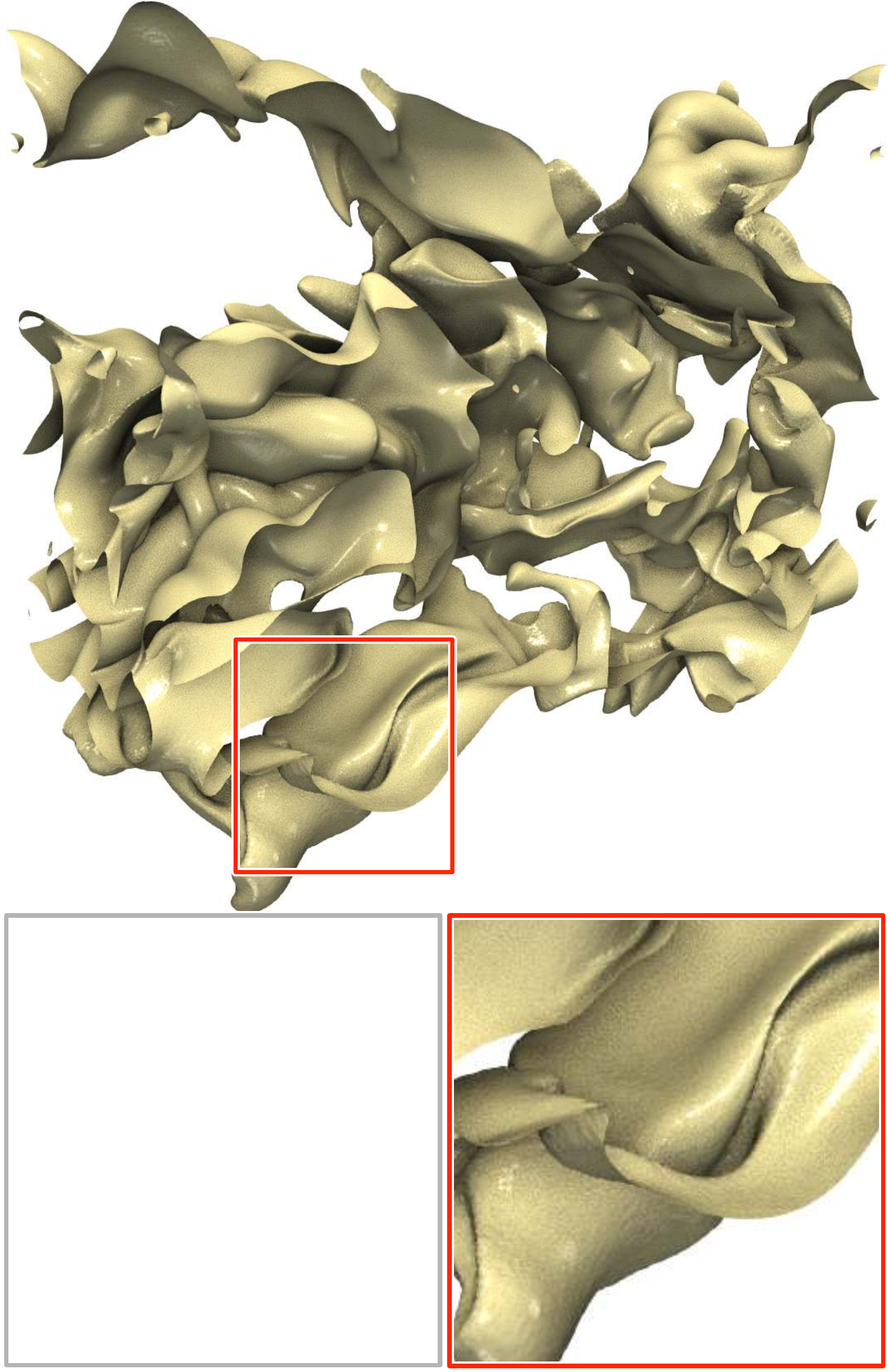}\\
 \includegraphics[width=0.1925\linewidth]{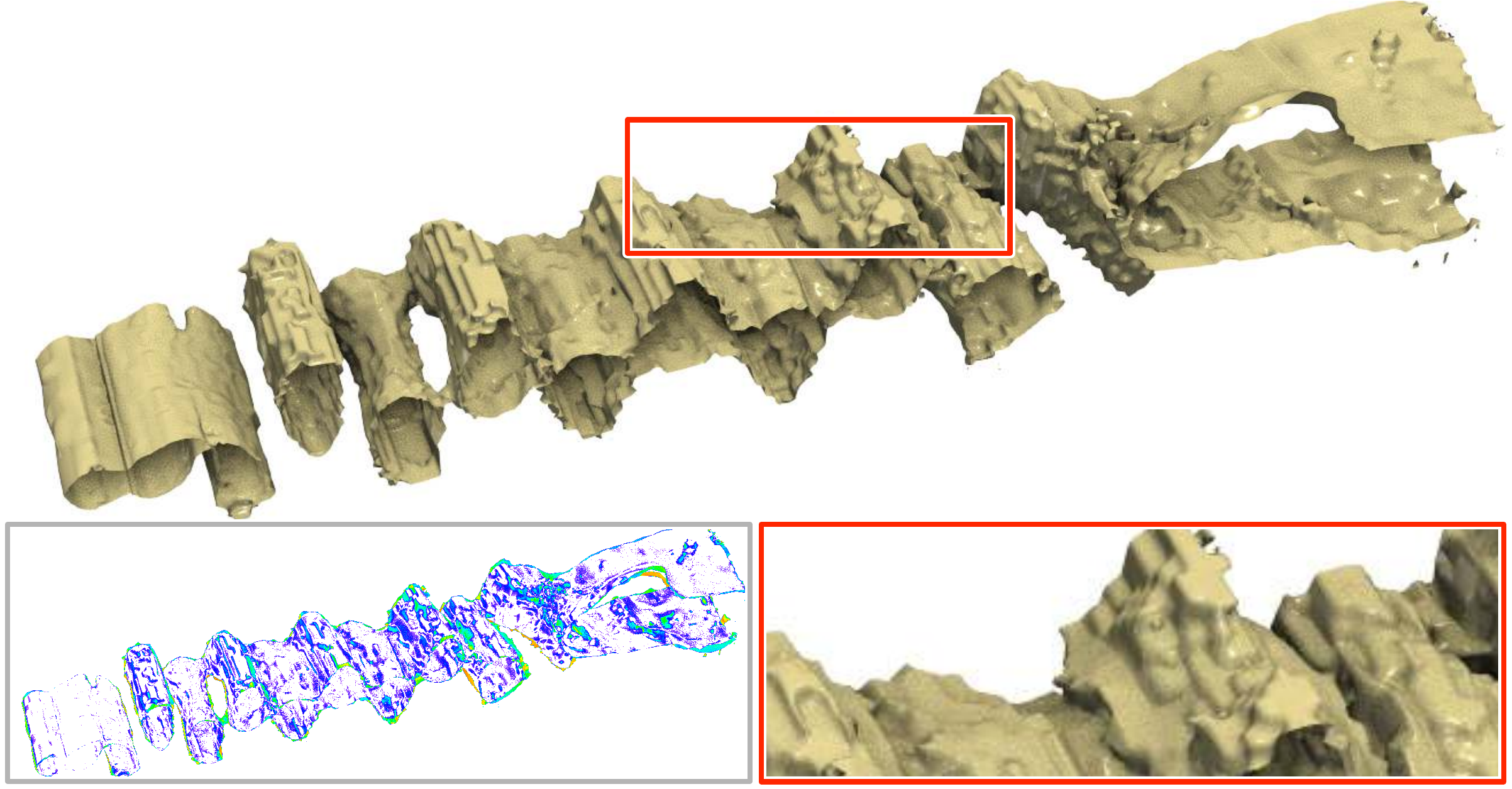}&
 \includegraphics[width=0.1925\linewidth]{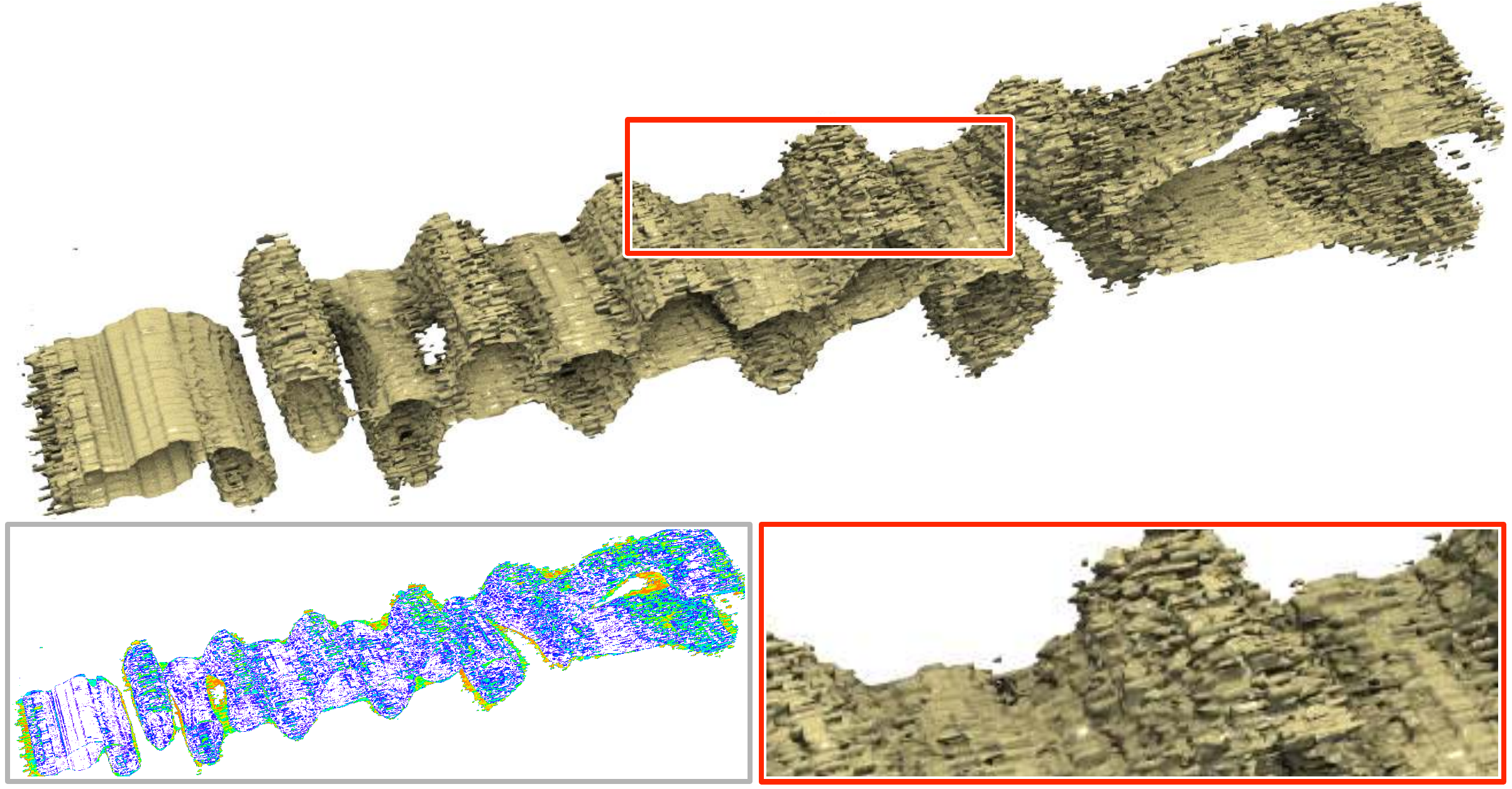}&
 \includegraphics[width=0.1925\linewidth]{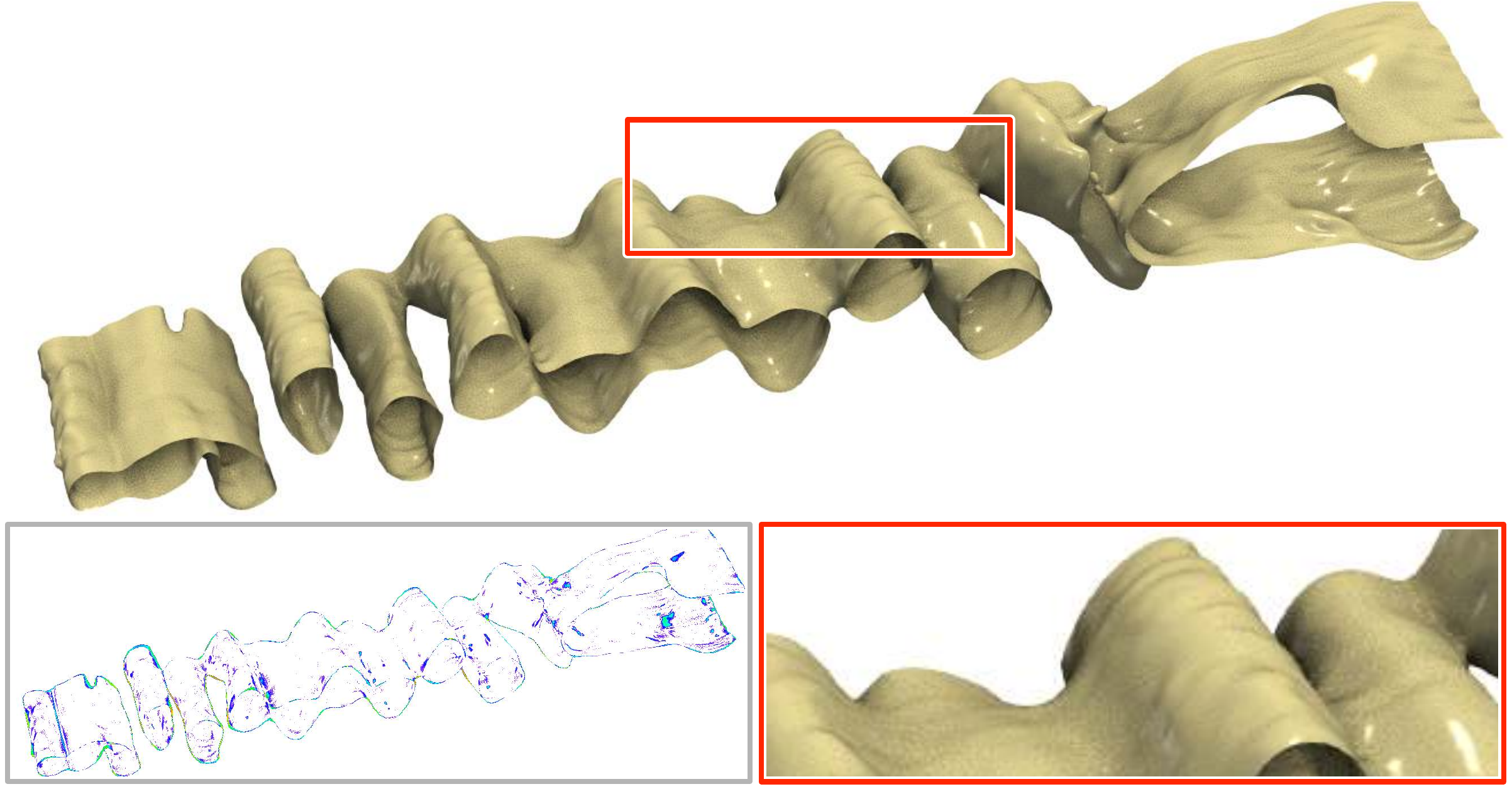}&
  \includegraphics[width=0.1925\linewidth]{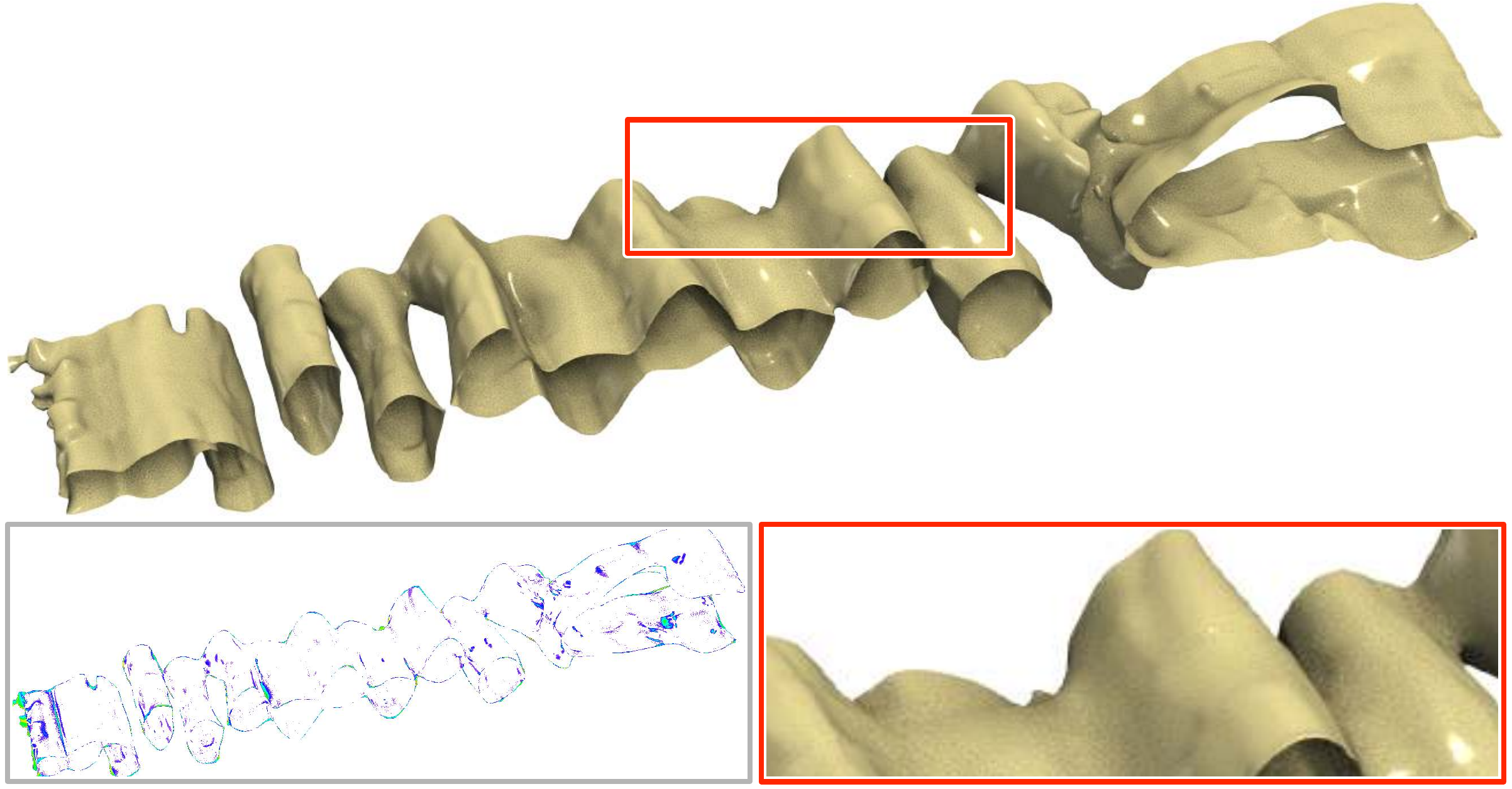}&
\includegraphics[width=0.1925\linewidth]{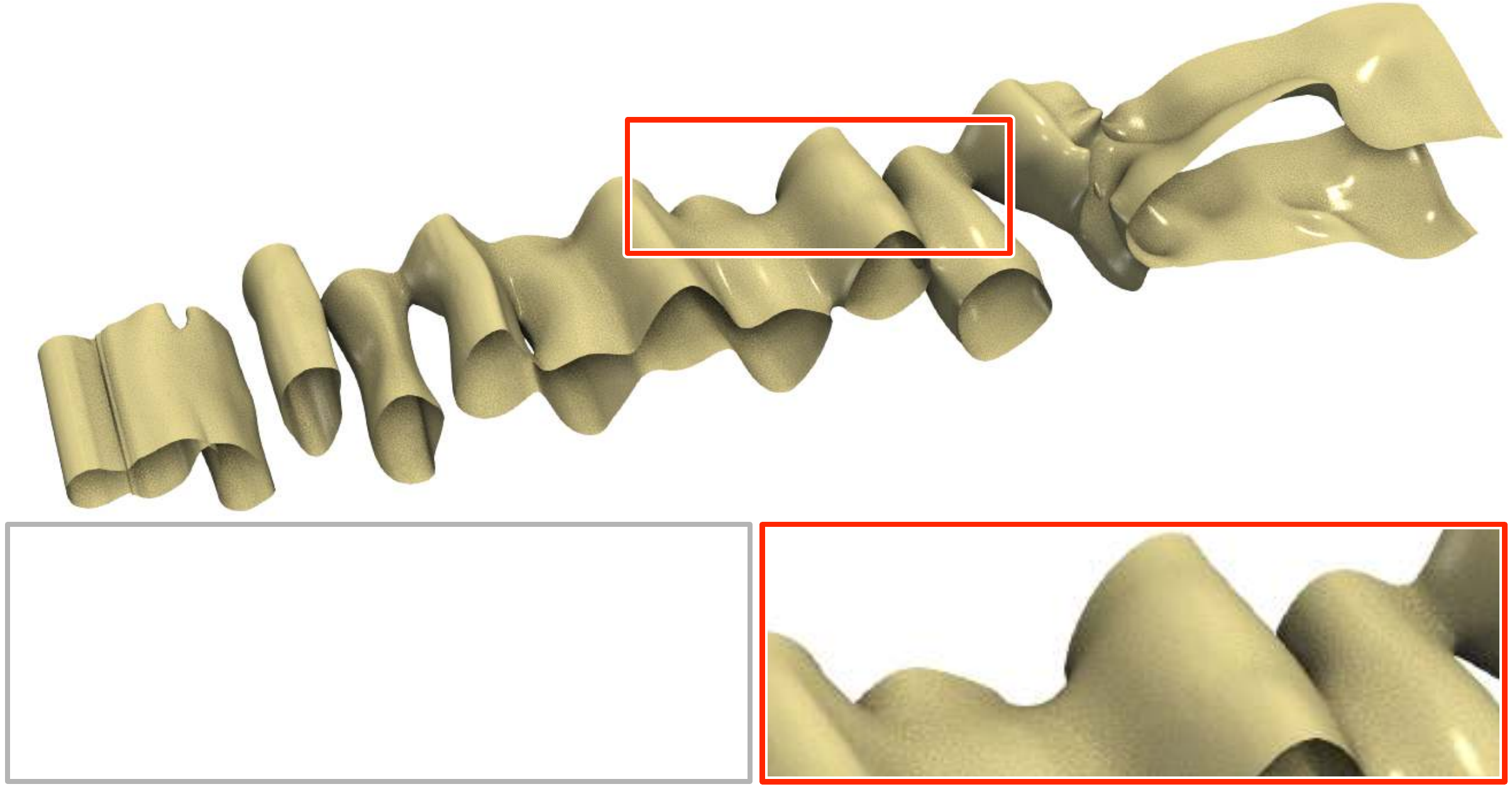}\\
 \includegraphics[width=0.1925\linewidth]{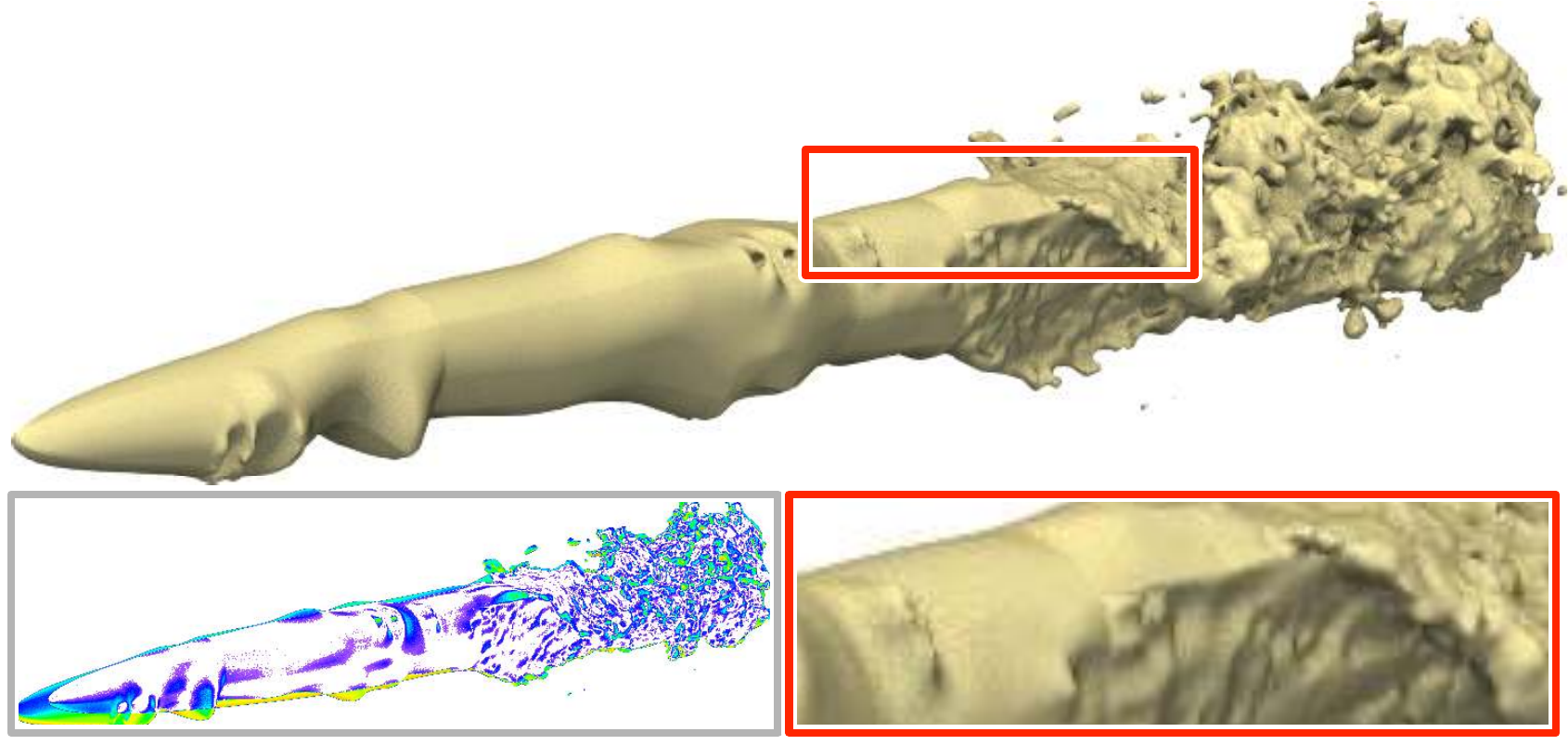}&
 \includegraphics[width=0.1925\linewidth]{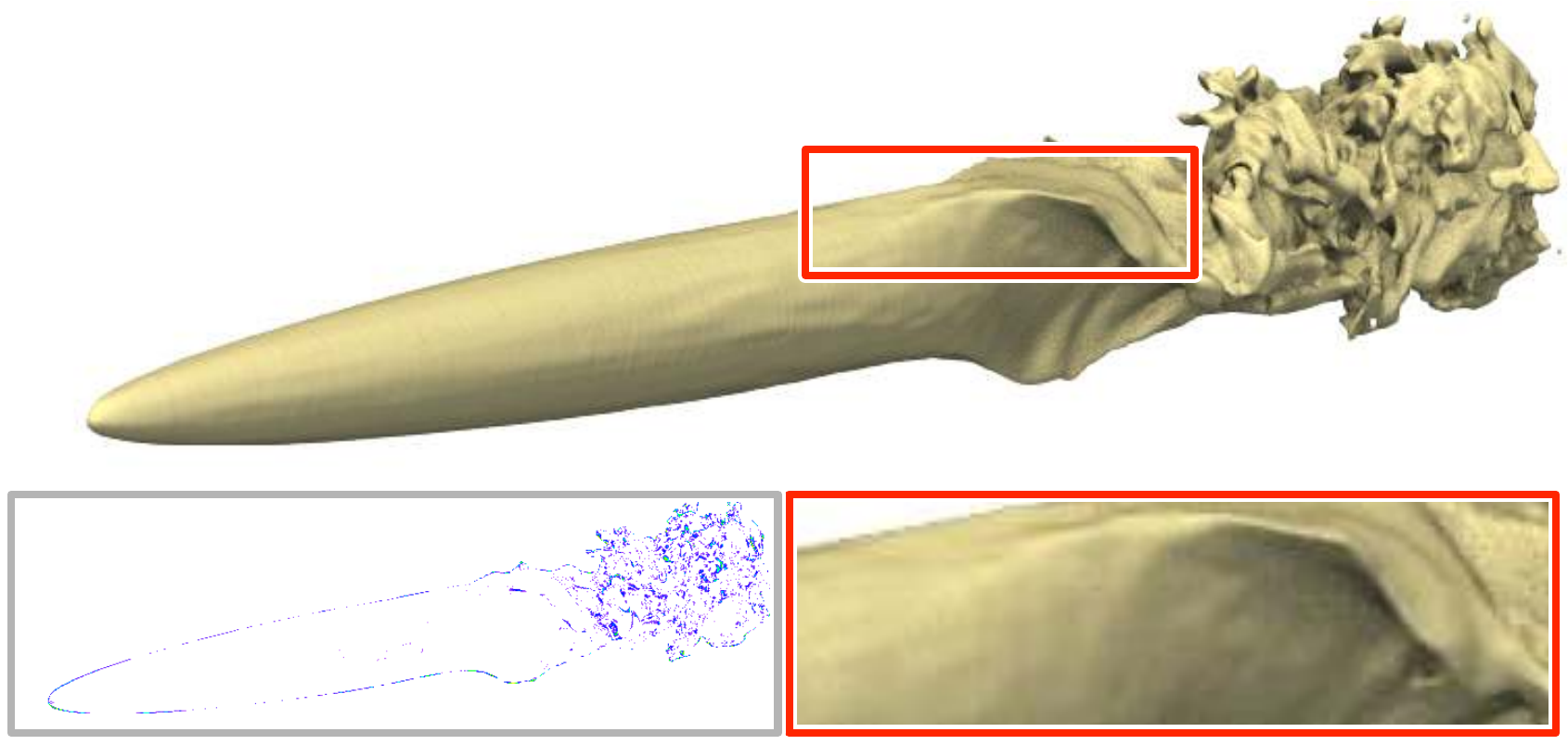}&
 \includegraphics[width=0.1925\linewidth]{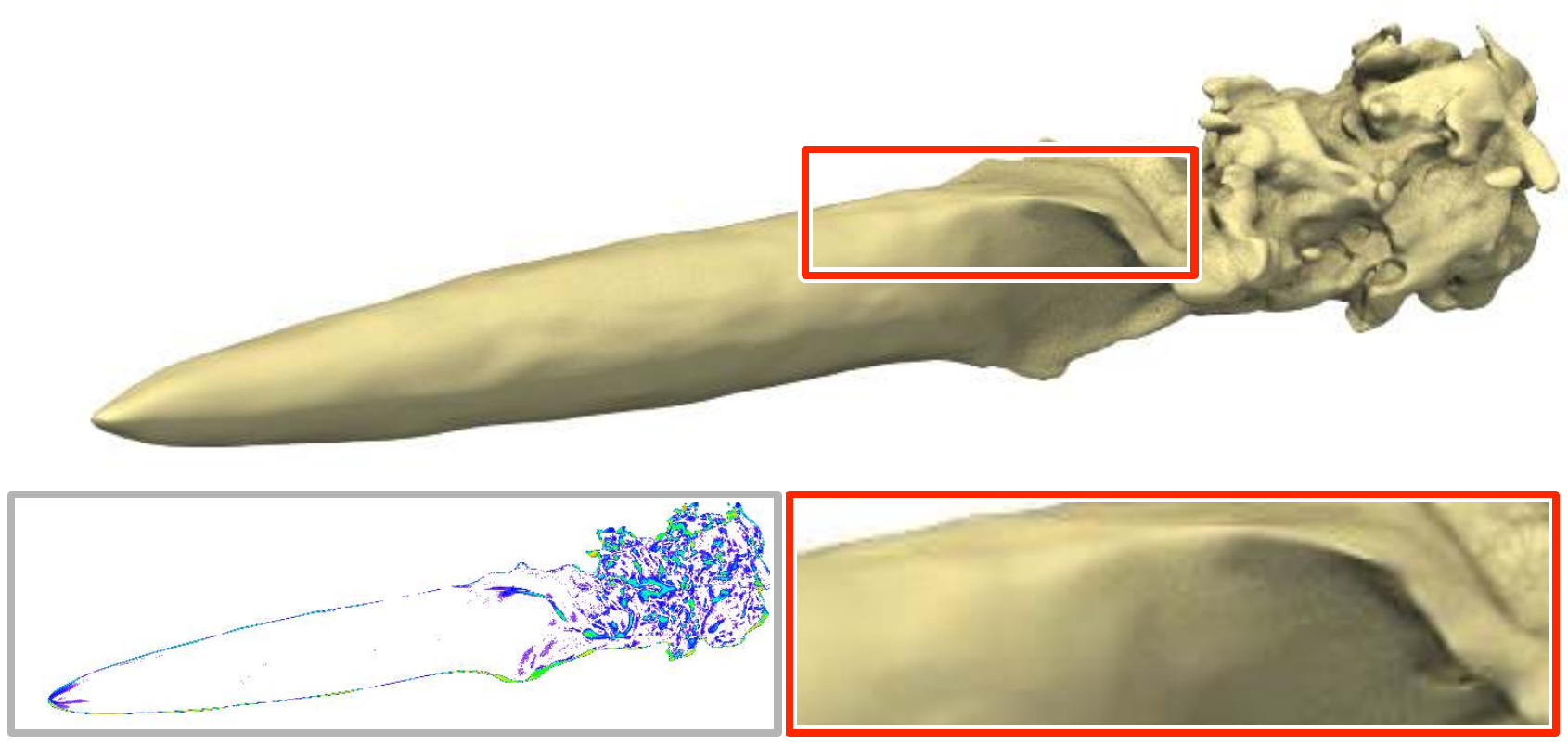}&
  \includegraphics[width=0.1925\linewidth]{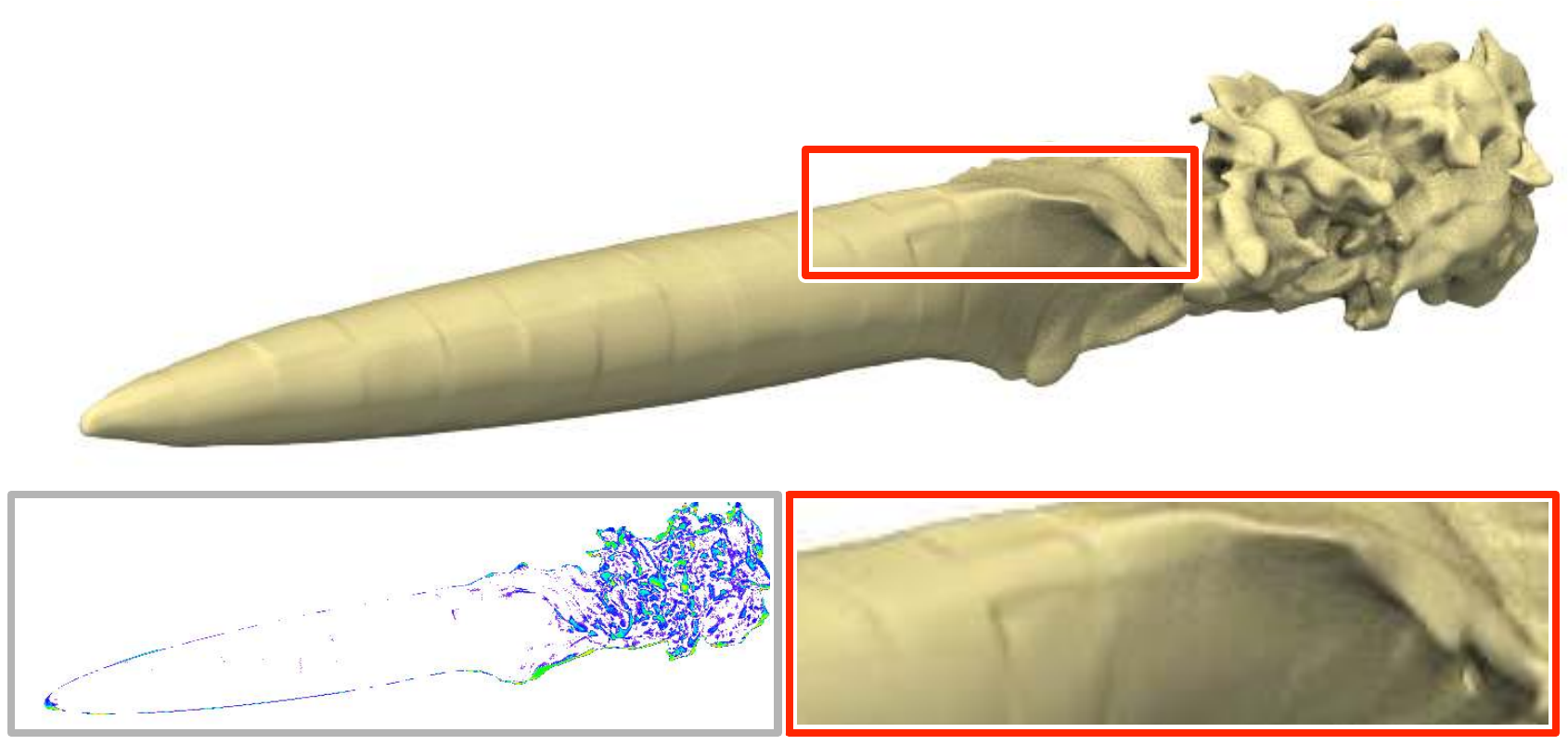}&
\includegraphics[width=0.1925\linewidth]{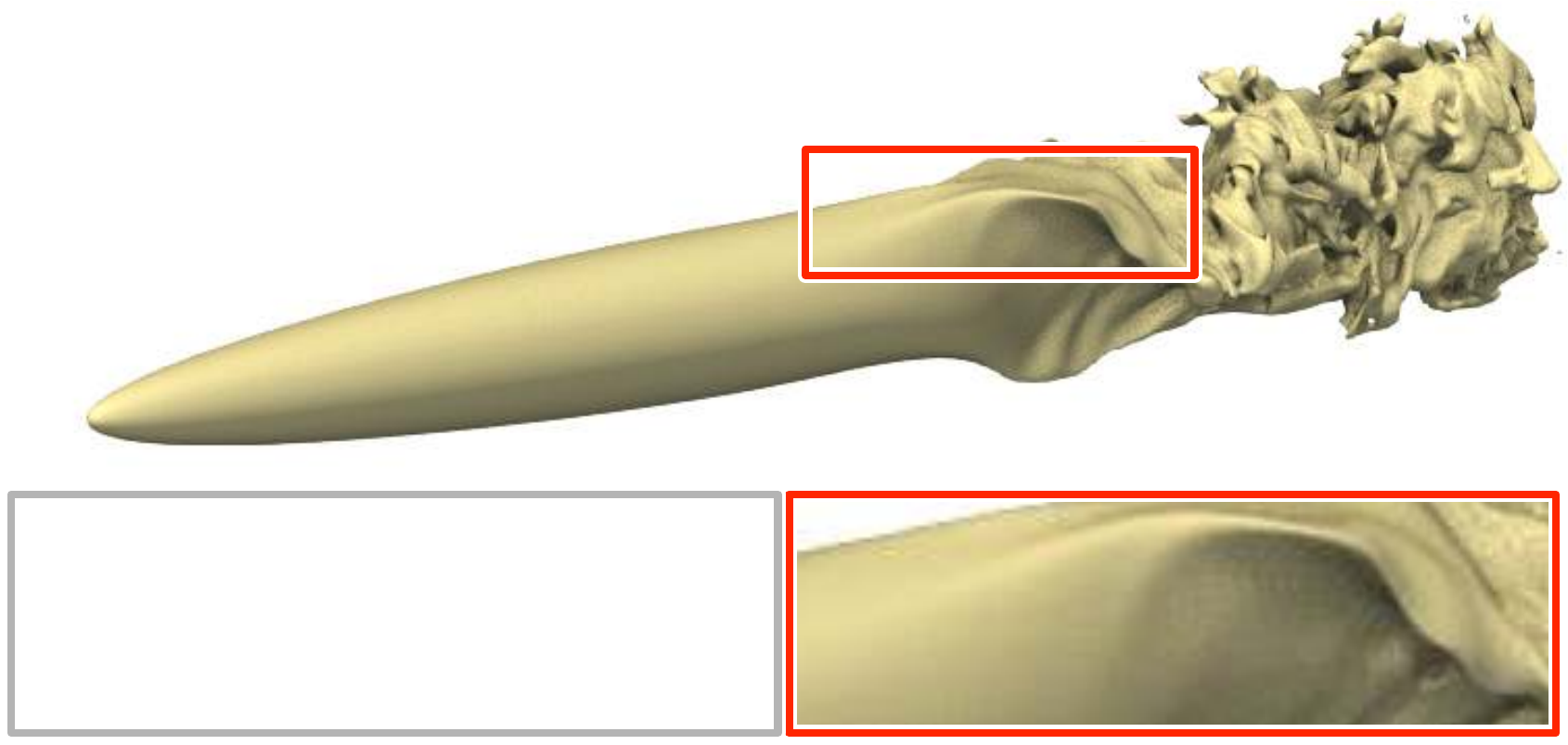}\\
 \includegraphics[width=0.1925\linewidth]{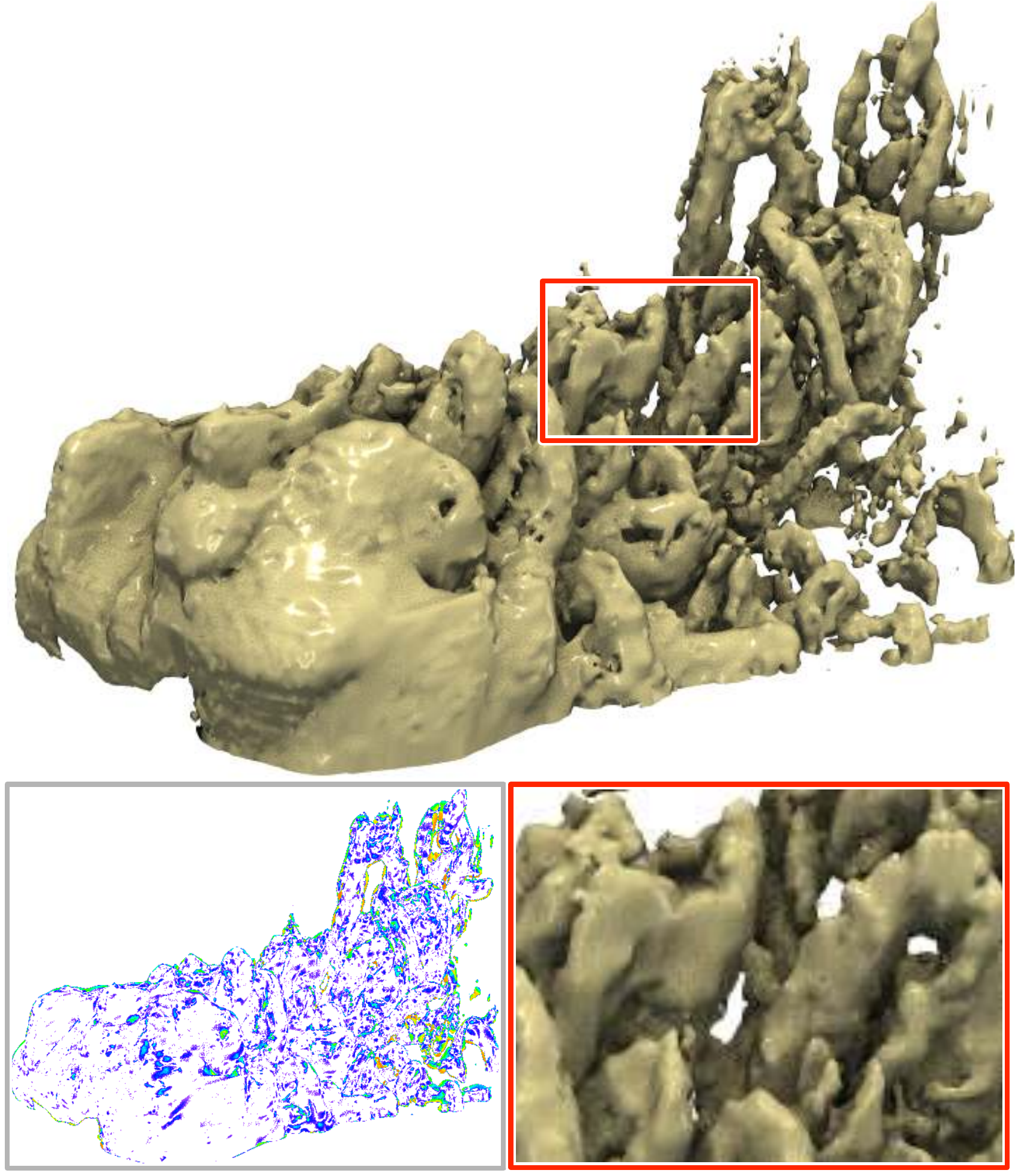}&
 \includegraphics[width=0.1925\linewidth]{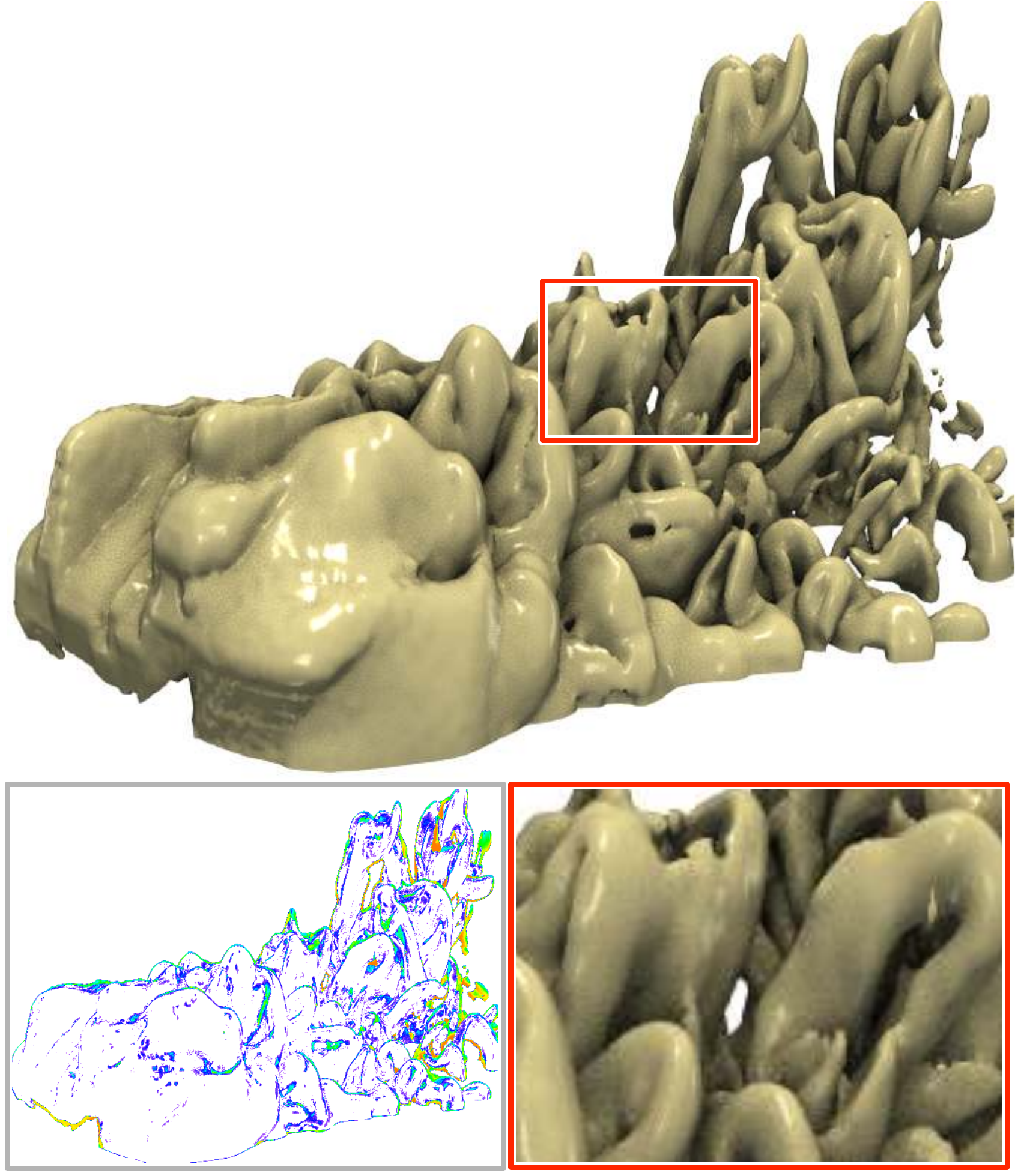}&
 \includegraphics[width=0.1925\linewidth]{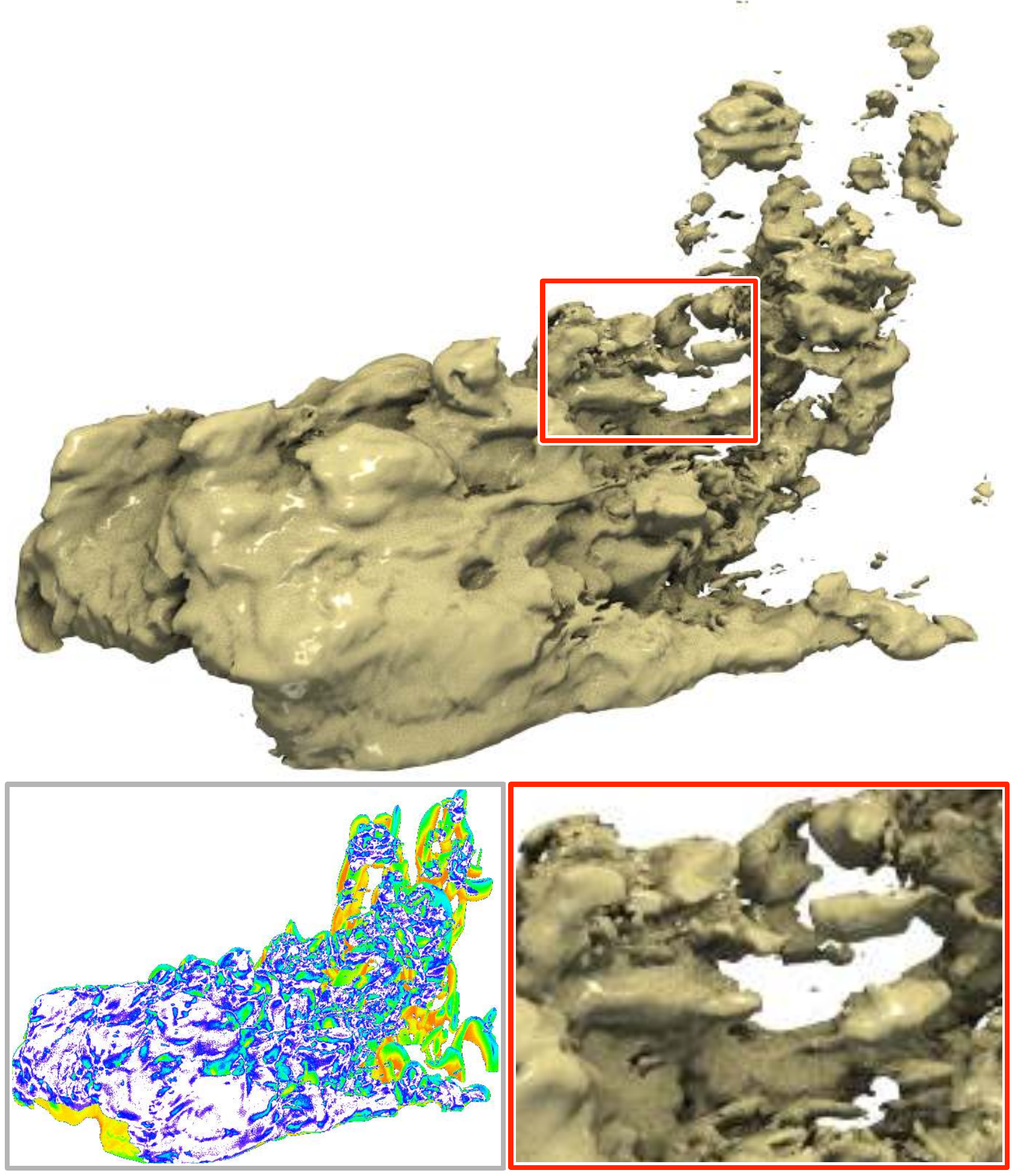}&
  \includegraphics[width=0.1925\linewidth]{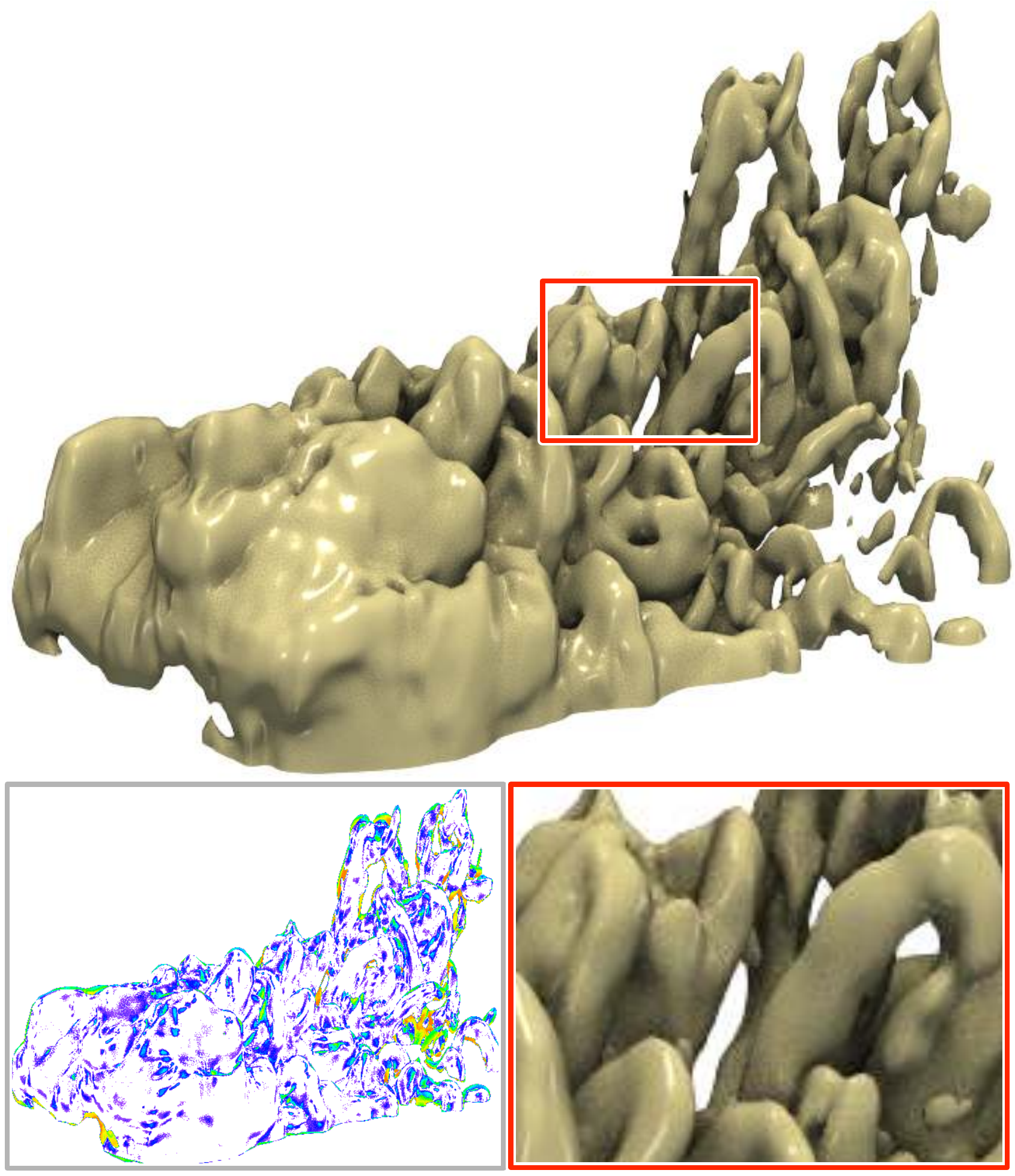}&
\includegraphics[width=0.1925\linewidth]{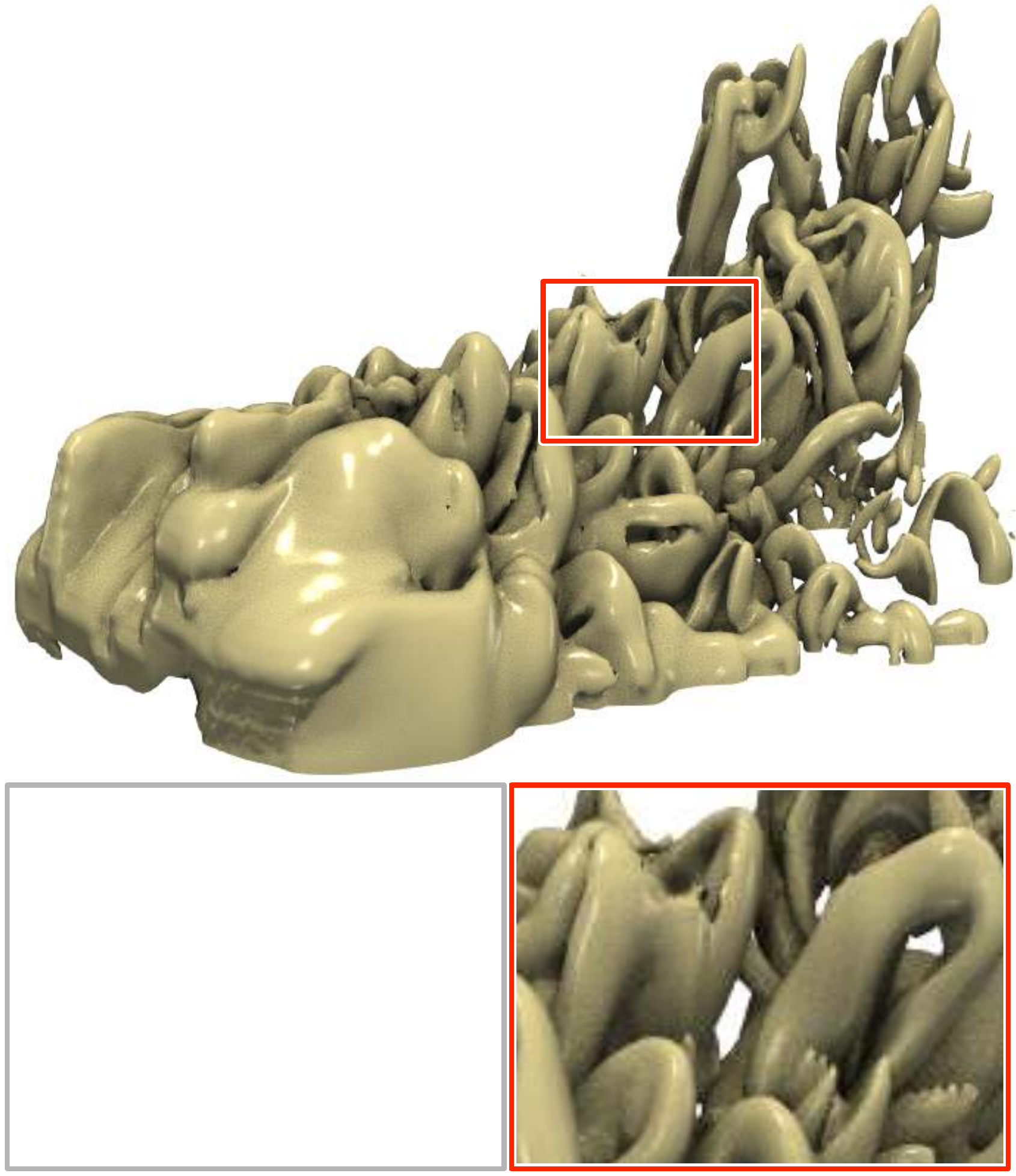}\\
\mbox{(a) SZ3} & \mbox{(b) TTHRESH} & \mbox{(c) neurcomp} &\mbox{(d) ECNR} &\mbox{(e) GT} 
\end{array}$
\end{center}
\vspace{-.25in} 
\caption{Isosurface rendering results of SZ3, TTHRESH, neurcomp, ECNR, and GT. From top to bottom: combustion, half-cylinder, solar plume, and Tangaroa. The chosen isovalues are reported in Table~\ref{tab:comp-time-varying}.} 
\label{fig:ir-results}
\end{figure*}

\vspace{-0.05in}
\subsection{Results}

We present the primary quantitative and qualitative comparison results of ECNR and three baseline methods (SZ3, TTHRESH, and neurcomp). Additional comparison with \hot{traditional and grid-based (MHT, fV-SRN, and APMGSRN) methods} and further discussion of ECNR's streaming reconstruction capability and neurcomp's unstable optimization are presented in the appendix. 

{\bf Baselines.} 
We compared our ECNR with three baseline solutions:
\begin{myitemize}
\vspace{-0.05in}
  \item SZ3~\cite{Liang-TBD22}: an error-bounded lossy compression method identifies sparse representations within the domain of locally spanning splines. 
  \item TTHRESH~\cite{Ripoll-TVCG20}: a lossy compressor using the Tucker decomposition, a higher-order singular value decomposition.
  \item neurcomp~\cite{Lu-CGF21}: an implicit neural network that utilizes MLPs for volumetric data representation, achieving a desirable CR by limiting the number of network parameters. 
\vspace{-0.05in}
\end{myitemize}
Both SZ3 and TTHRESH support data compression at higher dimensions. For neurcomp, we modified the original code by adding a time dimension to the network input. 
Hence, all methods compress a time-varying volumetric dataset as a space-time 4D volume. 

SZ3 and TTHRESH accept an error tolerance as the input parameter. To achieve a fair comparison, after we obtained the CR of ECNR, we adjusted the input tolerance of these two methods to match the CR of ECNR roughly. For neurcomp, we adapted the number of neurons in each hidden layer to reach a certain CR. We performed inference multiple rounds during training, each after a certain number of training iterations. We stopped the training when neurcomp reached a PSNR similar to ECNR or the network converged. The reported encoding time refers to the training time, excluding the time spent on multiple rounds of inference. 

{\bf Evaluation metrics.} 
Three metrics were considered in our evaluation. We used {\em peak signal-to-noise ratio} (PSNR) at the data level. The calculation is based on the original data and the compressed and decompressed version of data generated from one of the methods. At the image level, we utilized {\em learned perceptual image patch similarity} (LPIPS)~\cite{Zhang-CVPR18}. The calculation is based on the volume rendering images produced from the corresponding versions of data. At the feature level, {\em chamfer distance} (CD)~\cite{Barrow-IJCAI77} is measured on the isosurfaces extracted from the corresponding versions of data.

\begin{table}[htb]
\caption{Model size (MS, in MB) and GPU and CPU memory (GB) consumption for compressing the combustion dataset.}
\vspace{-0.1in}
\centering
{\scriptsize
\begin{tabular}{c|llll}
 method  &MS$\downarrow$  &GPU-mem$\downarrow$  &CPU-mem$\downarrow$  \\ \hline
 SZ3 &4.6 &--- &120.7  \\
 TTHRESH &5.3 &--- &115.3 \\
 neurcomp  &4.5  &15.97 & 202.1 \\ \hline
 ECNR scale 3  &0.068  &2.35  &5.6  \\
 ECNR scale 2  &0.434  &2.38  &8.4  \\
 ECNR scale 1  &4.2  &8.14  &34.3  \\
\end{tabular}
}
\label{tab:memory}
\end{table}

{\bf Quantitative results.}
In Table~\ref{tab:comp-time-varying}, we report quantitative results of applying the four methods to compress the time-varying datasets. 
Regarding the three quality metrics, ECNR achieves the best for all datasets, only losing to TTHRESH on the LPIPS of the combustion dataset and CD of the solar plume dataset. 
The CR of ECNR ranges from 838 to 6,262, depending on the underlying data complexity. 
The CRs of the other three methods reference those of ECNR. 
Regarding encoding and decoding time, SZ3 is the clear winner, followed by TTHRESH. 
The training speedup of ECNR over neurcomp is 2.09$\times$ to 3.18$\times$, 
and the inference speedup is 12.97$\times$ to 29.57$\times$. 
This shows the advantage of ENCR as an efficient INR-based solution compared with neurcomp. 
Note that the training speedup is lowest for the combustion dataset because if we train neurcomp further, the PSNR unexpectedly drops below 21 dB, which is undesirable. 
Even though ECNR's encoding time falls behind SZ3 and TTHRESH by a large margin, \hot{its decoding time is comparable to that of TTHRESH}. 
Furthermore, ECNR achieves better quality; for example, the average gain of PSNR is 5.87 dB over SZ3 and 7.18 dB over TTHRESH. 

Table~\ref{tab:memory} shows these four methods' model size and GPU memory and CPU memory (i.e., RAM) consumption for compressing the combustion dataset. 
The model size for SZ3 or TTHRESH simply refers to the compressed data size. 
SZ3 and TTHRESH do not utilize GPU, and thus we only report their CPU memory consumption. 
ECNR processes these scales sequentially, and we list them scale by scale. 
The memory consumption is capped at the finest scale (scale 1). 
Compared with neurcomp, the efficiency gain of ECNR is about 2$\times$ on GPU memory usage and 5.8$\times$ on CPU memory usage. 
ECNR takes less GPU memory due to its smaller parameter size and less CPU memory because its multiscale representation skips processing blocks with low residuals at a finer scale. 
The smaller memory footprint allows ECNR to process volume datasets with a large spatial extent and/or long temporal sequence. 

Furthermore, we report the performance of ECNR under different CRs for the time-varying datasets experimented. 
We controlled the CR of ECNR by adjusting its block dimension. 
Note that we only compared ECNR with SZ3 and TTHRESH in this experiment. 
Since neurcomp is far from convergence when controlling its training time to align with that of ECNR for a fair comparison. 
Figure~\ref{fig:diff-CR} shows the performance curves. 
The results indicate that ECNR achieves better accuracy than SZ3 and TTHRESH under high CRs. 
SZ3 is superior to TTHRESH except for the solar plume dataset. 
In the extreme case of the solar plume dataset, ECNR could still maintain a PSNR value above 35 dB under a CR higher than 17,500, while SZ3 already suffers large distortion when CR reaches 7,500.

{\bf Qualitative results.}
In Figures~\ref{fig:vr-results}~and~\ref{fig:ir-results}, we show volume rendering and isosurface rendering images at a selected timestep for different methods. We compute the pixel-wise difference images (i.e., the Euclidean distance in the CIELUV color space) to better perceive the differences between each method and GT. These different images are shown on the bottom-left side. On the bottom-right side, we show a zoomed-in view for close-up comparison. 
Overall, ECNR achieves the best visual quality in both volume rendering and isosurface rendering images. 
SZ3 and TTHRESH are undesirable, yielding severe block- and strip-like artifacts. 
Nevertheless, TTHRESH has the best quality for the combustion dataset as the specified transfer function or chosen isovalue does not reveal the lower value range where it suffers the most quality loss. 
neurcomp produces similar results as ECNR but still falls behind upon close examination. 

\begin{figure}[htb]
	\begin{center}
	$\begin{array}{c@{\hspace{0.02in}}c@{\hspace{0.02in}}c}
				\includegraphics[height=0.75in]{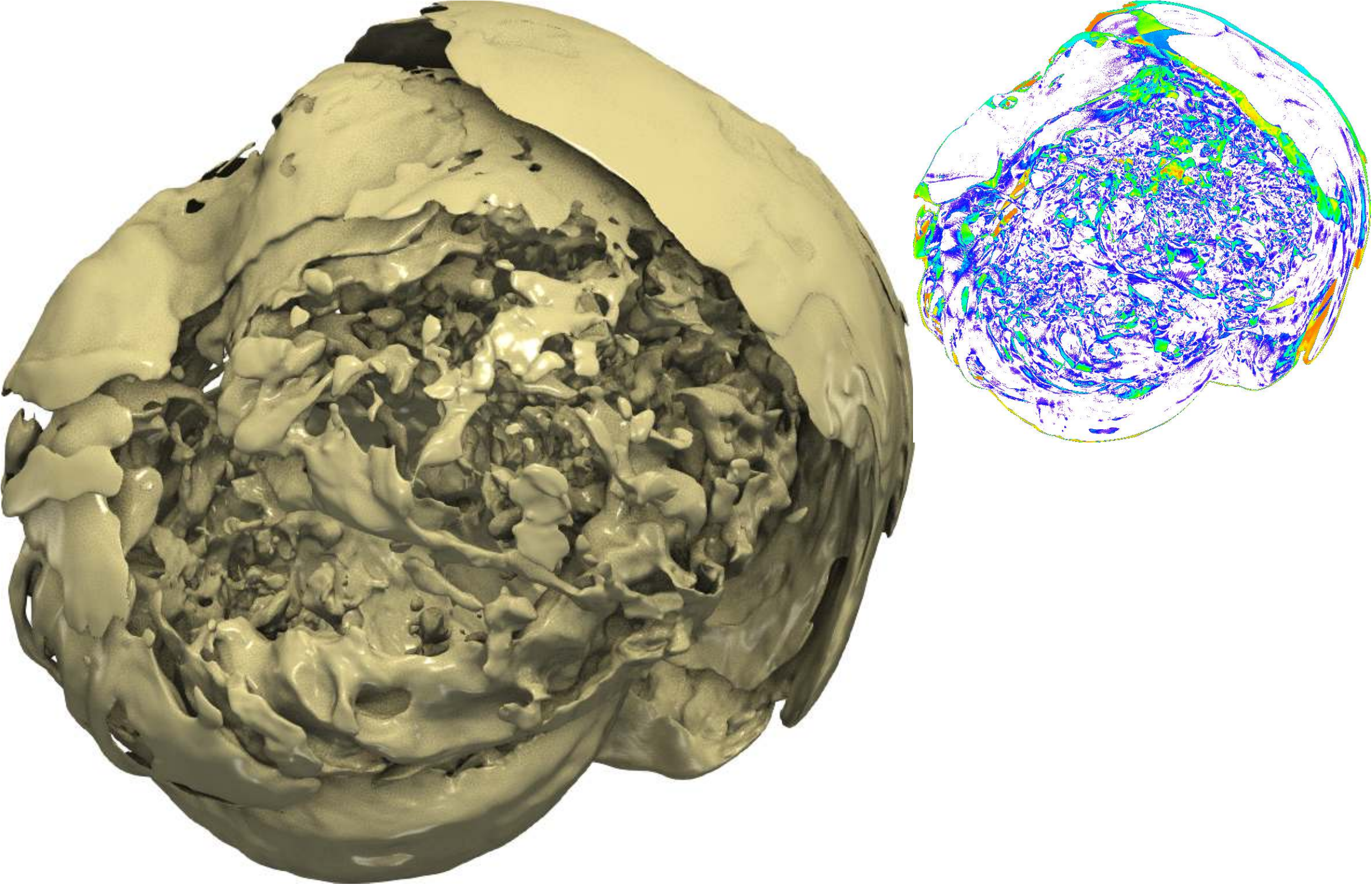}&
				\includegraphics[height=0.75in]{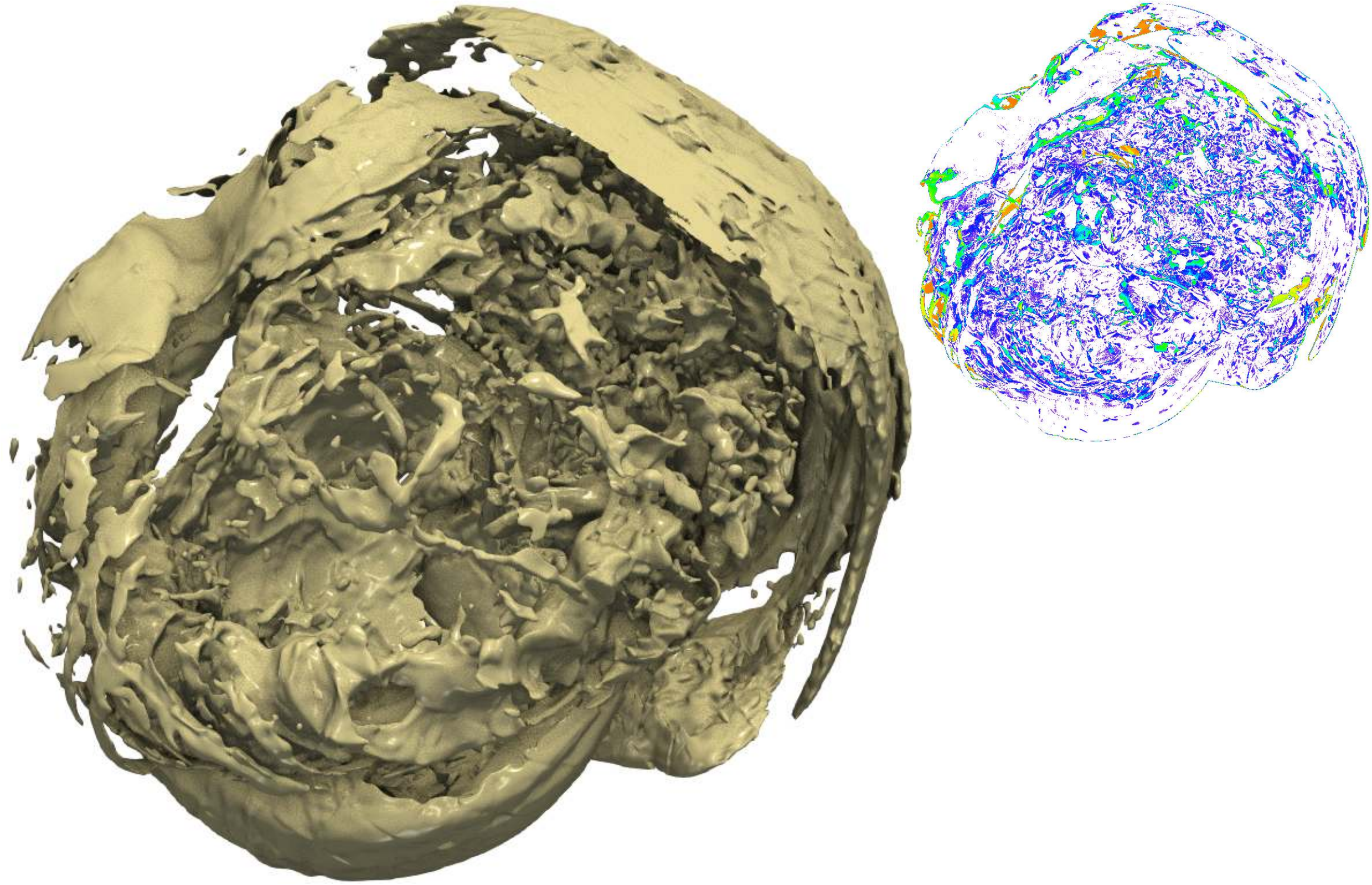}&
				\includegraphics[height=0.75in]{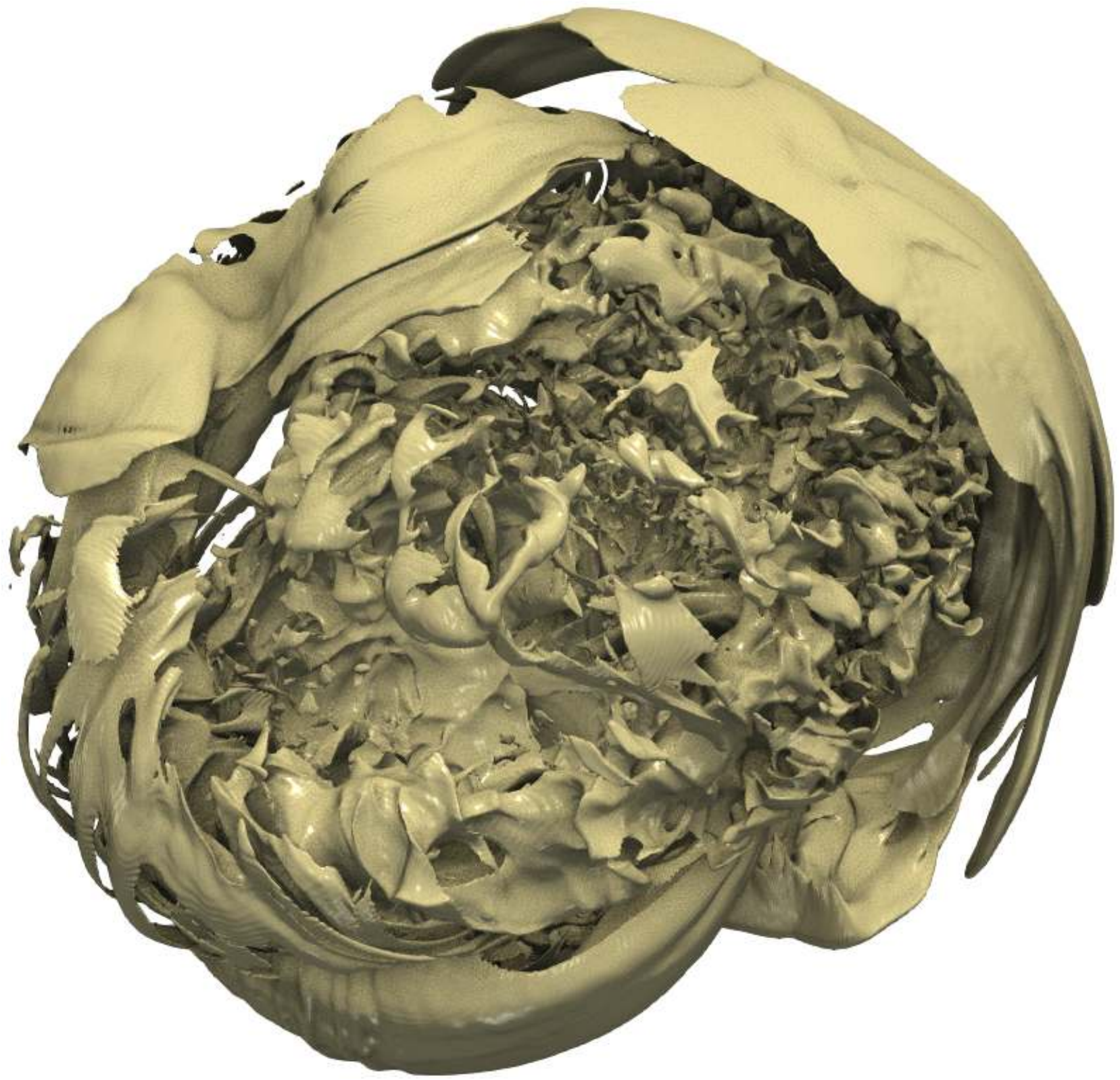}\\
				\mbox{\footnotesize (a) neurcomp} &
				\mbox{\footnotesize (b) ECNR} &
				\mbox{\footnotesize (c) GT}\\
	\end{array}$
	\end{center}	
	\vspace{-.25in}
	\caption{Isosurface rendering ($v = 0.45$) results of neurcomp and ECNR using the supernova-2 dataset.}
	\label{fig:largeSpatial}			
\end{figure}

{\bf Large spatial extent.}
In addition to the above results for time-varying data compression, we experimented with one large spatial static volume, supernova-2, with the dimension of $1200 \times 1200 \times 1200$, to evaluate the performance differences between ECNR and neurcomp. 
For ECNR, we set $s=4$ due to its large spatial extent. From the coarsest to finest scales, the numbers of target blocks per MLP were adjusted to 1, 2, 4, and 8 because the dataset is static. 
We optimized the model 225, 175, 150, and 125 epochs for each scale. 
\hot{Block-guided pruning} happened at 75 and 115 epochs with a target sparsity of 20\% and 30\%. 
After training, we quantized the MLP parameters with $\beta=9$ bits and fine-tuned 20 epochs. 
For neurcomp, we trained the model long enough until it converged. 
ECNR achieves better reconstruction accuracy based on the quantitative results reported in Table~\ref{tab:largeTemporalSpatial}. 
The isosurface rendering results shown in Figure~\ref{fig:largeSpatial} indicate that ECNR outperforms neurcomp in terms of visual quality.


\begin{table}[htb]
\caption{PSNR (dB), LPIPS, and CD ($v = 0.45$) values, as well as \hot{model size (MS, in MB)} of the supernova-2 dataset.}
\vspace{-0.1in}
\centering
{\scriptsize
\begin{tabular}{c|lll|l}
method 	& PSNR$\uparrow$	& LPIPS$\downarrow$ &CD$\downarrow$	& MS$\downarrow$	 \\ \hline
neurcomp &43.98 &\textbf{0.025} 	&4.24 &1.1  			  \\ 
ECNR  	&\textbf{47.65} &0.028  &\textbf{2.94} &\textbf{0.99}\\ 
\end{tabular}
}
\label{tab:largeTemporalSpatial}
\end{table}

\begin{figure}[htb]
	\begin{center}
		$\begin{array}{c@{\hspace{0.02in}}c@{\hspace{0.02in}}c@{\hspace{0.02in}}c}
				\includegraphics[height=0.8in]{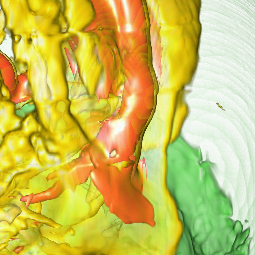}&
				\includegraphics[height=0.8in]{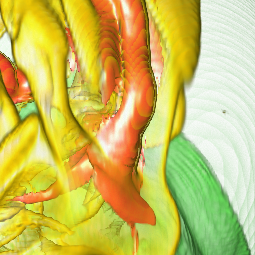} &
				\includegraphics[height=0.8in]{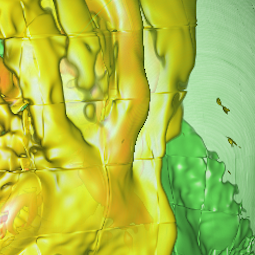} &
				\includegraphics[height=0.8in]{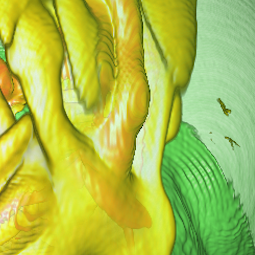}    \\
				\mbox{\footnotesize (a) \hot{ECNR}} &
				\mbox{\footnotesize (b) \hot{GT}} &
				\mbox{\footnotesize (c) \hot{ECNR}} &
				\mbox{\footnotesize (d) \hot{GT}} \\
		\end{array}$
	\end{center}
	\vspace{-.25in}
	\caption{\hot{Zoomed-in volume rendering results of ECNR with datasets of different spatial resolutions. (a) and (b): supernova-1. (c) and (d): supernova-3.}}
	\label{fig:diff-resolution}
\end{figure}

\hot{{\bf Evaluation of different spatial resolutions.}
To analyze the performance of ECNR on datasets with different spatial resolutions, we compressed supernova-1 and supernova-3 (64$\times$ that of supernova-1). 
We set the block dimension of supernova-3 4$\times$ that of supernova-1 and kept other hyperparameters identical.
Table~\ref{tab:diff-resolution} shows that supernova-3 needs a significantly longer time than supernova-1 for both encoding and decoding, while the model size of supernova-3 is even less than supernova-1. 
The corresponding CRs are 205 and 13,122 for supernova-1 and supernova-3, respectively. 
This is because supernova-3 employs proportionally larger blocks than supernova-1, making each MLP iterate more coordinates in reconstruction and less sensitive to small fitting errors when obtaining effective blocks. 
As for visual quality, we show zoomed-in volume rendering in Figure~\ref{fig:diff-resolution}.
For larger datasets and block dimensions, the reconstruction results tend to be smoother with more obvious boundary artifacts due to the limited capacity of local MLPs.}

\begin{table}[htb]
\caption{\hot{PSNR (dB) and LPIPS values, as well as model size (MS, in MB), encoding time (ET), and decoding time (DT).}}
\vspace{-0.1in}
\centering
{\scriptsize
\hot{
\begin{tabular}{c|lll|ll}
dataset & PSNR$\uparrow$ & LPIPS$\downarrow$ & MS$\downarrow$ & ET$\downarrow$ & DT$\downarrow$ \\ \hline
supernova-1 &48.55 &0.029 &1.5 &2.2m &2.2s\\
supernova-3 &49.94 &0.052 &1.3 &3.1h &8.3m\\ 
\end{tabular}
}}
\label{tab:diff-resolution}
\end{table}

\begin{table}[htb]
\caption{PSNR (dB), LPIPS, and CD values, as well as \hot{model size (MS, in MB)}.}
\vspace{-0.1in}
\centering
{\scriptsize
\begin{tabular}{cc|lll|l}
dataset &method 	& PSNR$\uparrow$	& LPIPS$\downarrow$ &CD$\downarrow$	& MS$\downarrow$	 \\ \hline
asteroids			&MINER &37.56  &0.256	&7.56 	&0.53	  			  \\ 
($v = 0.0$)	&MINER+ 	& 39.92 &0.118 	&2.80 &0.50  			  \\ 			
 			&ECNR  	&\textbf{40.59}	&\textbf{0.107}  &\textbf{2.18} &\textbf{0.49}\\\hline
supernova-1		&MINER 	&\textbf{50.47}  &\textbf{0.026}	&0.75 	&14.2	  			  \\ 
($v = 0.0$)	&MINER+ 	& 49.03 	&0.035 	&\textbf{0.67} &3.7  			  \\ 
 			&ECNR  	&	48.55 	& 0.029 &0.74 &\textbf{1.5}\\ 
\end{tabular}
}
\label{tab:block-assignment}
\end{table}

\begin{figure}[htb]
	\begin{center}
		$\begin{array}{c@{\hspace{0.01in}}c@{\hspace{0.01in}}c@{\hspace{0.01in}}c}
				\includegraphics[height=0.65in]{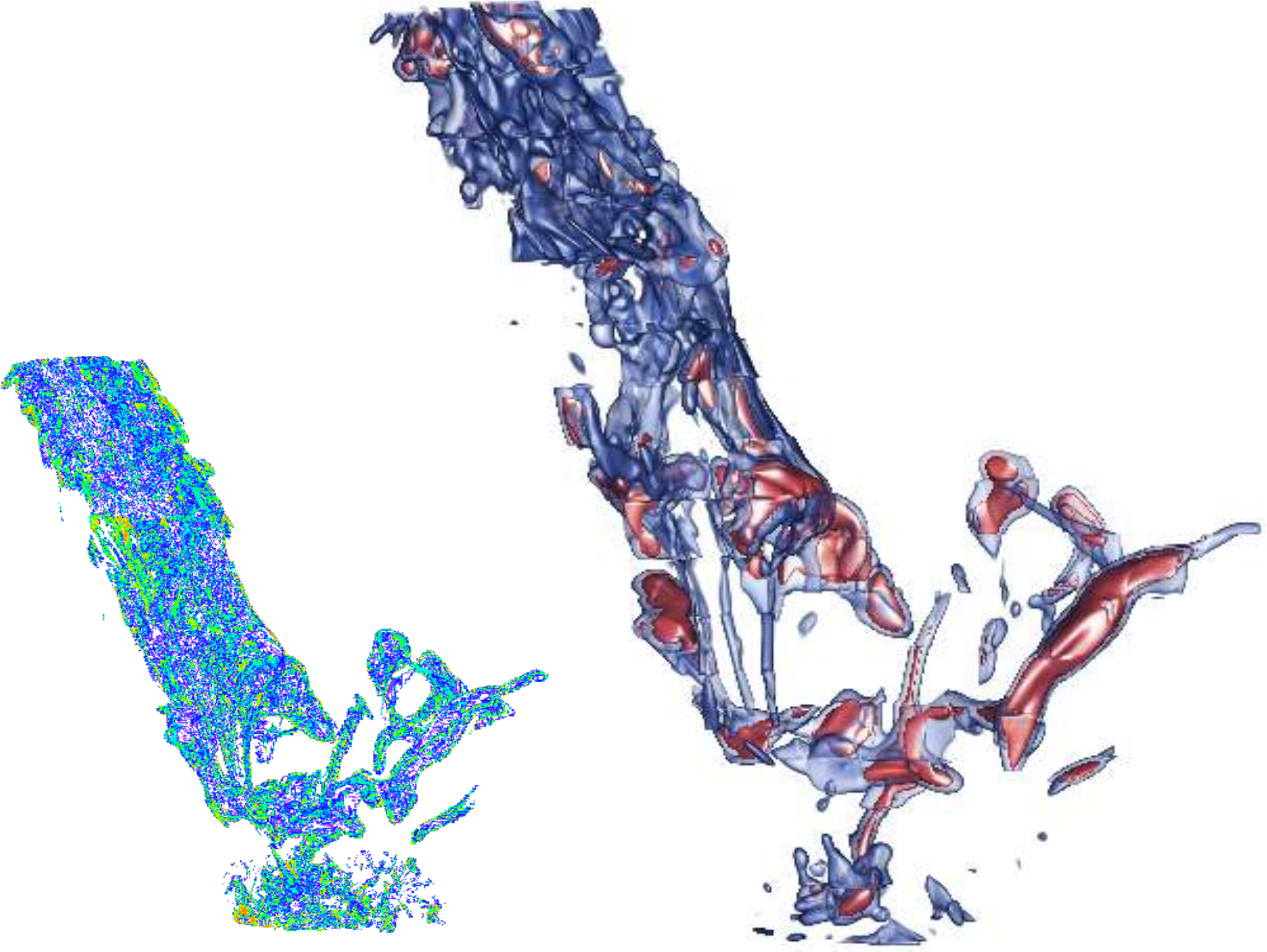}&
				\includegraphics[height=0.65in]{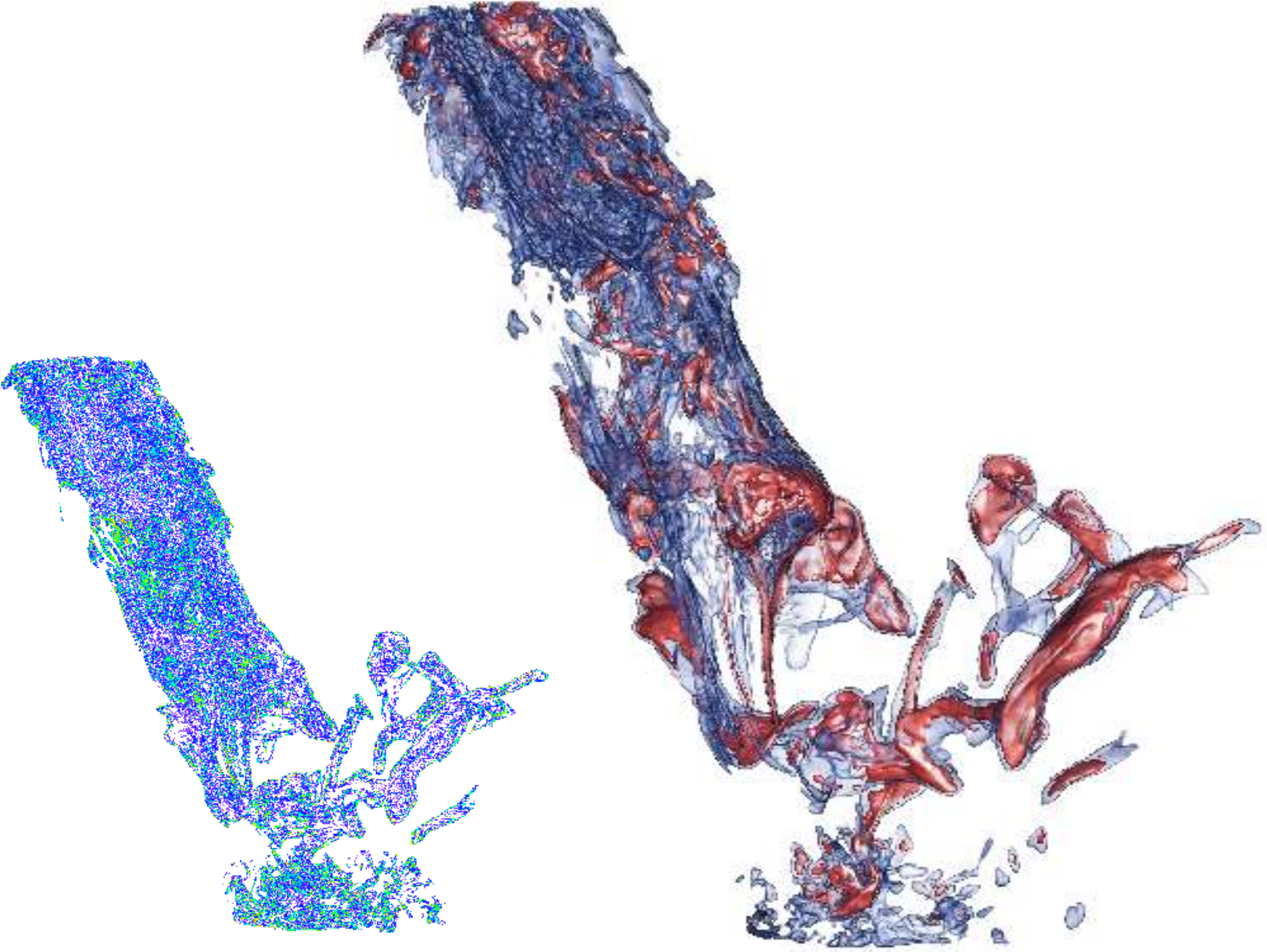}&
				\includegraphics[height=0.65in]{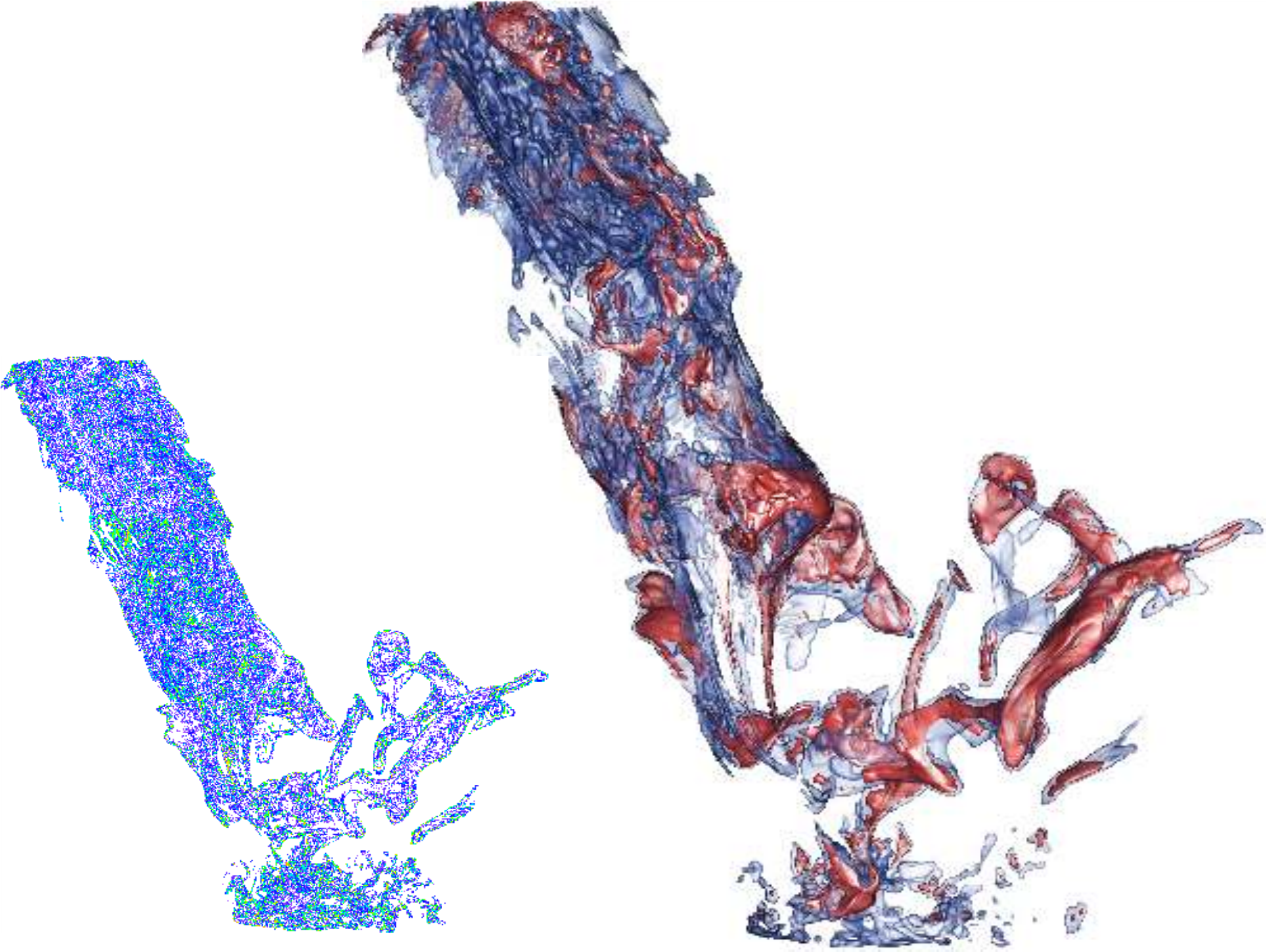} &
				\includegraphics[height=0.65in]{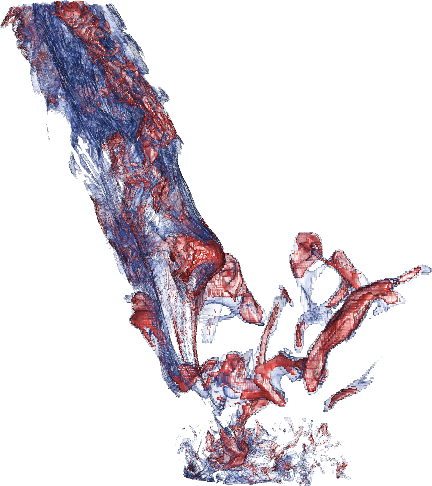}    \\
				\includegraphics[height=0.675in]{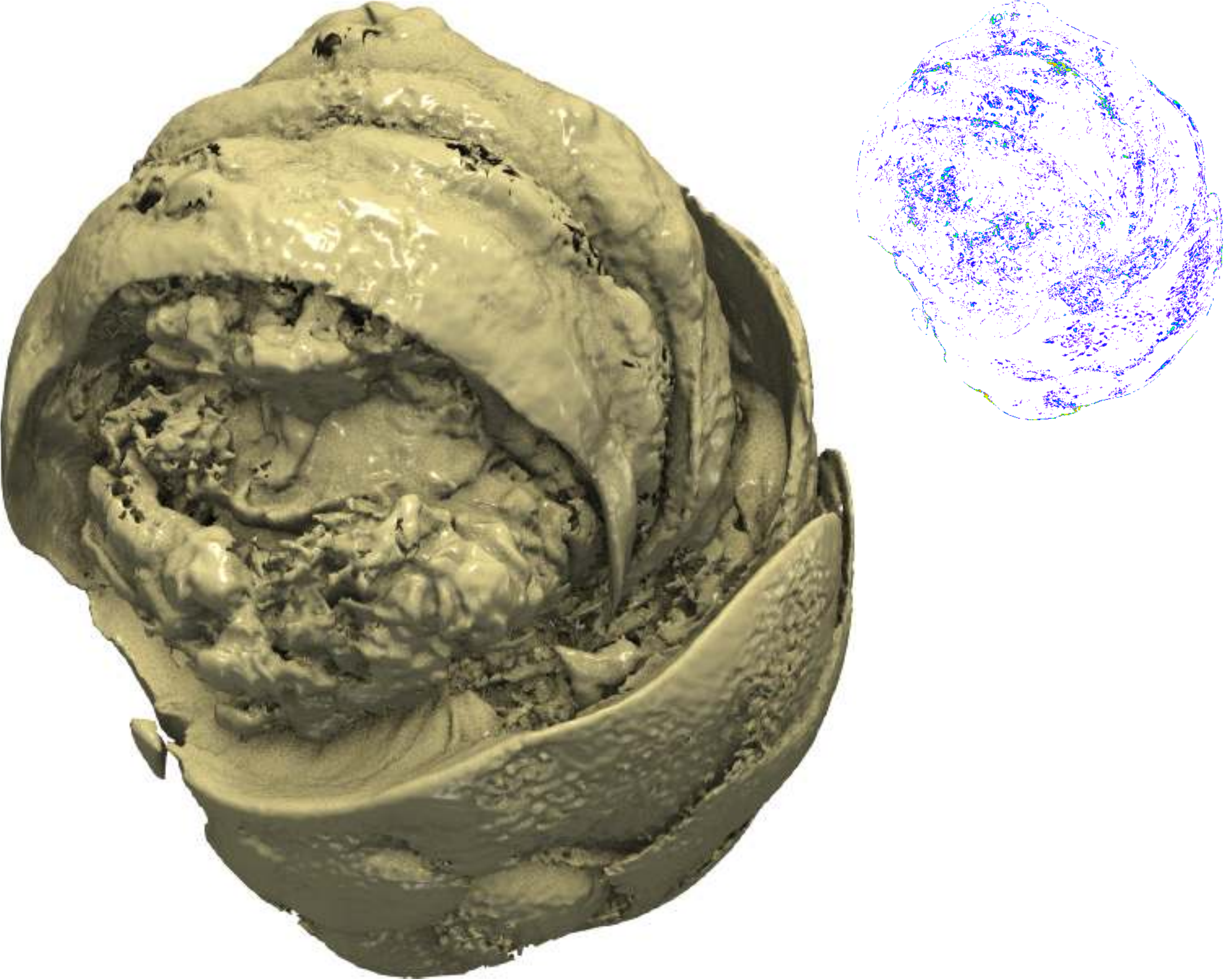}&
				\includegraphics[height=0.675in]{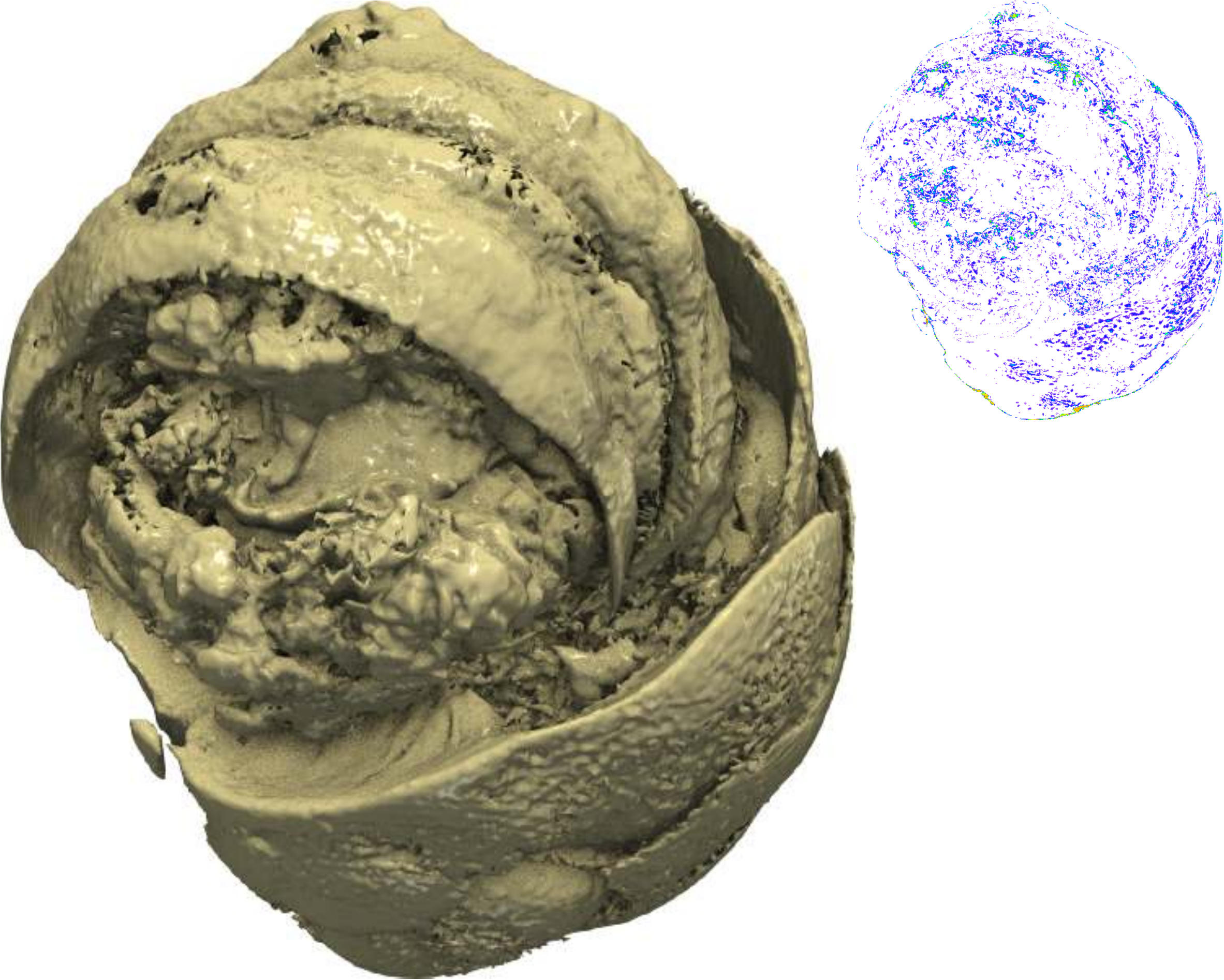}&
				\includegraphics[height=0.675in]{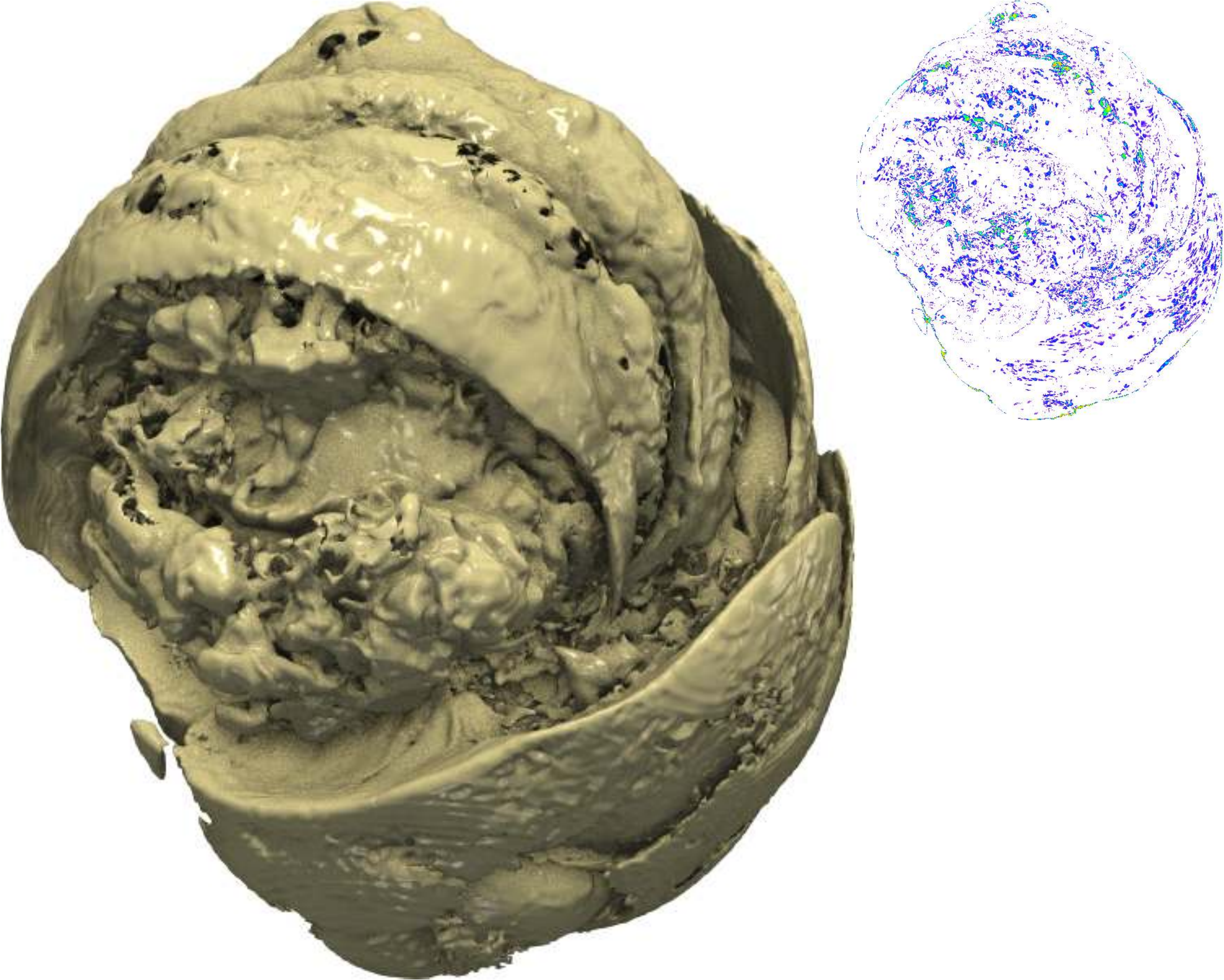} &
				\includegraphics[height=0.675in]{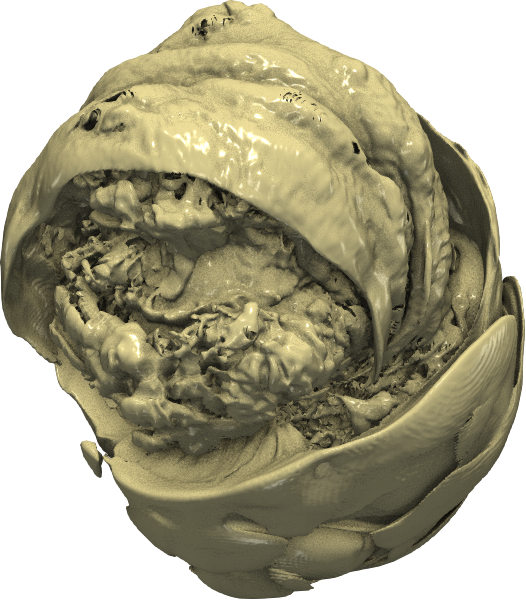}    \\
				\mbox{\footnotesize (a) MINER}   & \mbox{\footnotesize (b) MINER+} & \mbox{\footnotesize (c) ECNR} & \mbox{\footnotesize (d) GT} 
			\end{array}$
	\end{center}
	\vspace{-.25in}
	\caption{Rendering results of MINER, MINER+, and ECNR.
	Top: volume rendering using the asteroids dataset. 
	Bottom: isosurface rendering ($v = 0.0$) using the supernova-1 dataset.}
	\label{fig:comp-MINER}
\end{figure}

\begin{table}[htb]
\caption{Average PSNR (dB) and LPIPS values across all timesteps, as well as \hot{encoding time (ET), decoding time (DT), and model size (MS, in MB)} of the half-cylinder dataset.}
\vspace{-0.1in}
\centering
{\scriptsize
\begin{tabular}{c|ll|ll|l}
scheme 	& PSNR$\uparrow$	& LPIPS$\downarrow$ 	& ET$\downarrow$	& DT$\downarrow$	 & MS$\downarrow$	 \\ \hline
w/o CNN 	& 44.61 			& 0.036	 			& \textbf{4.7h}			& \textbf{25.3s}	 &\textbf{1.65}  			  \\ 
 w/ CNN  	&	\textbf{45.12}	& \textbf{0.028} 	& 4.8h			& 78.6s			 &1.75\\ 
\end{tabular}
}
\label{tab:CNN_eval}
\end{table}

\begin{figure}[htb]
	\begin{center}
				\includegraphics[height=0.425in]{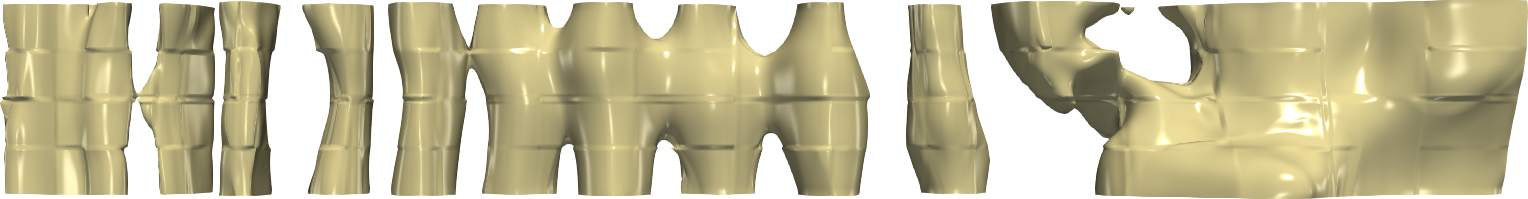}\\
				\mbox{\footnotesize (a) w/o CNN} \\			
				\includegraphics[height=0.425in]{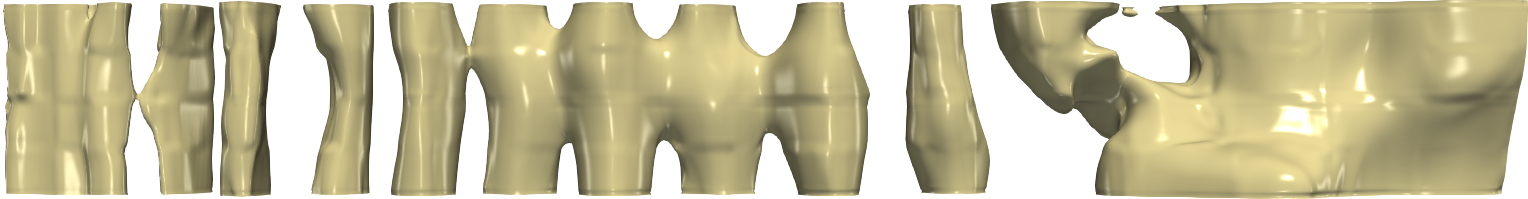}\\
				\mbox{\footnotesize (b) w/ CNN} \\			
				\includegraphics[height=0.425in]{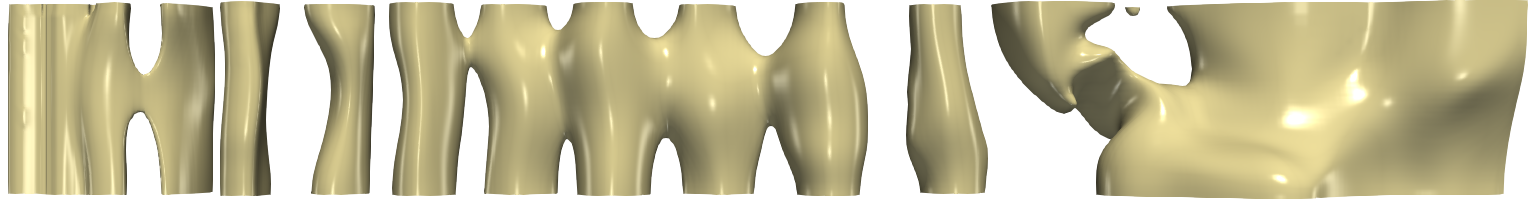}\\
				\mbox{\footnotesize (c) GT}\\			
	\end{center}
	\vspace{-.25in}
	\caption{Isosurface rendering ($v = 0.1$) results of ECNR using the half-cylinder dataset with or without CNN post-processing.}
	\label{fig:CNN_eval}
\end{figure}

\vspace{-0.05in}
\subsection{Network Analysis}

For network analysis of ECNR, we investigate block assignment and lightweight CNN. \hot{Additional network analysis in six aspects, including sparse vs.\ dense model, deep compression ablation study, comparison with standard pruning and local quantization, block dimension, choice of $s$, and $\lambda_b$ in block-guided pruning, is presented in the appendix.} 

{\bf Block assignment.}
To investigate the impact of block assignment, we compared ECNR with and without block assignment on compressing the asteroids and supernova-1 datasets. Without block assignment, the network processes the data like MINER, where each MLP is assigned to encode a single block. MINER does not optimize or store latent codes since no multiple blocks are assigned to an MLP. So, besides this basic version, we employed the same deep compression strategy (Section~\ref{sec:deep-comp}) for MINER to make the comparison fair and name it MINER+. 
For the asteroids dataset, we keep the model sizes of MINER and MINER+ similar to that of ECNR, and therefore, all three methods will have similar CRs. 
For the supernova-1 dataset, we keep the same model parameters (i.e., block dimension, neuron number) for all three methods. The resulting model size of MINER is 3.84$\times$ that of MINER+, and 9.47$\times$ that of ECNR. 
We report quantitative results in Table~\ref{tab:block-assignment} and corresponding rendering results in Figure~\ref{fig:comp-MINER}.
The results show that with the same model size, ECNR yields the best result and MINER+ outperforms MINER. 
With the same model parameters, all three achieve similar results while ECNR gains the highest CR. 

{\bf Lightweight CNN.}
We conducted an ablation study on the half-cylinder dataset to investigate the cost and effectiveness of the lightweight CNN for reducing boundary artifacts. 
The CNN only adds about five minutes to the training time and 100 KB to the model size (refer to Table~\ref{tab:CNN_eval}). Even though decoding takes an additional 50 seconds due to the sequential inference of CNN, it is still significantly faster than neurcomp, and the inference result (refer to Figure~\ref{fig:CNN_eval}) shows less pronounced boundary artifacts. 

%

\vspace{-0.05in}
\subsection{\hot{Discussion and Limitations}}

\hot{As a neural representation solution based on GPUs, the training and inference performance of ECNR, while outperforming neurcomp, still lags significantly behind SZ3 and TTHRESH (refer to Table~\ref{tab:comp-time-varying}). Thanks to the unified 4D space-time partitioning strategy, ECNR can handle large time-varying datasets as long as the memory can hold the effective blocks at the finest scale. 

One caveat is that due to the multiscale spatiotemporal representation, ECNR does not process each timestep independently, which may limit its use. A concurrent work of KD-INR~\cite{Han-TVCG23} uses a flat block partition to lift this constraint. 
%
Second, our current ECNR implementation enforces the same number of scales for both spatial and temporal domains in the multiscale block partitioning. This may not work ideally for volumetric datasets with a small spatial dimension and a large temporal dimension. An easy solution may be to manually separate the dataset into several temporal intervals and compress them individually. However, such a scenario may hinder the model from fully utilizing temporal coherence or redundancy. A strategy that handles these domains separately can address this issue. 
Third, the effectiveness of ECNR, like any other compression method, would drop when the time-varying dataset exhibits consistently super-rich spatiotemporal content and/or variations. In such a scenario, the percentage of effective blocks remains high at fine scales, limiting the compression efficacy. 
}

Additionally, we use the Euclidean distance to measure the similarity between blocks in the block assignment. Even though the Euclidean distance can be computed cost-effectively, it does not capture the distribution similarity between blocks. Using the Jensen-Shannon divergence or a more advanced statistical measure may offer a more appropriate solution. 
Finally, the limited capacity of the lightweight CNN does not guarantee the removal of boundary artifacts that exist in the MLP-decoded results when the distortion is significant. The remaining artifacts in the decompressed data may still negatively impact the visual quality of the rendering results.

\vspace{-0.05in}
\section{Conclusions and Future Work}

We have presented ECNR, an efficient compressive neural representation solution for compressing time-varying volumetric datasets. As an INR-based method, ECNR abandons the ``one-size-fits-all" notion of nercomp and advocates using multiple small MLPs to encode local blocks in a multiscale fashion. This enables us to consume only about half of the GPU memory and less than 20\% of the CPU memory as required by neurcomp, allowing successful processing of time-varying datasets with a large spatial extent and/or long temporal sequence. 
The multiscale strategy taken by ECNR leads to significant encoding and decoding speedups compared with neurcomp, positioning it as a practical solution for deployment. Experimental results show that under a similar CR, ECNR achieves better quality (PSNR at the data level, LPIPS at the image level, and CD at the feature level) than neurcomp and conventional compression methods (SZ3 and TTHRESH). 

Our future work includes the following. 
First, the current implementation is built with PyTorch only, without utilizing advanced CUDA techniques to accelerate MLP computations. We plan to integrate tiny-cuda-nn to improve the encoding and decoding efficiency. 
Second, multiscale block partitioning produces a series of hierarchical bounding boxes, potentially serving as stratified sampling criteria in the rendering process. We will leverage this property and evaluate the neural rendering performance of ECNR as an SRN like fV-SRN~\cite{Weiss-CGF22}.
Third, we would like to extend ECNR to handle multivariate time-varying datasets by ingesting variable-specific latent vector information into the encoding and decoding process. 

\vspace{-0.05in}
\section*{Acknowledgements}
This research was supported in part by the U.S.\ National Science Foundation through grants IIS-1955395, IIS-2101696, OAC-2104158, and the U.S.\ Department of Energy through grant DE-SC0023145. The authors would like to thank the anonymous reviewers for their insightful comments.

\vspace{-0.05in}
\bibliographystyle{abbrv}
\bibliography{template-abbv}

\begin{thebibliography}{10}

\bibitem{Ripoll-TVCG20}
R.~Ballester-Ripoll, P.~Lindstrom, and R.~Pajarola.
\newblock {TTHRESH}: Tensor compression for multidimensional visual data.
\newblock {\em IEEE TVCG}, 26(9):2891--2903, 2020.

\bibitem{Barrow-IJCAI77}
H.~G. Barrow, J.~M. Tenenbaum, R.~C. Bolles, and H.~C. Wolf.
\newblock Parametric correspondence and chamfer matching: Two new techniques for image matching.
\newblock In {\em Proc. IJCAI}, pages 659--663, 1977.

\bibitem{Berger-TVCG19}
M.~Berger, J.~Li, and J.~A. Levine.
\newblock A generative model for volume rendering.
\newblock {\em IEEE TVCG}, 25(4):1636--1650, 2019.

\bibitem{Biswas-TVCG21}
A.~Biswas, S.~Dutta, E.~Lawrence, J.~Patchett, J.~C. Calhoun, et~al.
\newblock Probabilistic data-driven sampling via multi-criteria importance analysis.
\newblock {\em IEEE TVCG}, 27(12):4439--4454, 2021.

\bibitem{Diaz-C&G20}
J.~D{\'\i}az, F.~Marton, and E.~Gobbetti.
\newblock Interactive spatio-temporal exploration of massive time-varying rectilinear scalar volumes based on a variable bit-rate sparse representation over learned dictionaries.
\newblock {\em C\&G}, 88:45--56, 2020.

\bibitem{Dutta-PVIS17}
S.~Dutta, J.~Woodring, H.-W. Shen, J.-P. Chen, and J.~P. Ahrens.
\newblock Homogeneity guided probabilistic data summaries for analysis and visualization of large-scale data sets.
\newblock In {\em Proc. PacificVis}, pages 111--120, 2017.

\bibitem{Gu-CG23}
P.~Gu, D.~Z. Chen, and C.~Wang.
\newblock {NeRVI}: Compressive neural representation of visualization images for communicating volume visualization results.
\newblock {\em C\&G}, 116:216--227, 2023.

\bibitem{Han-VIS19}
J.~Han and C.~Wang.
\newblock {TSR-TVD}: Temporal super-resolution for time-varying data analysis and visualization.
\newblock {\em IEEE TVCG}, 26(1):205--215, 2020.

\bibitem{Han-TVCG22}
J.~Han and C.~Wang.
\newblock {SSR-TVD}: Spatial super-resolution for time-varying data analysis and visualization.
\newblock {\em IEEE TVCG}, 28(6):2445--2456, 2022.

\bibitem{Han-TVCG}
J.~Han and C.~Wang.
\newblock {CoordNet}: Data generation and visualization generation for time-varying volumes via a coordinate-based neural network.
\newblock {\em IEEE TVCG}, 29(12):4951--4963, 2023.

\bibitem{Han-TVCG23}
J.~Han, H.~Zheng, and C.~Bi.
\newblock {KD-INR}: Time-varying volumetric data compression via knowledge distillation-based implicit neural representation.
\newblock {\em IEEE TVCG}, 2023.
\newblock Accepted.

\bibitem{Han-VIS21}
J.~Han, H.~Zheng, D.~Z. Chen, and C.~Wang.
\newblock {STNet}: An end-to-end generative framework for synthesizing spatiotemporal super-resolution volumes.
\newblock {\em IEEE TVCG}, 28(1):270--280, 2022.

\bibitem{Han-arXiv15}
S.~Han, H.~Mao, and W.~J. Dally.
\newblock Deep compression: Compressing deep neural networks with pruning, trained quantization and {Huffman} coding.
\newblock {\em arXiv preprint arXiv:1510.00149}, 2015.

\bibitem{Han-NIPS15}
S.~Han, J.~Pool, J.~Tran, and W.~Dally.
\newblock Learning both weights and connections for efficient neural network.
\newblock In {\em Proc. NeurIPS}, pages 1135--1143, 2015.

\bibitem{He-VIS19}
W.~He, J.~Wang, H.~Guo, K.-C. Wang, H.-W. Shen, et~al.
\newblock {InSituNet}: Deep image synthesis for parameter space exploration of ensemble simulations.
\newblock {\em IEEE TVCG}, 26(1):23--33, 2020.

\bibitem{Huffman-IRE52}
D.~A. Huffman.
\newblock A method for the construction of minimum-redundancy codes.
\newblock {\em Proc. IRE}, 40(9):1098--1101, 1952.

\bibitem{Iverson-ECPP12}
J.~Iverson, C.~Kamath, and G.~Karypis.
\newblock Fast and effective lossy compression algorithms for scientific datasets.
\newblock In {\em Proc. ECPP}, pages 843--856, 2012.

\bibitem{lakshminarasimhan-ECPP11}
S.~Lakshminarasimhan, N.~Shah, S.~Ethier, S.~Klasky, R.~Latham, et~al.
\newblock Compressing the incompressible with {ISABELA}: In-situ reduction of spatio-temporal data.
\newblock In {\em Proc. ECPP}, pages 366--379, 2011.

\bibitem{Li-TVCG}
H.~Li, X.~Yang, H.~Zhai, Y.~Liu, H.~Bao, et~al.
\newblock {Vox-Surf}: Voxel-based implicit surface representation.
\newblock {\em IEEE TVCG}, 2022.
\newblock Accepted.

\bibitem{Li-CGF18}
S.~Li, N.~Marsaglia, C.~Garth, J.~Woodring, J.~P. Clyne, and H.~Childs.
\newblock Data reduction techniques for simulation, visualization and data analysis.
\newblock {\em CGF}, 37(6):422--447, 2018.

\bibitem{Liang-ICBD18}
X.~Liang, S.~Di, D.~Tao, S.~Li, S.~Li, et~al.
\newblock Error-controlled lossy compression optimized for high compression ratios of scientific datasets.
\newblock In {\em Proc. BigData}, pages 438--447, 2018.

\bibitem{Liang-TBD22}
X.~Liang, K.~Zhao, S.~Di, S.~Li, R.~Underwood, et~al.
\newblock {SZ3}: A modular framework for composing prediction-based error-bounded lossy compressors.
\newblock {\em IEEE TBD}, 9(2):485--498, 2022.

\bibitem{Lin-ICLR20}
T.~Lin, S.~U. Stich, L.~Barba, D.~Dmitriev, and M.~Jaggi.
\newblock Dynamic model pruning with feedback.
\newblock In {\em Proc. ICLR}, 2020.

\bibitem{Lu-CGF21}
Y.~Lu, K.~Jiang, J.~A. Levine, and M.~Berger.
\newblock Compressive neural representations of volumetric scalar fields.
\newblock {\em CGF}, 40(3):135--146, 2021.

\bibitem{Martel-TOG21}
J.~N.~P. Martel, D.~B. Lindell, C.~Z. Lin, E.~R. Chan, M.~Monteiro, et~al.
\newblock {ACORN}: Adaptive coordinate networks for neural scene representation.
\newblock {\em ACM ToG}, 40(4):58:1--58:13, 2021.

\bibitem{Molchanov-CVPR19}
P.~Molchanov, A.~Mallya, S.~Tyree, I.~Frosio, and J.~Kautz.
\newblock Importance estimation for neural network pruning.
\newblock In {\em Proc. CVPR}, pages 11264--11272, 2019.

\bibitem{Muller-TOG22}
T.~M{\"u}ller, A.~Evans, C.~Schied, and A.~Keller.
\newblock Instant neural graphics primitives with a multiresolution hash encoding.
\newblock {\em ACM ToG}, 41(4):102:1--102:15, 2022.

\bibitem{Muraki-VIS92}
S.~Muraki.
\newblock Approximation and rendering of volume data using wavelet transforms.
\newblock In {\em Proc. Vis}, pages 21--22, 1992.

\bibitem{Rapp-TVCG20}
T.~Rapp, C.~Peters, and C.~Dachsbacher.
\newblock Void-and-cluster sampling of large scattered data and trajectories.
\newblock {\em IEEE TVCG}, 26(1):780--789, 2020.

\bibitem{Reiser-ICCV21}
C.~Reiser, S.~Peng, Y.~Liao, and A.~Geiger.
\newblock {KiloNeRF}: Speeding up neural radiance fields with thousands of tiny {MLPs}.
\newblock In {\em Proc. ICCV}, pages 14315--14325, 2021.

\bibitem{saragadam-ECCV22}
V.~Saragadam, J.~Tan, G.~Balakrishnan, R.~G. Baraniuk, and A.~Veeraraghavan.
\newblock {MINER}: Multiscale implicit neural representation.
\newblock In {\em Proc. ECCV}, pages 318--333, 2022.

\bibitem{sitzmann-NIPS20}
V.~Sitzmann, J.~Martel, A.~Bergman, D.~Lindell, and G.~Wetzstein.
\newblock Implicit neural representations with periodic activation functions.
\newblock In {\em Proc. NeurIPS}, pages 7462--7473, 2020.

\bibitem{Soler-PVIS18}
M.~Soler, M.~Plainchault, B.~Conche, and J.~Tierny.
\newblock Topologically controlled lossy compression.
\newblock In {\em Proc. PacificVis}, pages 46--55, 2018.

\bibitem{Suter-CGF13}
S.~K. Suter, M.~Makhynia, and R.~Pajarola.
\newblock {TAMRESH}--tensor approximation multiresolution hierarchy for interactive volume visualization.
\newblock {\em CGF}, 32(3):151--160, 2013.

\bibitem{takikawa-CVPR21}
T.~Takikawa, J.~Litalien, K.~Yin, K.~Kreis, C.~Loop, et~al.
\newblock Neural geometric level of detail: Real-time rendering with implicit {3D} shapes.
\newblock In {\em Proc. CVPR}, pages 11358--11367, 2021.

\bibitem{Tang-CG}
K.~Tang and C.~Wang.
\newblock {STSR-INR}: Spatiotemporal super-resolution for time-varying multivariate volumetric data via implicit neural representation.
\newblock {\em C\&G}, 2024.
\newblock Accepted.

\bibitem{Wang-TVCG23}
C.~Wang and J.~Han.
\newblock {DL4SciVis}: A state-of-the-art survey on deep learning for scientific visualization.
\newblock {\em IEEE TVCG}, 29(8):3714--3733, 2023.

\bibitem{Wang-TVCG08}
C.~Wang and K.-L. Ma.
\newblock A statistical approach to volume data quality assessment.
\newblock {\em IEEE TVCG}, 14(3):590--602, 2008.

\bibitem{Wang-PVIS17}
K.-C. Wang, K.~Lu, T.-H. Wei, N.~Shareef, and H.-W. Shen.
\newblock Statistical visualization and analysis of large data using a value-based spatial distribution.
\newblock In {\em Proc. PacificVis}, pages 161--170, 2017.

\bibitem{Weiss-CGF22}
S.~Weiss, P.~Herm{\"u}ller, and R.~Westermann.
\newblock Fast neural representations for direct volume rendering.
\newblock {\em CGF}, 41(6):196--211, 2022.

\bibitem{Wen-NIPS16}
W.~Wen, C.~Wu, Y.~Wang, Y.~Chen, and H.~Li.
\newblock Learning structured sparsity in deep neural networks.
\newblock In {\em Proc. NeurIPS}, pages 2074--2082, 2016.

\bibitem{Wu-arXiv23}
Q.~Wu, D.~Bauer, Y.~Chen, and K.-L. Ma.
\newblock {HyperINR}: A fast and predictive hypernetwork for implicit neural representations via knowledge distillation.
\newblock {\em arXiv preprint arXiv:2304.04188}, 2023.

\bibitem{Wu-TVCG}
Q.~Wu, D.~Bauer, M.~J. Doyle, and K.-L. Ma.
\newblock Interactive volume visualization via multi-resolution hash encoding based neural representation.
\newblock {\em IEEE TVCG}, 2023.
\newblock Accepted.

\bibitem{Wurster-TVCG}
S.~W. Wurster, H.~Guo, H.-W. Shen, T.~Peterka, and J.~Xu.
\newblock Deep hierarchical super resolution for scientific data.
\newblock {\em IEEE TVCG}, 29(12):5483--5495, 2023.

\bibitem{Wurster-VIS23}
S.~W. Wurster, T.~Xiong, H.-W. Shen, H.~Guo, and T.~Peterka.
\newblock Adaptively placed multi-grid scene representation networks for large-scale data visualization.
\newblock {\em IEEE TVCG}, 30(1):965--974, 2024.

\bibitem{Xie-CGF22}
Y.~Xie, T.~Takikawa, S.~Saito, O.~Litany, S.~Yan, et~al.
\newblock Neural fields in visual computing and beyond.
\newblock {\em CGF}, 41(2):641--676, 2022.

\bibitem{Ye-arXiv18}
S.~Ye, T.~Zhang, K.~Zhang, J.~Li, J.~Xie, et~al.
\newblock A unified framework of {DNN} weight pruning and weight clustering/quantization using {ADMM}.
\newblock {\em arXiv preprint arXiv:1811.01907}, 2018.

\bibitem{Yeo-TVCG95}
B.-L. Yeo and B.~Liu.
\newblock Volume rendering of {DCT}-based compressed {3D} scalar data.
\newblock {\em IEEE TVCG}, 1(1):29--43, 1995.

\bibitem{Zhang-CVPR18}
R.~Zhang, P.~Isola, A.~A. Efros, E.~Shechtman, and O.~Wang.
\newblock The unreasonable effectiveness of deep features as a perceptual metric.
\newblock In {\em Proc. CVPR}, pages 586--595, 2018.

\end{thebibliography}

\newpage

\section*{Appendix}
\setcounter{section}{0}
\setcounter{figure}{0}
\setcounter{table}{0}
\setcounter{algocf}{0}

\begin{figure*}[htb]
	\begin{center}
		$\begin{array}{c@{\hspace{0.02in}}c@{\hspace{0.02in}}c@{\hspace{0.02in}}c}
				\includegraphics[height=1.0in]{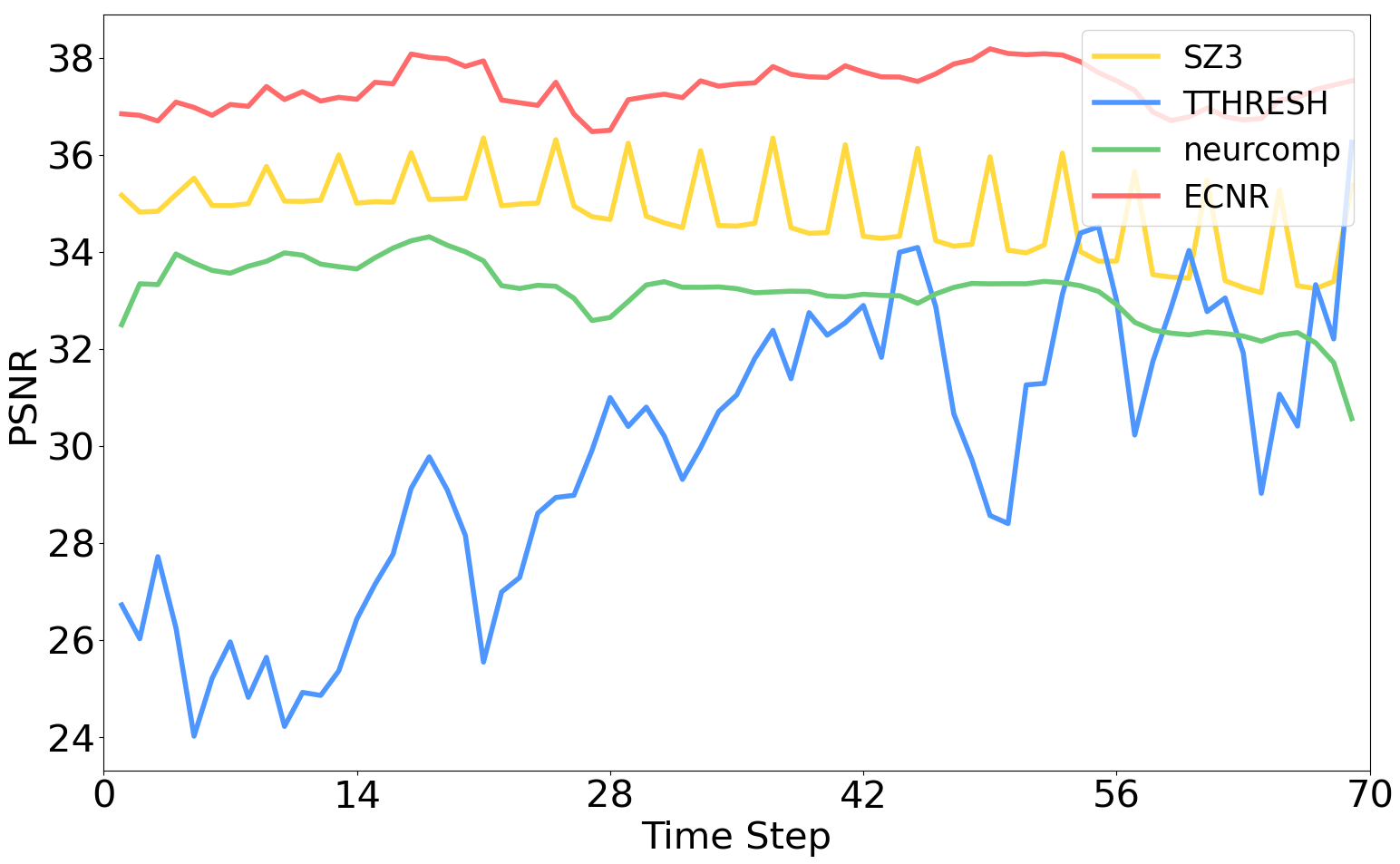} &
				\includegraphics[height=1.0in]{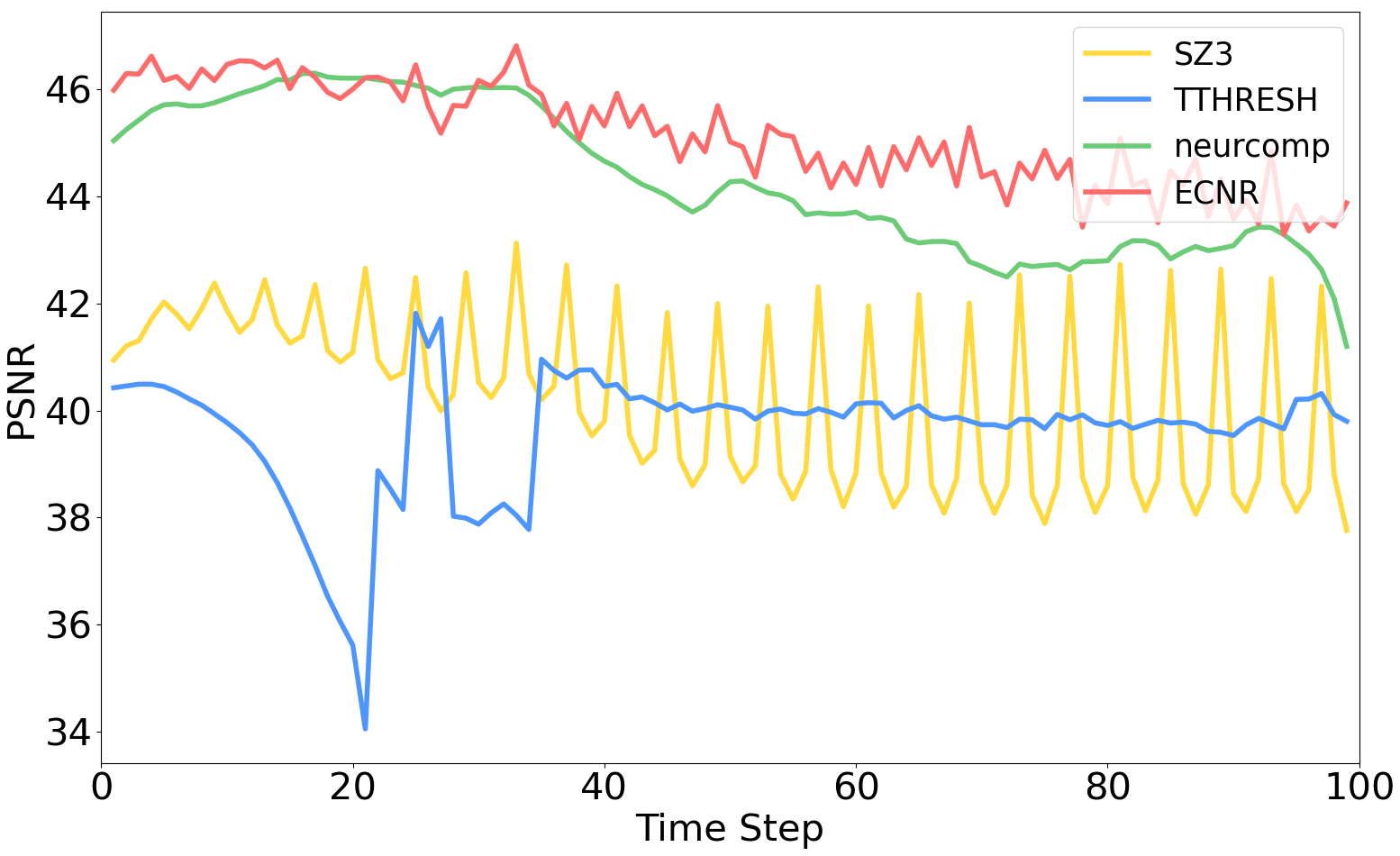} &
				\includegraphics[height=1.0in]{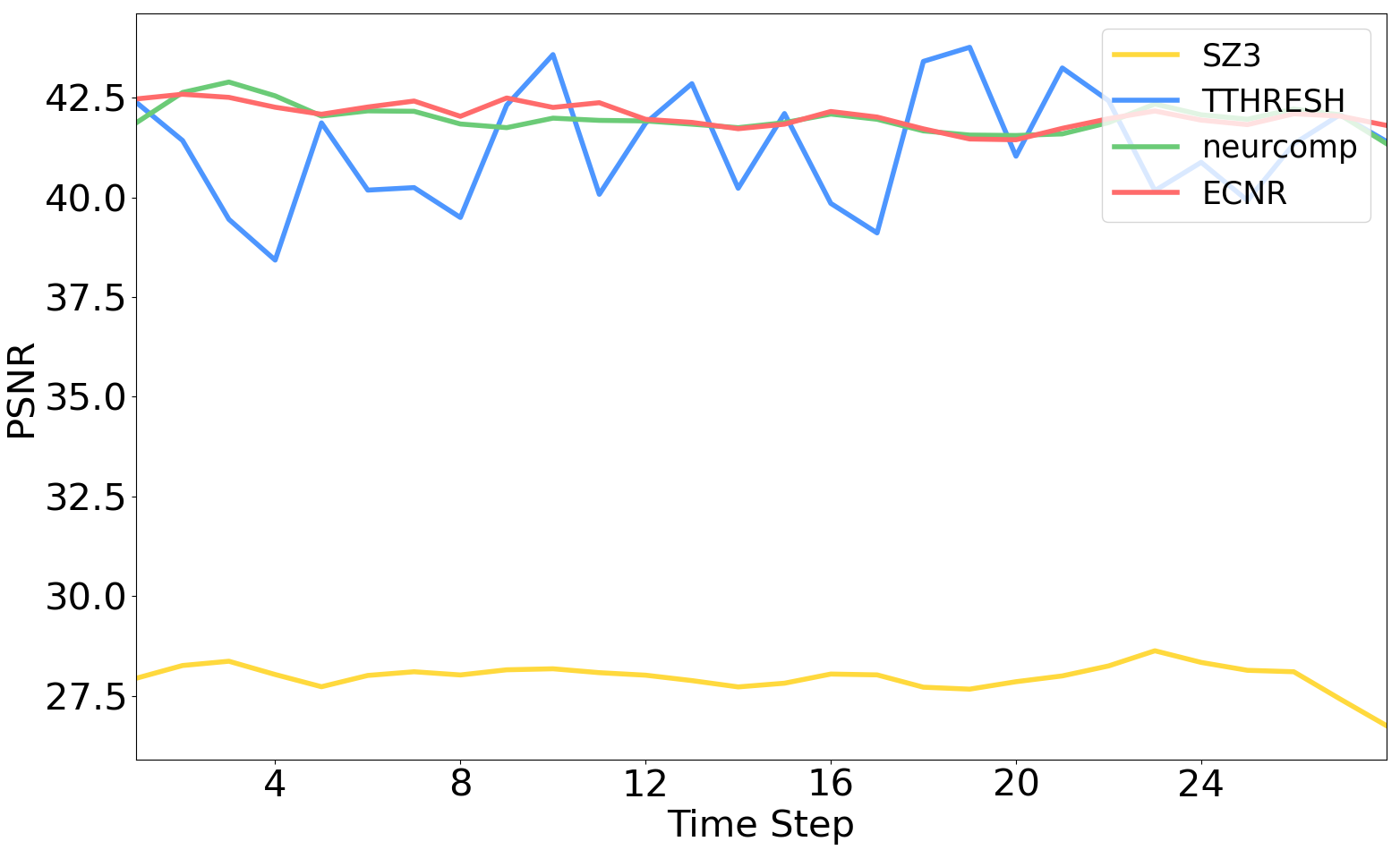} &
				\includegraphics[height=1.0in]{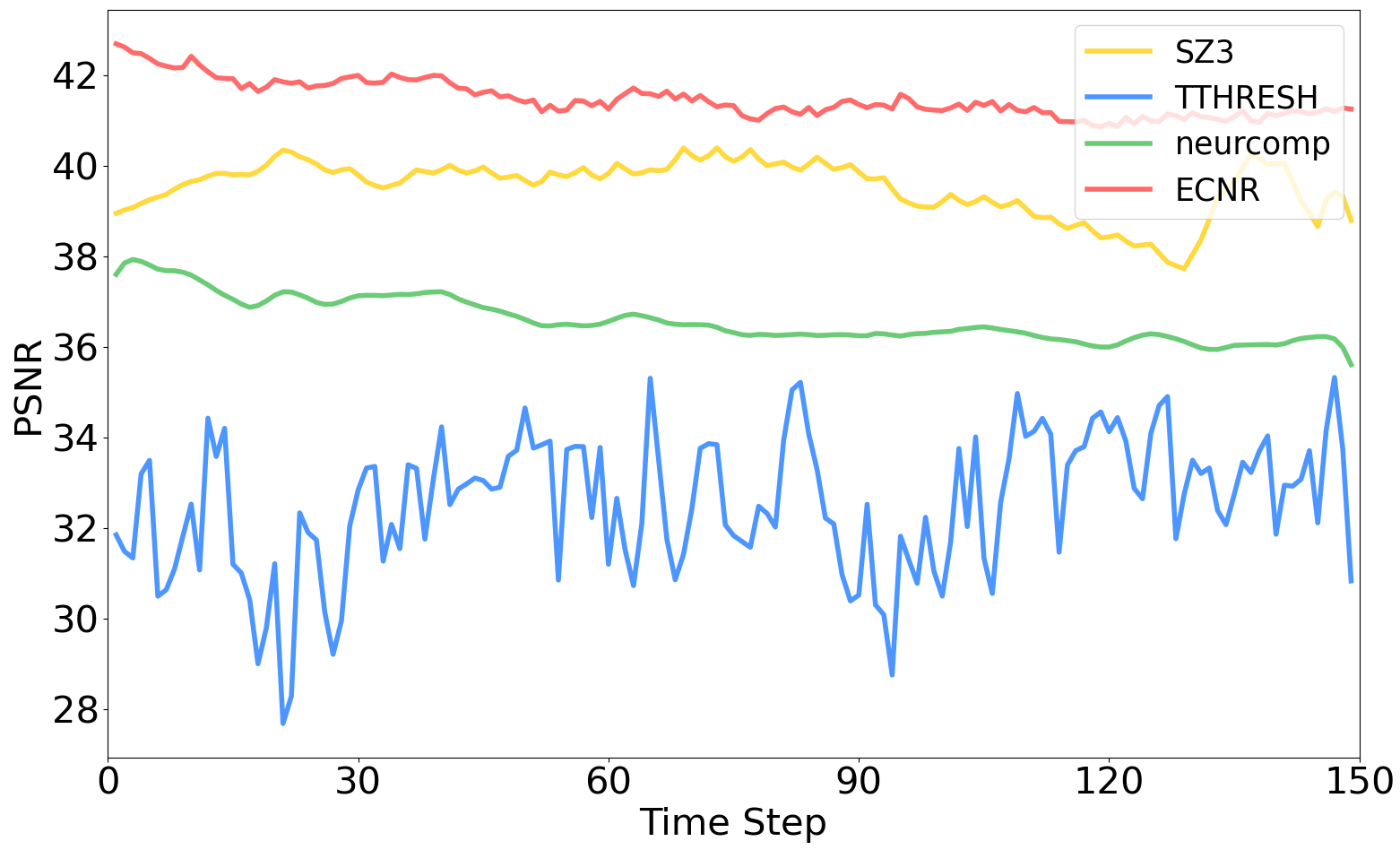} \\
				\includegraphics[height=1.0in]{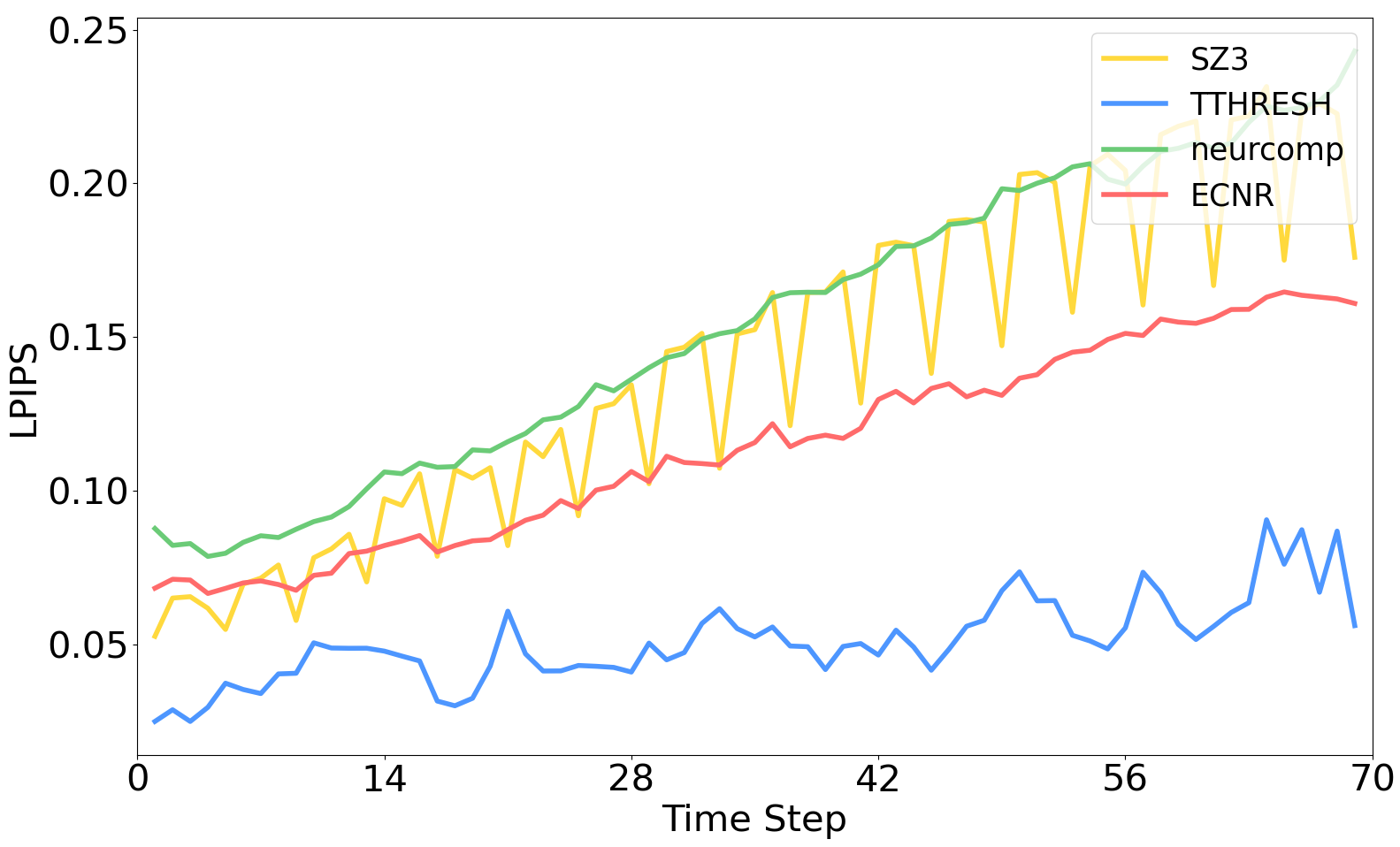} &
				\includegraphics[height=1.0in]{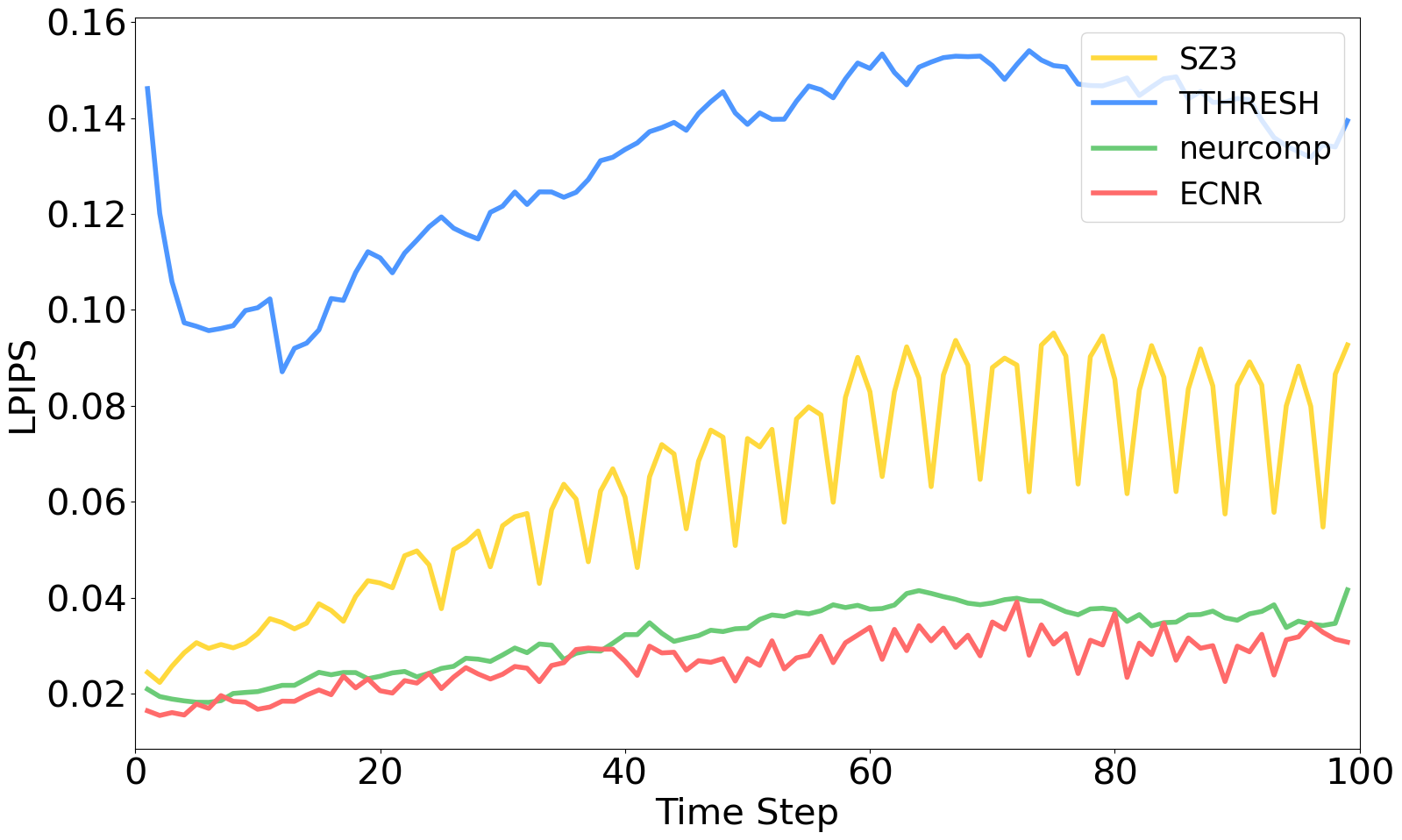} &
				\includegraphics[height=1.0in]{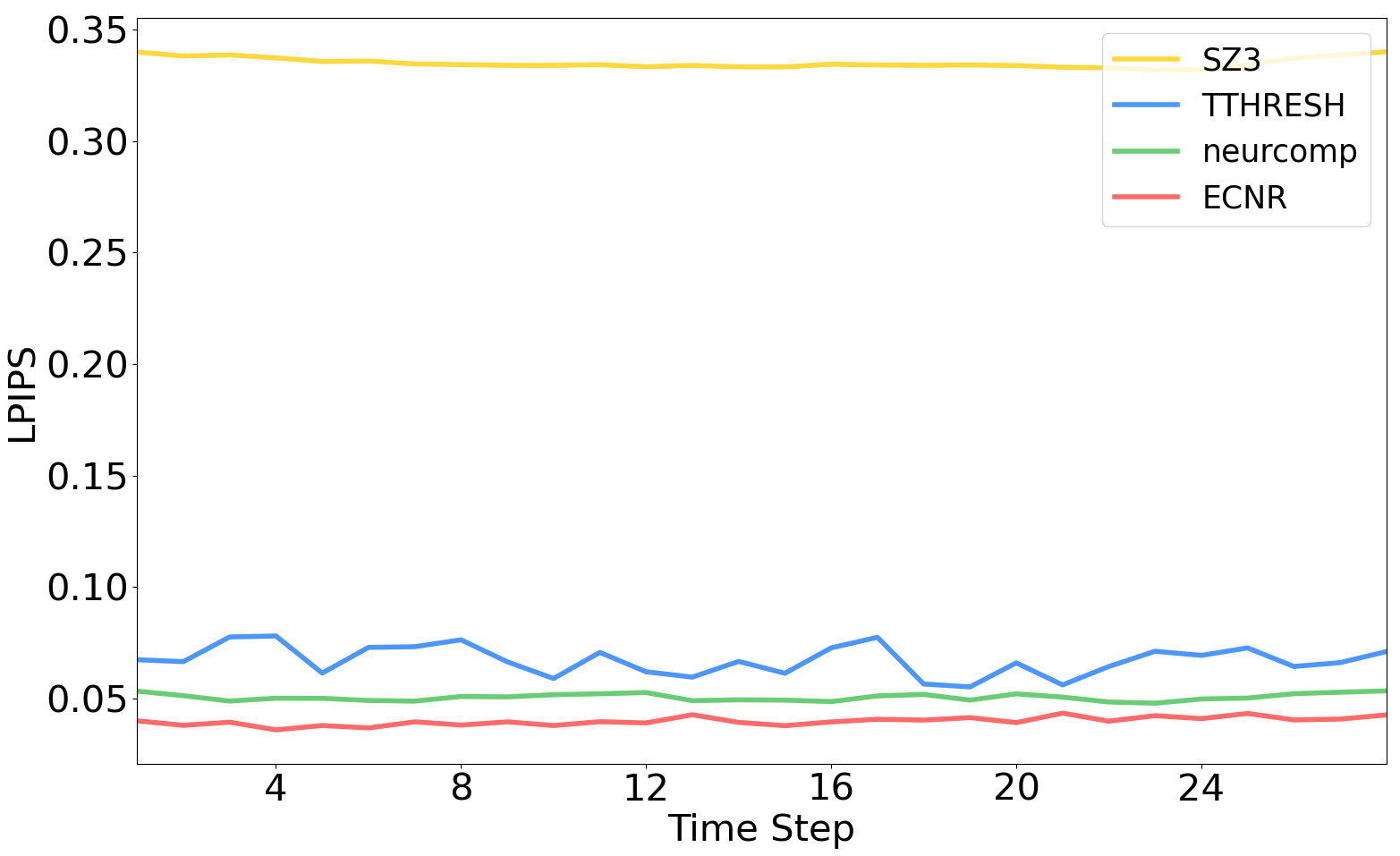} &
				\includegraphics[height=1.0in]{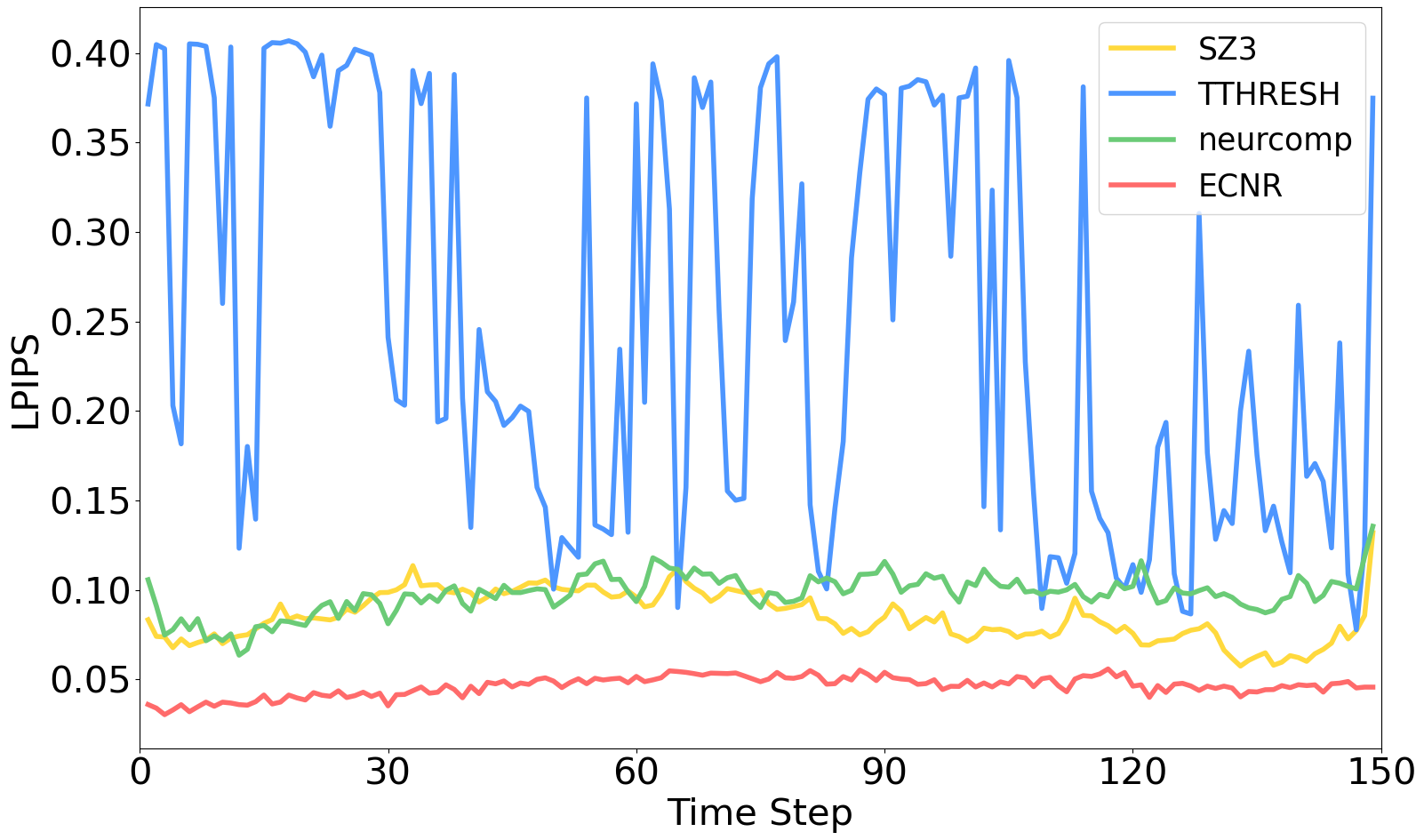} \\
				\includegraphics[height=1.0in]{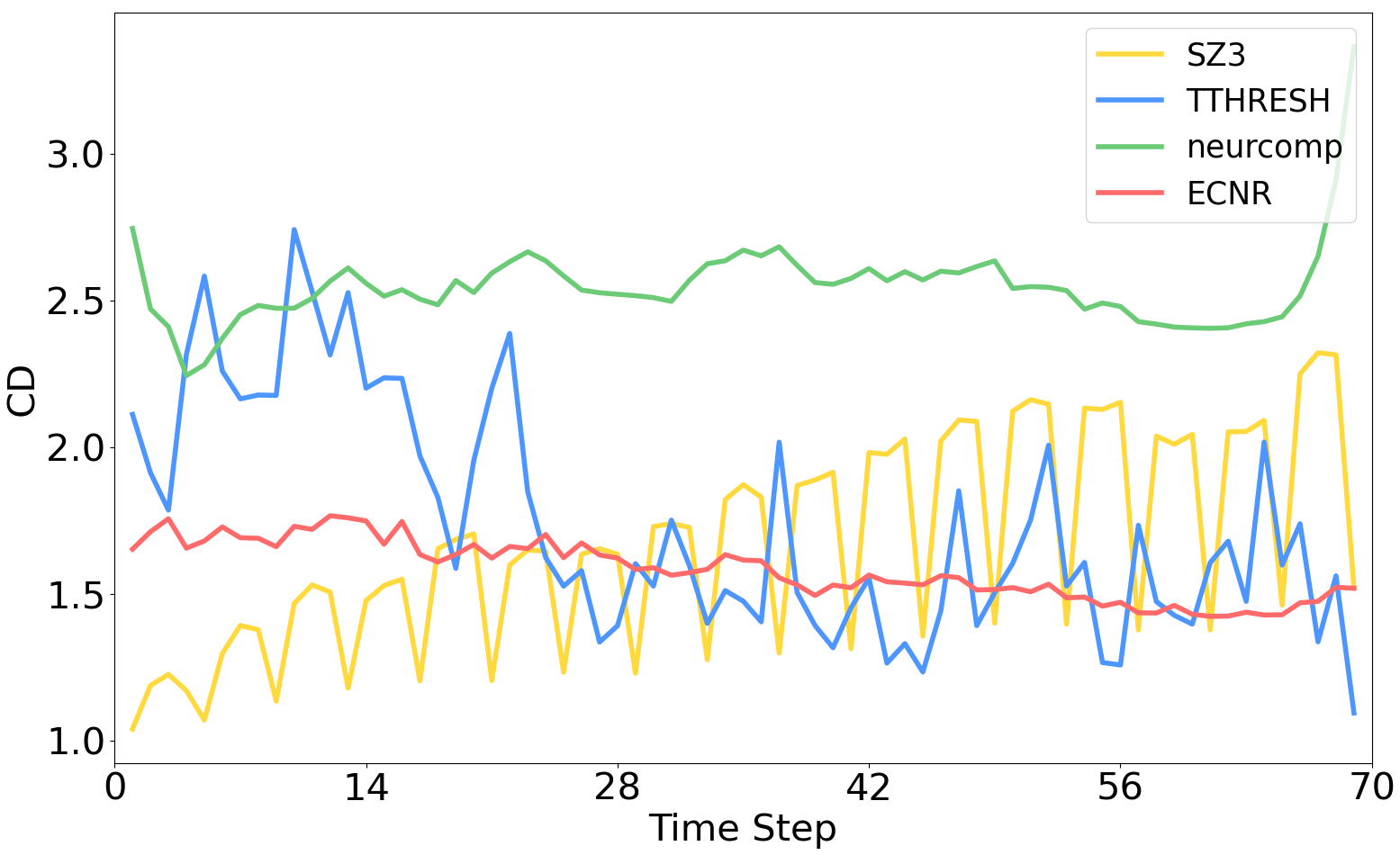} &
				\includegraphics[height=1.0in]{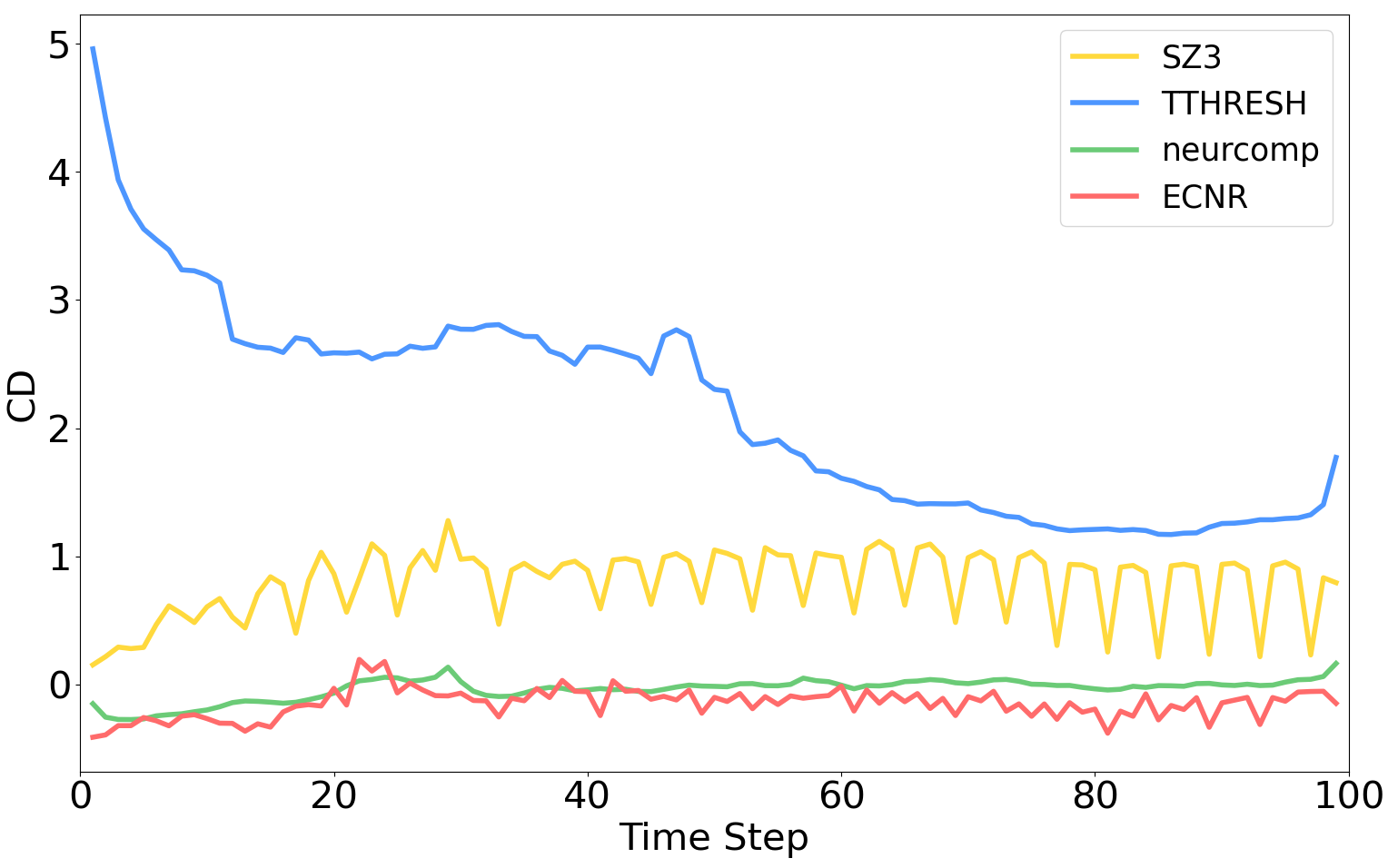} &
				\includegraphics[height=1.0in]{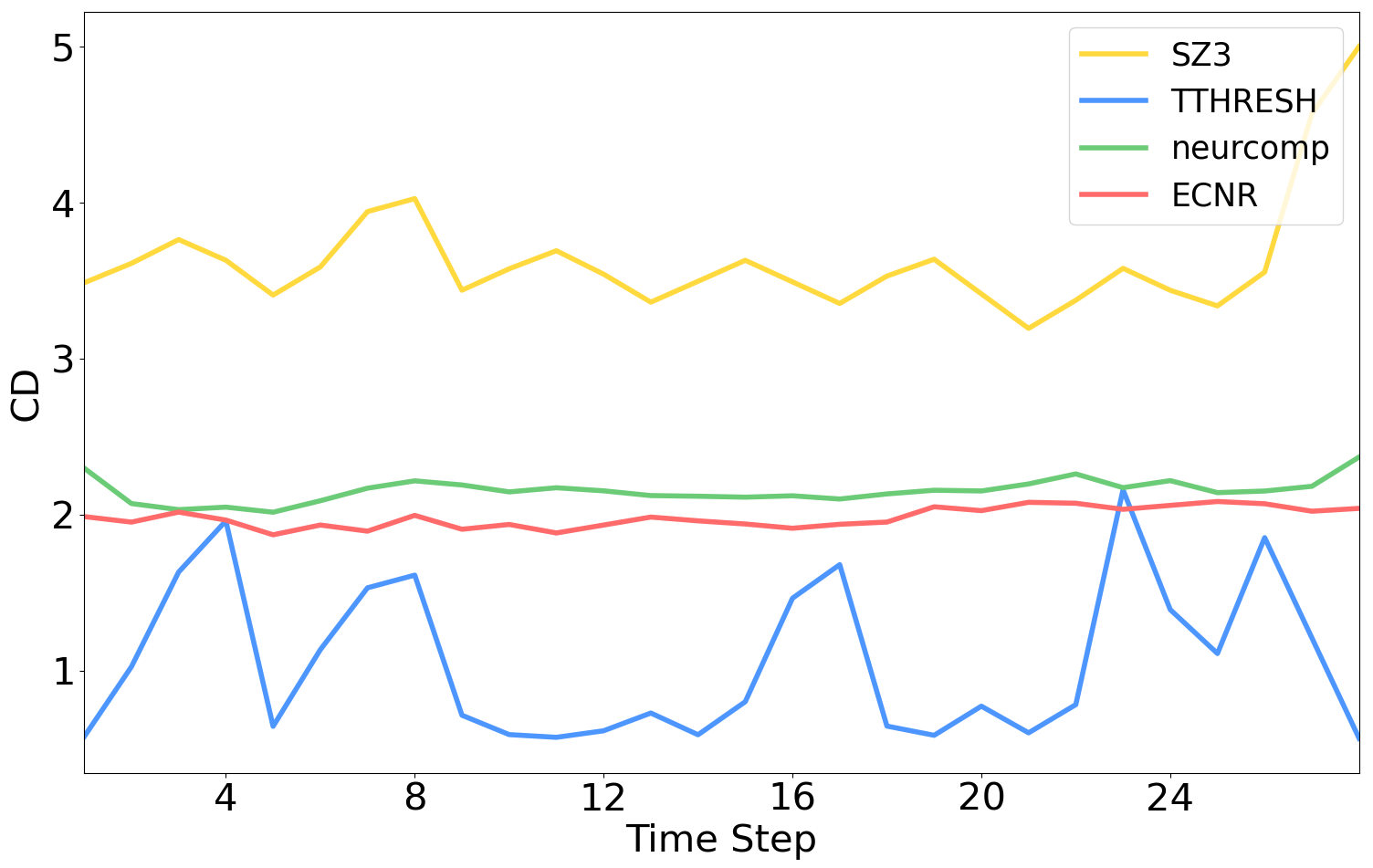} &
				\includegraphics[height=1.0in]{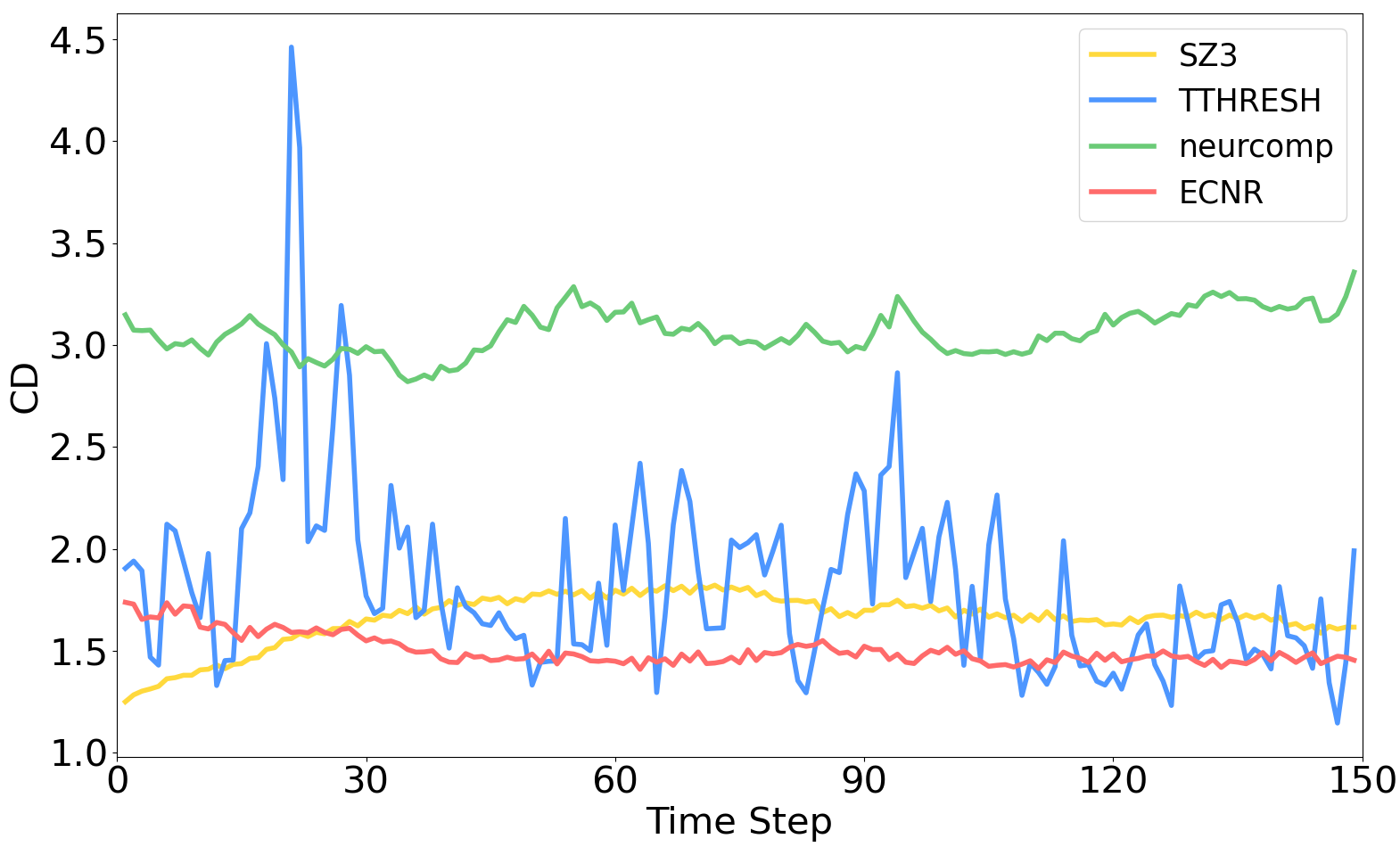} \\
				\mbox{\footnotesize (a) combustion}
				& \mbox{\footnotesize (b) half-cylinder}
				& \mbox{\footnotesize (c) solar plume}
				& \mbox{\footnotesize (d) Tangaroa}
			\end{array}$
	\end{center}
	\vspace{-.25in}
	\caption{Top to bottom: PSNR (dB), LPIPS, and CD values over timesteps. For CD, the chosen isovalues are reported in Table 2 in the paper.}
	\label{fig:BaselinesPlot}
\end{figure*}

\section{Implementation Details}

For easy reproducing ECNR, we list the pseudocode of one linear layer of ECNR in Algorithm~\ref{alg:LinearLayer} and provide the details of training MLPs in parallel. 
Let $m$ denote the number of MLPs, $\mathcal{F}_{\IN}$ and $\mathcal{F}_{\OUT}$ are the input and output feature dimensions of this linear layer. 
Unlike the conventional linear layer representing its weight and bias by two 2D matrices, the linear layer of ECNR is constructed by two 3D tensors where the extra dimension denotes which MLP to use. 
In the encoding process, we train all MLPs simultaneously. 
Thus MLPs\_indices would be an array with values in $[0,\ m-1]$. 
For ECNR with only one layer, its input would be a tensor with shape $(m,\  N,\ \mathcal{F}_{\IN})$, where $N$ is the number of voxels in one block. bmm$(\cdot)$ is a PyTorch function that computes matrix-matrix product. 
The linear layer can output a tensor with shape $(m,\  N,\ \mathcal{F}_{\OUT})$. 
Since in PyTorch, all matrix multiplications in the feedforward process can be handled in parallel, parallel training of multiple MLPs can be accomplished without resorting to complex coding beyond the PyTorch implementation.

\begin{algorithm}[htb]

\SetAlgoLined
class ECNRLinearLayer():\\
\Indp
	def init(self,\ $\mathcal{F}_{\IN}$,\ $\mathcal{F}_{\OUT}$,\ $m$):\\
		\Indp
		\tcc{ initialize weight and bias }
    	 self.weight = Parameter($m$,\ $\mathcal{F}_{\IN}$,\ $\mathcal{F}_{\OUT}$)\\
    	 self.bias = Parameter($m$,\ 1,\ $\mathcal{F}_{\OUT}$)\\
		\texttt{\\}
 		\Indm
\Indm
\Indp
	def forward(self,\ input,\ MLPs\_indices):\\
		\Indp
		\tcc{ select MLPs parameters for forwarding}
		 MLPs\_weight = self.weight[MLPs\_indices, $\cdots$]\\
		 MLPs\_bias = self.bias[MLPs\_indices, $\cdots$]\\
		 return bmm(input,\ MLPs\_weight) + MLPs\_bias\\
		\Indm
\Indm    	 
\caption{ECNR linear layer in PyTorch-like format}
\label{alg:LinearLayer}
\end{algorithm}

\section{Additional Results}

{\bf Quantitative performance across timesteps.} 
In addition to the average performance results reported in Table 2 in the paper, we plot PSNR, LPIPS, and CD curves over timesteps, as shown in Figure~\ref{fig:BaselinesPlot}. 
For the combustion dataset, the increasing LPIPS values over time for SZ3, neurcomp, and ECNR are due to the increasingly turbulent nature of the combustion process. As the data gets more complex, the volume rendering results show larger differences with respect to the GT. 
For the half-cylinder dataset, the low performance on PSNR and CD of TTHRESH in the beginning 21 timesteps is due to data sparsity. 

{\bf Compression file storage.}
During encoding, extra space is necessary to store the metadata, including the volume dimensions and network configurations. 
We used an array to map the MLPs to their assigned blocks in the multiscale block partition and assignment step. 
We encapsulated this information and metadata in the header of our output file. 
Besides the header, we stored the block latent codes with floating-point precision. 
After applying \hot{block-guided pruning} and network quantization, we need to save three components: a binary mask that indicates unpruned parameters, floating-point precision shared parameters, and the parameter indices (with precision $\beta$ bits) that map the unpruned parameters (after \hot{block-guided pruning}) to the shared parameters (after network quantization). 
Our output file contains all the necessary information for decoding. 

\begin{figure}[htb]
	\begin{center}
		$\begin{array}{c@{\hspace{0.02in}}c@{\hspace{0.02in}}c}
				\includegraphics[height=1.125in]{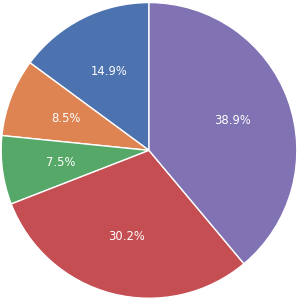}&
				\includegraphics[height=1.125in]{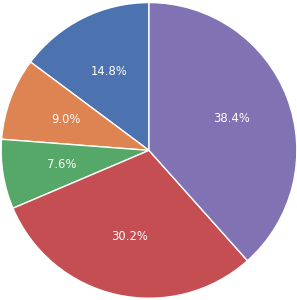} \\
				\mbox{\footnotesize (a) combustion}   & \mbox{\footnotesize (b) half-cylinder} \\
				\includegraphics[height=1.125in]{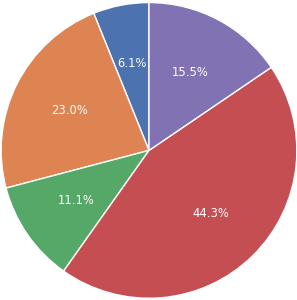}&
				\includegraphics[height=1.125in]{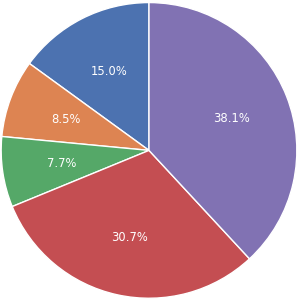}&
				\includegraphics[height=0.525in]{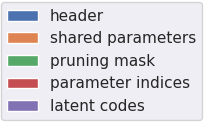}\\
				\mbox{\footnotesize (c) solar plume}   & \mbox{\footnotesize (d) Tangaroa} & \\ 
			\end{array}$
	\end{center}	
		\vspace{-.25in}
	\caption{ECNR compression file storage percentages for different components.}
	\label{fig:comp-storage-ratio}
\end{figure}

For each component in the output file, we show its occupied storage percentage in Figure~\ref{fig:comp-storage-ratio}. 
For the combustion, half-cylinder, and Tangaroa datasets, latent codes and parameter indices occupy most storage. 
Rather than using 32-bit floating-point precision for all parameters, combining a subset of floating-point precision shared parameters with $\beta$-bit parameter indices offers a more concise representation. 
Therefore, the large storage percentage of parameter indices suggests the effectivity of deep compression. 
Note that latent codes in the solar plume dataset take less space than others. 
This can be attributed to its relatively larger block dimension and fewer timesteps, leading to fewer blocks for encoding. 

\begin{table}[htb]
	\caption{\hot{PSNR (dB), LPIPS, and CD values, as well as encoding time (ET) and decoding time (DT)}.}
	\vspace{-0.1in}
	\centering
	\hot{
	\resizebox{\columnwidth}{!}{
		\begin{tabular}{c|c|lll|ll}
			dataset  & method   & PSNR$\uparrow$ & LPIPS$\downarrow$ & CD$\downarrow$ & ET$\downarrow$ & DT$\downarrow$ \\
			\hline	  &SZ3    &35.21 &\textbf{0.017} &\textbf{1.13} &\textbf{2.4s} &3.7s \\
					  &TTHRESH &38.49 &0.036 &2.87 &1.4m &15.7s \\
			asteroids &MHT &\textbf{40.46} &0.029 &1.85 &1.7m &\textbf{1.1s}  \\
		   ($v = 0.0$) &fV-SRN &37.49 &0.047 &5.66 &1.6m &2.1s  \\
			 		  &APMGSRN &38.29 &0.043 &4.46 &1.2m &1.8s \\
					  &neurcomp &36.78 &0.054 &8.67 &54.4m &23.8s  \\
			  		  &ECNR &38.85 &0.037 &3.72 &2.9m &3.1s \\
			 \hline 
					    &SZ3     &36.48 &0.261 &2.06 &\textbf{1.8s} &2.3s \\
					    &TTHRESH &43.73 &\textbf{0.221} &\textbf{1.52} &47.1s &7.5s \\
			supernova-1 &MHT &43.56 &0.229 &2.35 &1.7m &\textbf{0.6s}  \\
			($v = 0.0$) &fV-SRN  &40.38 &0.270 &3.42 &1.6m &1.2s  \\
			   			&APMGSRN &41.23 &0.262  &3.02  &1.2m &1.1s             \\
			  		    &neurcomp &43.23 &0.227 &2.17 &33.1m &15.2s  \\
			 			& ECNR &\textbf{44.08} &0.228  &1.77 &5.7m &2.1s \\
		\end{tabular}
	}}
	\label{tab:comp-single-timestep}
\end{table}
\begin{figure*}[htb]
	\begin{center}
		$\begin{array}{c@{\hspace{0.0125in}}c@{\hspace{0.0125in}}c@{\hspace{0.0125in}}c@{\hspace{0.0125in}}c@{\hspace{0.0125in}}c@{\hspace{0.0125in}}c@{\hspace{0.0125in}}c}
				\includegraphics[height=0.665in]{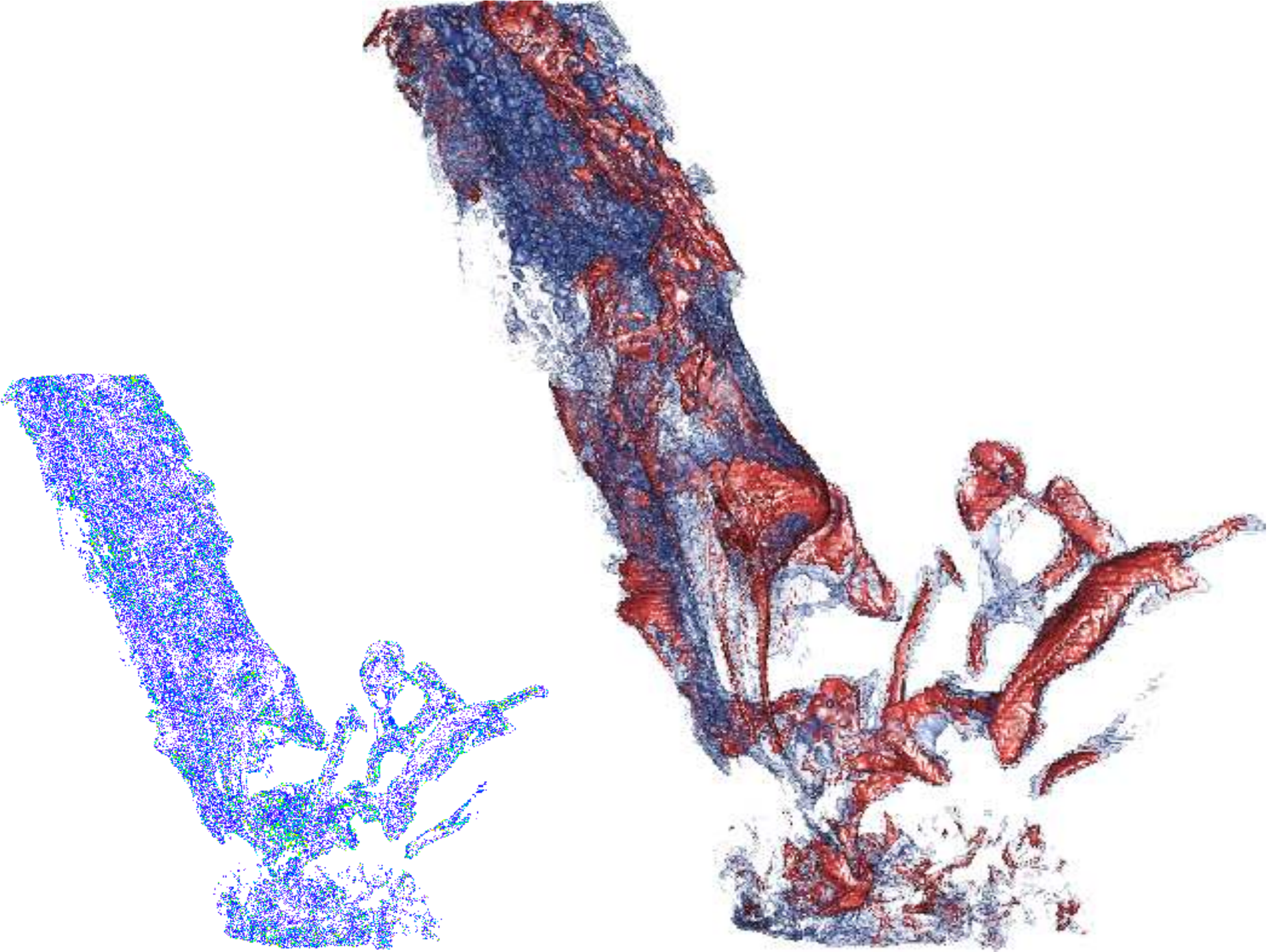} &
				\includegraphics[height=0.665in]{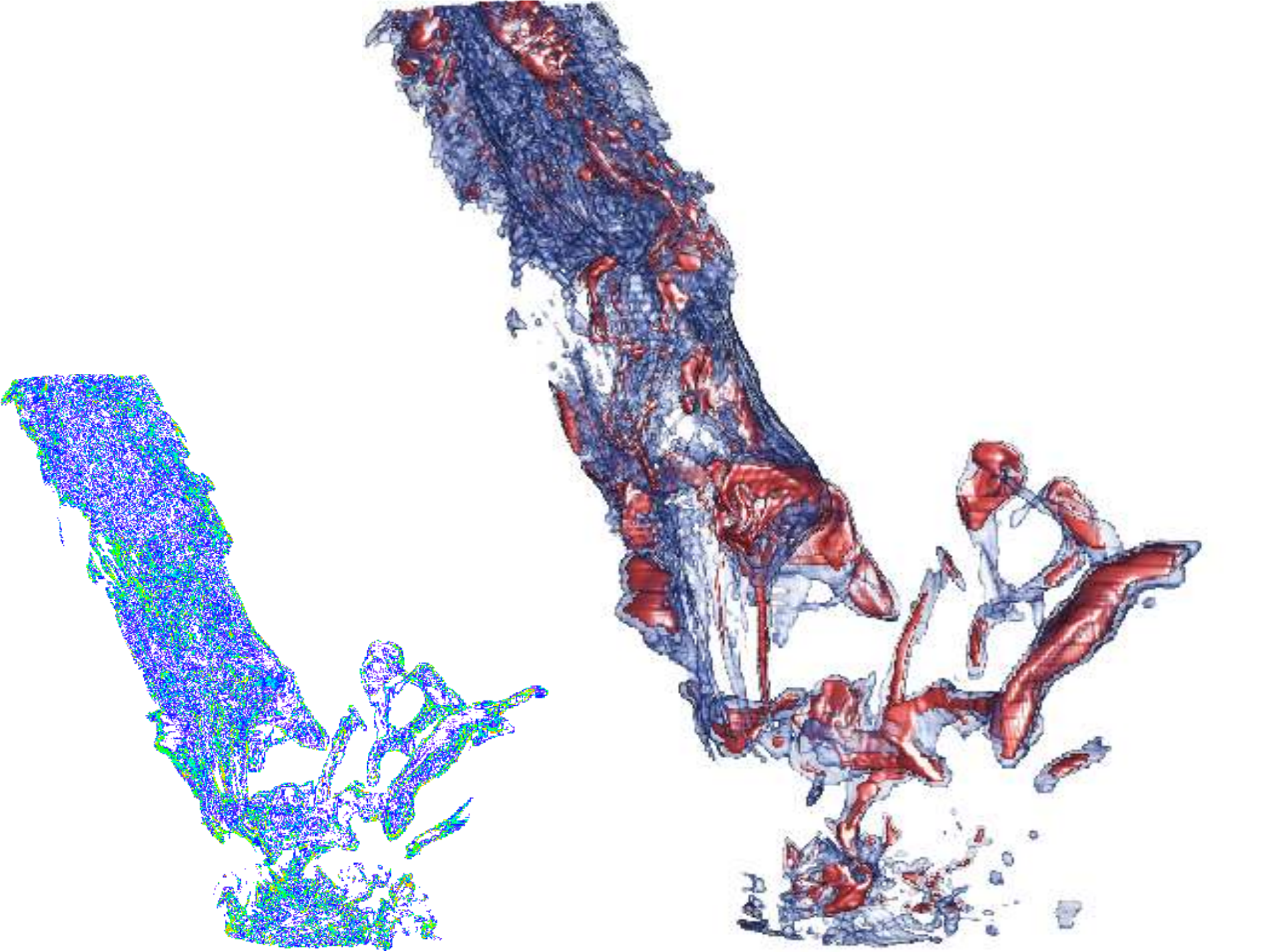} &
				\includegraphics[height=0.665in]{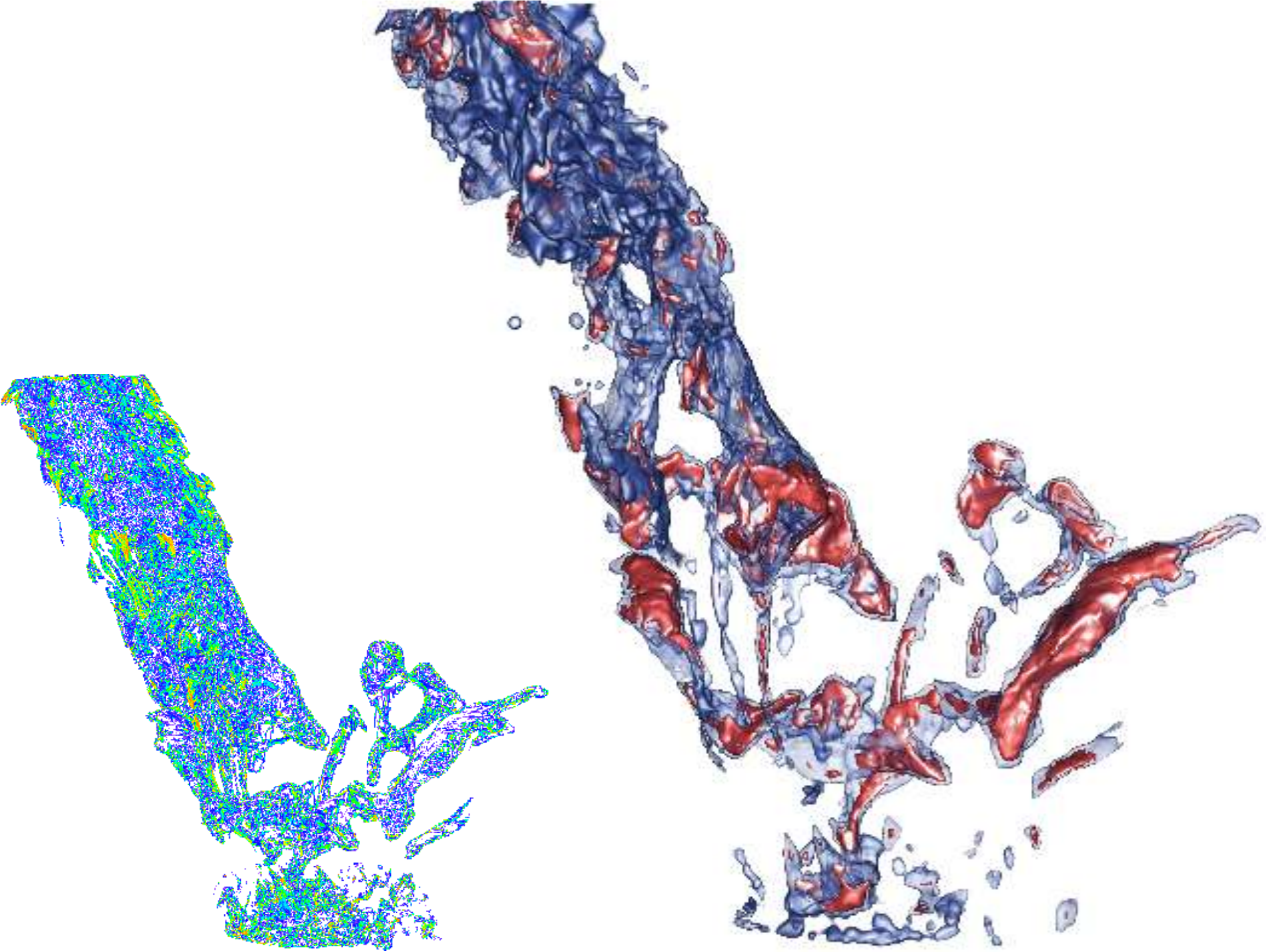} &
				\includegraphics[height=0.665in]{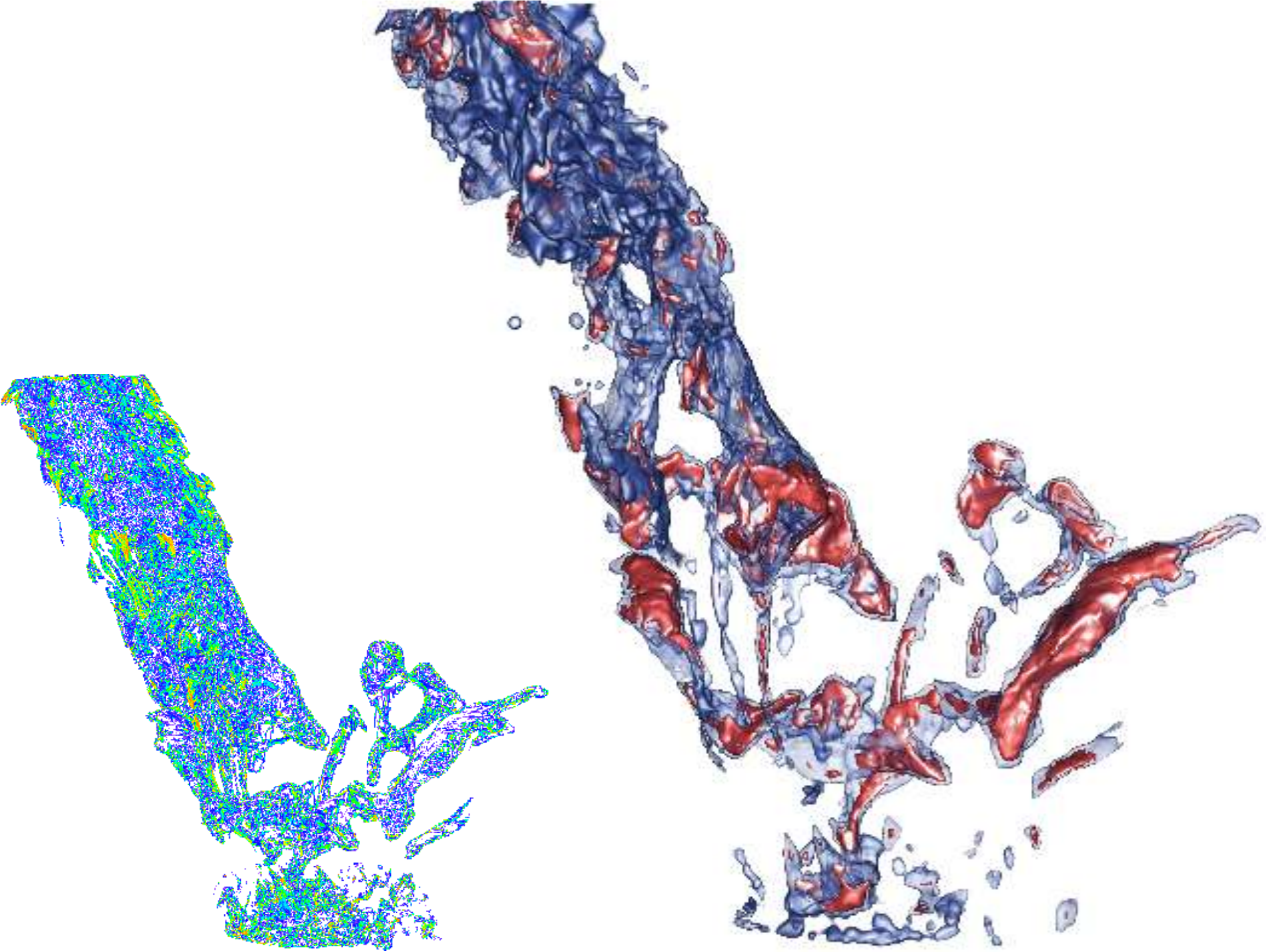} &
				\includegraphics[height=0.665in]{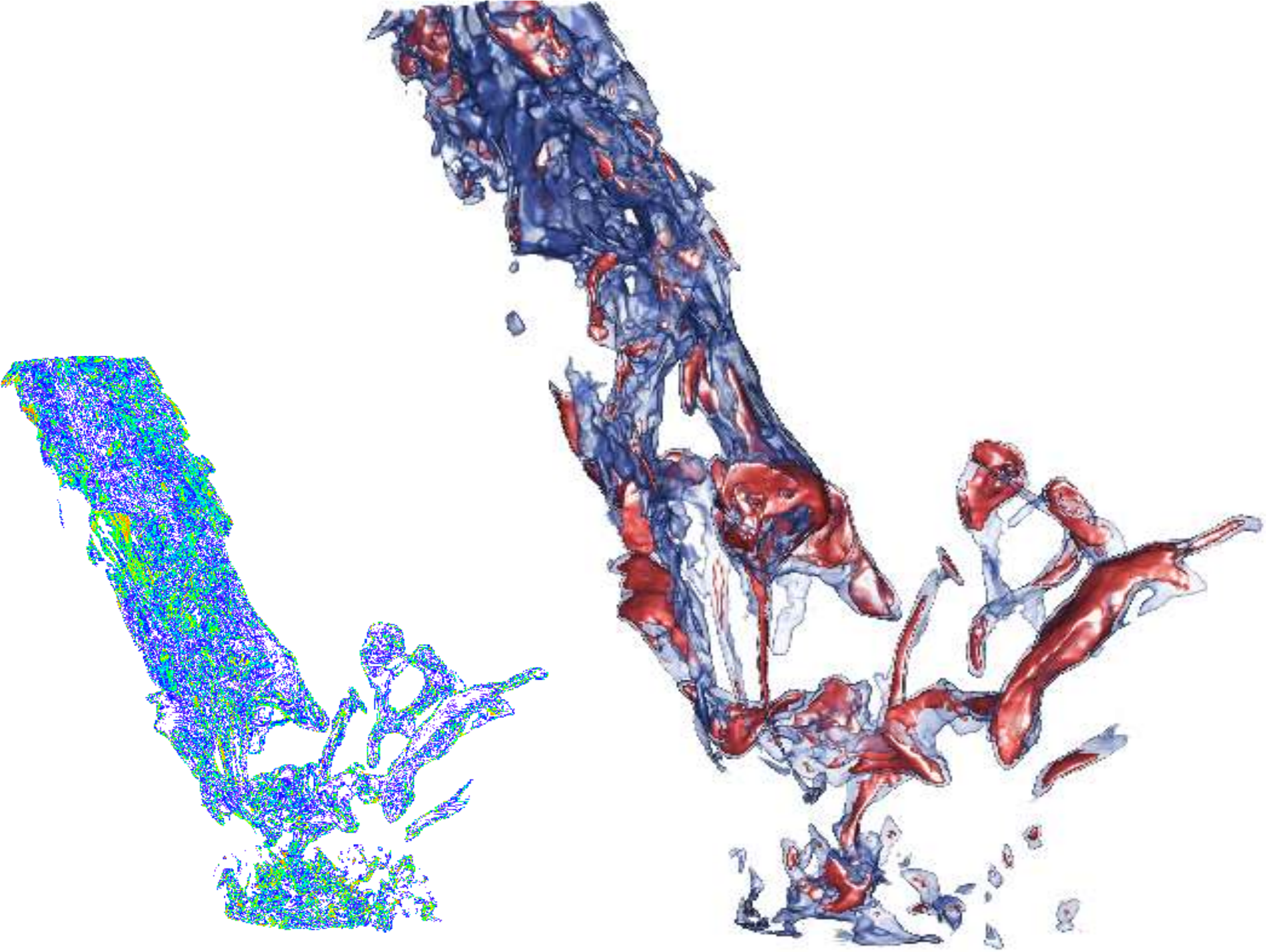} &
				\includegraphics[height=0.665in]{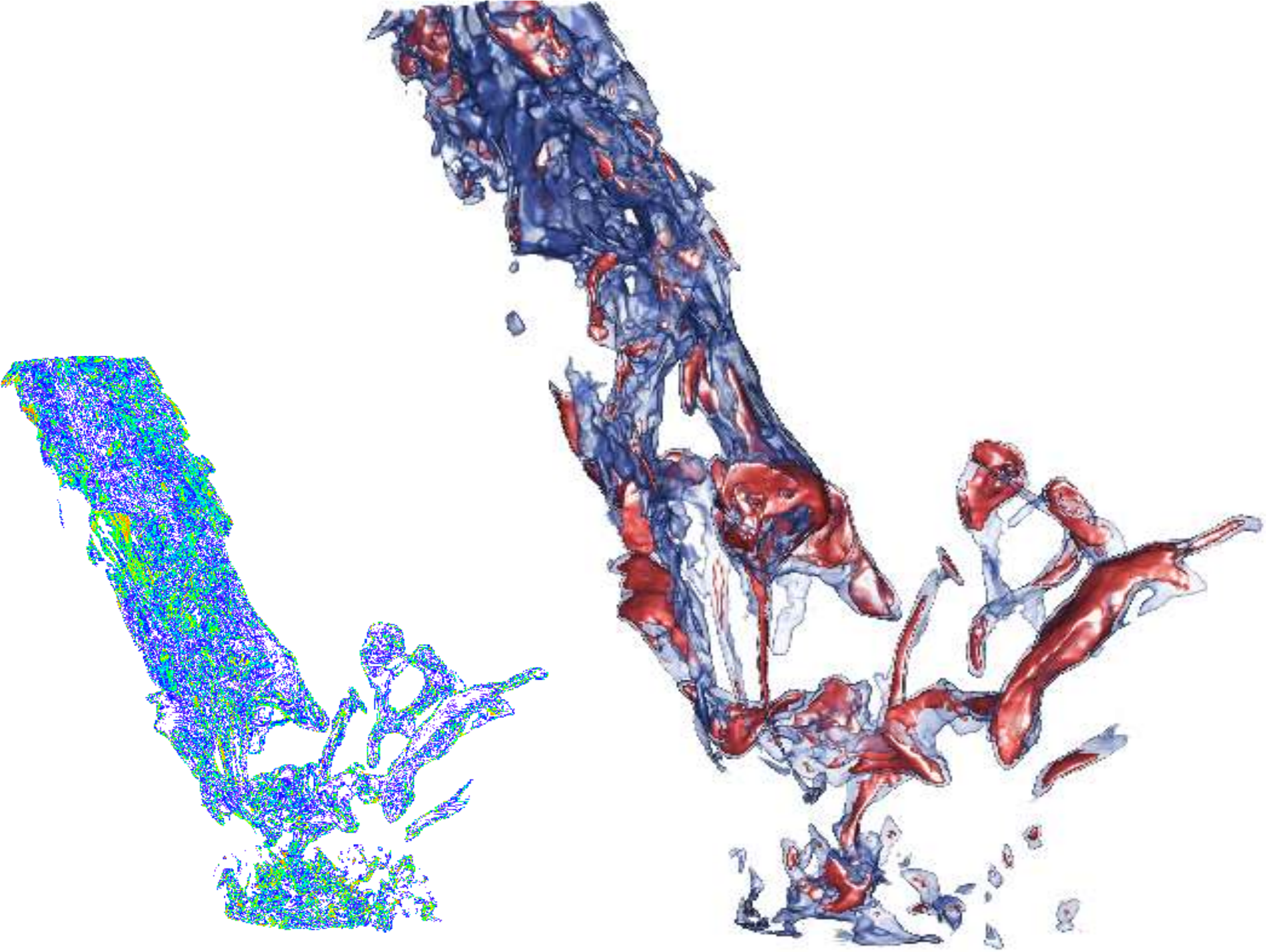} &
				\includegraphics[height=0.665in]{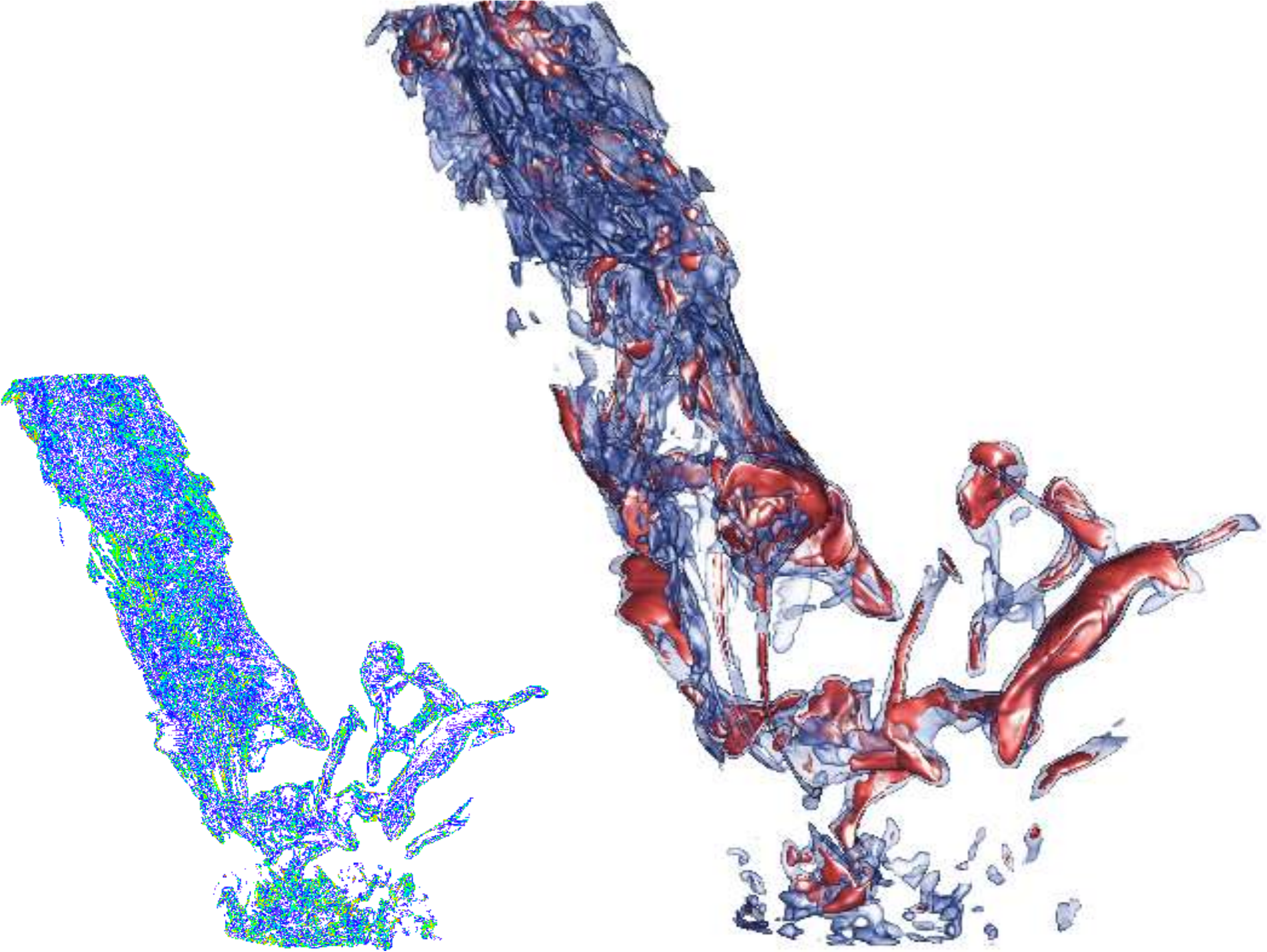} &
				\includegraphics[height=0.665in]{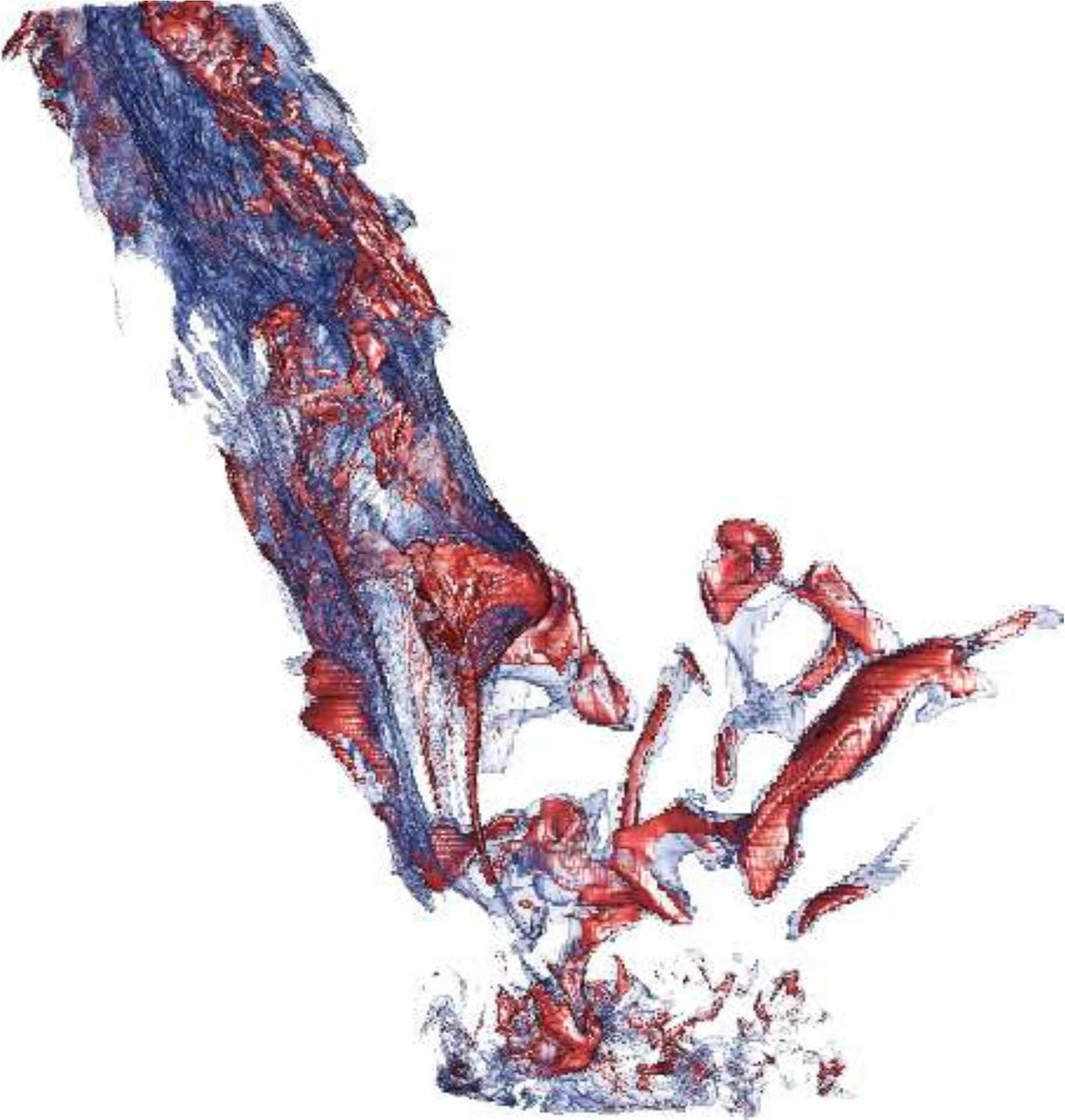}    \\
				\includegraphics[height=0.7in]{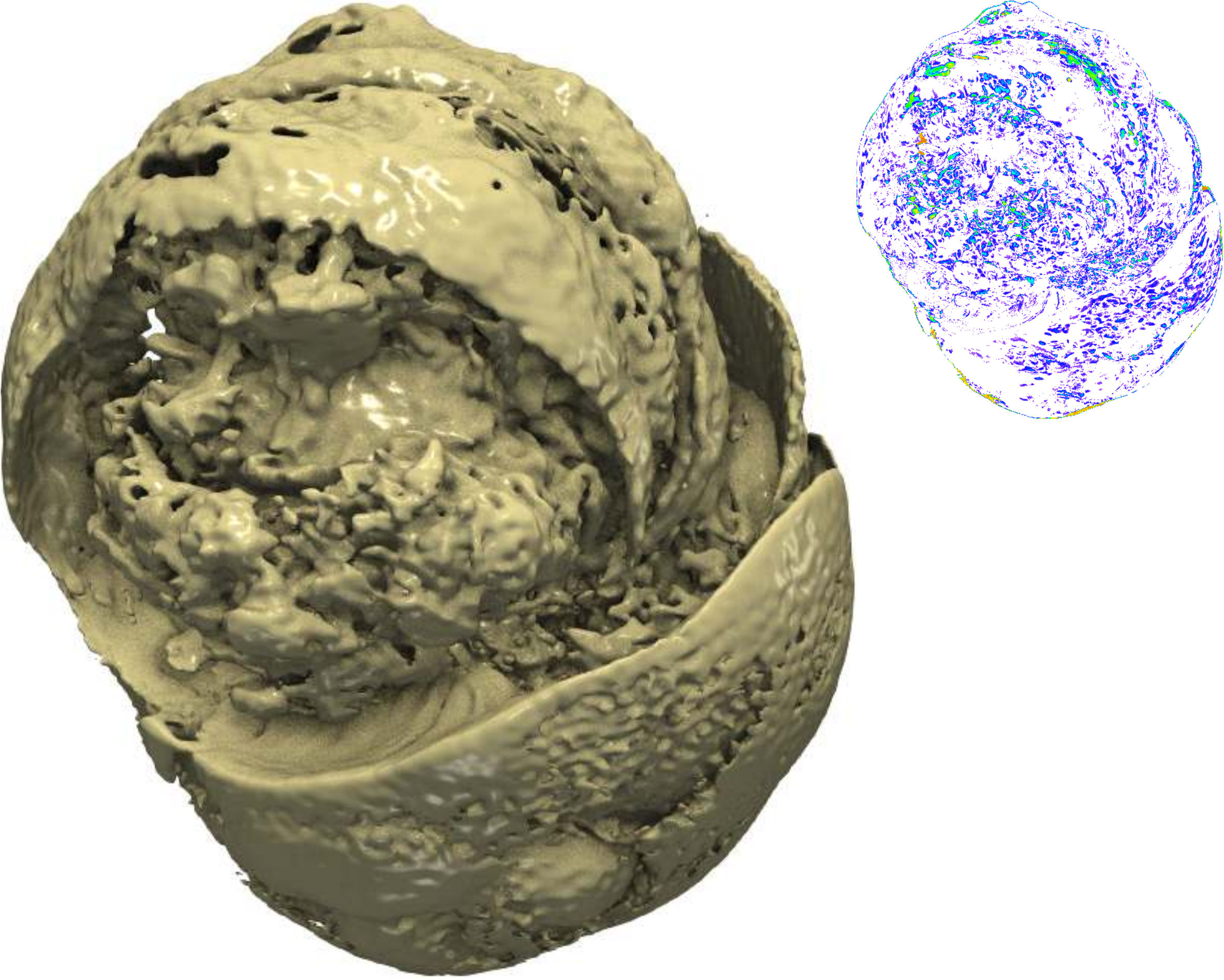} &
				\includegraphics[height=0.7in]{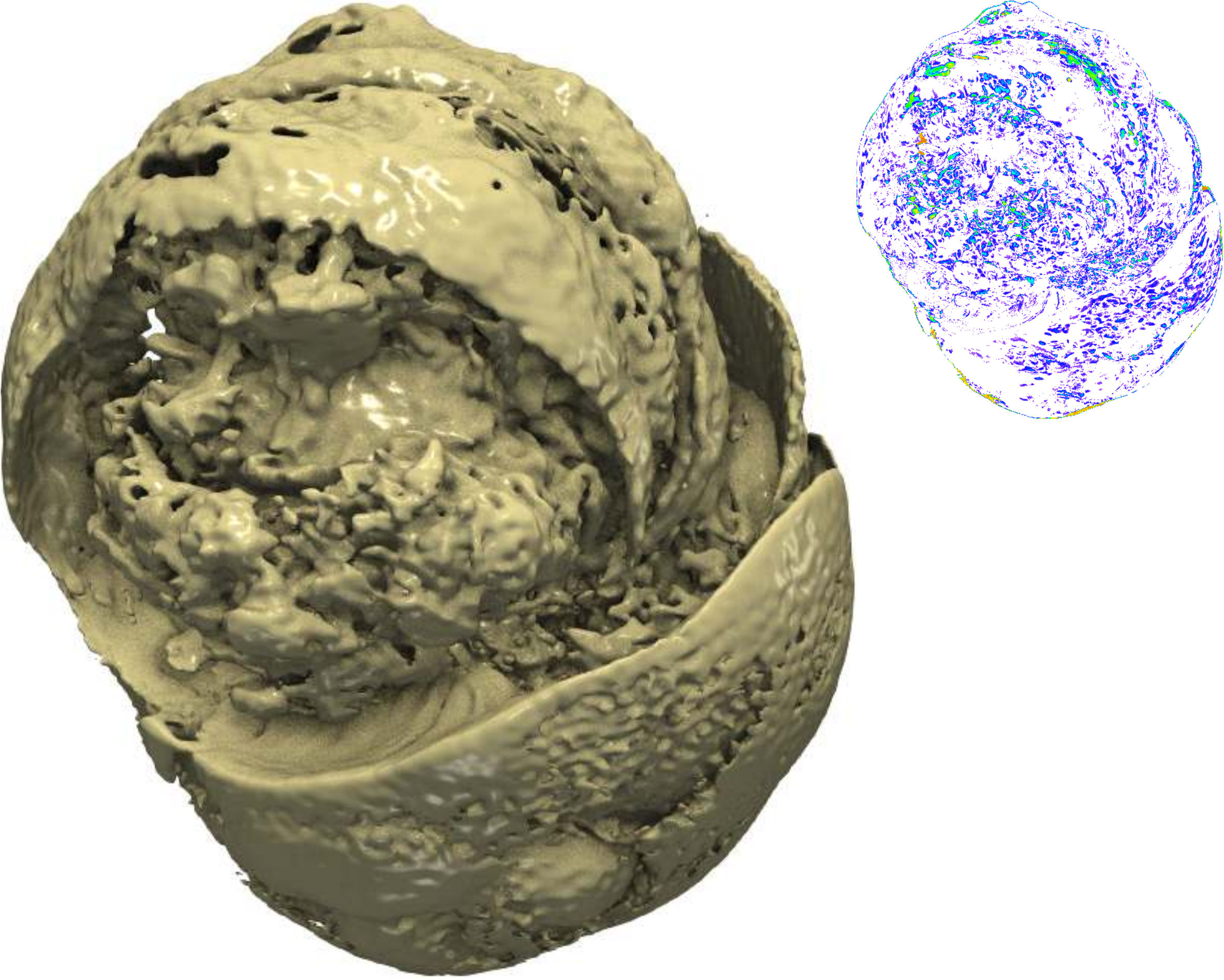} &
				\includegraphics[height=0.7in]{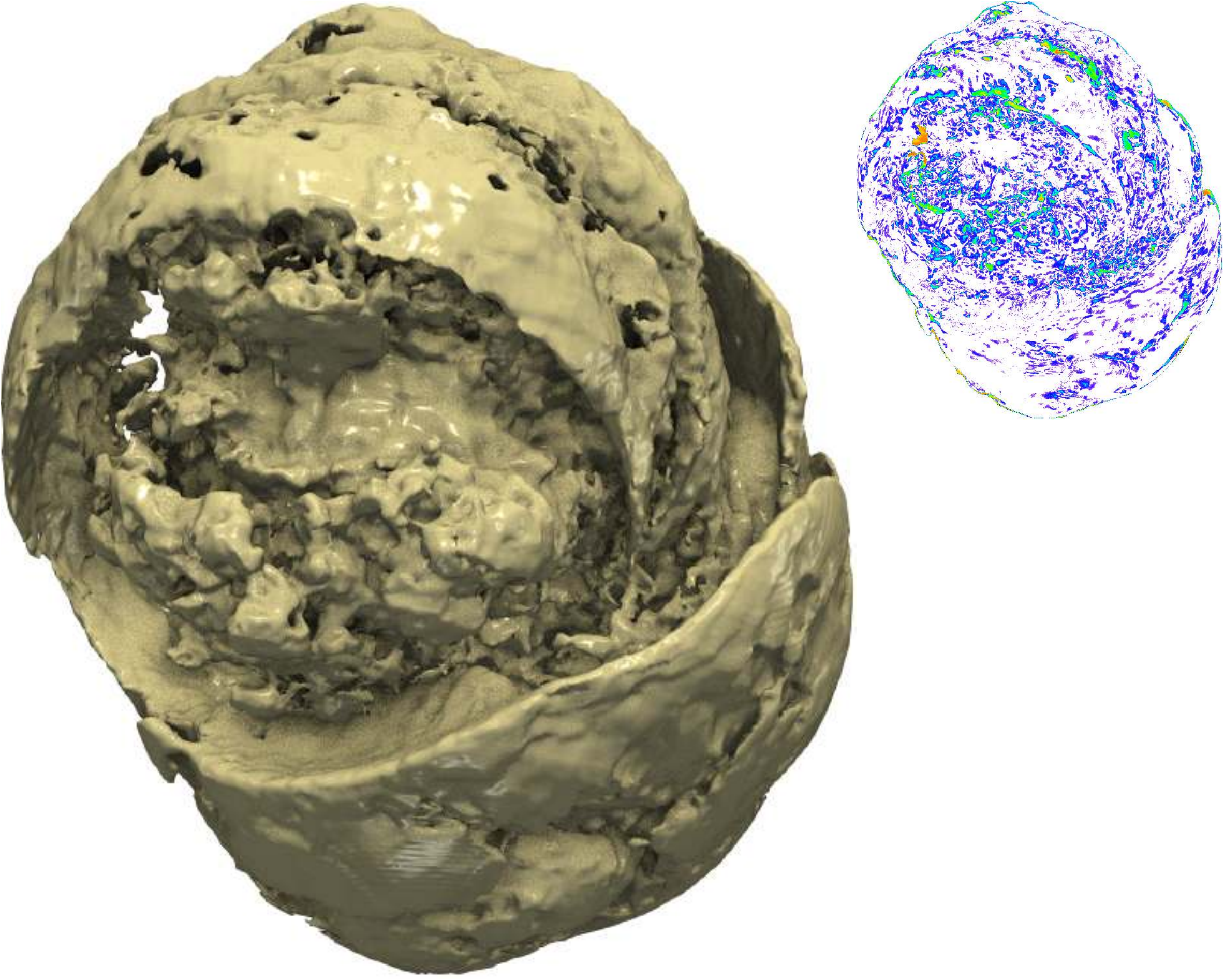} &
				\includegraphics[height=0.7in]{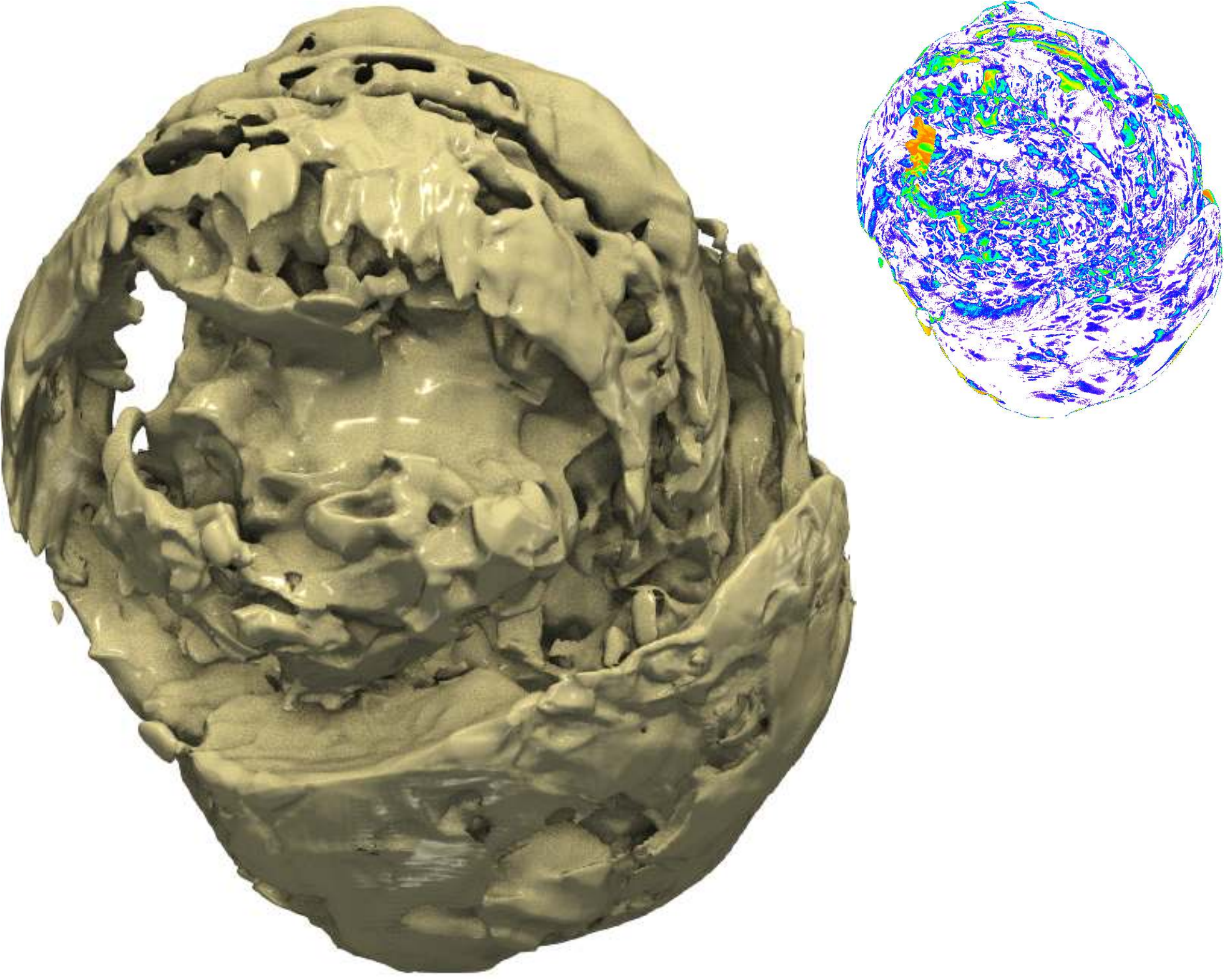} &
				\includegraphics[height=0.7in]{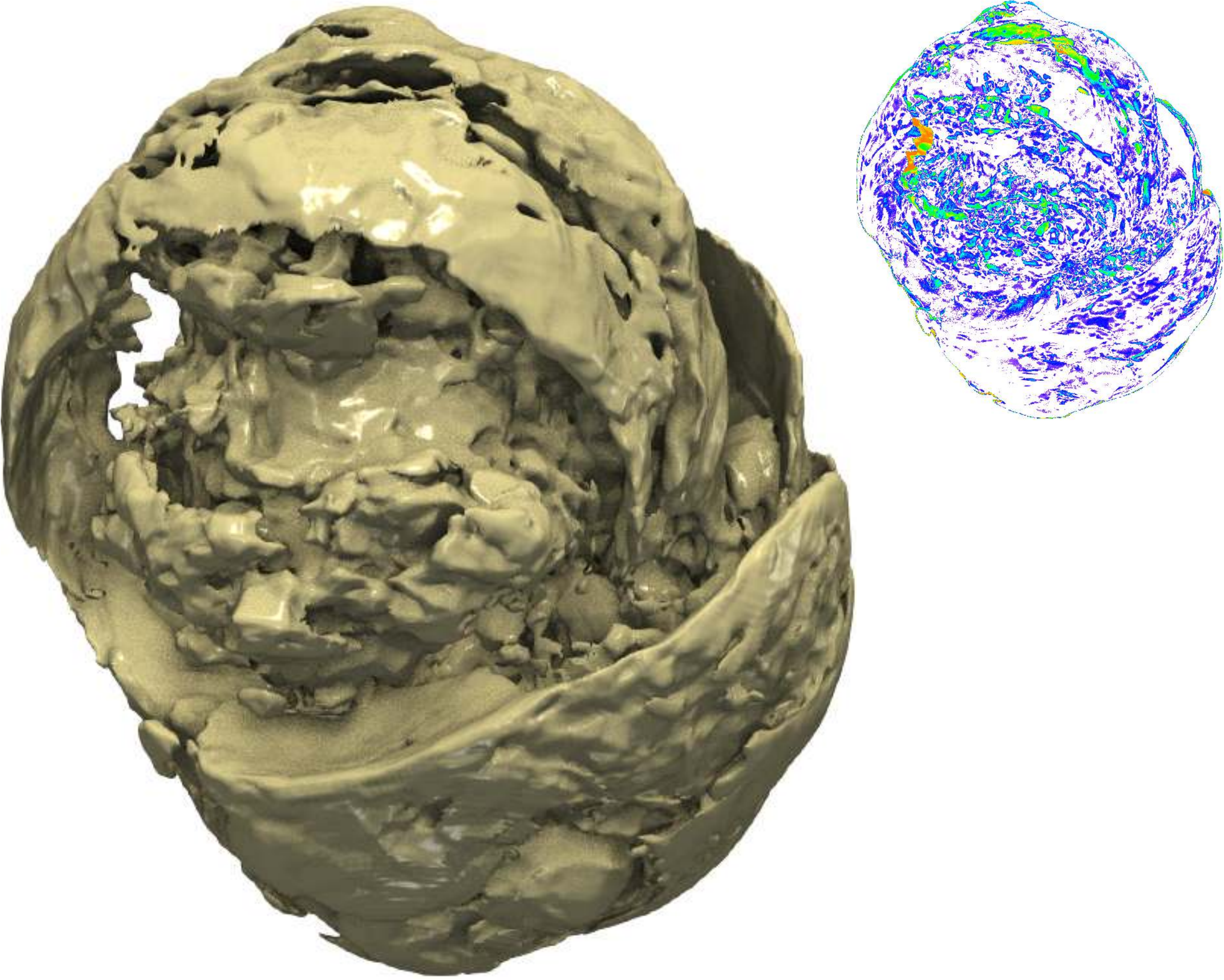} &
				\includegraphics[height=0.7in]{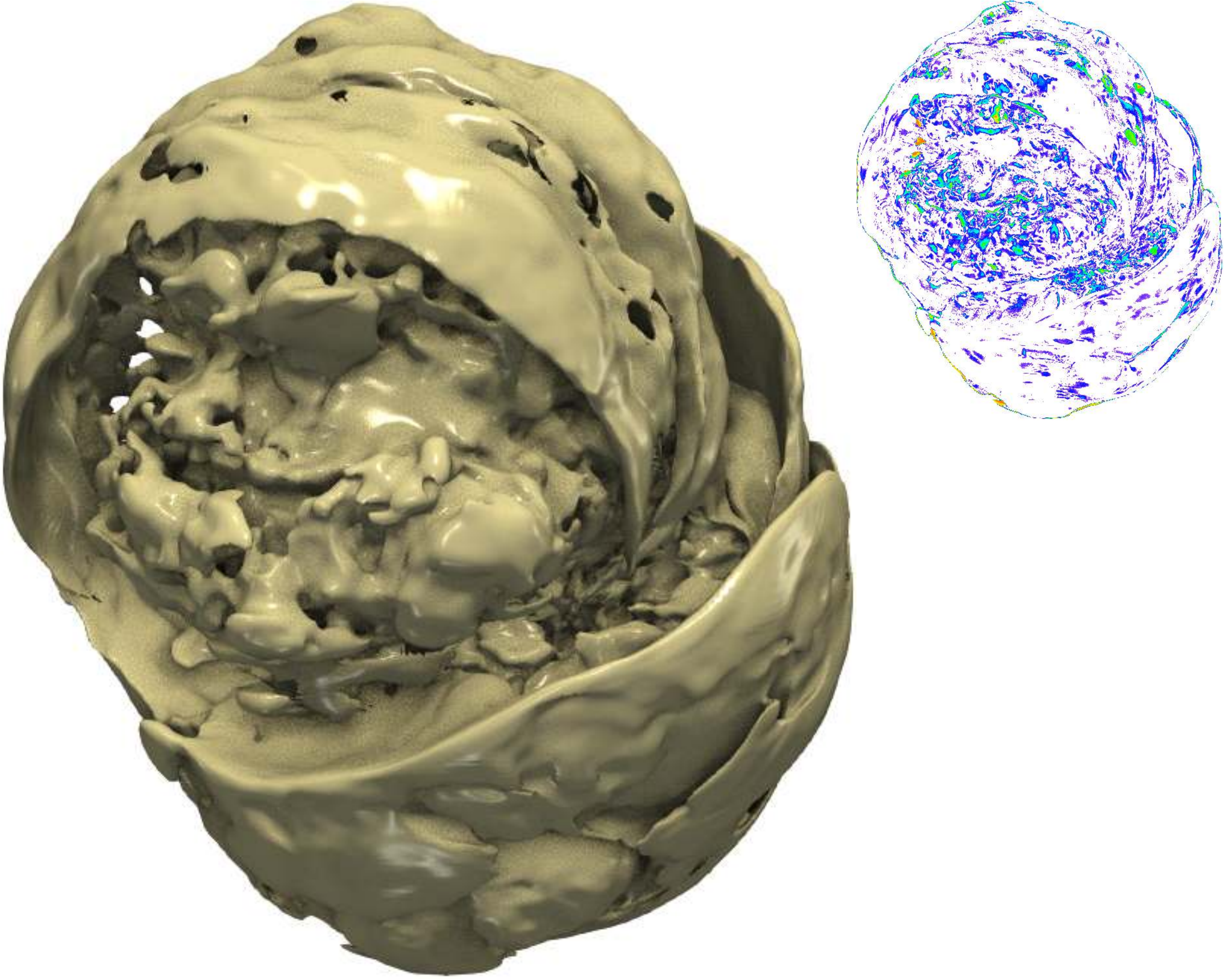} &
				\includegraphics[height=0.7in]{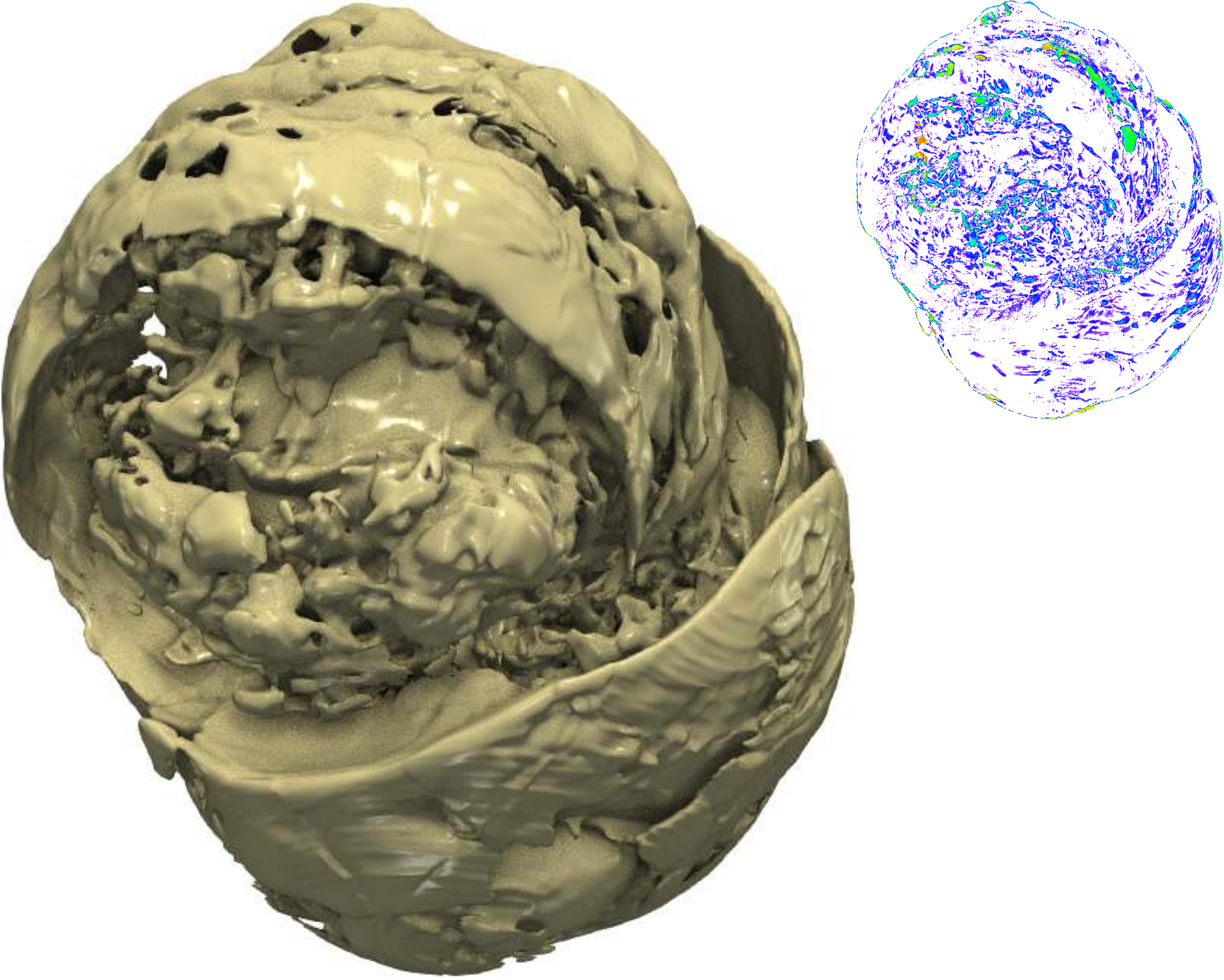} &
				\includegraphics[height=0.7in]{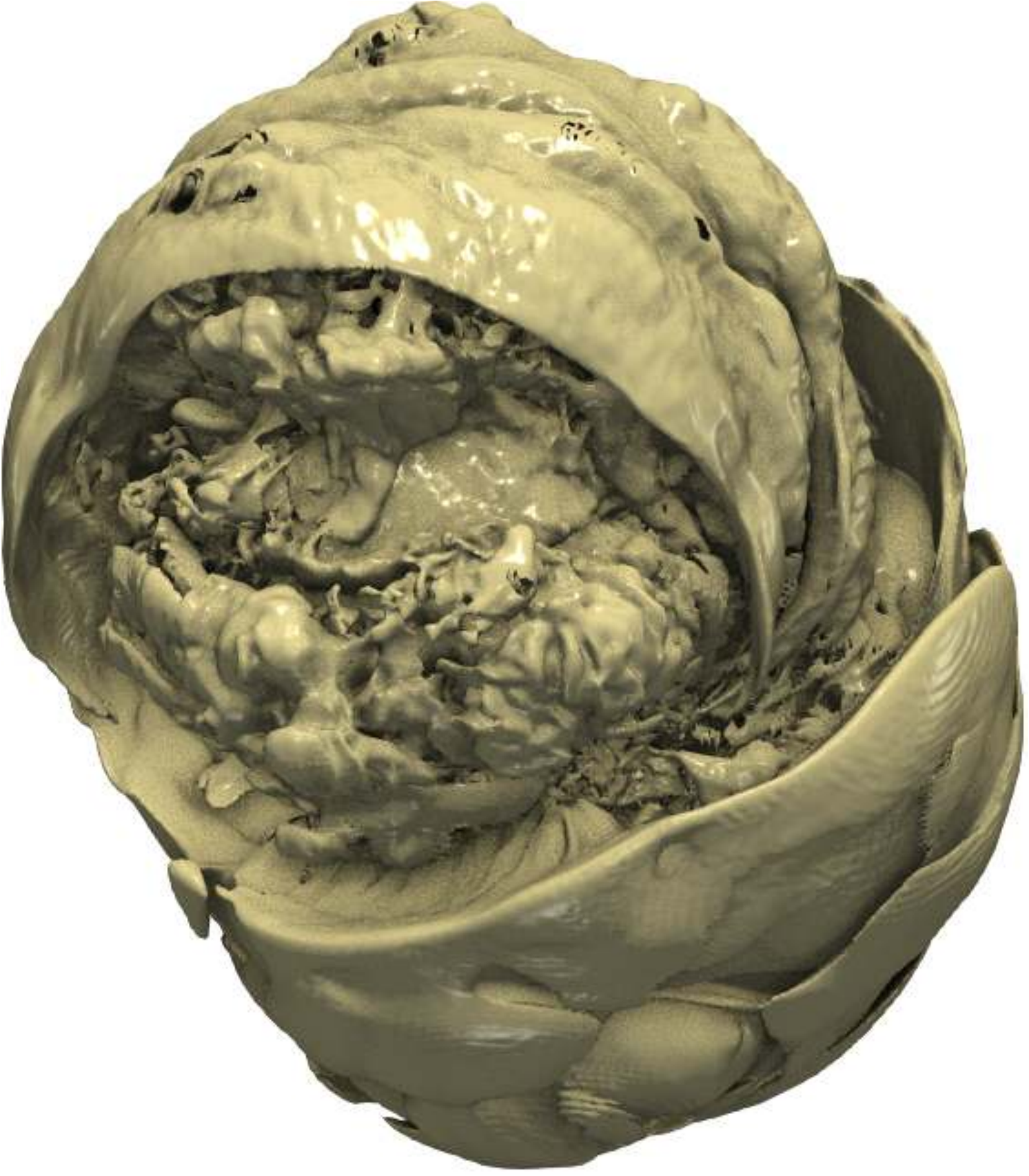}    \\
				\mbox{\footnotesize (a) \hot{SZ3}}   & \mbox{\footnotesize (b) \hot{TTHRESH}} & \mbox{\footnotesize (c) \hot{MHT}} &
					\mbox{\footnotesize (d) \hot{fV-SRN}} &
					\mbox{\footnotesize (e) \hot{APMGSRN}} &
					\mbox{\footnotesize (f) \hot{neurcomp}} &
					\mbox{\footnotesize (g) \hot{ECNR}} &
					\mbox{\footnotesize (h) \hot{GT}} 
			\end{array}$
	\end{center}
	\vspace{-.25in}
	\caption{\hot{Comparison of rendering results of different compression methods. Top and bottom: asteroids and supernova-1.}} 
	\label{fig:comp-APMGSRN}
\end{figure*}

\hot{{\bf Static volume compression.}}
\hot{To investigate the performance of ECNR on static volume compression, we compress asteroids and supernova-1 datasets with traditional methods (SZ3 and TTHRESH), grid-based methods (MHT, fV-SRN, and APMGSRN), neurcomp, and ECNR. The model size of all methods was controlled to 280 KB, which is the small storage setting of APMGSRN. We used the implementation in APMGSRN\footnote{https://github.com/skywolf829/APMGSRN} for the comparison of fV-SRN and MHT. We stopped neurcomp training when the network converged. 
For ECNR, we set the parameters (i.e., block dimension and neuron number, etc.) to control its model size. 
We still applied three scales for both datasets.} 
The number of target blocks per MLP from the coarsest to finest scales was adjusted to 1, 2, and 4 because the dataset is static. 
We trained the MLPs and latent codes from the coarsest to finest scales for 200, 150, and 125 epochs. 
\hot{Block-guided} pruning happened at 75 and 115 epochs with a target sparsity of 20\% and 30\%. 
The lightweight CNN was not applied since we did not observe obvious boundary artifacts from the MLP-decoded result.

In Table~\ref{tab:comp-single-timestep}, we report the quantitative results.  
\hot{Compared with traditional and grid-based methods, ECNR is slower in encoding and decoding except for the decoding of TTHRESH. 
However, ECNR shows faster convergence and inference compared with neurcomp, thanks to the multiple MLPs and multiscale representation. 
Since the CR is not high and both datasets are relatively sparse, traditional methods and MHT perform well on reconstruction quality.
For asteroids, SZ3 preserves the superior visual quality, and all deep-learning methods demonstrate similar reconstruction accuracy as shown in Figure~\ref{fig:comp-APMGSRN}.
For less sparse supernova-1, ECNR reaches the best or second-best reconstruction quality of PSNR, LPIPS, and CD.
The rendering results in Figure~\ref{fig:comp-APMGSRN} also show that ECNR yields less error compared with other deep-learning methods.}

\begin{figure}[htb]
	\begin{center}
		$\begin{array}{c@{\hspace{0.02in}}c@{\hspace{0.02in}}c@{\hspace{0.02in}}c}
				\includegraphics[height=0.9in]{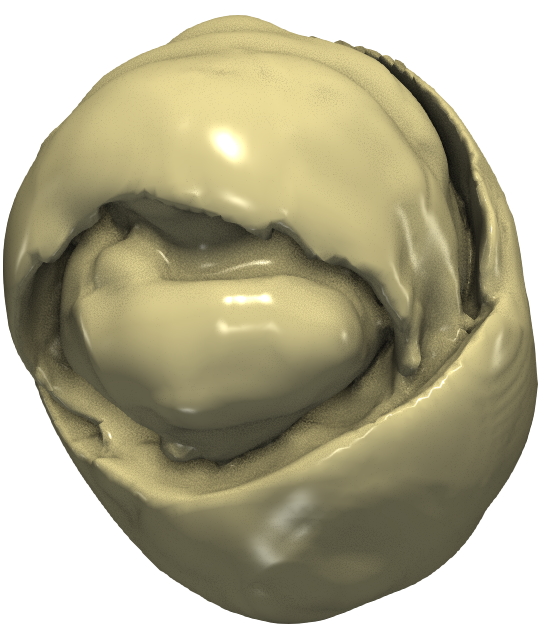}&
				\includegraphics[height=0.9in]{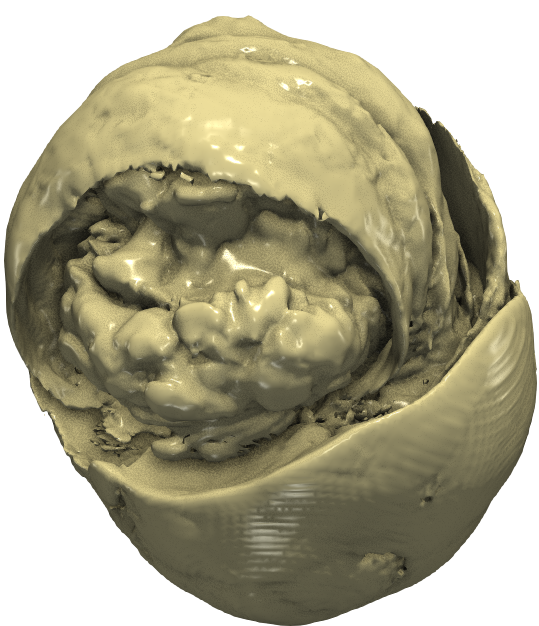} &
				\includegraphics[height=0.9in]{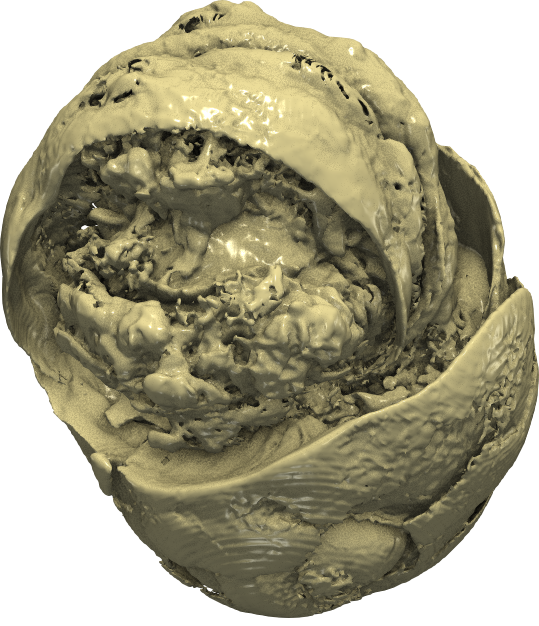} &
				\includegraphics[height=0.9in]{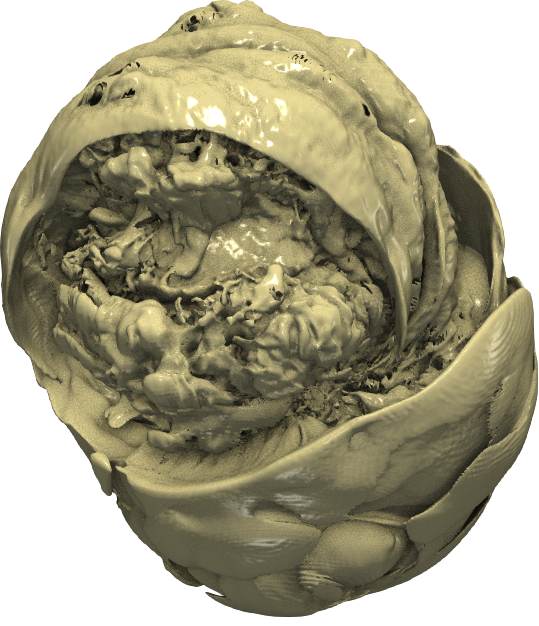}    \\
				\mbox{\footnotesize (a) scale 3} &
				\mbox{\footnotesize (b) scale 2 } &
				 \mbox{\footnotesize (c) scale 1} & 
				\mbox{\footnotesize (d) GT} \\
			\end{array}$
	\end{center}
	\vspace{-.25in}
	\caption{Isosurface rendering ($v = 0.0$) results of the supernova-1 dataset with ECNR under streaming reconstruction (from scale 3 to scale 1).}
	\label{fig:streaming}
\end{figure}

{\bf Streaming reconstruction.}
In ECNR, decoding the data at a coarse scale is independent of encoding the data at a finer scale. 
Considering the scenario of in-situ encoding, transmitting the coarse-scale compression file first to preview the simulation result by end users is possible. 
As shown in Table~\ref{tab:streaming} and Figure~\ref{fig:streaming}, the coarsest scale 3 model file has only 0.3 MB while preserving the primary structure of the supernova. 
Since scale 3 also takes the least time to encode and decode, it could serve as a preview of the compression target. 
The detailed structure could then be added along with encoding the subsequent scales.

\begin{table}[htb]
\caption{Encoding \hot{model size (MS, in MB), encoding time (ET), and decoding time (DT)} for each scale of the supernova-1 dataset.}
\vspace{-0.1in}
\centering
{\scriptsize
\begin{tabular}{c|lll}
scale & MS$\downarrow$	& ET$\downarrow$ & DT$\downarrow$\\ \hline
scale 3 &0.3 & 12s &0.04s\\
scale 2 &0.6 & 28s &1.2s\\
scale 1 &1.6 & 87s &2.7s\\
\end{tabular}
}
\label{tab:streaming}
\end{table}

\begin{figure}[htb]
	\begin{center}
		$\begin{array}{c@{\hspace{0.02in}}c@{\hspace{0.02in}}c}
				\includegraphics[height=1in]{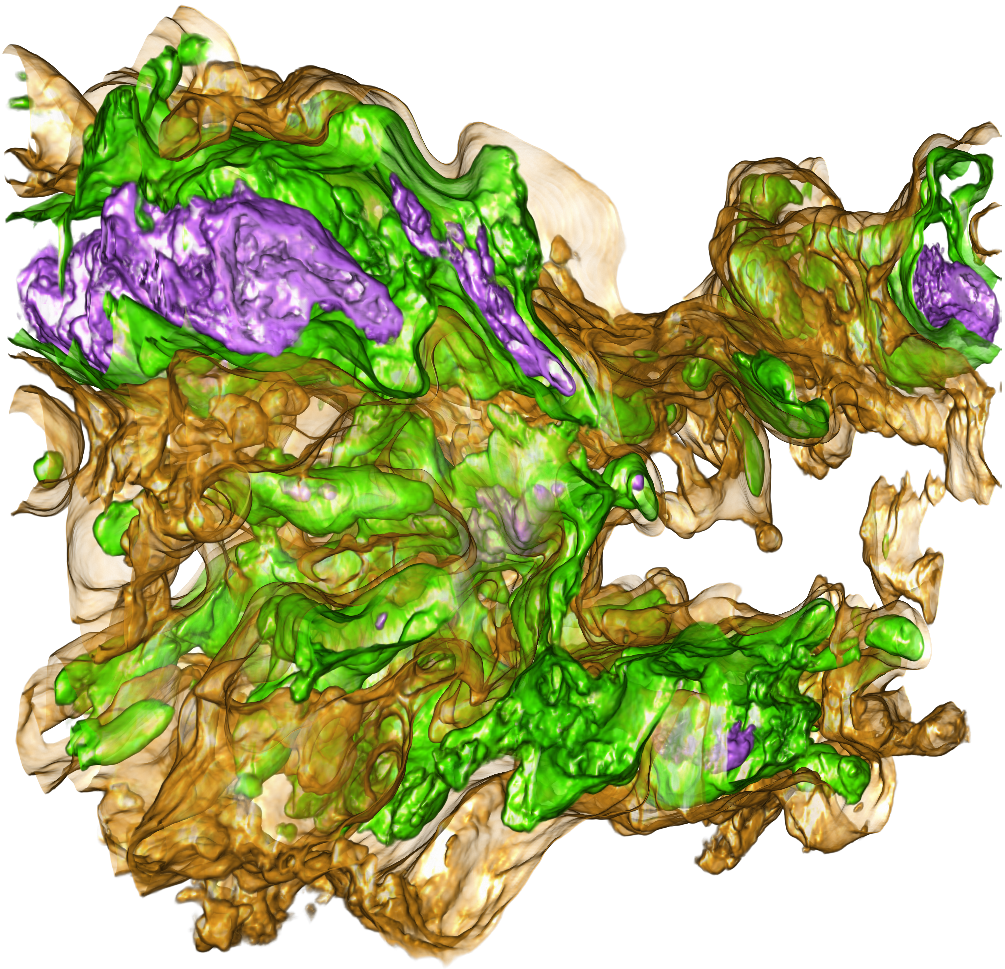}&
				\includegraphics[height=1in]{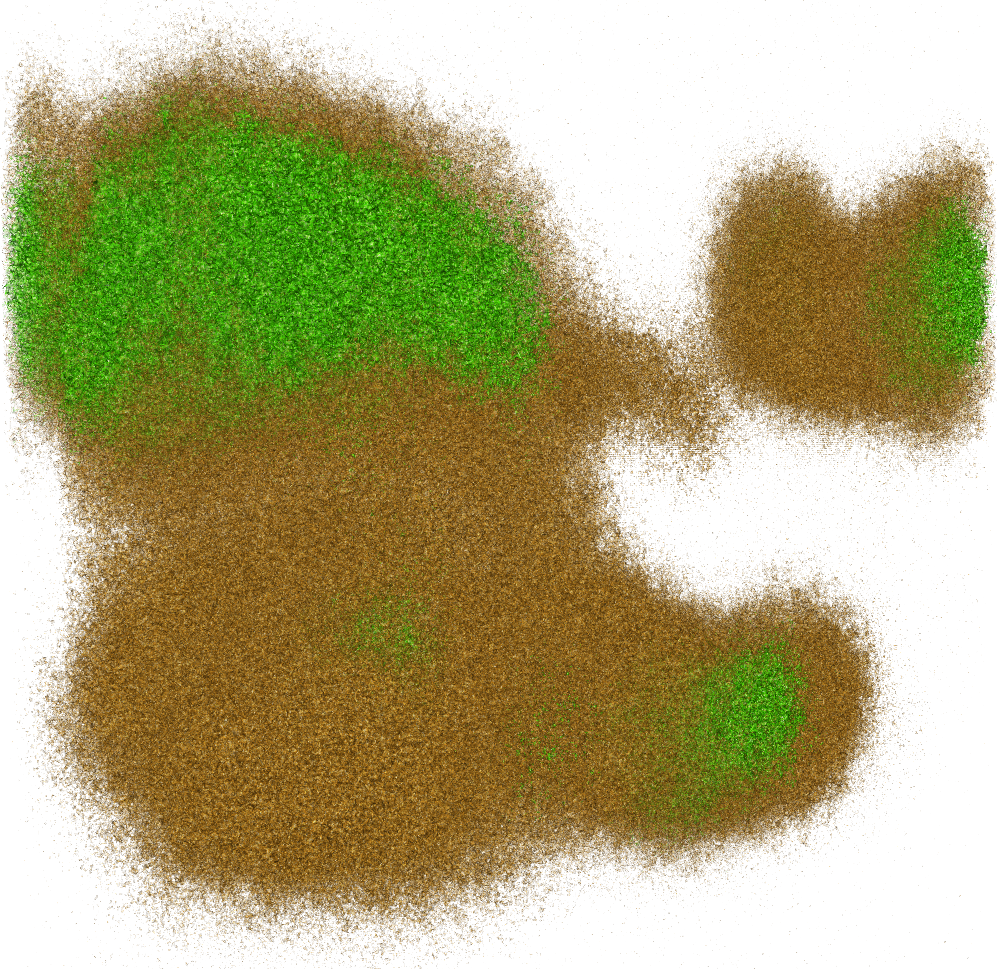} &
				\includegraphics[height=1in]{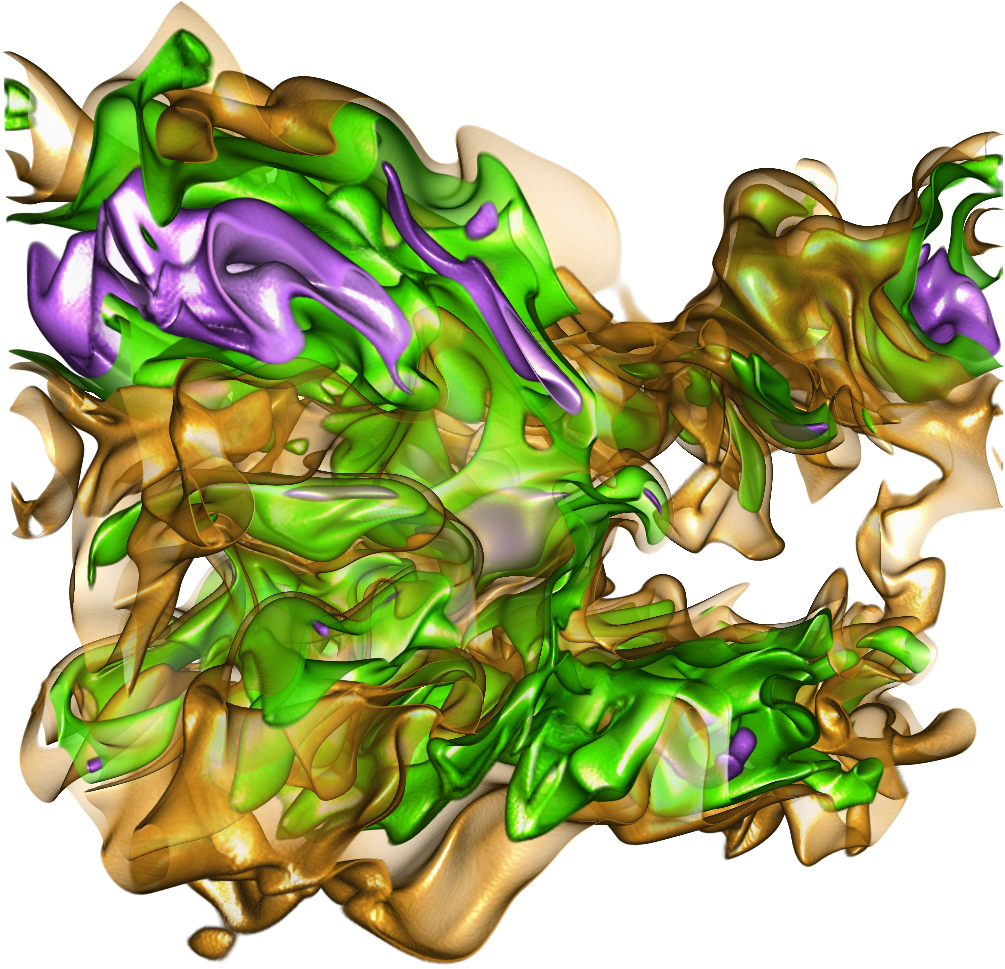}    \\
				\mbox{\footnotesize (a) $6,000$ iterations}   & \mbox{\footnotesize (b) $10^4$ iterations} & \mbox{\footnotesize (c) GT} 
			\end{array}$
	\end{center}
	\vspace{-.25in}
	\caption{Volume rendering results of neurcomp using the combustion dataset. Increasing the training iterations could lead to unstable outcomes.}
	\label{fig:failure-case}
\end{figure}

{\bf neurcomp unstable optimization.}
When training neurcomp on a large-scale dataset, the network could encounter unstable optimization, leading to unacceptable outcomes. 
This could be partly attributed to the limited network capacity for fitting complex distribution and rich content of 4D volume. 
Such an example with the combustion dataset is shown in Figure~\ref{fig:failure-case}. 
We can see that increasing the training epochs of neurcomp yields undesirable results. 
Unlike neurcomp, our ECNR is stable thanks to the adaptive increase of network capacity for the complex content in the multiscale fitting process. 

\section{\hot{Additional Network Analysis}}

\hot{Besides block assignment and lightweight CNN presented in the paper, we further study ECNR in six aspects.}

{\bf Sparse vs.\ dense \hot{model}.}
In ECNR, we prune unimportant network parameters to reduce the model size. 
\hot{Block-guided pruning} reaches 30\% sparsity for a large network with 24 neurons in each layer.  
Initializing and training a dense, small network with fewer neurons in each layer should result in a similar model size without pruning. 
We trained two ECNR models (ECNR-dense and ECNR-sparse) on the supernova-1 dataset to investigate which is better, assuming both should have a similar CR for a fair comparison. 
ECNR-dense denotes multiple MLPs initialized with 16 neurons in each layer and without \hot{block-guided pruning} in training. 
ECNR-sparse denotes multiple MLPs initialized with 24 neurons in each layer and with 30\% of parameters pruned in the training. 
To make a fair comparison and avoid the influence of other factors, we set $s=1$ without applying quantization or entropy encoding. The block dimension was set to $54 \times 54 \times 54$ in this experiment. In Table~\ref{tab:sparse-vs-dense}, we report PSNR and LPIPS values. ECNR-sparse shows higher accuracy than ECNR-dense under the same model size, indicating that it is necessary to employ a sparse network to achieve higher performance.

\begin{table}[htb]
\caption{PSNR (dB) and LPIPS values, as well as \hot{model size (MS, in MB)} of the supernova-1 dataset.}
\vspace{-0.1in}
\centering
{\scriptsize
\begin{tabular}{c|lll}
model 				& PSNR$\uparrow$	& LPIPS$\downarrow$ & MS$\downarrow$\\ \hline
ECNR-sparse  &\textbf{40.93} &\textbf{0.034} &1.5\\
ECNR-dense 	 &39.84 &0.038 &1.5\\
\end{tabular}
}
\label{tab:sparse-vs-dense}
\end{table}

\begin{figure}[htb]
	\begin{center}
		$\begin{array}{c@{\hspace{0.02in}}c@{\hspace{0.02in}}c}
				\includegraphics[height=0.585in]{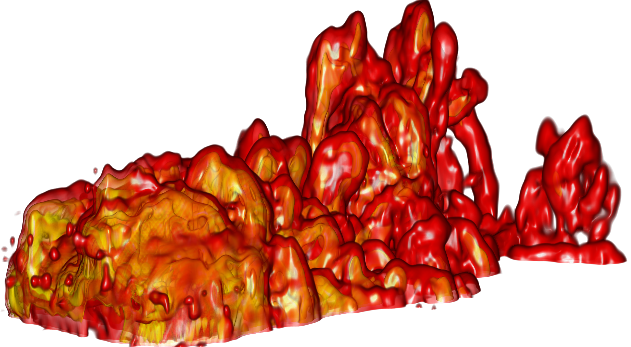} &
				\includegraphics[height=0.585in]{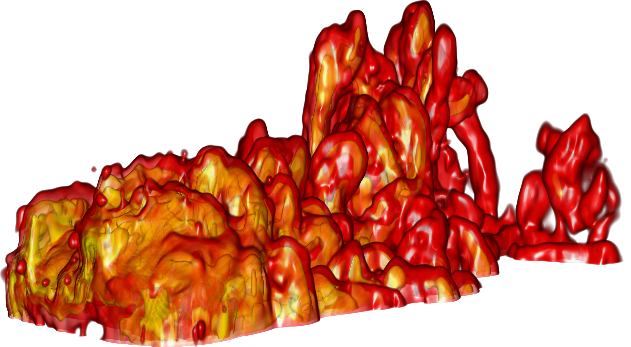} &
				\includegraphics[height=0.585in]{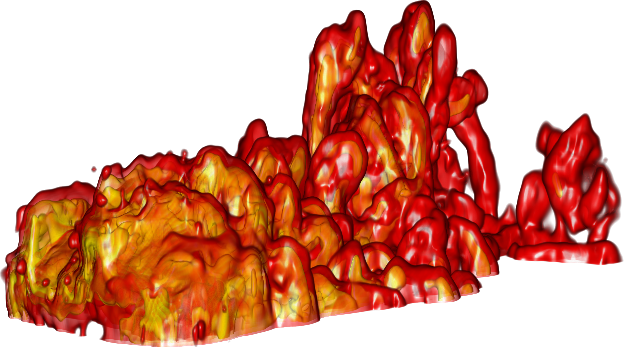} \\
				\mbox{\footnotesize (a)} & \mbox{\footnotesize (b)} & \mbox{\footnotesize (c)} \\
			\end{array}$
		$\begin{array}{c@{\hspace{0.02in}}c}
				\includegraphics[height=0.585in]{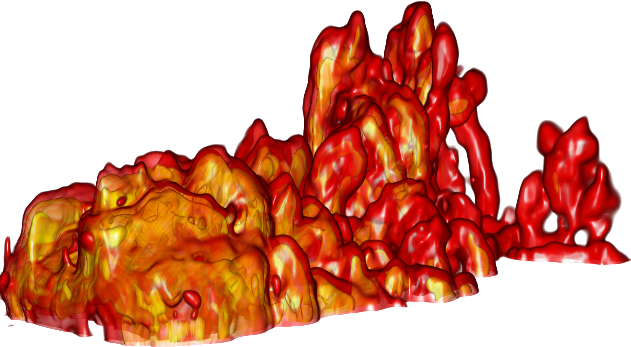} &
				\includegraphics[height=0.585in]{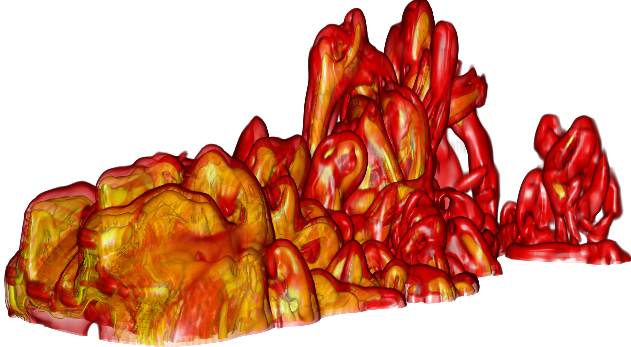} \\
				\mbox{\footnotesize (d)} & \mbox{\footnotesize (e)}\\
			\end{array}$
	\end{center}
	\vspace{-.25in}
	\caption{Volume rendering results of the Tangaroa dataset under different deep compression schemes: (a) w/o pruning and quantization, (b) w/ pruning but w/o quantization, (c) w/o pruning but w/ quantization, and (d) w/ pruning and quantization. (e) is the GT.}
	\label{fig:deep-comp}
\end{figure}

{\bf Deep compression ablation study.}
To investigate the impact of deep compression strategy on model size and reconstruction quality, we conducted an ablation study on the two steps, i.e., \hot{block-guided pruning} and network quantization, using the Tangaroa dataset. 
We did not consider entropy encoding since it is applied directly to compress the model file losslessly. 
Table~\ref{tab:deep-comp} reports the average PSNR and LPIPS values across all timesteps and the model size under different schemes. 
The results show that different schemes achieve a similar reconstruction accuracy.
As displayed in Figure~\ref{fig:deep-comp}, visual quality under different schemes is also close. 
Therefore, pruning and quantization can reduce the model size with a small reconstruction quality loss, and we chose the scheme with pruning and quantization in our experiments. 

\begin{table}[htb]
\caption{Average PSNR (dB) and LPIPS values across all timesteps, as well as \hot{model size (MS, in MB)} of the Tangaroa dataset.}
\vspace{-0.1in}
\centering
{\scriptsize
\begin{tabular}{c|lll}
scheme & PSNR$\uparrow$ & LPIPS$\downarrow$ & MS$\downarrow$ \\ \hline
w/o pruning; w/o quantization &42.33 &0.044 &12.9\\
w/ pruning; w/o quantization &\textbf{42.38} &0.047 &7.8\\ 
w/o pruning; w/ quantization  &41.29 &0.045 &5.5\\
w/ pruning; w/ quantization &41.47 &\textbf{0.039} &\textbf{4.7}
\end{tabular}
}
\label{tab:deep-comp}
\end{table}

\begin{figure}[htb]
	\begin{center}
		$\begin{array}{c@{\hspace{0.02in}}c@{\hspace{0.02in}}c}
				\includegraphics[height=0.85in]{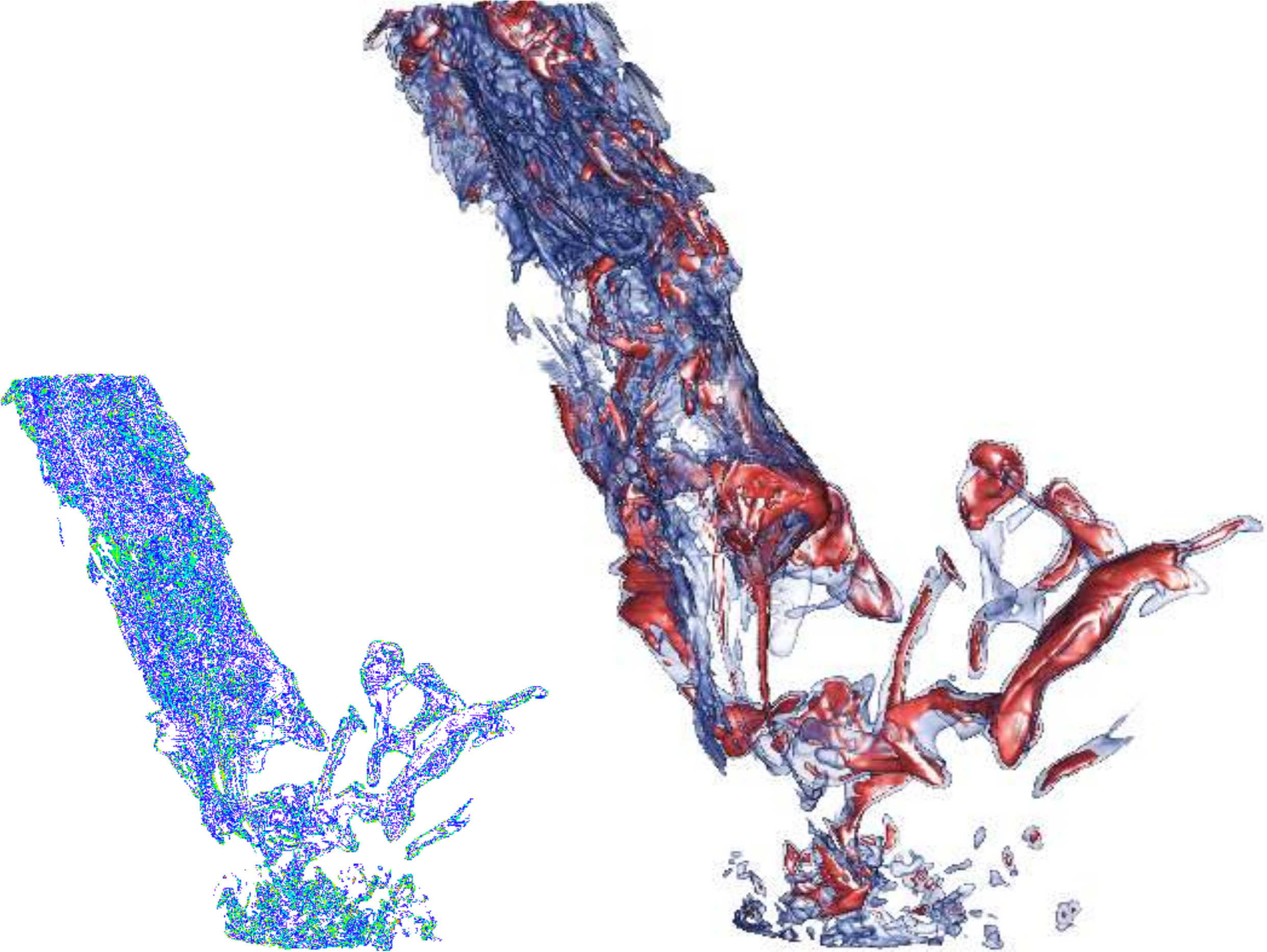} &
				\includegraphics[height=0.85in]{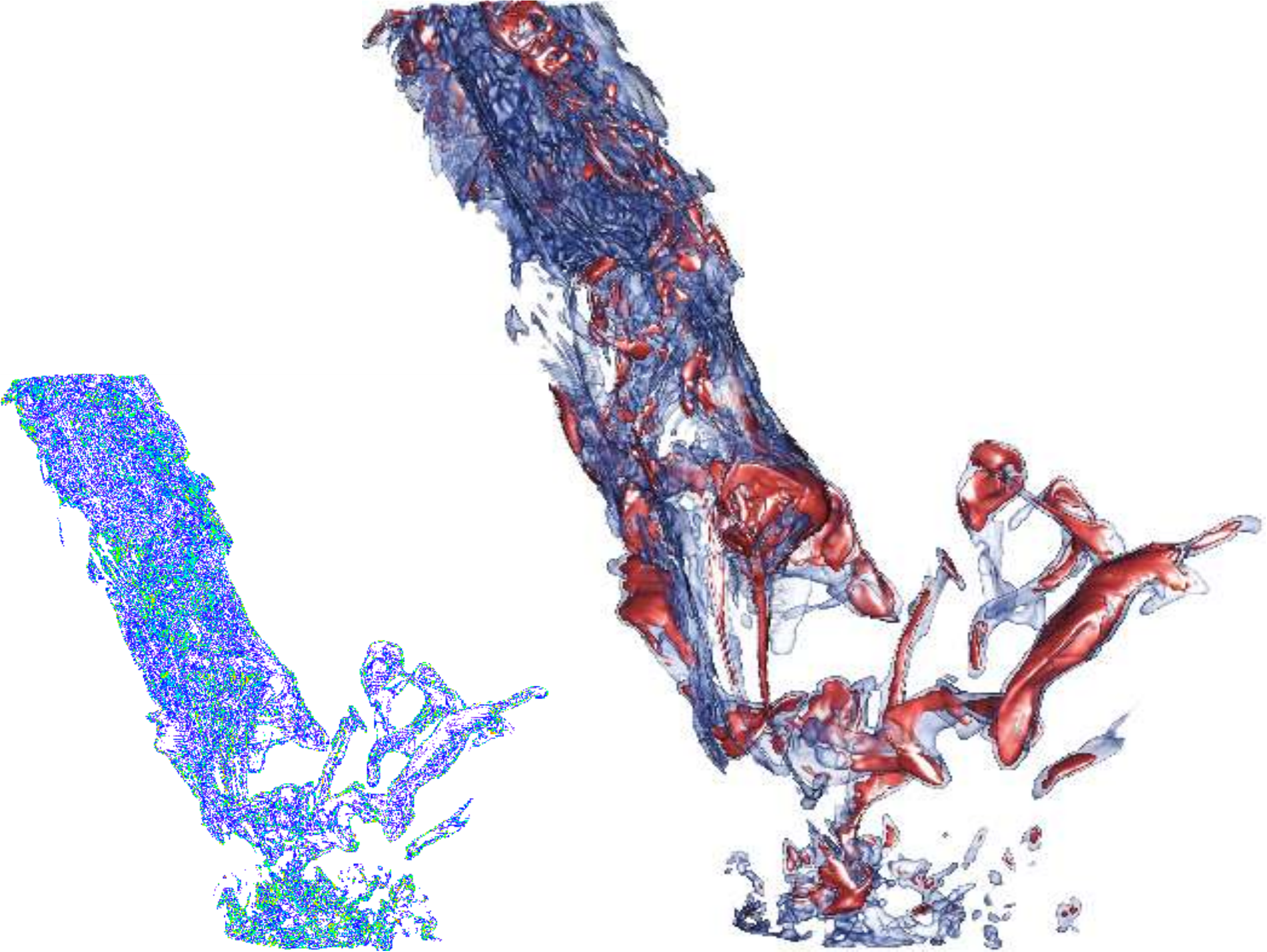} &
				\includegraphics[height=0.85in]{figures/asteroids-vol-GT.pdf}    \\
				\includegraphics[height=0.9in]{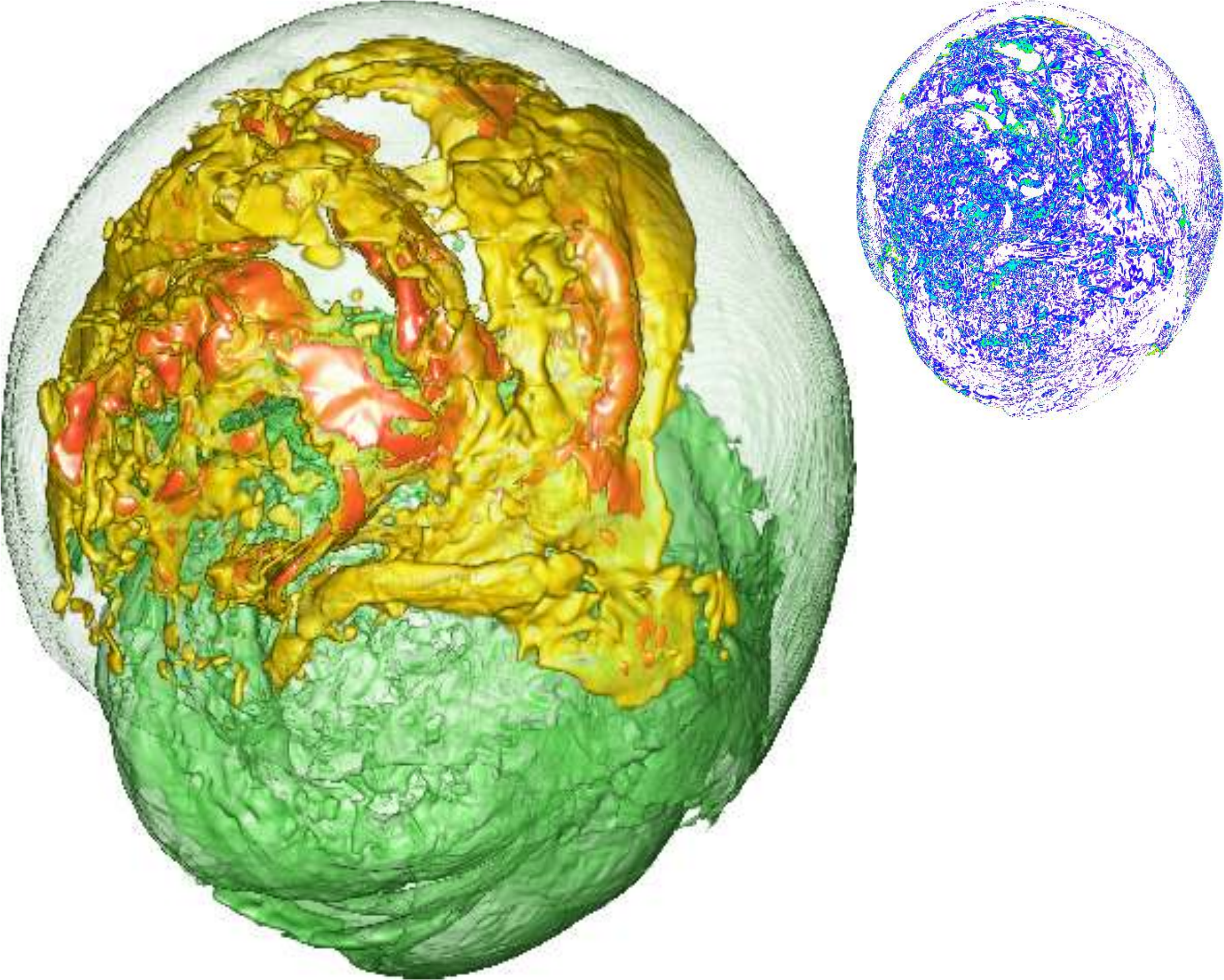} &
				\includegraphics[height=0.9in]{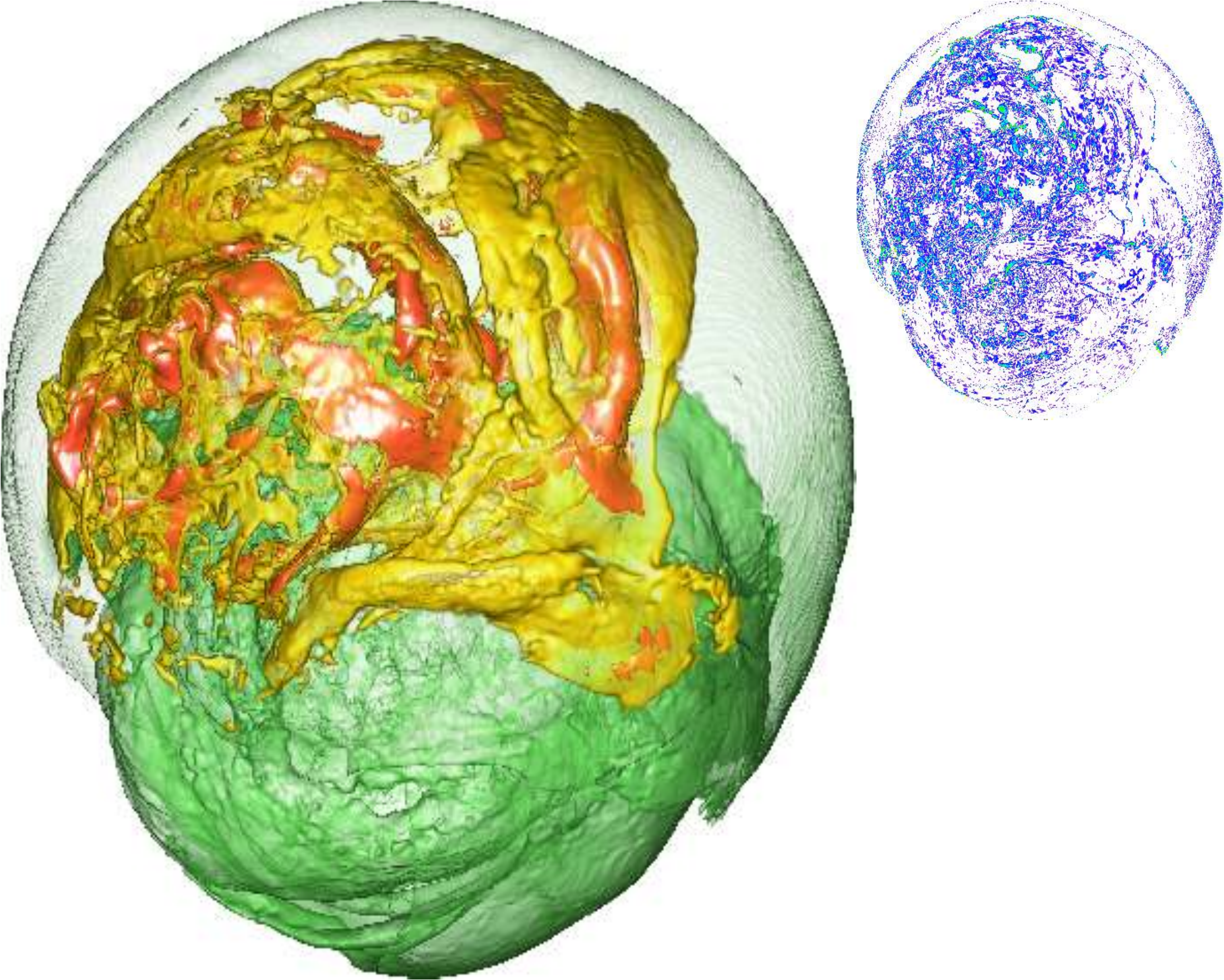} &
				\includegraphics[height=0.9in]{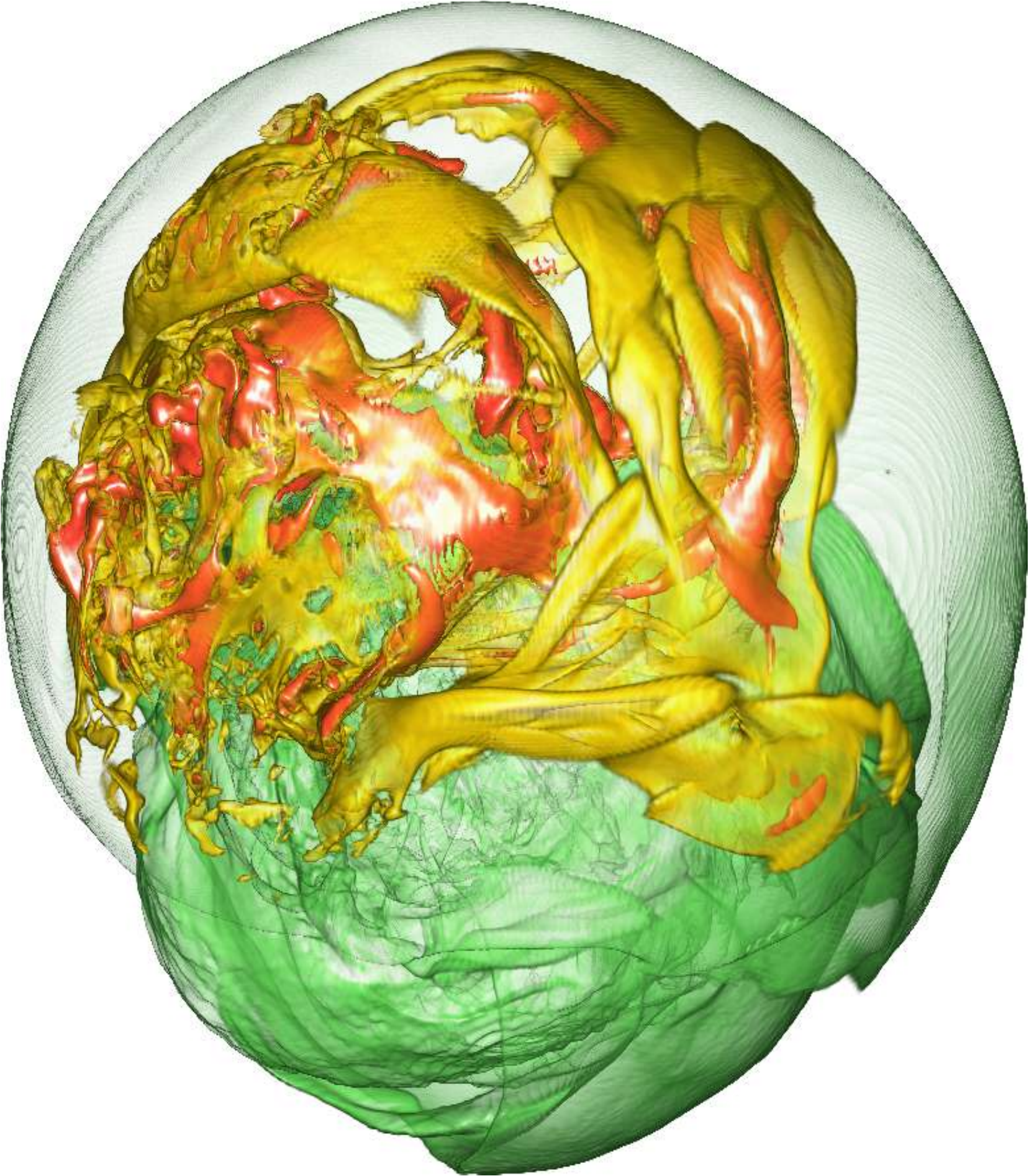}    \\
				\mbox{\footnotesize (a) \hot{ECNR$^{\star}$}} &
				 \mbox{\footnotesize (b) \hot{ECNR}} & 
				\mbox{\footnotesize (c) \hot{GT}} \\
			\end{array}$
	\end{center}
	\vspace{-.25in}
	\caption{\hot{Volume rendering results of ECNR$^{\star}$ and ECNR. Top and bottom: asteroids and supernova-1.}}
	\label{fig:vsStandardCompression}
\end{figure}

\hot{{\bf Comparison with standard pruning and local quantization.}
For block-guided pruning, we add block error information to guide the pruning intensity of different MLPs within the same scale.
Moreover, unlike most deep compression strategies that only operate on the weight of the neural networks~\cite{Han-NIPS15,Ye-arXiv18,Lin-ICLR20}, we prune and quantize both the weight and bias for each small MLP. 
Compared with traditional deep learning approaches, our model contains numerous independent small MLPs in which the number of weights per layer is limited, resulting in a relatively high bias proportion compared to a large MLP. 
For instance, given one hidden layer of an MLP, if its weight size is $512^2$, its proportion of bias is approximately 0.2\% for this hidden layer. 
However, for a small weight size of $24^2$, this proportion can increase to 4\%. 
Given the existence of thousands of such small MLPs, identifying a sparse bias structure becomes feasible and meaningful for model compression. 
Furthermore, prior quantization approaches only focused on a single neural network~\cite{Lu-CGF21}. 
Applying quantization to multiple networks has not been thoroughly investigated. 
Considering multiple independent small MLPs can choose the same shared parameters to represent effective blocks containing similar content. 
We quantize MLPs globally instead of locally and maintain only one global shared parameter table for all MLPs in the following fine-tuning stage. 

To highlight the difference between our scheme and standard compression (denoted as ECNR$^\star$), we evaluated our block-guided pruning and global quantization with the common practice of standard pruning and local quantization on the asteroids and supernova-1 datasets. 
We chose $\beta=7$ bits for local quantization due to the small size of local MLPs.
We controlled neuron numbers for two schemes to reach a similar PSNR for asteroids and a similar model size for supernova-1.
Table~\ref{tab:vsStandardCompression} shows the quantitative results. 
With a similar PSNR, the model size of ECNR$^\star$ is 1.9$\times$ that of ECNR. 
As shown in Figure~\ref{fig:vsStandardCompression}, ECNR achieves better reconstruction accuracy than ECNR$^\star$ under the same model size.}

\begin{table}[htb]
\caption{\hot{PSNR (dB) and LPIPS values, as well as model size (MS, in MB). ECNR$^{\star}$ denotes ECNR with standard pruning and local quantization.}}
\vspace{-0.1in}
\centering
{\scriptsize
\hot{
\begin{tabular}{c|c|lll}
dataset &scheme & PSNR$\uparrow$ & LPIPS$\downarrow$ & MS$\downarrow$ \\ \hline
asteroids&ECNR$^{\star}$ &40.24 &0.123 &0.93\\
&ECNR &\textbf{40.59} &\textbf{0.107} &\textbf{0.49}\\ \hline
supernova-1&ECNR$^{\star}$ &44.86 &0.042 &\textbf{1.5}\\
&ECNR &\textbf{48.55} &\textbf{0.029} &\textbf{1.5}\\
\end{tabular}
}}
\label{tab:vsStandardCompression}
\end{table}


\begin{figure}[htb]
	\begin{center}
	$\begin{array}{c@{\hspace{0.02in}}c@{\hspace{0.02in}}c}
				\includegraphics[height=0.585in]{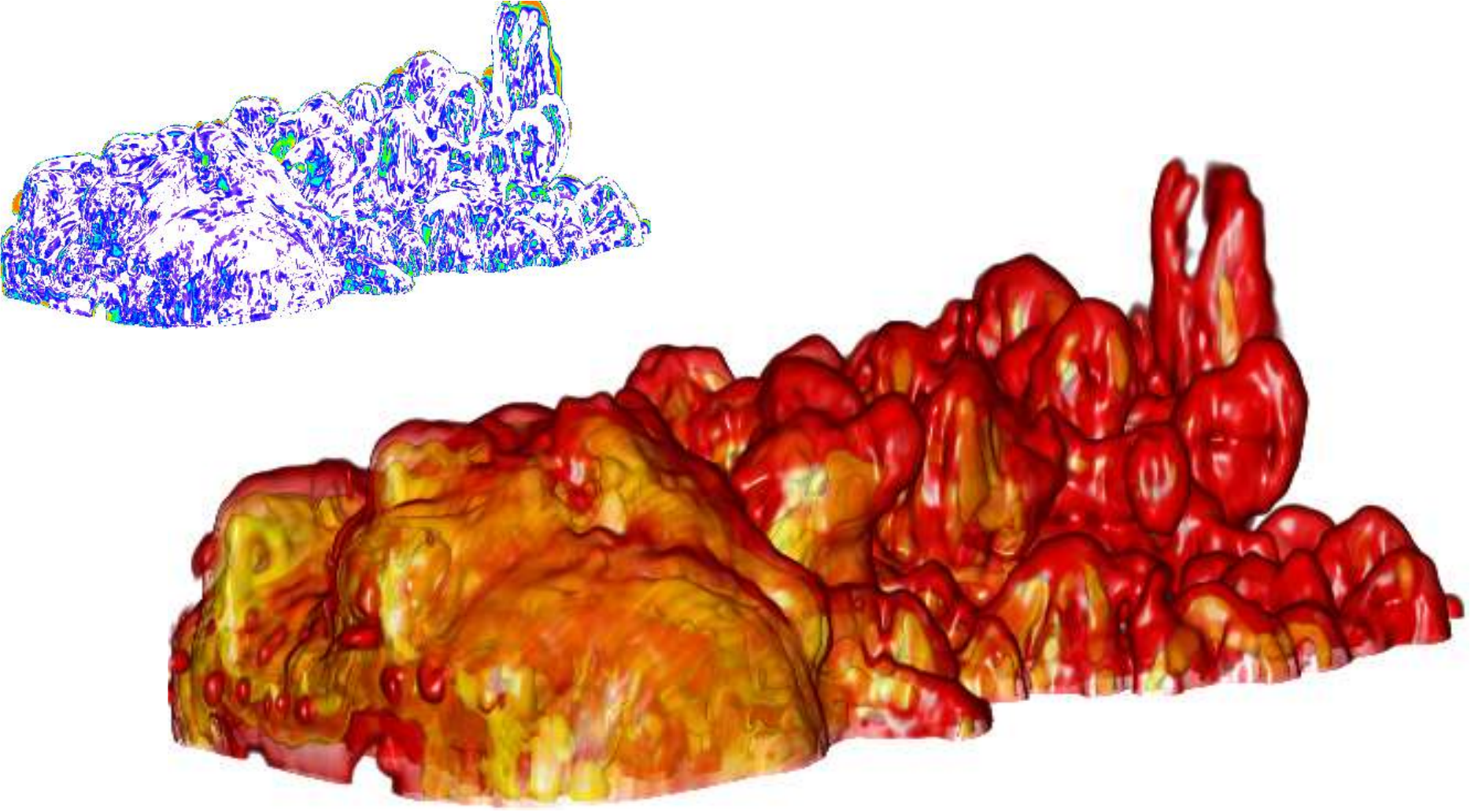}&
				\includegraphics[height=0.585in]{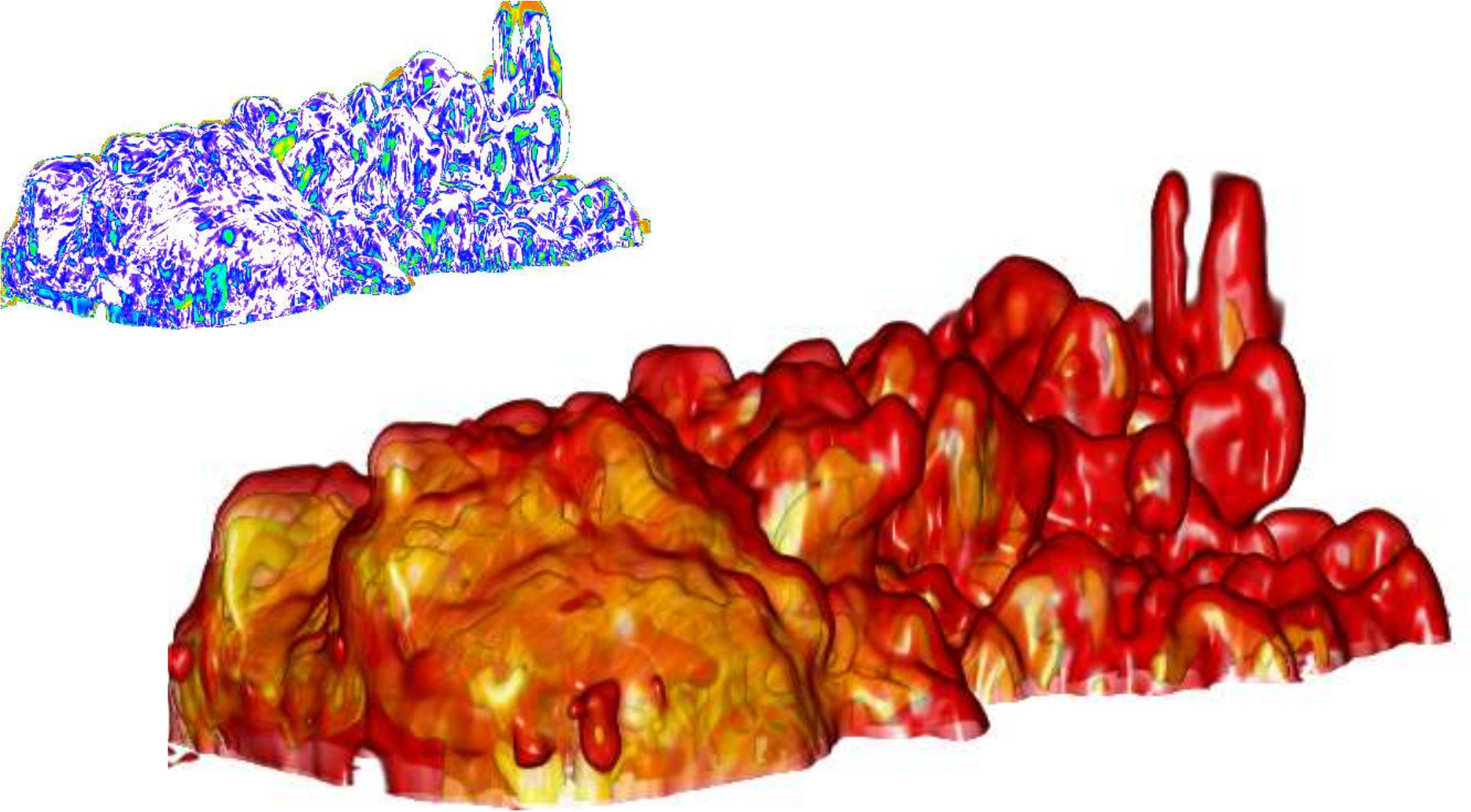}&
				\includegraphics[height=0.585in]{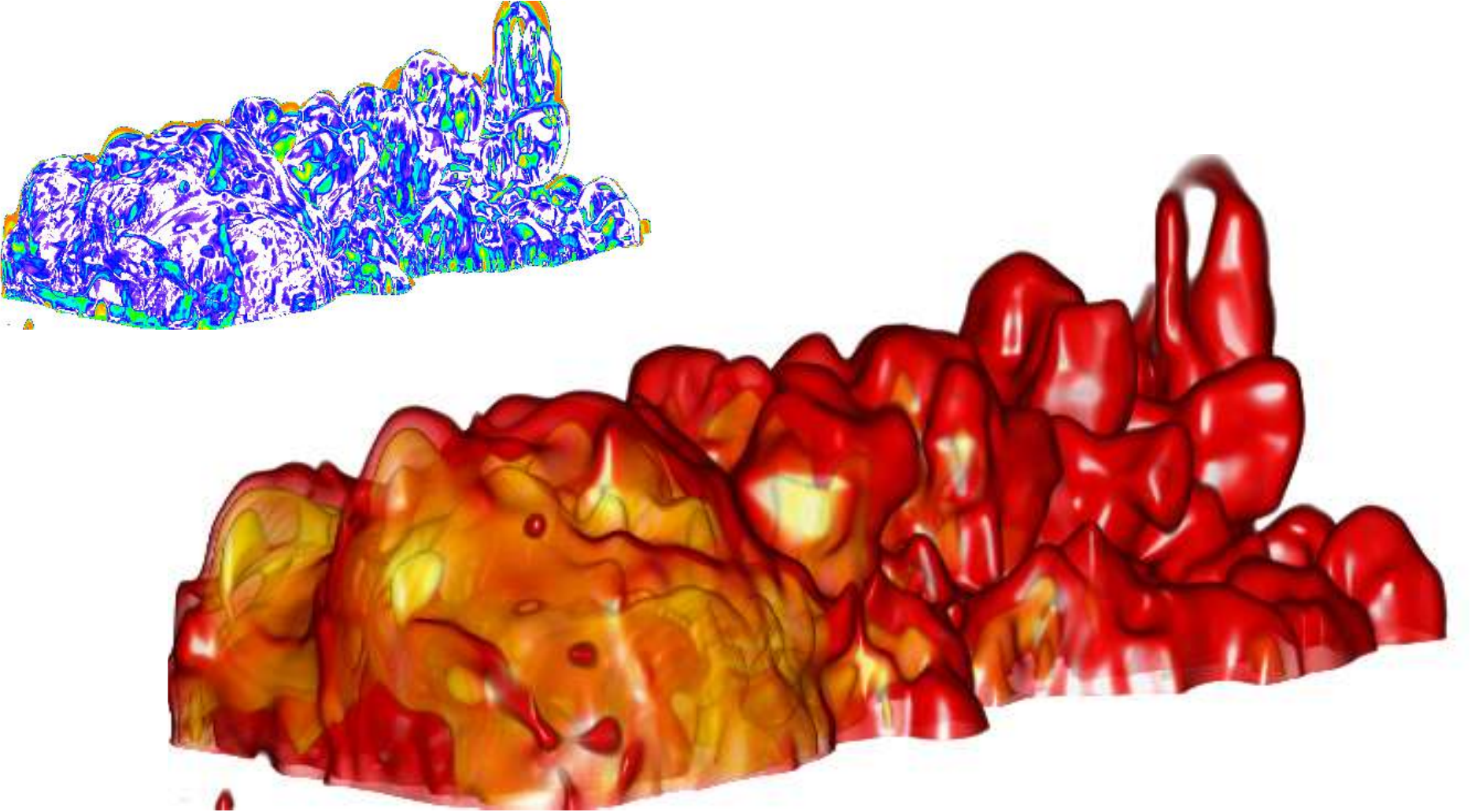}\\
				\includegraphics[height=0.585in]{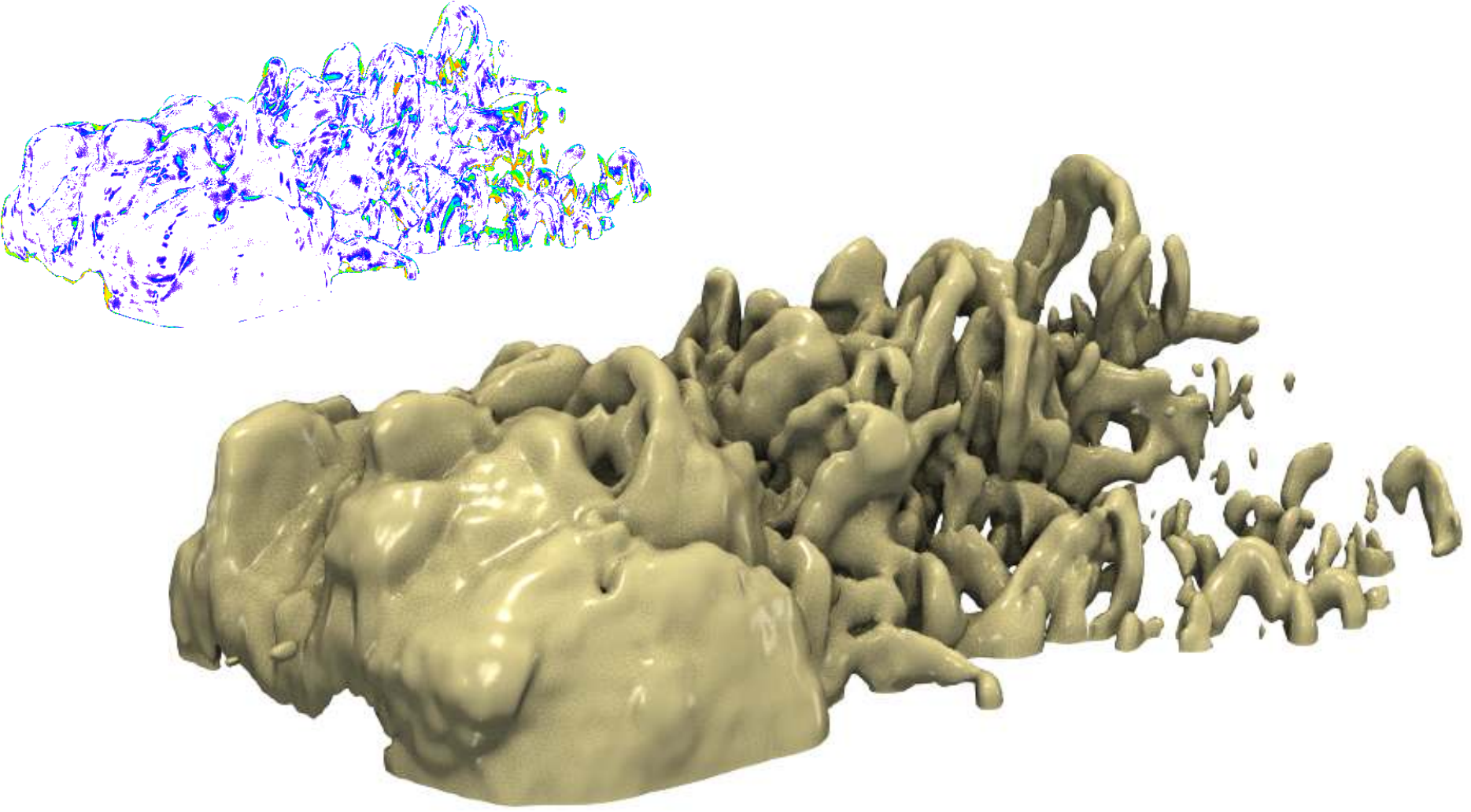}&
				\includegraphics[height=0.585in]{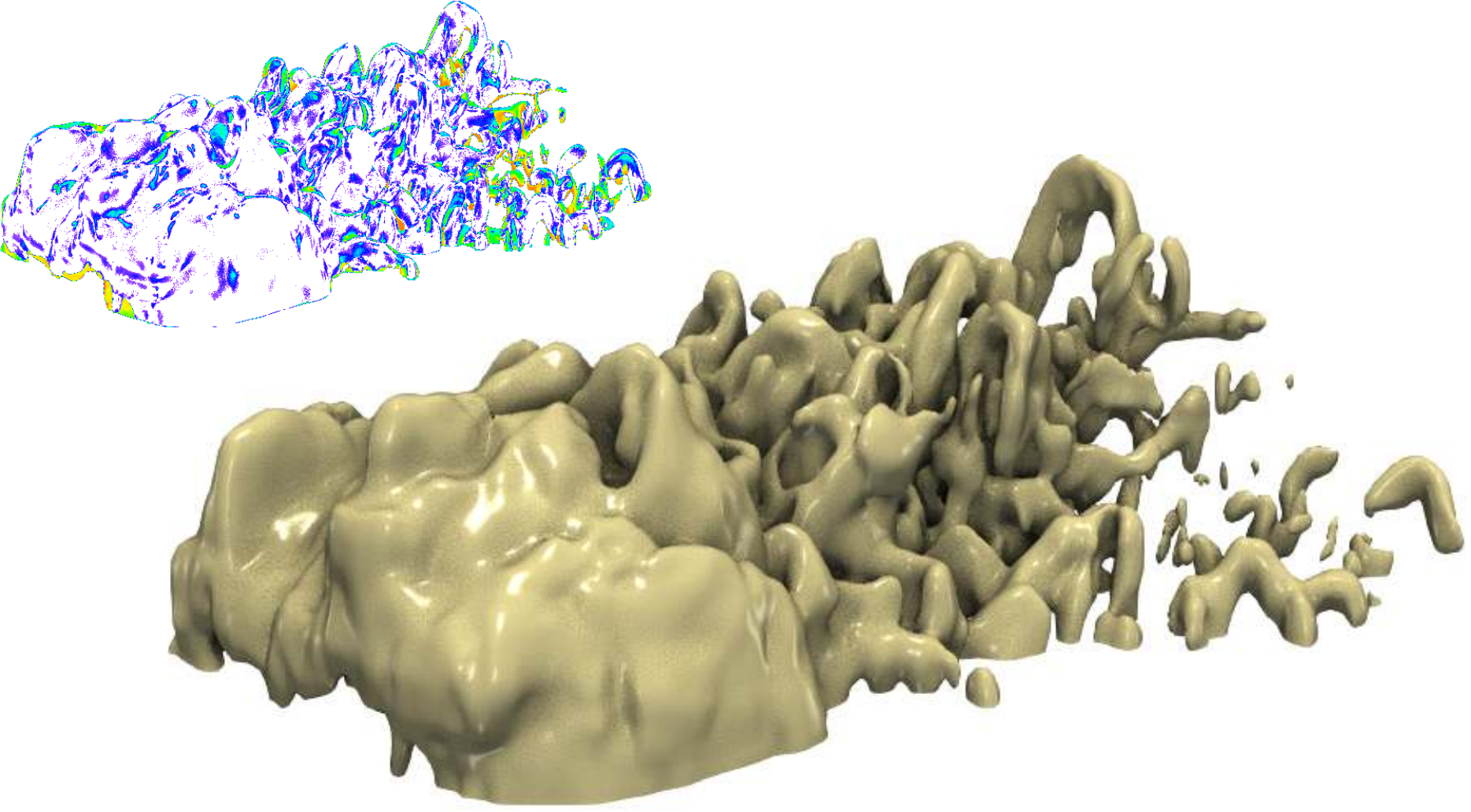}&
				\includegraphics[height=0.585in]{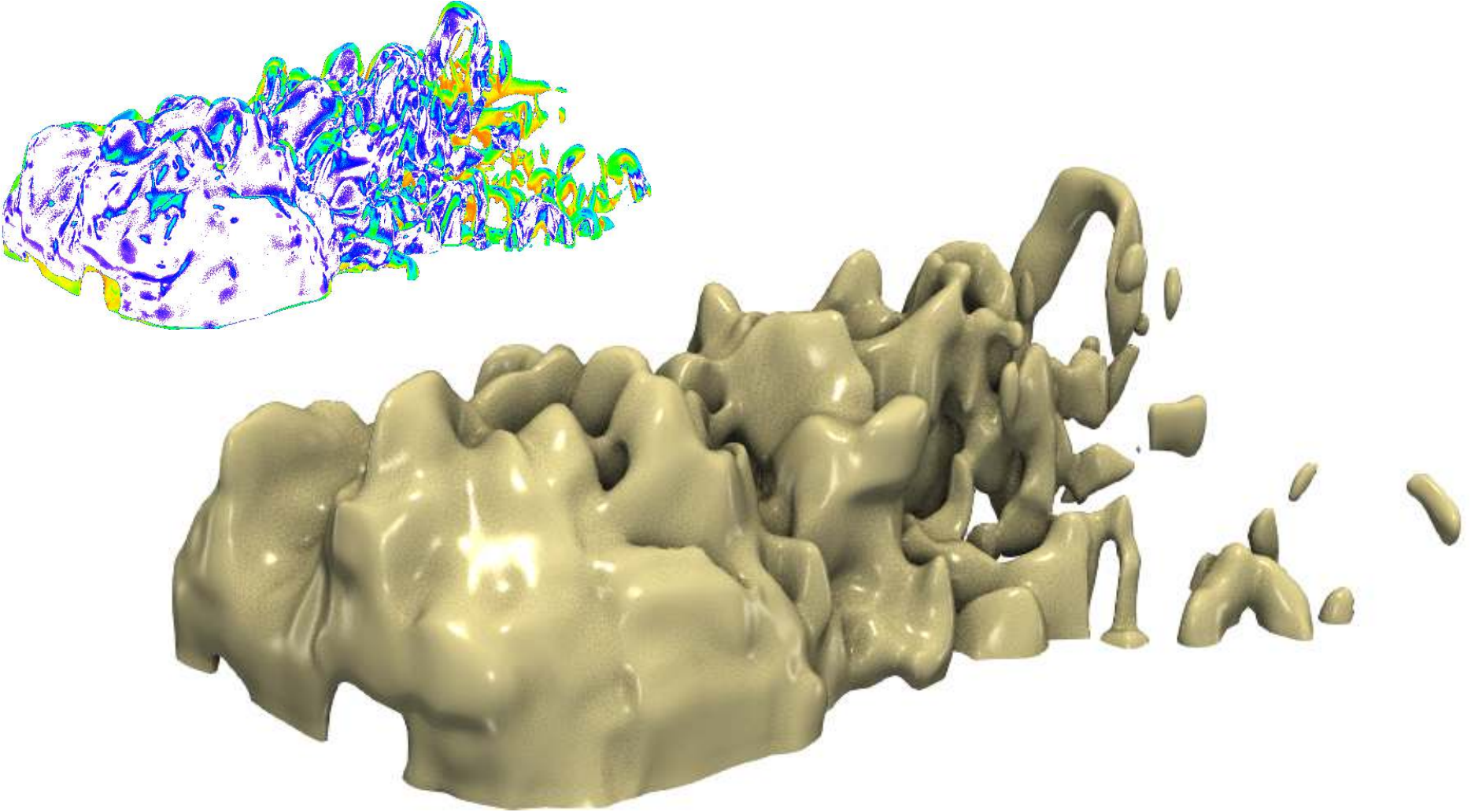}\\				
				\mbox{\footnotesize (a) $25 \times 5 \times 5$} &
				\mbox{\footnotesize (b) $25 \times 15 \times 10$} &
				\mbox{\footnotesize (c) $25 \times 45 \times 15$}\\
	\end{array}$
	\end{center}	
	\vspace{-.25in}
	\caption{Volume rendering and isosurface rendering ($v=-0.85$) results of ECNR using the Tangaroa dataset with different block dimensions.}
	\label{fig:block-dims}			
\end{figure}

{\bf Block dimension.}
Among ECNR's hyperparameters, block dimension emerges as a salient factor to adjust the CR. We evaluated ECNR on the Tangaroa dataset with different block dimensions to study quality performance under various CRs. We report quantitative results in Table~\ref{tab:block-dimension} and corresponding volume rendering and isosurface rendering results in Figure~\ref{fig:block-dims}. As block dimension increases, the CR increases, reconstruction quality decreases, and encoding time and decoding time increase because each MLP needs to represent larger blocks, which hurts parallel efficiency. We chose a block dimension of $25 \times 15 \times 10$ for the Tangaroa dataset as a tradeoff.


\begin{table}[htb]
\caption{Average PSNR (dB), LPIPS, and CD ($v = -0.85$) values across all timesteps, as well as \hot{encoding time (ET), decoding time (DT), and model size (MS, in MB)} of the Tangaroa dataset.}
\vspace{-0.1in}
\centering
{\scriptsize
\resizebox{\columnwidth}{!}{
\begin{tabular}{c|lll|ll|l}
block dimension 				& PSNR$\uparrow$	& LPIPS$\downarrow$ &CD$\downarrow$	& ET$\downarrow$	& DT$\downarrow$	 & MS$\downarrow$\\ \hline
$25 \times 5 \times 5$ 	&\textbf{43.84}  &\textbf{0.038} &\textbf{1.07} & \textbf{91.2m}		&\textbf{131.7s} 	 &20.1 \\ 
$25 \times 15 \times 10$ 	&41.47	&0.046 &1.50 & 98.5m & 133.2s &4.6 \\
$25 \times 45 \times 15$ 	&38.53 	&0.062 &2.74 &165.6m & 140.6s &\textbf{1.5} \\
\end{tabular}
}}
\label{tab:block-dimension}
\end{table}


\hot{{\bf Choice of  $s$.}
In Table~\ref{tab:diff-scales} and Figure~\ref{fig:diff-scales}, we compare the result of ECNR using different scales $s$ on the half-cylinder dataset with an identical block dimension $40 \times 15 \times 5$. When $s$ increases, the total block numbers will increase, but it may also filter out more non-effective blocks. The results show that $s=3$ could reach the lowest percentage of effective blocks and preserve good quality with the smallest model size. Therefore, we set $s=3$ for most datasets.}

\begin{table}[htb]
\caption{\hot{PSNR (dB), LPIPS, and CD ($v=0.1$) values, as well as model size (MS, in MB), effective blocks (EB), total blocks (TB), and percentage of effective blocks (EB/TB) of the half-cylinder dataset using different $s$ values.}}
\vspace{-0.1in}
\centering
{\scriptsize
\resizebox{\columnwidth}{!}{
\hot{
\begin{tabular}{c|llll|ccc}
$s$ & PSNR$\uparrow$ & LPIPS$\downarrow$ & CD$\downarrow$ & MS$\downarrow$ & EB$\downarrow$  & TB$\downarrow$ & EB/TB$\downarrow$\\ \hline
2 &\textbf{46.60} &0.016 &0.790 &11.2&141,377  &\textbf{435,200} &32.49\%\\
3 &46.34 &0.019 &\textbf{0.776} &\textbf{8.7} &\textbf{112,448}  &436,800 &\textbf{25.74\%}\\ 
4 &45.91 &\textbf{0.015} &0.814 &9.2 &118,075  &436,904 &27.03\%\\ 
\end{tabular}	
}
}}
\label{tab:diff-scales}
\end{table}

\hot{
\begin{figure}[htb]
	\begin{center}
		$\begin{array}{c@{\hspace{0.02in}}c@{\hspace{0.02in}}c@{\hspace{0.02in}}c}
				\includegraphics[height=0.535in]{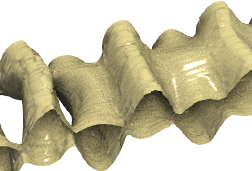}&
				\includegraphics[height=0.535in]{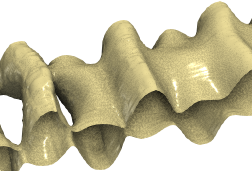}&
				\includegraphics[height=0.535in]{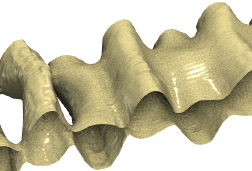} &
				\includegraphics[height=0.535in]{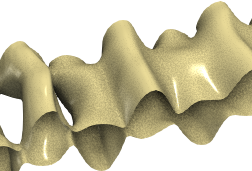} \\
				\mbox{\footnotesize (a) \hot{$s=2$}} &
				\mbox{\footnotesize (b) \hot{$s=3$}} &
				\mbox{\footnotesize (c) \hot{$s=4$}} &
				\mbox{\footnotesize (d) \hot{GT}} \\
		\end{array}$
	\end{center}
	\vspace{-.25in}
	\caption{\hot{Zoomed-in isosurface rendering ($v=0.1$) results of the half-cylinder dataset with ECNR under different scales.}}
	\label{fig:diff-scales}
\end{figure}
}

\begin{figure}[H]
	\begin{center}
		$\begin{array}{c@{\hspace{0.05in}}c@{\hspace{0.05in}}c}
				\includegraphics[height=1in]{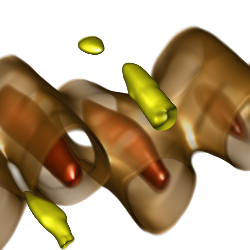} &
				\includegraphics[height=1in]{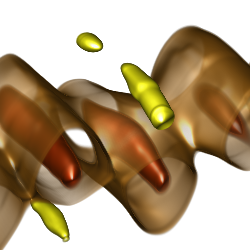} &
				\includegraphics[height=1in]{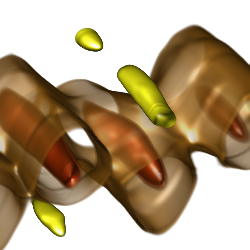} \\
				\mbox{\footnotesize (a) $\lambda_b=1$} & \mbox{\footnotesize (b) $\lambda_b=0$} & \mbox{\footnotesize (c) $\lambda_b=0.1$  } \\
			\end{array}$
		$\begin{array}{c@{\hspace{0.05in}}c}
				\includegraphics[height=1in]{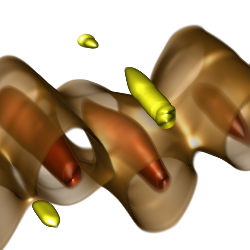} &
				\includegraphics[height=1in]{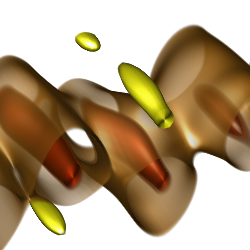} \\
				\mbox{\footnotesize (d) $\lambda_b=0.01$} & \mbox{\footnotesize (e) GT}\\
			\end{array}$
	\end{center}
	\vspace{-.25in}
	\caption{Zoomed-in volume rendering results of the half-cylinder dataset with ECNR under different $\lambda_b$.}
	\label{fig:lambdab}
\end{figure}

{\bf $\lambda_b$ in \hot{block-guided pruning}.}
To choose an appropriate weight $\lambda_b$ for pruning, we optimized our model on the half-cylinder dataset with different $\lambda_b$ values. 
In Table~\ref{tab:lambdab}, we report the average PSNR, LPIPS, and CD values using different $\lambda_b$. 
Figure~\ref{fig:lambdab} shows the zoomed-in volume rendering results. 
$\lambda_b=0.1$ achieves the overall best quality, and we chose this setting for all experiments.

\begin{table}[H]
\caption{Average PSNR (dB), LPIPS, and CD ($v=0.1$) values across all timesteps of the half-cylinder dataset.}
\vspace{-0.1in}
\centering
{\scriptsize
\begin{tabular}{c|lll}
$\lambda_b$ & PSNR$\uparrow$ & LPIPS$\downarrow$ &CD$\downarrow$\\ \hline
1.0 &43.41 &0.031  &1.04\\
0.0 &44.69 &0.027 &\textbf{0.86} \\ 
0.1 &\textbf{45.12} &\textbf{0.026} & 0.87 \\ 
0.01&44.98 &\textbf{0.026} &0.92
\end{tabular}
}
\label{tab:lambdab}
\end{table}

%


%
%

\end{document}